\documentclass[11pt]{article}
\newlength{\defbaselineskip}
\usepackage{lmodern}

\usepackage{amsmath}
\usepackage{amssymb}
\usepackage{amsfonts}

\usepackage{paralist}
\usepackage[utf8]{inputenc} 
\usepackage[T1]{fontenc}    
\usepackage{hyperref}       
\usepackage{url}            
\usepackage{booktabs}       
\usepackage{nicefrac}       

\usepackage{graphicx}
\usepackage{color}
\usepackage{xcolor}         
\usepackage{tikz}

\usepackage{algorithm}
\usepackage{algpseudocode}
\usepackage{framed}
\usepackage{cancel}
\usepackage{xspace}

\usepackage{enumitem}
\usepackage{subfigure}
\usepackage{makecell}
\usepackage{soul}
\usepackage{wrapfig}

\usepackage{rotating}
\usepackage{multirow}
\usepackage{longtable} 

\usepackage[multiple]{footmisc}
\usepackage[section]{placeins}

\usepackage{amsmath}
\usepackage{graphicx}

\hypersetup{
     colorlinks   = true,
     linkcolor    = blue,
     citecolor    = green
}
\usepackage{MnSymbol}

\newcommand{\argmin}{\text{argmin}}
\newcommand{\Probab}[1]{\mbox{}{\bf{Pr}}\left[#1\right]}

\newcommand{\Trace}[1]{\mbox{}\operatorname{Tr}\left[#1\right]}

\newcommand{\Expected}[2][]{\left\langle #2 \right\rangle_{#1}}

\newcommand{\LAMBDAPL}{\lambda^{PL}_{min}}
\newcommand{\LAMBDADETX}{\lambda^{\vert detX\vert=1}_{min}}

\definecolor{darkgreen}{rgb}{0.05, 0.65, 0.06}
\newcommand {\charles}[1]{}

\newcommand {\michael}[1]{}
\newcommand {\michaeladdressed}[1]{}
\newcommand {\michaelB}[1]{}
\newcommand {\charlesB}[1]{}

\newcommand {\cformike}[1]{}

\newcommand {\chris}[1]{}

\usepackage[normalem]{ulem}

\usepackage{xspace}
\newcommand{\TR}{\top}

\newcommand{\SPECTRALNORM}{\texttt{SpectralNorm}\xspace}
\newcommand{\LOGSPECTRALNORM}{\texttt{LogSpectralNorm}\xspace}

\newcommand{\ALPHA}{\texttt{Alpha}\xspace}

\newcommand{\ALPHAHAT}{\texttt{AlphaHat}\xspace}

\newcommand{\ALPHAHATEQN}{\hat\alpha}
\newcommand{\ALPHAHATLONG}{\alpha\log_{10}\lambda_{max}}

\newcommand{\RANDDIST}{\texttt{Rand-Distance}}

\newcommand{\WW}{\texttt{WeightWatcher}\xspace}

\newcommand{\HCIZtext}{\texttt{HCIZ}\xspace}
\newcommand{\SLT}{\texttt{SLT}\xspace}
\newcommand{\RMT}{\texttt{RMT}\xspace}
\newcommand{\HTRMT}{\texttt{HTRMT}\xspace}
\newcommand{\STATMECH}{\texttt{StatMech}\xspace}
\newcommand{\SETOL}{\texttt{SETOL}\xspace}
\newcommand{\SMOG}{\texttt{SMOG}\xspace}
\newcommand{\HTSR}{\texttt{HTSR}\xspace}

\newcommand{\SEMIEMP}{\texttt{SemiEmpirical}\xspace}

\newcommand{\PLKS}{\texttt{PL KS}}
\newcommand{\MPSOFTRANK}{\texttt{MP SoftRank}}

\newcommand{\TRACELOG}{\texttt{ERG}\xspace}
\newcommand{\ERG}{\texttt{ERG}\xspace}
\newcommand{\IFA}{\texttt{IFA}\xspace}
\newcommand{\ECS}{\texttt{ECS}\xspace}

\newcommand{\POWERLAW}{\texttt{PowerLaw}}

\newcommand{\SHAPE}{\emph{Shape}\xspace}
\newcommand{\SCALE}{\emph{Scale}\xspace}

\newcommand{\IZFE}{\beta\mathbf{F}^{IZ}}

\newcommand{\Q}{\bar{\mathcal{Q}}}
\newcommand{\QT}{\bar{\mathcal{Q}}^{2}}

\newcommand{\STG}{\beta\mathbf{\Gamma}^{ST}_{\Q}}
\newcommand{\IZG}{\beta\mathbf{\Gamma}^{IZ}_{\QT}}
\newcommand{\IZGINF}{\beta\mathbf{\Gamma}^{IZ}_{\QT,N\gg 1}}

\newcommand{\GN}{\mathcal{G(\lambda)}}
\newcommand{\GNI}{\mathcal{G}(\lambda_{i})}
\newcommand{\GNECSI}{\mathcal{G}(\LambdaECS_{i})}
\newcommand{\GEN}{Norm Generating Function}

\newcommand{\OVERLAP}{\mathbf{R}}
\newcommand{\OLAPTOLAP}{{\OVERLAP}^{\TR}\OVERLAP}
\newcommand{\OLAPSQD}{\Trace{\mathbf{R}^{\TR}\mathbf{R}}}
\newcommand{\THRMAVGIZ}[1]{\left\langle #1 \right\rangle_{\mathbf S}^{\beta}}
\newcommand{\THRMAVG}[1]{\left\langle #1 \right\rangle_{\mathbf s}^{\beta}}
\newcommand{\THRMAVGw}[1]{\left\langle #1 \right\rangle_{\mathbf w}^{\beta}}

\newcommand{\EFF}{ecs}
\newcommand{\HCIZ}{\mathbb{Z}^{IZ}}
\newcommand{\NDX}{\mathbf{x}^{n}}

\newcommand{\AVGE}{\bar{\mathcal{E}}}
\newcommand{\AVGTE}{\bar{\mathcal{E}}_{train}}
\newcommand{\AVGGE}{\bar{\mathcal{E}}_{gen}}

\newcommand{\TTE}{{\mathcal{E}}_{train}}
\newcommand{\TGE}{{\mathcal{E}}_{gen}}

\newcommand{\STGE}{{\mathcal{E}}^{ST}_{gen}}
\newcommand{\AVGSTTE}{\bar{\mathcal{E}}^{ST}_{train}}
\newcommand{\AVGSTGE}{\bar{\mathcal{E}}^{ST}_{gen}}

\newcommand{\AVGNNTE}{\bar{\mathcal{E}}^{NN}_{train}}
\newcommand{\AVGNNGE}{\bar{\mathcal{E}}^{NN}_{gen}}

\newcommand{\AVGEMPTE}{\bar{\mathcal{E}}^{emp}_{train}}
\newcommand{\AVGEMPGE}{\bar{\mathcal{E}}^{emp}_{gen}}

\newcommand{\Ntest}{N^{test}}
\newcommand{\DX}{\mathbf{x}}
\newcommand{\DXtrain}{\mathbf{x}^{train}}
\newcommand{\DXtest}{\mathbf{x}^{test}}
\newcommand{\DATA}{\mathbf{x}_{\mu}}
\newcommand{\DATAtrain}{\mathbf{x}_{\mu}^{train}}
\newcommand{\DATAtest}{\mathbf{x}_{\mu}^{test}}

\newcommand{\MY}{\mathrm{y}}

\newcommand{\Ys}{\MY_{\mu}^{S}}
\newcommand{\Yt}{\MY_{\mu}^{T}}
\newcommand{\Ytrue}{\MY_{\mu}^{true}}
\newcommand{\Ytrain}{\MY_{\mu}^{train}}
\newcommand{\Ytest}{\MY_{\mu}^{test}}
\newcommand{\YsVEC}{\mathbf{y}^{S}}
\newcommand{\YtVEC}{\mathbf{y}^{T}}
\newcommand{\Ymu}{\MY_{\mu}}
\newcommand{\YsIS}[1]{\MY_{\mu=#1}^{S}}
\newcommand{\YtIS}[1]{\MY_{\mu=#1}^{T}}

\newcommand{\XI}{\boldsymbol{\xi}}
\newcommand{\XImu}{\boldsymbol{\xi}_{\mu}}
\newcommand{\NDXI}{\boldsymbol{\xi}^{n}}
\newcommand{\NDXIn}{\boldsymbol{\xi}^{n}}
\newcommand{\AVGNDXI}{\overset{\rule{0.6em}{0.7pt}}{\scriptscriptstyle\boldsymbol{\xi}^{n}}}
\newcommand{\AVGNDXIn}{\overset{\rule{0.6em}{0.7pt}}{\scriptscriptstyle\boldsymbol{\xi}^{\ND}}}

\newcommand{\XItrain}{\boldsymbol{\xi}^{train}}

\newcommand{\SVEC}{{\mathbf{s}}}
\newcommand{\TVEC}{{\mathbf{t}}}
\newcommand{\WVEC}{{\mathbf{w}}}
\newcommand{\XVEC}{{\mathbf{x}}}

\newcommand{\NNOUT}{E^{out}_{NN}}
\newcommand{\TOUT}{E^{out}_{T}}
\newcommand{\SOUT}{E^{out}_{S}}

\newcommand{\EPSL}{\epsilon}
\newcommand{\EPSLR}{\epsilon(R)}

\newcommand{\EPSLw}{\epsilon(\WVEC)}

\newcommand{\EPSLSTx}{\epsilon(\SVEC, \TVEC )}

\newcommand{\DELBF}{\mathbf{\mathrm{E}}^{n}_{\mathcal{L}}}
\newcommand{\DELBFell}{\mathbf{\mathrm{E}}^{n}_{\mathcal{\ell}}}
\newcommand{\DET}{\mathrm{E}_{\mathcal{L}}(\mathbf{w},\XI)}
\newcommand{\DETmu}{{E}_{\mathcal{L}}(\mathbf{w},\XI_{\mu})}

\newcommand{\DETOPSTL}{{\mathrm{E}}^{n}_{\mathcal{L}}(\SVEC,\TVEC,\NDXI)}
\newcommand{\DETOPSTx}{{\mathrm{E}}^{n}_{\mathcal{L}}(S,T,\DX)}
\newcommand{\DETSTLL}{{E}_{\ell_2}(\SVEC,\TVEC,\XI)}
\newcommand{\DETOP}{{\mathrm{E}}^{n}_{\mathcal{L}}(\mathbf{w})}
\newcommand{\DETOPX}{{\mathrm{E}}^{n}_{\mathcal{L}}(\mathbf{w}, \NDX)}

\newcommand{\DETOPXY}{\mathrm{{E}}^{n}_{\mathcal{L}}(\mathbf{w}, \NDX, \MY^{n})}
\newcommand{\DEL}{{E}_{\mathcal{L}}}

\usepackage{bm}

\newcommand{\DETOPXI}{\mathbf{\mathrm{E}}^{n}_{\mathcal{L}}(\mathbf{w}, \XI^{n})}
\newcommand{\DETOPXILL}{\mathbf{\mathrm{E}}^{n}_{\ell_2}(\mathbf{w}, \XI^{n})}
\newcommand{\DETOPSTLL}{\mathbf{\mathrm{E}}^{n}_{\ell_2}(\SVEC,\TVEC, \XI^{n})}
\newcommand{\DETOPST}{\mathbf{\mathrm{E}}^{n}_{\ell_2}(\SMAT,\TMAT)}
\newcommand{\DETOPNN}{\mathbf{\mathrm{E}}^{n}_{\ell_2}(\SMAT,\TMAT, \NDXI)}

\newcommand{\DETOT}{\mathcal{E}(\mathbf{w})}

\newcommand{\ND}{n}

\newcommand{\SMAT}{{\mathbf{S}}}
\newcommand{\TMAT}{{\mathbf{T}}}
\newcommand{\WMAT}{{\mathbf{W}}}
\newcommand{\AECS}{\tilde{\mathbf{A}}}
\newcommand{\AECSM}{\tilde{\mathbf{A}}_{\text{\tiny M}}}
\newcommand{\AECSN}{\tilde{\mathbf{A}}_{\text{\tiny N}}}
\newcommand{\AMATM}{{\mathbf{A}}_{\text{\tiny M}}}
\newcommand{\AMATN}{{\mathbf{A}}_{\text{\tiny N}}}

\newcommand{\XECS}{\tilde{\mathbf{X}}}
\newcommand{\TECS}{\tilde{\mathbf{T}}}
\newcommand{\LambdaECS}{\tilde{\lambda}}
\newcommand{\LambdaMax}{\lambda_{max}}
\newcommand{\LambdaMin}{\lambda_{0}}
\newcommand{\EVEC}{\mathbf{E}^{out}}

\newcommand{\LambdaECSmin}{{\LambdaECS_{min}}}
\newcommand{\LambdaECSmax}{\lambda^{ECS}_{max}}
\newcommand{\LambdaPLmin}{\lambda^{PL}_{min}}

\newcommand{\LambdaCutoffMin}{\lambda_{0}}

\newcommand{\MECS}{\tilde{M}}

\newcommand{\AMAT}{\mathbf{A}}
\newcommand{\AHAT}{\hat{\mathbf{A}}}

\newcommand{\XMAT}{\mathbf{X}}
\newcommand{\DMAT}{\mathbf{D}}

\newcommand{\GAN}{H^{an}(\WVEC)}
\newcommand{\HAN}{H^{an}}
\newcommand{\GANR}{H^{an}(R)}
\newcommand{\GANMAT}{H^{an}(\mathbf{R})}
\newcommand{\GANHT}{H^{an}_{hT}(\WVEC)}
\newcommand{\GANHTR}{H^{an}_{hT}(R)}
\newcommand{\GANMATHT}{H^{an}_{hT}(\mathbf{R})}
\newcommand{\HANHT}{H^{an}_{hT}}
\newcommand{\HEFF}{\mathbf{H}^{ECS}_{\QT}}
\newcommand{\HBARE}{\mathbf{H}_{\QT}}
\newcommand{\HANPP}{h^{an}(\OVERLAP)}
\newcommand{\HANPPHT}{h^{an}_{hT}(\OVERLAP)}

\newcommand{\ADD}{\mathbf{D}}
\newcommand{\MDD}{\mathcal{D}}

\newcommand{\XINORM}{\mathcal{N}}

\newcommand{\ZAN}{Z^{an}_{\ND}}

\newcommand{\ZANHT}{Z^{an,hT}_{\ND}}

\newcommand{\HH}{\mathbf{H}}

\newcommand{\INTA}{\int_{\mathbf{A}}}
\newcommand{\INTS}{\int_{\SMAT}}
\newcommand{\INTsvec}{\int_{\SVEC}}
\newcommand{\INTAHAT}{\int_{\AHAT}}

\newcommand{\XHAT}{\hat{\mathbf{X}}}

\newcommand{\NORM}{\mathcal{N}}

\newcommand{\GNORM}{\mathbb{G}_{\AMAT}}
\newcommand{\GMAX}{\mathcal{G}^{max}}
\newcommand{\GFANCY}{\mathcal{G}(\XMAT)}
\newcommand{\ZIZ}{\mathbb{Z}^{IZ}}
\newcommand{\ZD}{\mathbb{Z}^{IZ}}
\newcommand{\EZDA}{\mathbb{E}_{\AMAT}[\ZIZ]}
\newcommand{\EZDX}{\mathbb{E}_{\XMAT}[\ZIZ]}

\newcommand{\EZDAONE}{\mathbb{E}_{\AMAT_{M}}[\ZIZ]}
\newcommand{\EZDATWO}{\mathbb{E}_{\AMAT_{N}}[\ZIZ]}

\newcommand{\EQN}{Eqn.}
\newcommand{\Det}[1]{\mbox{}{\text{det}}\left(#1\right)}
\newcommand{\DeltaMu}{\vartheta_{\mu}}
\newcommand{\IM}{\mathbf{I}_{\text{\tiny M}}} 
\newcommand{\IMm}{\mathbf{I}_m} 
\newcommand{\IH}{\mathbb{I}_{\text{\tiny H}}} 
\newcommand{\DETX}{\texttt{DetX}\xspace}
\newcommand{\ETA}{\eta\xspace}
\newcommand{\ETAw}{\eta(\WVEC)\xspace}
\newcommand{\ETAMLPXI}{\eta([\SMAT_l,\TMAT_l],\NDXI)\xspace}
\newcommand{\ETAMLPAVG}{\langle\eta([\SMAT_l,\TMAT_l],\NDXI)\rangle_{\AVGNDXI}\xspace}
\newcommand{\ETAMLP}{\eta([\SMAT_l,\TMAT_l])\xspace}
\newcommand{\ETAMAT}{\eta(\SMAT_l,\TMAT_l,\NDXI)\xspace}
\newcommand{\ETAMATAVG}{\langle\eta(\SMAT_l,\TMAT_l,\NDXI)\rangle_{\AVGNDXI}\xspace}
\newcommand{\AVGR}{R\xspace}
\newcommand{\LargeN}{Large-$\mathcal{N}\;$\xspace}
\newcommand{\NWICK}{\mathcal{N}_{\!\scalebox{0.6}{\textit{Wick}}}}
\providecommand{\PhasesOfTraining}{5+1 Phases of Training\xspace}
\providecommand{\SemiEmpirical}{Semi-Empirical\xspace}
\providecommand{\EmpiricalSpectralDensity}{Empirical Spectral Density\xspace}
\providecommand{\ThermodynamicLimit}{Thermodynamic limit\xspace}
\providecommand{\Thermodynamic}{Thermodynamic\xspace}
\providecommand{\WideLayer}{Wide Layer\xspace}
\providecommand{\Phenomenology}{phenomenology\xspace}

\providecommand{\Shape}{Shape\xspace}
\providecommand{\Scale}{Scale\xspace}
\providecommand{\IdealLearning}{Ideal Learning\xspace}
\providecommand{\Ideal}{Ideal\xspace}
\providecommand{\Typical}{typical\xspace}
\providecommand{\ATypical}{atypical\xspace}
\providecommand{\OverRegularized}{Over-Regularized\xspace}
\providecommand{\OverRegularization}{Over-Regularization\xspace}
\providecommand{\DragonKings}{Dragon Kings\xspace}

\providecommand{\HighTemperature}{High-Temperature\xspace}
\providecommand{\DataModel}{Data Model\xspace}

\providecommand{\Quenched}{Quenched\xspace}
\providecommand{\AnnealedApproximation}{Annealed Approximation\xspace}
\providecommand{\Annealed}{Annealed\xspace}
\providecommand{\SaddlePointApproximation}{Saddle Point Approximation\xspace}
\providecommand{\VolumePreservingTransformation}{Volume Preserving Transformation\xspace}
\providecommand{\EffectiveCorrelationSpace}{Effective Correlation Space\xspace}

\providecommand{\CorrelationMatrix}{Correlation Matrix\xspace}
\providecommand{\StudentTeacher}{Student-Teacher\xspace}
\providecommand{\AnnealedHamiltonian}{Annealed Hamiltonian\xspace}
\providecommand{\EnergyLandscape}{Energy Landscape\xspace}
\providecommand{\ThermalAverages}{Thermal average\xspace}
\providecommand{\ThermalAverage}{Thermal Average\xspace}
\providecommand{\LearningRate}{Learning Rate\xspace}

\providecommand{\PowerLaw}{Power Law\xspace}

\providecommand{\QualitySquared}{Quality-Squared\xspace}
\providecommand{\TotalEffectivePotential}{Total Annealed Error Potential\xspace}
\providecommand{\EffectivePotential}{Annealed Error Potential\xspace}
\providecommand{\HeavyTailedSelfRegularization}{Heavy-Tailed Self-Regularization\xspace}
\providecommand{\VeryHeavyTailed}{Very Heavy-Tailed\xspace}
\providecommand{\HeavyTailed}{Heavy-Tailed\xspace}
\providecommand{\StateOfTheArt}{State-of-the-Art\xspace}
\providecommand{\StatisticalLearningTheory}{Statistical Learning Theory\xspace}
\providecommand{\StatisticalMechanics}{Statistical Mechanics\xspace}
\providecommand{\StatisticalMechanicsOfGeneralization}{Statistical Mechanics of Generalization\xspace}
\providecommand{\CorrelationTraps}{Correlation Traps\xspace}
\providecommand{\CorrelationTrap}{Correlation Trap\xspace}

\providecommand{\AverageTrainingError}{Average Training Error\xspace}
\providecommand{\AverageGeneralizationError}{Average Generalization Error\xspace}
\providecommand{\AverageGeneralizationAccuracy}{Average Generalization Accuracy\xspace}

\providecommand{\AverageSTGeneralizationError}{Average ST Generalization Error\xspace}
\providecommand{\AverageLayerQualitySquared}{Average Layer Quality-Squared\xspace}
\providecommand{\LayerQualitySquared}{Layer Quality-Squared\xspace}

\providecommand{\ModelGeneralizationError}{Model Generalization Accuracy\xspace}

\providecommand{\GeneralizationAccuracy}{Generalization Accuracy\xspace}
\providecommand{\TotalDataSampleError}{Total Data Sample Error\xspace}

\providecommand{\DataSampleError}{Data Sample Error\xspace}

\providecommand{\TrainingError}{Training Error\xspace}
\providecommand{\GeneralizationError}{Generalization Error\xspace}
\providecommand{\GeneralizationGap}{Generalization Gap\xspace}
\providecommand{\ModelQuality}{Model Quality\xspace}
\providecommand{\LayerQuality}{Layer Quality\xspace}
\providecommand{\LayerQualities}{Layer Qualities\xspace}
\providecommand{\MomentGeneratingFunction}{Moment Generating Function\xspace}
\providecommand{\CumulantGeneratingFunction}{Cumulant Generating Function\xspace}
\providecommand{\GeneratingFunctions}{Generating Functions\xspace}
\providecommand{\GeneratingFunction}{Generating Function\xspace}
\providecommand{\SourceMatrix}{Source Matrix\xspace}

\providecommand{\RateFunction}{Rate Function\xspace}
\providecommand{\ExpectedValue}{Expected Value\xspace}
\providecommand{\ExpectedValues}{Expected Values\xspace}
\providecommand{\Glassy}{glassy\xspace}
\providecommand{\WickRotation}{Wick rotation\xspace}
\providecommand{\WickRotations}{Wick rotations\xspace}

\providecommand{\Replica}{Replica\xspace}
\providecommand{\PartitionFunction}{Partition Function\xspace}
\providecommand{\FreeEnergy}{Free Energy\xspace}
\providecommand{\FreeEnergies}{Free Energies\xspace}
\providecommand{\RandomMatrixTheory}{Random Matrix Theory\xspace}

\providecommand{\SelfOverlap}{Self-Overlap\xspace}
\providecommand{\IndependentFluctuationApproximation}{Independent Fluctuation Approximation\xspace}

\providecommand{\BoltzmannWeightedAverage}{Boltzmann-weighted average\xspace}
\providecommand{\RTransforms}{R-transforms\xspace}
\providecommand{\RTransform}{R-transform\xspace}

\providecommand{\GreensFunction}{Greens Function\xspace}
\providecommand{\InverseMP}{Inverse Marchenko-Pastur\xspace}

\providecommand{\FatTailed}{Fat-Tailed\xspace}
\providecommand{\MPBulk}{MP Bulk\xspace}
\providecommand{\RandomLike}{Random-Like\xspace}
\providecommand{\LevyWigner}{Levy-Wigner\xspace}
\providecommand{\Wigner}{Wigner\xspace}
\providecommand{\FreeCumulants}{Free Cumulants\xspace}
\providecommand{\Cumulants}{Cumulants\xspace}
\providecommand{\Cauchy}{Cauchy\xspace}
\providecommand{\CauchyStieltjes}{Cauchy-Stieltjes\xspace}
\providecommand{\MultiLayer}{Multi-Layer\xspace}
\providecommand{\LinearPerceptron}{Linear Perceptron\xspace}
\providecommand{\MultiLayerPerceptron}{Multi-Layer Perceptron\xspace}
\providecommand{\Perceptrons}{Perceptrons\xspace}
\providecommand{\Perceptron}{Perceptron\xspace}
\providecommand{\BraKet}{Bra-Ket\xspace}
\providecommand{\Quality}{Quality\xspace}
\providecommand{\Student}{Student\xspace}
\providecommand{\Teacher}{Teacher\xspace}
\providecommand{\Students}{Students\xspace}

\providecommand{\Universality}{Universality\xspace}
\providecommand{\VolumePreserving}{Volume-Preserving\xspace}
\providecommand{\ScaleInvariant}{Scale-Invariant\xspace}

\providecommand{\Bare}{bare\xspace}

\providecommand{\Hamiltonian}{Hamiltonian\xspace}
\providecommand{\EffectiveHamiltonian}{Effective Hamiltonian\xspace}
\providecommand{\QuantumChemistry}{Quantum Chemistry\xspace}
\providecommand{\NuclearPhysics}{Nuclear Physics\xspace}
\providecommand{\SelfAveraging}{Self-Averaging\xspace}
\providecommand{\RenormalizationGroup}{Renormalization Group\xspace}
\providecommand{\ExactRenormalizationGroup}{Exact Renormalization Group\xspace}
\providecommand{\WilsonExactRenormalizationGroup}{Wilson Exact Renormalization Group\xspace}

\providecommand{\SizeConsistency}{Size-Consistency\xspace}
\providecommand{\SizeExtensivity}{Size-Extensivity\xspace}
\providecommand{\SizeIntensivity}{Size-Intensivity\xspace}
\providecommand{\SizeConsistent}{Size-Consistent\xspace}
\providecommand{\SizeExtensive}{Size-Extensive\xspace}
\providecommand{\SizeIntensive}{Size-Intensive\xspace}

\providecommand{\TrueAccuracy}{True Accuracy\xspace}
\providecommand{\Precision}{Precision\xspace}
\providecommand{\GroundTruth}{Ground Truth\xspace}
\providecommand{\Interpolation}{Interpolation\xspace}

\begin{document}

\title{%
SETOL: A Semi-Empirical Theory of (Deep) Learning
}

\author{%
Charles H. Martin\thanks{Calculation Consulting, 8 Locksley Ave, 6B, San Francisco, CA 94122, \texttt{charles@CalculationConsulting.com}.} 
\and
Christopher Hinrichs\thanks{Onyx Point Systems, chris@onyxpointsystems.com}
}

\date{}
\maketitle

\begin{abstract}
We present a \SemiEmpirical Theory of Learning (\SETOL)
that explains the remarkable performance of \StateOfTheArt (SOTA) Neural Networks (NNs).
We provide a formal explanation of the origin of the
fundamental quantities in the phenomenological theory of  \HeavyTailedSelfRegularization (\HTSR), the 
\HeavyTailed \PowerLaw \LayerQuality metrics,
\ALPHAHAT $(\alpha)$ and \ALPHAHAT $(\hat{\alpha})$.
In prior work, these metrics have been shown to predict trends in the test accuracies of pretrained SOTA NN models,
and, importantly,  without needing access to the testing or even training data.
Our \SETOL
uses techniques from \StatisticalMechanics (\STATMECH) as well as advanced methods from \RandomMatrixTheory (\RMT) and Quantum Chemistry. Our derivation suggests new mathematical preconditions for \emph{\Ideal} learning, including the new \TRACELOG metric (which is equivalent to applying a single step of the Wilson Exact Renormalization Group).
We test the assumptions and predictions of our \SETOL on a simple 3-layer
\MultiLayer \Perceptron (MLP), demonstrating excellent agreement with the key theoretical assumptions. 
For SOTA NN models, we show how to estimate the individual layer Qualities of a trained NN by simply computing the \EmpiricalSpectralDensity (ESD) of the layer weight matrices and
then plugging this ESD into our \SETOL formulae.
Notably, we examine the performance of the HTSR $\alpha$ and the \SETOL \TRACELOG \LayerQuality metrics, and find that they align
remarkably well, both on our MLP and SOTA NNs.

\end{abstract}

\newpage
\tableofcontents


\newpage

\section{Introduction}
\label{sxn:intro}
Deep Neural Networks (DNNs)—models in the field of Artificial Intelligence (AI)—have driven remarkable advances in multiple fields of science and engineering. AlphaFold has made significant progress in solving the protein folding problem.\cite{AlphaFold} Notably, the 2024 Nobel Prize in Physics was awarded to Hopfield and Hinton for developing early approaches to AI using techniques from \emph{Statistical Mechanics} (\STATMECH), and Jumper and Hassabis, along with Baker, received the 2024 Nobel Prize in Chemistry for their contributions to AlphaFold and computational protein design.\cite{Nobel2024Physics, Nobel2024Chemistry} Self-driving cars now roam the streets of major metropolitan cities like San Francisco. Large Language Models (LLMs) like ChatGPT have gained worldwide attention and initiated serious conversations about the possibility of creating an Artificial General Intelligence (AGI). Clearly, not a single area of science or engineering has ignored these remarkable advances in the field of AI and Neural Networks (NNs).

Despite this remarkable progress in a research field spanning over 50 years, developing, training, and maintaining such complex models require staggering capital resources, limiting their development to only the largest and best-funded organizations. While many such entities have open-sourced some of their largest models (such as Llama and Falcon), using these models requires assuming they have been trained optimally, without significant defects that could limit, skew, or even invalidate their use downstream. Moreover, testing such models can be very expensive and complex to interpret.

Because training and evaluating NNs is so hard, significant issues can manifest in many obvious and non-obvious ways. A primary research goal is to improve the efficiency and reduce the cost of training large NNs. A lesser-known but critical issue arises in many industrial settings, specifically “selecting the best models to test.” This arises in industries such as ad-click prediction, search relevance, quantitative trading, and more. Frequently, one has several seemingly equally good models to choose from, but testing the model can be very expensive, time-consuming, and even risky to the business. Recently, researchers and practitioners have started to fine-tune such large open-source models using techniques such as LoRA and QLoRA. Such methods allow one to adapt a large, open-source NN to a small dataset, and very cheaply. However, in fine-tuning, one could unwittingly overfit the model to the small dataset, degrading its performance for its intended use. Despite these and many other problems, theory remains well behind practice, and there is an increasingly pressing need to develop \emph{practical predictive theory} to both improve the training of these very large NN models and to design new methods to make their use more reliable.

Before discussing these methods, however, let us explain \emph{What is a \SemiEmpirical Theory}

\subsection{Statistical Mechanics (\STATMECH) vs. Statistical Learning Theory (\SLT)}

Historically, there have been two competing theoretical frameworks for understanding NNs:
\emph{\StatisticalMechanics} (\STATMECH)~\cite{Eng01, EB01_BOOK, Gardner_1988,SST90, SST92, LTS90, Solla2023}; and 
\emph{\StatisticalLearningTheory} (\SLT)~\cite{Vapnik98}. 
\begin{itemize}
\item
\textbf{\StatisticalMechanics (\STATMECH).}
This framework has been foundational to the early development of NN models, such as the Hopfield Associative Memory (HAM)~\cite{Hop82}, Boltzmann Machines~\cite{AHS85},~\cite{HinSej86_relearn}, etc.
\STATMECH~has also been used to build early theories of learning, such as the \StudentTeacher model for the \Perceptron \GeneralizationError~\cite{Eng01,EB01_BOOK}, the Gardner model~\cite{Gardner_1988}, and many others.
Notably, the HAM was based on an idea by Little, who observed that, in a simple model, long-term memories are stored in the eigenvectors of transfer matrix~\cite{Lit74}.
(This general idea, but in a broader sense, is central to our approach below.)
Moreover, \STATMECH~predicts that NNs exhibit phase behavior.
This has recently been rediscovered as the Double Descent phenomenon~\cite{BHMM19,loog2020}, but it was known in \STATMECH long before it's recent rediscovery~\cite{Opper01}.
However, unlike other applied physics theories (e.g., \SemiEmpirical methods in quantum chemistry), \STATMECH only offers qualitative analogies, failing to provide testable quantitative predictions
about large, modern NN models.\cite{roberts2022principles}
\item
\textbf{\StatisticalLearningTheory (\SLT).}
\SLT and related approaches (VC theory, PAC bounds theory, etc.) have been developed within the context of traditional computational learning problems~\cite{Vapnik98}, and 
they are based on analyzing the convergence of frequencies to probabilities (over problem classes, etc.).
It was recognized early on, however, that they could not be directly applied to NNs~\cite{VLL94}.
Moreover, \SLT cannot even reproduce quantitative properties of learning curves~\cite{WRB93,DKST96} (whereas \STATMECH~is very successful at this~\cite{SST92}).
\SLT also failed to predict the ``Double Descent'' phenomenon~\cite{BHMM19}.
More recently, it has been shown that in practical settings
\SLT can give vacuous~\cite{DR17_nonvacuous_TR} or even opposite results
to those actually observed~\cite{MM21a_simpsons_TR}.
\end{itemize}

\noindent
Technically, \SLT focuses on obtaining bounds on a model’s worst-case behavior, while \STATMECH seeks a probabilistic understanding of typical behaviors across different states or configurations.
Unfortunately, neither of these general theoretical frameworks has proven particularly useful to NN practitioners.
\SETOL combines insights from both.
Rather than being purely phenomenological like the \HTSR~approach, \SETOL is derived from first-principles, and in the form of a \SemiEmpirical~theory.
As such, \SETOL offers a practical, \SemiEmpirical framework that bridges rigorous theoretical modeling and empirical observations for modern NNs.

\subsection{Heavy-Tailed Self-Regularization (\HTSR)}

\HTSR theory is an approach that combines ideas from \STATMECH~with those of \emph{\HeavyTailed} \emph{\RandomMatrixTheory} (\RMT),
providing eigenvalue-based quality metrics that correlate with model quality (i.e., out-of-sample performance).
\HTSR theory posits that well-trained models have extracted subtle correlations from the training data, and that these correlations manifest themselves in the \SHAPE and \SCALE of the eigenvalues of the layer weight matrices $\mathbf{W}$. 
In particular, if one computes the empirical distribution of the eigenvalues, $\lambda_i$, of an individual  $N \times M$ weight matrix, $\mathbf{W}$, then this density, $\rho^{emp}(\lambda)$, which is an ESD, is \HeavyTailed (HT) and can be well-fit to a \emph{\PowerLaw} (PL), i.e., $\rho(\lambda)\sim\lambda^{-\alpha}$, with exponent $\alpha$.
\HTSR~theory provides a \emph{\Phenomenology} for qualitatively-distinct phases of learning~\cite{MM18_TR_JMLRversion}.
It can, however, also be used to define \emph{Layer-level Quality metrics} and \emph{Model-level Quality metrics}: e.g., the \ALPHA~$(\alpha)$ and \ALPHAHAT~$(\ALPHAHATEQN)$ PL metrics, described below.

Not needing any training data, \HTSR~theory has many practical uses.
It can be directly applicable to large, open-source models where the training and test data may not be available.
Model quality metrics can be used, e.g., to predict trends in the quality of SOTA models in computer vision (CV)~\cite{MM20a_trends_NatComm} and natural language processing (NLP)~\cite{YTHx22_TR,YTHx23_KDD}, both during and after training, and without needing access to the model test or training data.
Layer quality metrics can be used to diagnose potential internal problems in a given model, or (say) to accelerate training by providing optimal layer-wise learning rates \cite{NEURIPS2023_CHM} or pruning ratios \cite{alphapruning_NEURIPS2024}.
Most notably, the \HTSR theory provides \emph{Universal} \LayerQuality metrics encapsulated in what appears to be a critical exponent, $\alpha=2$, that is empirically associated with optimal or \IdealLearning. Moreover, as argued below, the
value $\alpha=2$ appears to define a phase boundary between a generalization and overfitting, analogous
to the phase boundaries seen in \STATMECH theories of NN learning. 

These results both motivate the search for a first principles understanding of the \HTSR theory, 
and suggest a path for developing a practical predictive theory of Deep Learning.
For this, however, we need to go beyond the \Phenomenology provided by \HTSR theory, to relate it to some sort of (at least semi-rigorous/semi-empirical) derivations based on the \STATMECH theory of learning, and drawing
upon previous success (in Quantum Chemistry) in developing a first principles \SemiEmpirical theory.

\subsection{What is a Semi-Empirical Theory?}

Historically, one of the most well known \emph{\SemiEmpirical} methods comes from \NuclearPhysics.
The \SemiEmpirical Mass Formula, dating back to 1935, is based on the heuristic Liquid Drop Model of the nucleus,
and it was used to predict experimentally observed binding energies of nucleons. 
This model describes nuclear fission, and it was central to its development of the atomic bomb:
\begin{quote}
  Prior to WWII, \NuclearPhysics was a phenomenological science, which relied upon experimental data and descriptive
  models~\cite{Negele05}.
\end{quote}
In the Post-war era, the epistemological nature of nuclear theory changed,
as it saw the development of \SemiEmpirical shell models of the nucleus.
These models were formulated with rigor (in the physics sense)
but also relied on heuristic assumptions and experimental data for accurate predictions.
They captured the structure of atomic nuclei
and could accurately describe various nuclear properties~\cite{Ivanenko1932, GoeppertMayer1949, Jensen1949}.
The shell models, analogous to the electronic shell structure of atoms,
represented a shift toward a more rigorous understanding of nuclear phenomena.

About this time, \RMT~itself was also introduced by \emph{\Wigner}~\cite{Wigner55}
to model the statistical patterns of the nuclear energy spectra of strongly interacting heavy nuclei.
These patterns were universal, independent of the specific nucleus,
suggesting that a probabilistic approach would be fruitful.
In the following decades, \RMT saw many advances, including the development of
the Marchenko-Pastur model~\cite{MarchenkoPastur1967},
and numerous other applications in physics~\cite{Guhr1998}.
By the 1990s, \RMT~was further expanded when \emph{Zee} introduced the \emph{Blue Function},
and reinterpreted the \emph{R-transform} as a self-energy within the
framework of many-body / quantum field theory (QFT)~\cite{Zee1996}.
Also, so-called \HCIZtext integrals, integrals over random matrices,
were being used both to model disordered electronic spectra~\cite{SchultenRMT},
and, later, the behavior of spin glass models~\cite{Bouchaud1998,Cherrier_2003}.

Returning to the 1950s, and prior to the development of highly accurate, modern,
computational \emph{ab initio} theories of \QuantumChemistry, 
Theoretical chemists introduced the \SemiEmpirical PPP method
for
conjugated polyenes~\cite{PariserParr53}.
The PPP model recasts the electronic structure problem as an \emph{\EffectiveHamiltonian}
for the $\pi$-electrons.
The PPP model resembles the later developed tight-binding model of condensed matter physics\cite{Hubbard1963}.
For many year,s this and related \SemiEmpirical methods
worked remarkably well, even better than the existing \emph{ab initio}
theories~\cite{Dewar1975,Ridley1973,Stewart1990,Warshel1976,THIEL2005559}.
Most importantly, these methods could be \emph{fit}
on a broad set of empirical molecular data, and then applied to molecules not in the original training set.

Around the same time, \emph{Löwdin} first formalized the concept of the \EffectiveHamiltonian,
which allowed the reduction of complex many-body problems to simplified
\emph{Effective Potentials} that still captured the essential physics.
Then, in the late 1960s, Brandow 
developed an \EffectiveHamiltonian theory of
nuclear structure, 
leveraging the \emph{Linked Cluster Theorem} (LCT) (see \cite{Hubbard1959}) and quantum mechanical many-body theory
to describe the highly correlated effective interactions in a reduced model space.
\footnote{Note also that the LCT shows that the log partition function $(i.e., \ln Z)$ can be expressed
a sum of connected diagrams, which is very similar to our result below, which expresses the
log partition function here as a sum of matrix cumulants from RMT.}

Like modern NNs, these \SemiEmpirical methods of \QuantumChemistry
worked well beyond their apparent range of validity,
generalizing very well to out-of-distribution (OOD) data. This led to the
search for a \SemiEmpirical \emph{Theory} to explain the
remarkable performance of these phenomenological methods.
Building on Brandow’s many-body formalism, Freed and collaborators
\cite{freed1977, Freed1983}
developed an \emph{ab initio} \EffectiveHamiltonian
Theory of \SemiEmpirical methods to explain the remarkable success of the \SemiEmpirical methods.
Specifically, the values of the PPP empirical parameters could be directly computed by way of
effective interactions, including both renormalized self-energies and higher-order terms.
 Somewhat later, in the 1990s,
 Martin et. al.~\cite{MartinFreed1996, Martin1996, Martin1996_CPL, Martin1998}
 extended and applied this \EffectiveHamiltonian theory
 and demonstrated the \Universality of the \SemiEmpirical PPP parameters numerically.
 Indeed, it is this \Universality that enabled the `for a time' inexplicable
 OOD performance of these methods.
 Crucially, this decades-long line of work established a comprehensive
 analytic and numerical \emph{Theory} of \SemiEmpirical methods.
 That is, a framework that confirmed the empirically observed \Universality
 provided theoretical justification for this,
 and enabled systematic improvements of the methods using numerical techniques.

 Finally, it is important to mention the \EffectiveHamiltonian approach provided by the Wilson \emph{\RenormalizationGroup}
 (RG)~\cite{NobelPrizeRG,PhysRevLett.69.800}.
 The RG approach provides a powerful framework for studying strongly correlated systems across different scales,
 enabling the construction of an \EffectiveHamiltonian by \emph{integrating out} weakly-correlated degrees of freedom in a \ScaleInvariant way.
 It is particularly suited for critical points and phase boundaries --
 such as the phase boundary between generalization and memorization in spin glass models of neural networks --
 and, importantly, predicts the existence of Universal Power Law (PL) exponents .
For \SETOL, we take what amounts to a single step of the \emph{\ExactRenormalizationGroup} (\ERG), leading to an new empirical metric
that defines the 'Ideal' NN layer.

 \paragraph{Relevance to Deep Learning}
 In this sense, \SemiEmpirical theories of \NuclearPhysics and \QuantumChemistry,
 (as well as the \RenormalizationGroup approach), seem particularly appropriate
 for Deep Learning.
  DNN models are complex black boxes that defy statistical descriptions in that
  they are commonly pre-trained on a large set of data; and than applied to new data sets in new domains via transfer learning.
  Most recently, the inexplicable success of transfer-learning is seen
in the GPT (Generative Pre-Trained Transformer) models~\cite{Radford2018},
and motivated early work by Jumper et. al. on protein folding~\cite{JKS16_TR}.

In contrast, these \SemiEmpirical approaches differ from more
recently developed theoretical approaches to deep learning, which are typically based on SLT, rather than \STATMECH~\cite{Roberts2021}.
In particular, there have recently appeared several theories of deep learning, formulated using ideas from \RMT.
However, regarding realistic models, it has been explicitly stated that
``These networks are however too complex in general for developing a rigorous theoretical analysis on the spectral behavior''~\cite{LBNx17_TR}.
Even in recent work applying \RMT~to NNs, it has been noted
``\emph{that we make no claim about trained weights, only random weights}''~\cite{Yang2021}.
The weight matrices of a trained NN, however, are clearly \emph{not} simply random matrices---since they encode the specific correlations from the training data.

\subsection{A Semi-Empirical Theory of Learning (\SETOL)}

We propose \SETOL, a \SemiEmpirical Theory for Deep Learning Neural Networks (NNs),
as both a theoretical foundation for \HTSR \Phenomenology
and a novel framework for predicting the properties of complex NN models.
This unified framework offers a deeper understanding of DNN generalization
through a \SemiEmpirical approach inspired by many-body physics,
combined with a classic \STATMECH model for NN generalization.
Specifically, \SETOL combines theoretical and empirical insights to evaluate \ModelQuality,
showing that the weightwatcher layer \HTSR PL metrics (\ALPHA and \ALPHAHAT)
can be derived using a phenomenological \EffectiveHamiltonian approach.
This approach expresses the \HTSR \LayerQuality in terms of the RMT matrix cumulants
of the layer weight matrix $\WMAT$,
and is governed by a \ScaleInvariant transformation equivalent
to a single step of an Exact \RenormalizationGroup (\ERG) transformation.
Here, we derive this from first principles, requiring no previous knowledge of statistical physics.

The \SETOL approach unifies the \HTSR principles with
a broader theoretical framework for layer analysis.
The \HTSR theory identifies \Universality (e.g., $\alpha=2$) as a hallmark of the best-trained DNN layers,
and, here, our \SETOL introduces the closely related \emph{Exact Renormalization Group Condition}, a \ScaleInvariant or
\VolumePreserving transformation that reflects an underlying \emph{Conservation Principle}.
Together, these principles form the theoretical foundation for deriving \HTSR \LayerQuality metrics from first principles.
By leveraging techniques from \STATMECH and modern \RMT, \SETOL offers a rigorous framework
to connect empirical observations with theoretical predictions, advancing our understanding of generalization
in neural networks.

\begin{itemize}
\item
  \textbf{Derivation of the HTSR Layer Quality metrics $\ALPHA$ and  $\ALPHAHAT$}.
  The \SETOL approach takes as input the
  \EmpiricalSpectralDensity (ESD) of the layers
  of trained NN, and  derives an expression for the approximate \emph{\AverageGeneralizationAccuracy}
  of a multi-layer NN. We call this approximation the \emph{\ModelQuality}, denoted $\Q^{NN}$.
  This \ModelQuality is expressed as a product of individual \LayerQuality terms, $\Q^{NN}_{L}$,
  which themselves can then 
  be directly related to the \HTSR Power Law (PL) empirical $\ALPHA$ ($\alpha$)
  and $\ALPHAHAT$  ($\ALPHAHATEQN=\ALPHAHATLONG$) metrics.

  In particular, the \LayerQualitySquared, $\QT\approx[\Q^{NN}_{L}]^{2}$, is
  expressed as the logarithm of an \HCIZtext integral, which is the \ThermalAverage of an \EffectivePotential
  for a matrix-generalized form of the Linear Student-Teacher model of classical \STATMECH. This \HCIZtext
  integral evaluates into the sum of integrated \RTransforms
  from \RMT, or, equivalently, as a sum of integrated matrix cumulants.
  From this, the \HTSR $\ALPHAHAT$ metric can be derived in the special case of \IdealLearning.
  \footnote{The \SETOL approach to the \HTSR theory resembles
  in spirit the derivation of the \SemiEmpirical PPP models using
  the \EffectiveHamiltonian theory, where each phenomenological parameter is associated with a renormalized
  effective interaction, expressed as a sum of linked diagrams or clusters.\cite{Martin1996, Martin1998}}

   \item 
     \textbf{Discovery of a Mathematical Condition for Ideal Learning.}
     By \IdealLearning, we mean that the specific NN layer has optimally converged, capturing as
     much of the information as possible in the training data without overfitting to any part of it.
     In defining this and deriving our results, we have discovered (and are proposing) a new condition
     for \IdealLearning, which is associated with the \Universality of the \HTSR theory:
   \begin{itemize}
      \item 
        \textbf{\HTSR Condition for Ideal Learning.}
        This \HTSR theory states that a NN layer is \Ideal  when the ESD can be well fit to a
        Power Law (PL) distribution, with PL exponent $\alpha = 2$. Importantly, 
        this appears to be a Universal property of all well-trained NNs, independent of the training data,
        model architecture, and training procedure.
      \item 
        \textbf{\SETOL \TRACELOG Condition for Ideal Learning.}
        The \SETOL condition for \IdealLearning states that the 
        dominant eigencomponents associated with the ESD 
        of layer form a reduced-rank \emph{Effective Correlation Space} (\ECS) that satisfies
        a new kind of Conservation Principle,  \emph{\ScaleInvariant} \emph{\VolumePreservingTransformation} where the largest eigenvalues $\LambdaECS_{i}$ of the~\ECS satisfy
        the condition  $\ln \prod \LambdaECS_{i} = \sum \ln \LambdaECS_{i} = 0$.  
        This is called the \emph{\TRACELOG Condition}.  The \TRACELOG Condition is equivalent to the taking a single step of the Wilson Exact Renormalization (ERG).
   \end{itemize}

   The \HTSR Condition has been proposed and analyzed previously~\cite{MM18_TR_JMLRversion,MM20a_trends_NatComm,YTHx23_KDD}; but
   the \TRACELOG Condition is new, based on our \SETOL theory.
   When these two conditions align, we propose the NN layer is in the \Ideal state.

   \item 
   \textbf{Experimental Validation.} 
   We present detailed experimental results on a simple model, along with observations on large-scale pretrained NNs, to demonstrate that the \HTSR conditions for ideal learning $(\alpha = 2)$ are experimentally aligned with the independent \SETOL condition for ideal learning
   $(\Det{\XECS}=1)$. 
   See Section.~\ref{sxn:empirical-trace_log}.
   Our primary objective here is not to demonstrate performance improvements on SOTA NNs---this has been previously established \cite{NEURIPS2023_CHM}. 
   Instead, our aim is to \textbf{validate the theoretical assumptions} of \SETOL, test the \textbf{predictions of the \SETOL framework}, and examine the \textbf{new, independent learning conditions} we discovered---on a model that is sufficiently simple that we can evaluate and stress test the theory.

   \item 
   \textbf{Observations on Overfit Layers ($\alpha < 2$).} 
   Being a \SemiEmpirical theory, \SETOL can also identify violations of it's assumptions.
   For example, when empirical results show $\alpha < 2$ for a single layer, the layer's ESD falls into the \HTSR \VeryHeavyTailed (VHT) Universality class.
   (See Section~\ref{sxn:hysteresis_effect}.)
   When this happens, the layer may be slightly overfit to the training data, resulting in \textbf{suboptimal performance} and potentially even exhibiting \textbf{hysteresis-like effects} (memory effects)---that we observe empirically.
   These effects indicate that overfit layers may retain memory-like behavior, affecting learning dynamics and generalization.
\end{itemize}

\newpage
\section{Heavy-Tailed Self-Regularization (\HTSR)}
\label{sxn:htsr}

In this section, we provide an overview of the \HTSR \Phenomenology.
\footnote{We may also refer to the \HTSR \Phenomenology as the \HTSR Theory; we use the term \Phenomenology to emphasize its empirical nature, and to distinguish it from the analytical methods used in the \SETOL approach.}
\HTSR has been presented in detail previously~\cite{MM19_HTSR_ICML,MM20_SDM,MM18_TR_JMLRversion}.%
Here, we provide a self-contained summary, with an emphasis on certain technical issues that will be important for our
\SETOL. We highlight its practical application for interpreting observed behaviors in trained weight matrices, and 
we distinguish the \HTSR \Phenomenology from the analytical methods used in the \SETOL approach.
In Section~\ref{sxn:htsr_setup}, we summarize the basic \HTSR setup and results;
in Section~\ref{sxn:guass_ht_univ}, we summarize Gaussian (for \RMT) and Heavy-Tailed (for \HTRMT) Universality; and
in Section~\ref{sxn:htsr-metics}, we describe \SHAPE metrics and \SCALE metrics that arise from \HTSR.
(Here, we focus on basic methods for identifying HT correlations in the ESDs of pre-trained weight matrices; 
in Sections~\ref{sxn:empirical-test_acc},~\ref{sxn:empirical-correlation_trap} and~\ref{sxn:hysteresis_effect}, we show detailed experiments using theoretical constructs from the \HTSR \Phenomenology.)


\subsection{The \HTSR Setup}
\label{sxn:htsr_setup}

We can write the \EnergyLandscape function 
(or NN output function) for a 
NN with $L$ layers as
\begin{equation}
\label{eqn:dnn_energy}
\NNOUT:=h_{L}(\mathbf{W}_{L}\times h_{L-1}(\mathbf{W}_{L-1}\times(\cdots)+\mathbf{b}_{L-1})+\mathbf{b}_{L}) 
\end{equation}
with activation functions $h_{l}(\cdot)$, and with weight matrices and biases $\mathbf{W}_{l}$ and $\mathbf{b}_{l}$.%
\footnote{The Energy Landscape function $\NNOUT$ acts on a data instance and generates a list of energies, or un-normalized probabilities;
~\cite{MM18_TR_JMLRversion}.
This notation was chosen to make an analogy with Random Energy Models (REM)
from spin-glass and protein folding theories~\cite{DerridaREM1981, BryngelsonWolynesPNAS1987}.
}
For simplicity of exposition here (\HTSR can be applied much more broadly), we ignore the structural details of the layers (dense or not, convolutions or not, residual/skip connections, etc.).
We also ignore the biases $b_{l}$ (because they can be subsumed into the weight matrices),
and we treat each layer as though it contains a single weight matrix $\mathbf{W}_{L}$.
We imagine training (or fine-tuning) this model on labeled data $\{\mathbf{x}_{\mu},y_{\mu}\}\in\mathcal{D}$, where $ \mathbf{x}_\mu $ is the $\mu$-th input vector and $y_\mu$ is its corresponding label (e.g., for binary classification, $y_{\mu}\in\{-1,1\}$).
We expect to use backprop via some variant of stochastic gradient descent (SGD) to minimize some loss functional, $\mathcal{L}$ (such as $\ell_2$, cross-entropy, etc.):  
\begin{equation}
\underset{\mathbf{W}_{l},\mathbf{b}_{l}}{\argmin}\;\sum_{\mu}\mathcal{L}[\NNOUT(\mathbf{x}_{\mu}),y_{\mu}]+\Omega ,
\label{eqn:dnn_opt}
\end{equation}
where $\Omega$ denotes some explicit regularizer (such as an $\ell_1$ or $\ell_2$ constraint on layer weight matrices) or some implicit regularization procedure (such as clipping the weight matrices or applying dropout).

Given a real-valued $N\times M$ layer weight matrix $\mathbf{W}$ (dropping the subscript), let $\mathbf{X}$ be the $M \times M$ layer \emph{\CorrelationMatrix}:
\begin{equation}
\mathbf{X}:=\dfrac{1}{N}\mathbf{W}^{\top}\mathbf{W} .
\label{eqn:X}
\end{equation}

\noindent
The \EmpiricalSpectralDensity (ESD) of $\mathbf{W}$, denoted $\rho^{emp}(\lambda)$, is formed from the M eigenvalues $\lambda_{j}$ of $\mathbf{X}$:
\begin{equation}
\rho_{emp}(\lambda):=\sum_{j=1}^{M}\delta(\lambda-\lambda_{j}) .
\label{eqn:rho}
\end{equation}
Given a model, we can compute the ESDs of all of its layers, as well as other metrics below,
with the open-source \WW tool~\cite{WW}.%
\footnote{For practical purposes, the \WW tool computes $\rho^{emp}(\lambda)$ by forming the Singular Value Decomposition (SVD) of the layer weight matrices $\mathbf{W}$, computing the eigenvalues $\lambda=\sigma^{2}$ from the singular values $\sigma$, and (when useful) smoothing them with a Kernel Density Estimator (KDE). For some calculations, such as the \TRACELOG condition, we must also select the appropriate normalization of $\mathbf{W}$.}

\begin{figure}[t] 
    \centering
    \subfigure[Log-Log ESD]{
      \includegraphics[width=7cm]{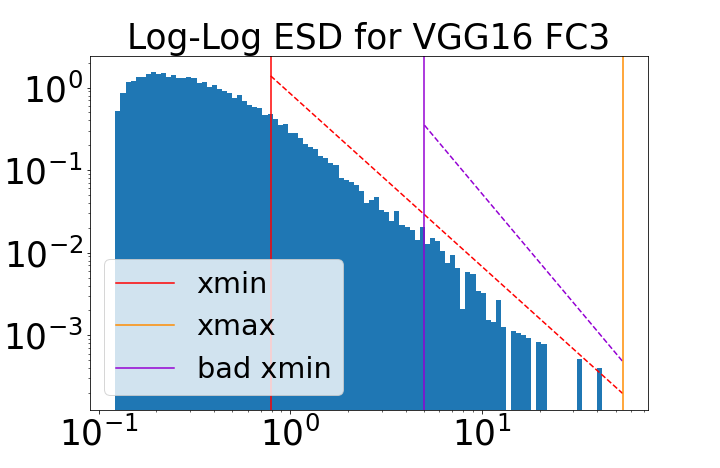}
      \label{fig:log-log-esd}
    }
    \subfigure[Lin-Lin ESD]{
      \includegraphics[width=7cm]{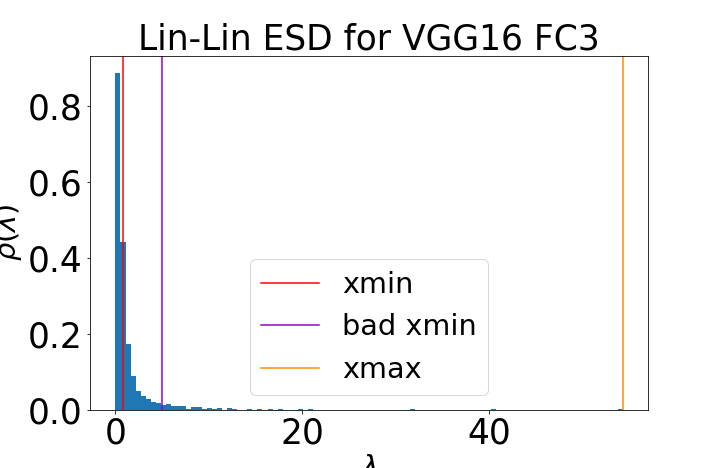}
      \label{fig:lin-lin-esd}
    }
    \subfigure[Log-Lin ESD]{
      \includegraphics[width=7cm]{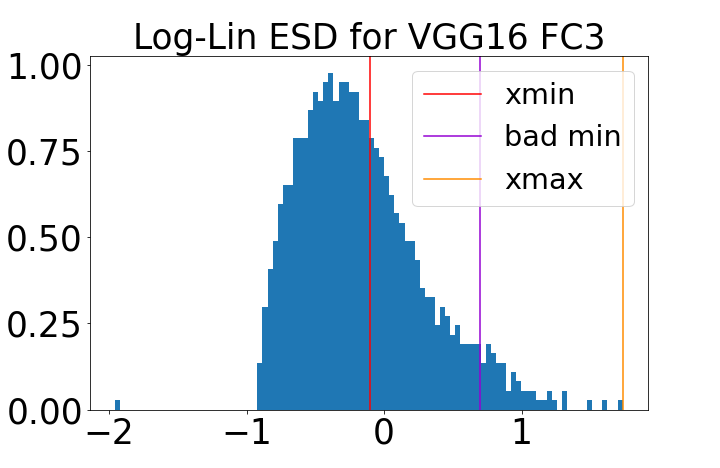}
      \label{fig:log-lin-esd}
    }
    \subfigure[$\LambdaMin$ vs KS distance]{
      \includegraphics[width=7cm]{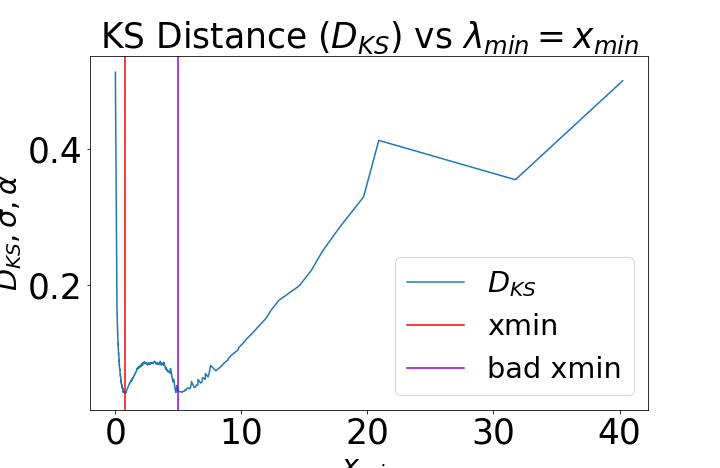}
      \label{fig:xmin-DKS-esd}
    }                                     
    \caption{\textbf{Fitting ESDs within \HTSR.} 
         Depiction of the ESD and results of PL fits for a typical well-trained layer of a modern NN (FC3 of VGG19), including both the actual and good PL fit (red) and a hypothetical bad PL fit (purple). 
         The same ESD is plotted on a Log-Log (a), Lin-Lin (b) and Log-Lin (c) scales.
         (d) depicts how the start of the PL tail, $\LambdaMin$, varies with the quality of the PL fit (the $D_{KS}$ distance). 
         All plots are generated using the open-source \WW tool. 
         See the main text for details.
    }
  \label{fig:log-esds}
\end{figure}

Based on empirical results based on thousands of pretrained models and tens of thousands of layers~\cite{MM18_TR_JMLRversion,MM20a_trends_NatComm,MM21a_simpsons_TR,YTHx23_KDD}, 
it is generally observed that the 
best performing NNs have ESDs that are HT, and the \emph{tails} of these ESDs, $\rho_{tail}(\lambda)$, can be well fit to a PL, beyond some cutoff $\lambda\ge\LambdaMin$.%
\footnote{Doing a large meta-analysis like this is tricky; but see~\cite{MM18_TR_JMLRversion,MM20a_trends_NatComm,MM21a_simpsons_TR,YTHx23_KDD}.  The \WW~tool provides a systematic, reproducible way to compute a PL fit (using an MLE method of Clauset et al.~\cite{CSN09_powerlaw}), as well other model metrics, including the \SPECTRALNORM, \RANDDIST, and \ALPHAHAT metrics~\cite{MM20a_trends_NatComm}.  Also, the ESD $\rho(\lambda)_{tail}$ is sometimes better fit by a Truncated \PowerLaw (TPL), due to finite-size effects.
(Again, this is important in practice, but we ignore this complexity in this initial discussion of \SETOL. ) }
For a PL fit,
\begin{equation}
\rho_{tail}(\lambda):=\rho^{emp}(\lambda\ge\LambdaMin)\sim\lambda^{-\alpha} ,
\label{eqn:rho_tail}
\end{equation}
where $\LambdaMin$ is where the tail of the ESD starts (i.e., it is not the minimum eigenvalue, but the minimum eigenvalue in the tail of the ESD). 
See Figure~\ref{fig:log-esds}.
As such, the tail of the ESD ``starts'' at some value $\LambdaMin$, called $xmin$ here, and it continues until the 
maximum eigenvalue $\LambdaMax$, called $xmax$ here
(labeled $xmax$ in the figure, shown by the orange line).
We estimate $xmin$ and $\alpha$ jointly, using the method of~\cite{CSN09_powerlaw}, 
as implemented in the \texttt{powerlaw} python package~\cite{ABP14}, 
which is also integrated into the open-source \WW tool~\cite{MM20a_trends_NatComm, WW}.%
\footnote{The authors of~\cite{Thamm2022} failed to find evidence of a PL-like distribution in NN weight matrices, which is 
likely to be the case when $\alpha$ and $xmin$ are not estimated \emph{jointly}, as can be seen in Figure~\ref{fig:xmin-DKS-esd}.  }


\paragraph{Fitting ESDs.}
Choosing the start of the tail, $\LambdaMin$, is important for \HTSR (and it will be very important for \SETOL, as we will describe below).
See Figure~\ref{fig:log-esds} for a depiction of how this was done within \HTSR theory.
Figures~\ref{fig:log-log-esd}-\ref{fig:log-lin-esd} show the results of both a ``good fit'' and a ``bad fit'' on the same ESD,
while Figure~\ref{fig:xmin-DKS-esd} indicates the quality of fit.
For the good fit, the start of the tail is the optimal value $\LambdaMin=xmin$ (in red); and for the bad fit, it is a suboptimal $bad\;xmin$ (purple).
Figure~\ref{fig:xmin-DKS-esd} depicts how the best fit is determined; it plots $xmin=\LambdaMin$ versus the $D_{KS}$ 
value, which is the Kolmogorov-Smirnov (KS) distance between the PL fit and the empirical data~\cite{CSN09_powerlaw}.
Notice that there are two nearly degenerate minima on Figure~\ref{fig:xmin-DKS-esd}, corresponding to the good fit and the bad fit. 
It is not uncommon to face such practical challenges, as real-world ESDs are often slightly deformed from a perfect PL density, e.g.,
they may have two or more near-degenerate solutions on the KS plots (d).
(They may also have anomalously large eigenvalues; this is discussed in more detail in Section~\ref{sxn:HT_ESDs}.)

When one finds a good PL fit for the ESD of a layer $\mathbf{W}$,
it provides information about the \SHAPE~and \SCALE~of the ESD of that layer.
In particular: 
the \SPECTRALNORM, $\lambda_{max}$, being a matrix norm, is a measure of the size \SCALE~of the ESD~\cite{MM21a_simpsons_TR}; 
the fitted PL exponent \ALPHA, $\alpha$, being the slope of the tail of the ESD on a Log-Log plot, describes the \SHAPE~of the ESD; and
the \WW \ALPHAHAT~metric combines \SHAPE~and \SCALE~information.
Also, as opposed to other applications of PL fits~\cite{CSN09_powerlaw,BouchaudPotters03}, in our analysis,
the start of the tail, $\LambdaMin=\LambdaPLmin$, plays a particularly important role because it
identifies the subspace of the strongest generalizing eigencomponents (i.e., $\XECS$, below) in each layer.

\begin{table}[t] 
\begin{center}
  \begin{tabular}{| l | c | c | r | }
    \hline
    HT/\RMT Universality class & $\mu$ range   & $\alpha$ range    & Best Fit  \\ \hline \hline
    RandomLike            & NA            & NA                & MP        \\ \hline
    Bulk+Spikes           & NA            & NA                & MP+Spikes \\ \hline
    Weakly Heavy Tailed   & $\mu > 4$     & $\alpha>6$        & PL        \\ \hline
    Heavy (Fat) Tailed    & $\mu\in(2,4)$ & $\alpha\in(2,6)$  & PL        \\ \hline
    Very Heavy Tailed     & $\mu\in(0,2)$ & $\alpha\in(1,2) $ & (T)PL     \\ \hline
    Rank Collapse       & NA            & NA                & NA        \\ \hline
  \end{tabular}
\end{center}
\caption{\HTSR~Heavy-Tailed Universality classes of \RMT. See Table~1 of \cite{MM18_TR_JMLRversion} for more details.    }
\label{tab:Uclass}
\end{table}

\subsection{Gaussian and Heavy-Tailed Universality}
\label{sxn:guass_ht_univ}

The \HTSR \Phenomenology uses \RMT to classify of the ESD of a layer $\mathbf{W}$ into one of 5+1 Phases of Training, 
each roughly corresponding to a (Gaussian or HT) Universality class (of \RMT or \HTRMT).
This is summarized in Table~\ref{tab:Uclass}. 
A Universality class is a set of matrices having a common limiting spectral distribution, regardless of the other properties of their entries. 
Of those, the most familiar is the Gaussian class, characterized by the Marchenko Pastur (MP) results from traditional \RMT~\cite{EW13,potters_bouchaud_2020}. 
The Gaussian Universality class, however, is particularly poorly suited for analyzing realistic NNs---precisely
because the ESDs of SOTA NNs are well-fit by HT distributions.
This should not be surprising: weight matrices of realistic NNs do \emph{not} have independent (i.i.d.)
entries---their entries are strongly-correlated precisely because they provide a view into the correlated training data.

To model strongly-correlated NN layer matrices, the \HTSR \Phenomenology characterizes NN layer weight matrices
in terms of their ESDs (when a good PL fit can be found) by postulating that the (tail of the) eigenvalue spectrum $\rho(\lambda)$ 
determines how each layer contributes to the overall generalization.  
To do this, the \HTSR approach models the strong-correlated layer weight matrices \emph{as if} they are actually i.i.d. HT random
(i.e., entry-wise uncorrelated) matrices. 
By doing this, one can  associate each $\rho(\lambda)$ with 
the corresponding HT Universality class, according to the PL exponent $\alpha$ fitted from the ESD. 
As we will see in 
Section~\ref{sxn:HT_ESDs}, it can be critical to distinguish  when the ESD is HT\emph{Correlation-wise} vs HT \emph{Element-wise}.

To understand Table~\ref{tab:Uclass} better, we first review basic results.  

\subsubsection{\RandomMatrixTheory (\RMT):  Marchenko-Pastur (MP) Theory and Tracy-Widom (TW) Fluctuations}

\begin{figure}[t] 
    \centering  
    \subfigure[MP, varying $Q=\tfrac{N}{M}$]{ 
      \includegraphics[width=7.0cm]{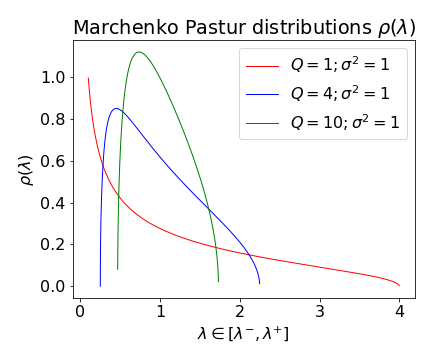}
      \label{fig:MP-esds-a}
    }                               
    \subfigure[Tracy Widom fluctuations]{
      \includegraphics[width=7.0cm]{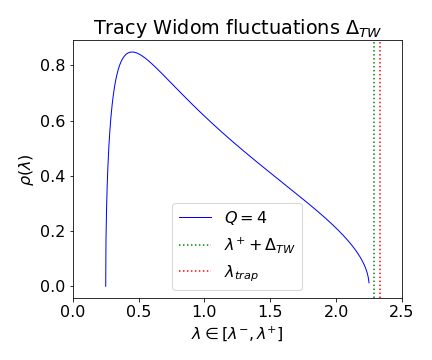}
      \label{fig:MP-esds-b}
    }                    
    \caption{MP distributions for different aspect ratios $Q$ and variance scales $\sigma^2$, and an example of the finite-sized TW fluctuation $\Delta_{TW}$. }
   \label{fig:MP-esds}
\end{figure}

The Marchenko-Pastur (MP) distribution predicts the (limiting) \SHAPE of an ESD, $\rho_{MP}(\lambda)$, when the layer weight matrix has elements that are i.i.d. random from the Gaussian \Universality class.
In particular, the ESD will be MP when the matrix elements are drawn from a Normal distribution $W_{i,j}\in  N(0,\sigma^{2})$, e.g., as is typical at initialization, before NN training begins.
Figure~\ref{fig:MP-esds} (from Figure~4 of \cite{MM18_TR_JMLRversion}) displays the MP distribution for different aspect ratios $Q=\tfrac{N}{M}$ and variances $\sigma^{2}$.  
Notice that the \SHAPE is characterized by a well-defined, compact envelope with sharp edges.

The MP distribution also predicts the \SCALE of an ESD, again when the layer weight matrix has elements that are i.i.d. random from the Gaussian \Universality class.
In particular, an MP distribution, $\rho_{MP}(\lambda)$, has very crisp, well-defined lower and upper bounds $\lambda^{-},\lambda^{+}$~\cite{MM18_TR_JMLRversion}, and (importantly) the upper bound $\lambda^{+}$ exhibits finite-size Tracy-Widom (TW) fluctuations, $\Delta_{TW}(\lambda)$, which are on the order of $\mathcal{O}(M^{-2/3})$. 
Thus, any layer eigenvalue with \SCALE greater than this, i.e., $\lambda>[\lambda^{+}+\Delta_{TW}(\lambda)]$, is an ``outlier'' or a ``spike.''

According to the \HTSR \Phenomenology, these spikes carry significant generalizing information. 
(This is well-known for Bulk-Plus-Spike models~\cite{MM18_TR_JMLRversion}, but the \HTSR \Phenomenology generalizes this concept.)
Relatedly, for layer matrices $\mathbf{W}$ with aspect ratio $Q>1$ (i.e., rectangular matrices, where $N>M$), MP \RMT predicts there should be no zero eigenvalues, i.e., $\lambda_i>0$, for all $i$. 
Generally speaking, for well trained NNs, for layers with $Q>1$, all eigenvalues are strictly larger than zero, i.e., 
well-trained layer weight matrices, with $Q>1$, should have full rank and exhibit no ``rank collapse.'' 
\HTSR places random Gaussian and ``Bulk-Plus-Spike'' matrices into the first two rows of Table~\ref{tab:Uclass}.
The essential feature of Gaussian random matrices is that their entries have no correlations. 
When some correlations are injected, a few large spike eigenvalues form, without otherwise disturbing the shape of the ESD. 
To really understand how individual NN layers converge, we need to understand when and why their ESDs become HT.

\subsubsection{\HeavyTailed \RandomMatrixTheory (\HTRMT) and \PowerLaw (PL) fits}
\label{sxn:htsr_pl_fits}

For very well-trained NN layers, ESDs are \emph{not} MP at all.
Frequently, if not always, their ESDs are HT---and they are HT \emph{because} they are strongly-correlated matrices.  
Importantly, they are \emph{not} HT element-wise.
Instead, their entries have a scale, and they have ESDs that are HT due to correlations learned during training. 
Existing theoretical approaches, including \SLT and even \STATMECH, cannot readily model such strongly-correlated systems.%
\footnote{For example, such theoretical approaches typically deal better with \emph{\Scale} information (such as $\lambda_{max}$) than with 
\emph{\Shape} information (such as $\alpha$), e.g., by characterizing an ``eigen-gap'' separating large eigenvalues from 
``noise''~\cite{bach2006_JMLR} according to a noise plus low-rank perturbation model~\cite{BFR11}.}

Such strongly-correlated systems, however, do frequently arise in other, related scientific domains, including
in the \STATMECH of self-organizing systems~\cite{bak97a,SornetteBook}, 
in electronic structure theory~\cite{Martin1996,Martin1998,Martin1996_CPL}, and
in quantitative finance~\cite{bouchaud1999,bouchaud2005,potters_bouchaud_2020}. 
In these (and other) domains, correlated systems frequently exhibit characteristic PL signatures; and it is common practice to \emph{model} correlated systems as random (uncorrelated) systems by using HT statistics (e.g., Levy distributions or PL random matrices), fully understanding that such systems are by no means actually i.i.d. random.
The \HTSR \Phenomenology builds upon this longstanding practice by 
delimiting families of HT NN weight matrices based on the corresponding Universality classes of Pareto matrices. 

We explain briefly how to interpret Table~\ref{tab:Uclass} with respect to \HTRMT. The 5+1 Phases of Training can be 
identified by fitting ESDs to MP or PL distributions, whichever gives the best fit, as shown in the last column.
In case the PL distribution is a better fit, \HTSR \Phenomenology treats the layer weight matrix as 
equivalent to an i.i.d. random matrix $\mathbf{W}(\mu)$, whose elements have been drawn from a Pareto distribution 
with exponent $\mu$. 

\paragraph{Heavy-Tailed Universality Classes of Random Pareto Matrices}
For such an element-wise HT matrix, the theoretical \emph{limiting} ESD of a Pareto matrix is also PL,
which allows us to related the fitted PL $\alpha$ with exponent $\alpha=a\mu+b$, to the Pareto exponent $\mu$.
Ideally, for an infinite width matrix ,  $a=\tfrac{1}{2}$ and $b=1$, but due to finite-size effects, however,
we have found we must take $a\ge \tfrac{1}{2}$ and $b\ge 1$, giving
\begin{align}
W_{i,j}(\mu)\sim\dfrac{C}{x^{\mu +1}},\;\;\;\rho(\lambda)\sim\lambda^{-(a\mu+b)}.
\end{align}
According to the above relation, 
we can use either the fitted PL exponent $\alpha$, or the Pareto exponent $\mu$,
to index the HT Universality classes,
Note, however, that the finite-size effects strongly depend on  the and aspect ratio  $Q=N/M$,
at least when applied to i.i.d random Pareto matrices, and 
the (Clauset MLE) PL fit may overestimate the $\alpha$ of the ESD.
Table~\ref{tab:Uclass} delimits the HT matrices 
into sub-categories (as shown in the bottom four rows)  based on the behaviors of $\alpha$ as a function of $\mu$.

\begin{figure}[h]
    \centering  
    \subfigure[Heavy Tailed ESDs]{ 
      \includegraphics[width=7cm]{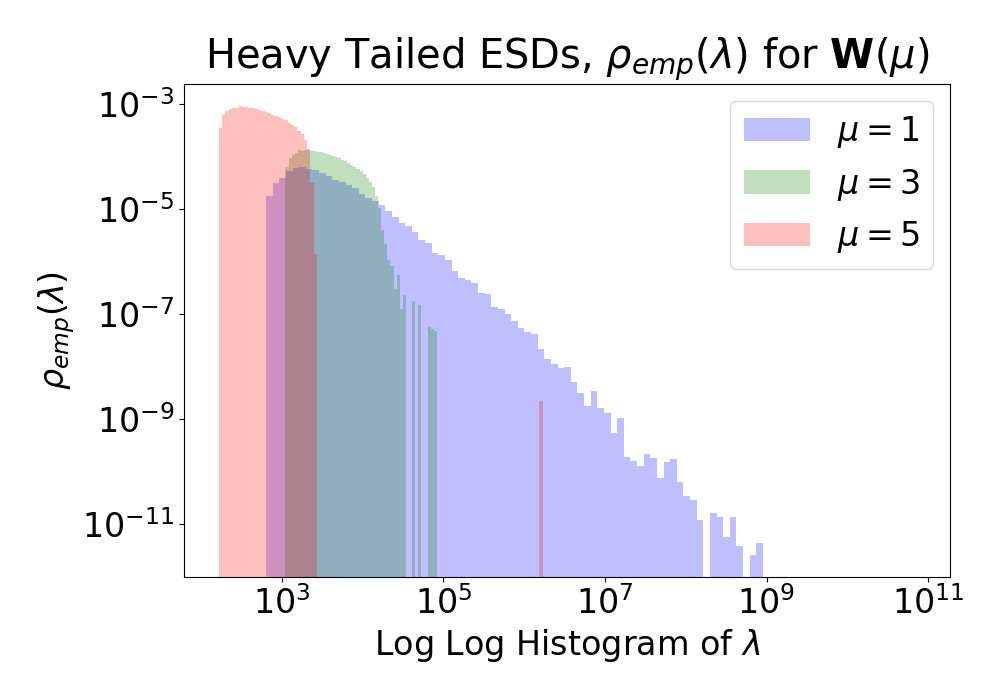}
      \label{fig:HT-esds-a}
    }                               
    \subfigure[PL $\alpha$ vs HT $\mu$ exponent]{
      \includegraphics[width=7cm]{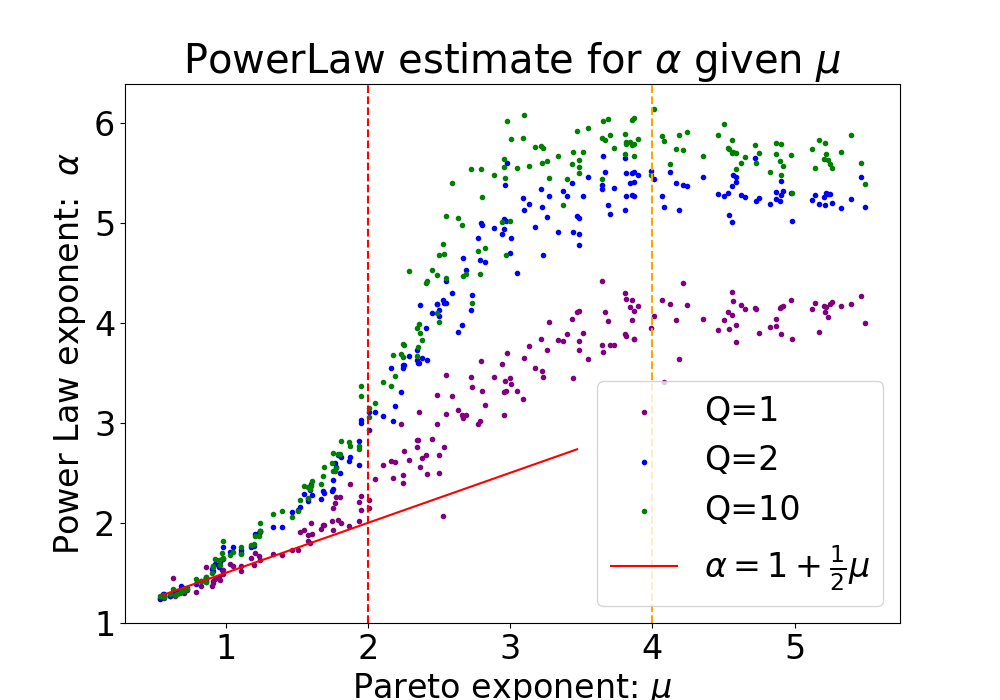}
      \label{fig:HT-esds-b}
    }                                                                                                                            
    \caption{Comparison of ESDs and Power Law (PL) exponents $\alpha$ from Heavy-Tailed (Pareto) 
weight matrices $\mathbf{W}(\mu)$. 
Subfigure (a) depicts 3 \Typical ESDs with Pareto exponent $\mu=1,3,5$, each decreasing in \SHAPE~and \SCALE.
Subfigure (b) shows how the exponent $\alpha$ of the PL fit varies with $\mu$, with significant finite-size
effects emerging for $\mu>2$ and $\alpha>2$.
}
   \label{fig:HT-esds}
\end{figure}

Figure~\ref{fig:HT-esds} illustrates how the fitted PL exponent $\alpha$ corresponds to the actual Pareto exponent $\mu$ 
for different aspect ratios $Q=M/M$.
Figure~\ref{fig:HT-esds-a} displays the ESDs of three different i.i.d. $1000\times1000$ HT random matrices, with $\mu=1,3,5$, on a Log-Log scale.  
Notice that smaller $\mu$, and therefore smaller $\alpha$, corresponds to heavier (i.e., larger) tails.
Figure~\ref{fig:HT-esds-b} shows how the empirically fit PL exponent $\alpha$ can vary with the theoretical $\mu$ for an associated $\mathbf{W}(\mu)$.
For $\mu<2$ and $Q=1$, the fitted $\alpha$ follows the linear relation $\alpha=\frac{1}{2}\mu+1$,
albeit with some error.
In contrast, for the more relevant $\mu \in (2,4)$ regime, the relation now depends far more 
strongly on the aspect ratio $Q$, and $\alpha\in[2,6]$.
For $\mu>4$, the fitted $\alpha$ saturates for each specific value of $Q$.

We emphasize that we only model the ESDs of the NN layer weight matrices using the
same Universality class to that associated with the ESD of a random, i.i.d, HT Pareto matrix.
In fact, the elements $W_{i,j}$ do not at all appear as if they have been
drawn from an HT Pareto distribution, and, in contrast, are almost always well fit to a Laplacian distribution.
Also, despite these strong finite-size effects, empirically one finds that the ESDs arising large, well trained,
modern NNs can frequently be well fit to a PL (or TPL), and that the fitted $\alpha\in [2,6]$ for $80-90\%$
of NN layers.  Notably, we rarely find $\alpha<2$ in the best performing, open source, pretrained DNNs.

\charles{Not sure where to put this..see text also}
As there is \emph{no} ground truth whatsoever as to the limiting spectral density of a strongly-correlated NN weight matrix
(especially without HT elements)
the \HTSR \Phenomenology uses Pareto matrices as a guide. 
However, as we will see in Section~\ref{sxn:HT_ESDs}, this analogy should be treated with caution because there are 
cases where it breaks down.

No matter why a matrix ESD is HT, it can be difficult to reliably estimate the $\alpha$ parameter when
the true $\alpha$ is large. For Pareto matrices of the size investigated here, 
an observed $\alpha$ above $6$ is uninformative --- the tail will decay very rapidly indeed,
leaving very little of it to study.
In this sense, the HT Universality classes are \emph{larger} than the set of only strongly-correlated 
matrices or Pareto random matrices.


There is a particularly important boundary between Universality classes where $\alpha = 2$. Recall that one of the 
properties of power law distributions $\rho(\lambda)\sim \lambda^{-\alpha}$ is that if $\alpha<2$, than the variance of 
$\rho(\lambda)$ is infinite. In such cases, the variance cannot be estimated empirically, making $\rho(\lambda)$ in 
some sense \emph{\ATypical}. This implies that the NN will have substantially greater difficulty in applying any further 
load to such a weight matrix. Thus, the value of $\alpha = 2$ is a \emph{critical value}. (See 
Figure~\ref{fig:mlp3-FC1-alpha-overloaded} in Section~\ref{sxn:hysteresis_effect} for an empirical study of this effect 
in a small MLP.)

Smaller PL exponent $\alpha$ values correspond to heavier tails, $\rho_{tail}(\lambda)$; and
the \HTSR \Phenomenology observes that smaller PL exponents $\alpha$ (at least for $\alpha\in(2,6)$) tend to correspond to better models.
This is the key idea of the \HTSR:
the generalizing components of a layer matrix $\mathbf{W}$ concentrate in larger singular vectors associated with the tail, and 
so that better models have more slowly-decaying (i.e., larger) ESD tails.
This differs significantly than simply taking a general low-rank approximation to $\mathbf{W}$, where the rank
is chosen without insight from the \HTSR \Phenomenology. 
The \SETOL theory formalizes this observation as a key assumption. We will revisit these model selection questions in 
Section~\ref{sxn:setol_overview} below.

\subsection{Data-Free \SHAPE and \SCALE \Quality Metrics}
\label{sxn:htsr-metics}

The \HTSR \Phenomenology provides quality metrics for both individual layers and (by averaging layers) for an entire NN model.

\paragraph{Layer-wise Quality Metrics.}
Using the \HTSR \Phenomenology, we can define several other \SHAPE and/or \SCALE based layer (quality) metrics.
These are available in the \WW tool, and they work very well in practice.
\begin{itemize}
\item 
\ALPHA 
$(\alpha)$: $\rho_{tail}(\lambda)\sim\lambda^{-\alpha}$. 
A \SHAPE-based quality metric.
\item
\LOGSPECTRALNORM: $\log_{10}\lambda_{max}$.
A \SCALE-based quality metric.
\item 
\ALPHAHAT 
$(\hat{\alpha})$: $\alpha\log_{10}\lambda_{max}$.
A \SCALE-adjusted \SHAPE-based quality metric.
\item
\RANDDIST: $JSD[\rho^{emp}|(\rho_{rand}^{emp})]$.
A \SHAPE-based, non-parametric quality metric, suitable for highly-accurate, epoch-by-epoch analysis.%
\footnote{JSD is the Jensen-Shannon Divergence between the original ESD and the ESD of the layer weight matrix, randomized element-wise.}
\item
\PLKS: $D_{KS}$.
The KS-distance, or quality-of-fit, of the PL fits.  
For transformers, foundation models, and large, complex, modern NNs, this is frequently an even better model quality metric than the $\alpha$ of the PL fit itself.
\item
\MPSOFTRANK: $\mathcal{R}_{MP}$.
The MP-SoftRank, defined in~\cite{MM18_TR_JMLRversion}, can be used to identify problems such as when there is significant label or data noise that causes spuriously small $\alpha$, and also when it is difficult to fit a PL law.%
\footnote{The~\WW tool also implements the WW-SoftRank, which is like the MP-SoftRank, but replaces $\lambda_{bulk}^{+}$ with $\lambda_{rand}^{max}$; these are mostly equivalent for large matrices, but they can be different for very small matrices.}
\end{itemize}

\noindent
Each of these quality metrics provide a simple characterization of the \SHAPE and/or \SCALE of the tail of the ESD of a given layer $\mathbf{W}$.
These metrics are related to each other, and they have various trade-offs in practice~\cite{MM20a_trends_NatComm, MM21a_simpsons_TR, YTHx23_KDD}.
Of particular interest here in our development of \SETOL are the PL-based \WW~\ALPHA and  \ALPHAHAT~metrics.

\paragraph{From Layer-wise Quality Metrics to Layer-Averaged Model Quality Metrics.}
One can use the \HTSR \Phenomenology to go beyond individual \LayerQuality metrics, to construct model quality metrics by averaging \LayerQuality metrics (over all layers that are not very small). 
Existing \HTSR model quality metrics implicitly require that all layers are statistically independent, so that the average model quality is just the average of the contributions from each weight matrix $\mathbf{W}$.%
\footnote{This independence assumption, clearly a mathematical convenience, gets us closer to a workable theory. One could go beyond a ``single layer theory'' by adding in intra-layer correlations empirically. The \WW tool does support this, but doing so is outside the scope of this work.}
Given a \LayerQuality metric, $\Q^{NN}_{L}(\mathbf{W})$, one can define the \emph{\ModelQuality} $\Q^{NN}$ metric for an entire model as 
\begin{align}
\label{eqn:ProductNorm}
\Q^{NN}&:=\underset{L}{\prod}\;\Q^{NN}_{L}(\mathbf{W}) ,
\end{align}
a product of each independent \LayerQuality $\Q^{NN}_{L}$, and then consider the layer average as the log \emph{\LayerQuality},
\begin{align}
\label{eqn:LogProductNorm}
\log\Q^{NN}=\frac{1}{N_{L}}\underset{L}{\sum}\;\log\Q^{NN}_{L}=\langle\log\Q^{NN}_{L}\rangle_{\bar{L}}
\end{align}
where $\langle\;\cdots\;\rangle_{\bar{L}}$ denotes the layer average.

In particular, prior work has used the following metrics:
\begin{itemize}
\item
The layer-averaged model quality metric \ALPHA, $\log\Q^{NN}=\langle\alpha\rangle_{\bar{L}}$, describes the \SHAPE of the ESDs.
One can use the averaged \ALPHA when studying a single model, and only varying the regularization hyperparameters, although \ALPHA also works very well as a model quality metric when comparing different transformer models~\cite{YHTx21_TR}.
\item
The layer-averaged model quality metric \LOGSPECTRALNORM, $\log\Q^{NN}=\langle\log\lambda_{max}\rangle_{\bar{L}}$, describes the \SCALE of the ESDs.
The averaged \LOGSPECTRALNORM does work as a model quality metric, but not as well as \ALPHA~(or \ALPHAHAT).
Notably, \SLT predicts that smaller, not larger, \LOGSPECTRALNORM should be correlated with model quality; the opposite is observed in practice!
This is because a smaller layer $\alpha$ generally, but not always, corresponds to a larger $\lambda_{max}$.%
\footnote{The \LOGSPECTRALNORM can exhibit a Simpson's paradox when segmenting models by quality)~\cite{MM21a_simpsons_TR}.  Nevertheless, this metric may be useful when a PL fit can not be obtained, say, when $N\gg M$ and $M$ is very small, as with LSTMs,  U-Net architectures, etc.}
\item
The layer-averaged model quality metric \ALPHAHAT, $\log\Q^{NN}=\langle\alpha\log\lambda_{max}\rangle_{\bar{L}}=\langle\hat{\alpha}\rangle_{\bar{L}}$, incorporates both \SHAPE and \SCALE information.
This can compensate for anomalies that can arise when (say) comparing models of different sizes or model qualities~\cite{MM21a_simpsons_TR} or when other issues cause unusually large $\lambda_{max}$. See Section~\ref{sxn:Traps}).
\end{itemize}

\noindent
The layer-averaged \ALPHAHAT model quality metric has been applied in a large meta-analysis of hundreds of SOTA 
pre-trained publicly-available NN models in CV and NLP~\cite{MM20a_trends_NatComm,YTHx22_TR,YTHx23_KDD,MM19a_TR}. 
Generally speaking, \HTSR shape-based metrics, when used appropriately, outperform all other metrics studied (including 
those from \SLT, and with access to the training/testing data,) for predicting the quality of SOTA pre-trained publicly available NN models.  
The \HTSR theory predicts that the best-performing NN models have layers with $\ALPHA\in[2,6]$, and with $\alpha=2$
indicating optimal performance.
Moreover, prior empirical results show that the \ALPHA and \ALPHAHAT~metrics can predict trends in the \Quality 
(i.e., the \GeneralizationAccuracy), of SOTA NN models---\emph{even without access to any training or testing data}~\cite{MM20a_trends_NatComm}.

\newpage
\section{A \SemiEmpirical Theory of (Deep) Learning (\SETOL)}
\label{sxn:setol}
Based on prior empirical results, and the success of the \ALPHA and \ALPHAHAT metrics that are based on the \HTSR \Phenomenology, this leads to the deeper question: 
\begin{quote}
\emph{Why do the \ALPHA and \ALPHAHAT metrics work so well as NN model quality metrics for SOTA NN~models?}
\end{quote}
That is, why do NN models with heavier-tailed layer Empirical Spectral Distributions  (ESDs) tend to generalize better when compared to related models, and how can single-layer metrics predict model quality so well ?
Relatedly, can we derive these metrics from first principles?
(If so, then under what conditions do they hold, and under what conditions do they fail?)

\noindent
To answer these questions, we will derive a general expression for the \LayerQuality, $\Q$, of an NN.
Although many modern NNs have many layers, we adopt a single-layer viewpoint (like a matrix-generalized Student–Teacher) because in \StatisticalMechanicsOfGeneralization (\SMOG) theory~\cite{SST92,STS90} the multi-layer generalization can be factorized or approximated.
For this, we will obtain by simple averaging our model quality metrics, under effectively a single layer approximation, that correspond to \ALPHA and \ALPHAHAT.

In deriving these quantities, we will introduce to NN theory a new \SemiEmpirical approach that combines techniques from \STATMECH and \RMT in a novel way.must depend on the spectral density
The \LayerQuality $\Q$ will estimate the contribution that an individual NN layer makes to the overall quality of a trained NN model.
In deriving $\Q$, we have discovered a new \LayerQuality metric, called the \TRACELOG condition,
which indicates that the generalizing components of the layer must concentrate into a low-rank subspace which we term the \emph{\EffectiveCorrelationSpace}, or~\ECS. That is, the better this condition is met, the more the generalizing components concentrate in this \ECS, and this tendency provides a new layer quality metric, derived in a totally independent way from \ALPHA and \ALPHAHAT. We will provide strong empirical justification that both metrics are acceptable Quality metrics by showing that they converge as model quality increases.

Importantly, we have conducted detailed experiments to show 
that the empirical estimates of the \SETOL \TRACELOG condition align remarkably well with predictions from the \HTSR
theory under \Ideal conditions (see Sections~\ref{sxn:empirical-test_acc}) and, then, 
that the key assumptions of our \SETOL theory are valid
(see Sections~\ref{sxn:empirical-effective_corr_space} and \ref{sxn:empirical-trace_log}).
In Section~\ref{sxn:empirical_comp_r_transforms}, we demonstrate how to apply theory directly using explicit calculations of the \RMT layer cumulants.
We next examine how the \HTSR predictions (i.e., the HT PL exponent $\alpha$) behave under non-\Ideal conditions (see Sections~\ref{sxn:empirical-correlation_trap} and \ref{sxn:hysteresis_effect}).
In the following, we will outline key conceptual aspects of \SETOL.
In Section~\ref{sxn:setol_overview}, we give an overview of \SETOL;
In Section~\ref{sxn:ideal_learning}, we describe the conditions of \Ideal learning under \SETOL and how they differ from those of \HTSR; and
In Section~\ref{sxn:HT_ESDs} we describe conditions that deviate from this.

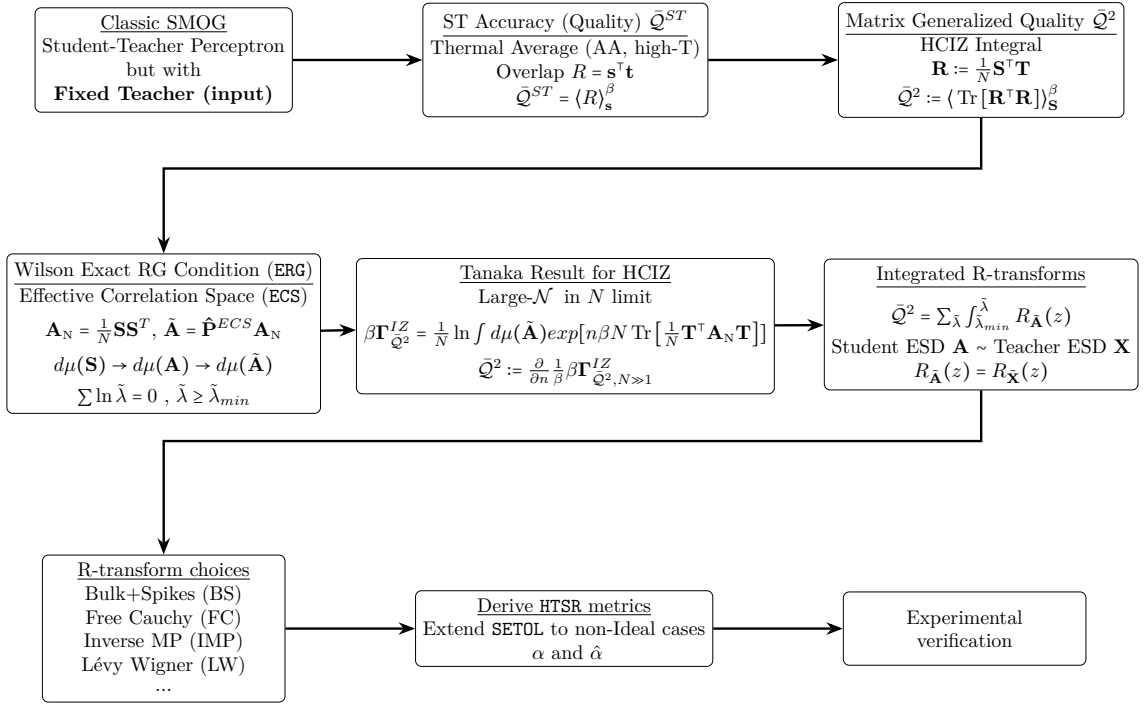
\begin{figure}[htbp]                 
  \centering
  \scalebox{0.75}{
\usetikzlibrary{arrows.meta, positioning, calc}

\tikzset{
  box/.style={
    draw, rectangle, rounded corners=3pt,
    minimum width=4.3cm, minimum height=1.3cm,
    align=center, font=\small
  },
  arrow/.style={-Stealth, very thick},
}

\begin{tikzpicture}[
  node distance = 2.5cm and 2.3cm,
  >=Stealth
]

\node[box] (smog)    {\underline{Classic SMOG}\\Student-Teacher Perceptron\\
but with  \\
\textbf{Fixed Teacher (input)}};
\node[box, right=of smog] (overlap)
      { \underline{ST Accuracy (Quality) $\Q^{ST}$} \\
    Thermal Average (AA, high-T) \\
        Overlap $R=\SVEC^{\top}\TVEC$ \\
      $\Q^{ST}=\THRMAVG{R}$ 
      };
\node[box, right=of overlap] (hciz)
      {\underline{Matrix Generalized Quality $\QT$}\\
      HCIZ Integral \\
$\OVERLAP:=\tfrac{1}{N}\SMAT^{\top}\TMAT$ \\
$\QT :=   \THRMAVGIZ{\Trace{\OVERLAP^{\top}\OVERLAP}}$  \
      };

\node[box, below=of smog] (rg)
      {
        \underline{Wilson Exact RG Condition (\ERG)}\\
      {Effective Correlation Space (\ECS)}\\[4pt]
      $\AMATN=\tfrac{1}{N}\SMAT\SMAT^{T}$, $\AECS=\mathbf{\hat{P}}^{ECS}\AMATN$ \\[4pt]
      $d\mu(\SMAT)\rightarrow  d\mu(\AMAT) \rightarrow d\mu(\AECS)$ \\[4pt]
      $\sum \ln \LambdaECS = 0 $  , $\LambdaECS \ge \LambdaECSmin $
      };

\node[box, below=of overlap] (tanaka)
      {\underline{Tanaka Result for HCIZ}\\
      \LargeN in $N$ limit \\[6pt]
      $\IZG = \tfrac{1}{N}\ln \int d\mu(\AECS)
      exp[n\beta N \Trace{\tfrac{1}{N}\TMAT^{\top}\AMATN\TMAT}]$\\[4pt] 
      $\QT := \tfrac{\partial }{\partial n}\tfrac{1}{\beta}\IZGINF$
      };

\node[box, below=of hciz] (cumul)
  {\underline{Integrated \RTransforms} \\[4pt]    
     $\QT = \sum_{\LambdaECS}\int_{\LambdaECS_{min}}^{\LambdaECS}R_{\AECS}(z)$ \\[4pt]  
     Student ESD $\AMAT$ $\sim$ Teacher ESD $\XMAT$ \\
   $R_{\AECS}(z) = R_{\XECS}(z)$
   };

\node[box, below=of rg] (rforms)
      {\underline{\RTransform choices}\\
      Bulk+Spikes (BS) \\
      Free Cauchy (FC)  \\ 
      Inverse MP (IMP) \\
      L\'evy Wigner (LW) \\
      $\cdots$
};

\node[box, right=of rforms] (alpha)
      {\underline{Derive \HTSR metrics}\\
      Extend \SETOL to non-Ideal cases \\\
      $\alpha$ and $\hat{\alpha}$};

\node[box, right=of alpha] (expv)
      {Experimental\\verification};

\draw[arrow] (smog)   -- (overlap);
\draw[arrow] (overlap) -- (hciz);

\draw[arrow] (rg)     -- (tanaka);
\draw[arrow] (tanaka) -- (cumul);

\draw[arrow] (rforms) -- (alpha);
\draw[arrow] (alpha)  -- (expv);

\coordinate (hcizDrop) at ($(hciz.south)+(0,-0.9)$);          
\coordinate (hcizLeft) at ($(hcizDrop -| rg.north)$);         
\draw[arrow] (hciz.south) -- (hcizDrop)                       
             -- (hcizLeft)                                    
             -- (rg.north);                                   

\coordinate (cumulDrop) at ($(cumul.south)+(0,-0.9)$);
\coordinate (cumulLeft) at ($(cumulDrop -| rforms.north)$);
\draw[arrow] (cumul.south) -- (cumulDrop)
             -- (cumulLeft)
             -- (rforms.north);

\end{tikzpicture}
}   

  \caption{Flowchart of the theoretical concepts used to construct SETOL.}
  \label{fig:cetl-flow}
\end{figure}
\subsection{\SETOL Overview}
\label{sxn:setol_overview}

\charles{Some old text here I need to clean up}

Our \SETOL formulates a parametric expression for the \LayerQuality $\Q$ using a matrix-generalization of the classic \StudentTeacher (ST)
model from the \SMOG theory of the 1990s~\cite{SST92,STS90}, 
but with a \SemiEmpirical twist in that the \Teacher is  now an actual, trained NN that is input to theory.
Recent advances in the evaluation of so-called HCIZ random matrix integrals~\cite{potters_bouchaud_2020,Tanaka2007, Tanaka2008},
such that the final expression for $\Q$ to be written in terms of empirically measured statistical properties of the layer ESD.
We summarize our basic approach here; see Section~\ref{sxn:matgen} for a detailed derivation, and see Section~\ref{sxn:empirical} for a detailed empirical analysis.  For reference here and later, Figure~\ref{fig:cetl-flow} displays a flowchart of the conceptual development of \SETOL.

As is done in the \StudentTeacher (ST) model~\cite{SST92}, 
we first formulate the \GeneralizationError $(\AVGE^{ST}_{gen})$ of the linear \Perceptron (in the \emph{\AnnealedApproximation},
and in the \emph{\HighTemperature} limit; see Section ~\ref{sxn:SMOG_main}), and 
we then generalize this to the case of a NN $(\AVGE^{ST}_{gen}\rightarrow\AVGE^{NN}_{gen})$, so that we can analyze the \Quality of each layer.
For the \Perceptron, the \GeneralizationError is an Energy, given as $\AVGE^{ST}_{gen}:=\THRMAVG{1-R}$, where $R$ is the ST vector overlap,
and $\THRMAVG{\cdots}$ is a \emph{\ThermalAverage} (defined in Section~\ref{sxn:mathP}),
a Boltzmann-weighted average.
In this case, the model \Quality, $\Q^{ST}$ is exactly the AA, high-T \AverageGeneralizationAccuracy 
$\Q^{ST}:=1-\AVGE^{ST}_{gen}=\THRMAVG{R}$.
For an MLP or general NN, each layer's Energy is (negatively) associated with a
\LayerQuality $\Q$, which we identify as the average contribution an
individual layer contributes to the overall generalized accuracy,
(i.e $1-\AVGE^{NN}_{gen}$) for a multilayer perceptron (MLP).

\textbf{Importantly, we deviate from the traditional approach in that we take the \Teacher as a fixed, empirical input to the theory.}  The Teacher contribution is then formulated in terms of the actual the layer eigenvalues, or Empirical Spectral Density (ESD).  In this way, the theory becomes \SemiEmpirical.

For technical reasons (below), we will seek the
\emph{\LayerQuality (Squared)} $\QT$, which is defined as the \ThermalAverage
of the matrix-generalized overlap ($\Trace{\OVERLAP^{2}}$),
\begin{align}
  \label{eqn:QT}
  \QT :=   \THRMAVGIZ{\Trace{\OVERLAP^{2}}}
\end{align}
where $\OVERLAP^{2}$ can be thought of as a \Hamiltonian for the \QualitySquared  $(\HBARE=\OVERLAP^{\top}\OVERLAP)$.

In \EQN~\ref{eqn:QT}, the so-called \emph{\Teacher} ($T$) is the NN model under consideration,
and 
$\mathbf{R}:=\frac{1}{N}{\SMAT^{\TR}\TMAT}$
denotes the ST overlap operator 
between the \Teacher layer weight matrix $\TMAT$ and a similar \emph{\Student} ($S$) layer weight matrix $\SMAT$.
The notation $\THRMAVGIZ{\cdots}$ denotes a \ThermalAverage over
all \Student weight matrices $\SMAT$ that resemble the \Teacher weight matrix $\TMAT$.
By ``resemble'', the \SETOL approach assumes that 
the ESD of $\SMAT$ has the same \emph{limiting} form as $\TMAT$, placing them in the same \HTSR Universality class.
This is made more precise below.

\paragraph{The Overlap $\OVERLAP$, and the Inner and Outer forms of $\AMAT$, $\AMATM$ and $\AMATN$, resp.}
Let us now express the average matrix-matrix overlap $\mathbf{R}$ in squared form using:
\begin{align}
  \label{eqn:R2}
\Trace{\OVERLAP^{2}} &:=\OLAPSQD  \\ \nonumber
&=\frac{1}{N^{2}}\Trace{\TMAT^{\TR}\SMAT\SMAT^{\TR}\TMAT}
=\frac{1}{N}\Trace{\TMAT^{\TR}\AMATN\TMAT}
\end{align}
where $\AMATN$ is called the \emph{Outer} \Student correlation matrix, and is  the $N\times N$ form of the \Student correlation matrix,
$\AMATN:=\frac{1}{N}\mathbf{SS}^{\TR}$.
We will also define the $M\times M$ \emph{Inner} \Student correlation matrix $\AMATM=\frac{1}{N}\SMAT^{\TR}\SMAT$, which is 
used later. Note that $\AMATM$ also has a $\frac 1 N$ factor, so that its non-zero eigenvalues will be the same as $\AMATN$, but it has dimensions $M \times M$. We will denote the Outer $\AMATN$ and Inner $\AMATM$ forms when the context demands it, and will simply use the general form $\AMAT$ when they can be interchanged.


\paragraph{The Quality-Squared Generating Function $\IZG$.}
As explained in Section~\ref{sxn:mathP}, this \QualitySquared is more readily obtained as 
the derivative of the \LayerQualitySquared \GeneratingFunction, $\IZG$, defined as
\begin{align}
  \label{eqn:IZG_QT}
  \QT := \dfrac{1}{\beta}\dfrac{\partial }{\partial \ND} \lim_{N\gg 1}\IZG
\end{align}
where $\ND$ is the number of training examples used to train the model.

$\IZG$ is essentially ($\beta$ times) a \emph{negative \FreeEnergy} for the (approximate) \LayerQualitySquared
(see Section~\ref{sxn:SMOG_main}, and the Appendix, Section~\ref{sxn:quality}).
For more details, see Section~\ref{sxn:matgen}, and the Appendix, Section~\ref{sxn:summary_sst92}).

We can write $\IZG$ as an HCIZ Integral, 
\begin{align}
  \label{eqn:QT_dS}
  \IZG  &= \tfrac{1}{N} \ln\int d\mu(\SMAT)\exp\left(\ND \beta N\Trace{\tfrac{1}{N}\TMAT^{\TR}\AMATN\TMAT}\right),
\end{align}
where $n$ is the number of training examples used to train the models.

The \SETOL approach then seeks to express $\IZG$ in~\EQN~\ref{eqn:QT_dS} as an HCIZ integral (and in the \WideLayer \LargeN limit in $N$$)~$\cite{potters_bouchaud_2020,Tanaka2007,Tanaka2008}.
We evaluate this at large-$N$ in $N$, and write
\begin{equation}
  \IZGINF:=\lim_{N\gg 1}\IZG
\end{equation}
The result is effectively expressed in the limit of fixed \emph{layer load} $\ND/N$, analogous to a renormalized mean-field  (i.e., \SemiEmpirical) theory over $N$ interacting (feature) vectors of length $M$.
In other words, in taking the \LargeN in $N$ limit, we assume implicitly that $\IZG$ is in the \ThermodynamicLimit (and at high-T).

With these definitions in place, moving forward, the following key assumptions, which can be tested empirically, must hold:
\begin{itemize}
  \item
  \textbf{The Effective Correlation Space (\ECS) Condition.}
  The generalizing components of the \Student (and \Teacher) layer weight matrices concentrate into a lower rank subspace---the~\ECS---spanned by the
  eigenvectors associated with the (heavy) tail of the layer ESD $\rho_{tail}(\lambda)$, such that the test error can 
  be reproduced with only these components. 
  We write $\AECS$ to denote the projection of the correlation
  matrix $\AECS:=\mathbf{P}_{ecs}\AMAT$, onto this subspace, now with rank $\MECS\ll M$.
  This restricts the measure $d\mu(\AMAT)$ to the~\ECS, $(d\mu(\AMAT)\rightarrow d\mu(\AECS))$.
  This assumption will be empirically examined using real-world \Teacher weight matrices $\TMAT=\WMAT$
  in Section~\ref{sxn:empirical-effective_corr_space}. 
  \item
  \textbf{The \TRACELOG Condition.}
  The Effective Student correlation matrix $\AECS$ 
  satisfies the \ERG condition that $\Trace{\ln\AECS}=\ln\Det{\AECS}=0$,
  so that the change of measure $d\mu(\SMAT) \rightarrow d\mu(\AECS) $ is Volume Preserving.
  This condition is derived explicitly in terms of Inner form,  $\AECSM$, and will also  hold for the Outer form, $\AECSN$
  Practically, this implies that the $\MECS$ eigenvalues $\LambdaECS$
  of the tail of the ESD must satisfy $\sum_{i=1}^{\MECS}\ln\LambdaECS_{i}\approx 0$.
  Experiments will test this assumption explicitly in Section~\ref{sxn:empirical-trace_log}.
\end{itemize}
Remarkably, both conditions hold best empirically when the \HTSR PL quality metric $\alpha\gtrsim 2$ is \Ideal. Motivated from these empirical observations, we have:
\begin{itemize}
  \item
  \textbf{$\IZGINF$ is expressed as an HCIZ integral, at large-$N$.}
  We have
  \begin{align}
  \label{eqn:IZGINF_HCIZ}
  \IZGINF = \lim_{N\gg 1}\tfrac{1}{N}\ln\int d\mu(\AECS)\exp\left(\ND\beta\Trace{\TMAT^{\TR}\AECSN\TMAT}\right) 
  \end{align}
  where  measure $d\mu(\AECS)$ lets us average over all (Outer) \Student Correlation matrices $\AECSN$ explicitly, which
  lie in the~\ECS space and which ``resemble'' the \Teacher.
By ``resemble'' we mean that they share the same functional form in
  their limiting ESDs,
   $\rho^{\infty}_{\AECS}(\lambda)\sim\rho^{\infty}_{\XECS}(\lambda)$.
  \item
  \textbf{The Layer Quality (Squared) $\QT$ is a~\GEN.}
  The final expression for $\QT$ can be written as the derivative of $\IZGINF$  as
  \begin{align}
    \label{eqn:QT_result}
    \QT = \dfrac{1}{\beta}\dfrac{\partial}{\partial \ND}\IZGINF = \sum_{i=1}^{\MECS}\GNI
  \end{align}
  where $\GNI$ is, albeit imprecisely, a \emph{\GEN}, and is  defined as the integrated \emph{\RTransform} $R(z)$ of the \Teacher
  layer ESD (where $z\in\mathbb{C}$), such that $\GN:= \int_{\LambdaECSmin}^{\lambda}R(z)dz$
  and $\LambdaECSmin$  encapsulates  the~\ECS (and, for some choices of $R(z)$,  selects the desired branch-cut of $R(z)$
  so that it is both analytic and well-behaved) .
\end{itemize}

To apply the theory, one must choose an \RTransform $R(z)$ for the \Teacher that models 
the tail of the ESD $\rho^{emp}_{T}(\lambda)$, and that can be
parameterized by some measurable property.
This may include the number of Spikes $\lambda^{spike}$, the fitted PL exponent $\alpha$,
the maximum eigenvalue $\lambda_{max}$, or even the entire tail $\rho^{tail}_{T}(\lambda)$.
This may be a formal expression, a computational procedure, or some combination. 

To integrate $R(z)$, however, and to have a physically meaningful result,
one must ensure that $R(z)$ is both
analytic and single-valued on the domain of interest, namely, the \ECS (and therefore
the (PL) tail of the ESD),  $\Re[z] \ge \LambdaECSmin$.
\footnote{$\Re[z]$ is the Real part of $z$.}
Because the ESD is frequently \HeavyTailed (HT), this
\RTransform $R(z)$ may have a branch-cut, and it is expected that this will occur
at $\Re[z]\approx\LambdaECSmin$, corresponding to a point roughly at or before the start of the \ECS (i.e., at the peak of the ESD on a log scale).
Selecting the branch-cut $R(z)$ provides an additional physical meaning to the \ECS.

To complete the theory, we
will also show that the \HTSR PL \LayerQuality metrics \ALPHA $(\alpha)$ and \ALPHAHAT $(\ALPHAHATEQN)$
can be formally derived directly from the \SETOL \LayerQuality $\Q$ by selecting the appropriate
\RTransform $R(z)$ and making empirically motivated approximations. In Section~\ref{sxn:r_transforms} we provide several possible
models of $R(z)$ and the resulting \LayerQuality $\Q$.

\paragraph{Renormalization Group Effective Hamiltonian}
The formulation of \SETOL closely parallels the construction of an \EffectiveHamiltonian~$\HEFF$
via the \WilsonExactRenormalizationGroup~(\ERG) approach. Consider a \emph{\Bare} Hamiltonian $\HBARE$ for the \LayerQualitySquared,
defined as $\HBARE:= \mathbf{R}^{\TR}\mathbf{R}$.
We can express \EQN~\ref{eqn:QT_dS} in terms of this \Bare Hamiltonian $\HBARE$,
and rewrite \EQN~\ref{eqn:IZGINF_HCIZ} in terms of a \emph{Renormalized} \EffectiveHamiltonian~$\HEFF$
that spans the \EffectiveCorrelationSpace~(\ECS). Formally, we have:
\begin{align}
\label{eqn:RG}
\tfrac{1}{N}\ln \int d\mu(\SMAT)e^{\ND \beta N \operatorname{Tr}[\HBARE]} \;\xrightarrow{ERG}\; \lim_{N\gg 1}\tfrac{1}{N}\ln \int d\mu(\AECS)e^{\ND \beta N \operatorname{Tr}[\HEFF]} 
\end{align}
where the \ERG transformation is defined by the \ScaleInvariant change of measure,
applied in the \WideLayer \LargeN limit in $N$,
and where $\HEFF$ is defined implicitly through the result for $\QT$ (\EQN~\ref{eqn:QT_result}).
The result is, formally, a sum of the integrated \RTransforms $\GNI$.
In a sense, this result resembles (a non-perturbative form of) the Linked Cluster Theorem
in that the log \PartitionFunction is expressed as a sum of integrated matrix-generalized cumulants.  And in analogy with \SemiEmpirical theories of Quantum Chemistry, the \HTSR \ALPHA $(\alpha)$ and \ALPHAHAT $\ALPHAHATEQN)$ enter as renormalized empirical parameters.
Most importantly, the \ScaleInvariant \TRACELOG condition can be verified empirically (See Section~\ref{sxn:empirical}.)
Importantly, in analogy with the Wilson \ExactRenormalizationGroup theory, 
the \HTSR $\alpha=2$ resembles in spirit
an \ERG \emph{Universal Critical Exponent} at a phase boundary being between the \HeavyTailed 
(HT) and the \VeryHeavyTailed (VHT) phase of learning of the \HTSR theory.
This observation strengthens our argument that the \HTSR HT and VHT phases
are analogous to the generalizing and overfitting phases, respectively,
of the classical \SMOG theories of NN learning.

\subsection{Comparing SETOL with HTSR: Conditions for Ideal Learning}
\label{sxn:ideal_learning}


The \SETOL approach establishes a starting point for developing a first-principles theory for modern NNs. 
Among other things, by connecting with the \HTSR \Phenomenology, it lets us identify conditions for an \Ideal state of learning for an individual NN layer, under the Single Layer Approximation.
By \Ideal, we mean that the layer is being used most effectively i.e., in some sense it is at its optimal data load, and thus it is conjectured to result in the best model quality.

\textbf{The \Ideal State of Learning} is conjectured to be characterized by the following three conditions:
\begin{enumerate} 
\item \label{itm:ideal_1}
  the tail of ESD, $\rho^{emp}_{tail}(\lambda)$, can be well fit to a PL of $\alpha\approx 2$: $\rho^{emp}_{tail}(\lambda)\sim\lambda^{-2}$;
\item \label{itm:ideal_2}
  the eigenvalues in the tail, $\lambda_{i}$, satisfy the \TRACELOG (i.e., Trace-Log) condition: $\sum_{i}\ln\lambda_{i}=0$; and
\item \label{itm:ideal_3}
  the generalizing components of the layer concentrate in the singular vectors associated with the tail of the ESD, (whose span we call the \EffectiveCorrelationSpace).
\end{enumerate}
In Section~\ref{sxn:empirical}, we will test and justify this conjecture by showing that the induced measures of each condition converge to one another at the same rate, and reach convergence at the exact point beyond which accuracy begins to degrade.

These claims are fundamentally about NN learning itself. 
They are motivated by our formulation of the \SETOL approach in our search for a practical predictive theory behind the HTSR \Phenomenology.
When $(\ref{itm:ideal_1})$ and $(\ref{itm:ideal_2})$ conditions hold for any layer, we conjecture that $(3)$ holds as well.
Moreover, when $(\ref{itm:ideal_1}-\ref{itm:ideal_3})$ hold for all layers, we conjecture the
NN has the lowest \GeneralizationError (and highest \ModelQuality) possible for given model architecture and dataset.

Previous results have shown that the \HTSR quality metrics (\ALPHA,  \ALPHAHAT, etc.)  correlate very well with reported test accuracies,
as well as model quality on an epoch-by-epoch basis~\cite{MM18_TR_JMLRversion,KFWB13}.
These results hold because, as indicated by the \HTSR theory, the PL exponent $\alpha$ characterizes both the quality of the layer and provides an after-the-fact measure of the amount of regularization.%
\footnote{By ``after-the-fact'', we mean that it provides a measure of the regularization in a layer, along the lines of the self-regularization interpretation of \HTSR Theory~\cite{MM18_TR_JMLRversion}. However, we do \emph{not} recommend that it be used as an explicit regularization parameter. Informally, this is since the ``easiest way to obtain HT ESDs is to make weight matrices HT element-wise; but this is \emph{not} what is observed in practice, and thus this is precisely \emph{not} what \HTSR Theory and our new \SETOL approach are designed to model.}
However, the \HTSR approach says nothing about the \SETOL \TRACELOG condition; and neither does the \SETOL approach require a minimum of $\alpha=2$ to obtain the best model quality, as observed by the \HTSR \Phenomenology.
Remarkably, we can show that $(1)$ and $(2)$ do hold \emph{simultaneously}, both in carefully designed experiments on a small model,
as well as for many pre-trained, high quality open-source models (such as VGG, ResNet, Llama, Falcon, etc). 

\HTSR, however, has been developed as a \Phenomenology describing the best-trained, most accurate open-source models available.
As such, it may be biased towards such models, and it may not describe less optimal learning scenarios.
The key goal of this work is to derive independent conditions, both theoretical and experimental, that can
identify the conditions for \IdealLearning, and to stress-test these conditions in carefully designed,
reproducible experiments.

\subsection{Detecting Non-Ideal Learning Conditions}
\label{sxn:HT_ESDs}


The \HTSR \Phenomenology posits that SGD training reduces the \emph{\TrainingError} by accumulating correlations into the large eigenvalues
in NN layer weight matrices  $\mathbf{W}$ such that they \emph{self-organize} into a HT with a PL signature,
and that this successful self-organization leads to good model quality.
Conversely, it also posits that when training has gone awry in some way, the resulting ESD, $\rho^{emp}(\lambda)$, will
be deformed in some way.   
In many practical situations, there can be other, 
competing factors that give rise to large eigenvalues
that do not contribute to the generalization capacity of the model, and, consequently, 
can affect the \Scale (i.e., the largest eigenvalue(s) $\lambda_{max}$) 
and \Shape (i.e., the PL exponent $\alpha$, or goodness of fit $D_{KS}$) of the layer ESDs.
These large $\lambda$ could be \DragonKings~\cite{sornette2009dragonkings}, \emph{\CorrelationTraps}, or some other anomaly.

To apply the \HTSR \Phenomenology most effectively, one must be able to identify various spurious factors and
distinguish real correlations from any other large eigenvalues, including the effects of both
extreme eigenvalues $\lambda$, individual matrix elements $W_{i,j}$, and rank-$1$ perturbations in $\WMAT$.
In one case the ESD is HT primarily due to correlations that help the model generalize, whereas
in another when the ESD may be more HT than expected due to suboptimal training, mis-labeled data, etc.
In extreme cases, spurious eigenvalues can push the weight matrix
into the \VeryHeavyTailed \Universality class (i.e., $\alpha<2$. See Table~\ref{tab:Uclass}, Section~\ref{sxn:htsr}), or
disrupt the formation of a HT, resulting in a poor PL fit, undermining the core proposition of the  \HTSR approach.

When training a model with SGD, one may only achieve  a sub-optimal result
when using overly large learning rates / small batch sizes, (see Section~\ref{sxn:empirical}),
from poor hyper-parameter settings,
or simply because direct regularization fails. In such cases, the \HTSR approach allows one to detect
potential problems by looking for large eigenvalues~\emph{not} resulting from correlations~\cite{GSZ20_TR}.

Importantly, in the context of the \SETOL theory, we can now identify such empirical anomalies
due to \ATypical layer weight matrices $\mathbf{W}$, a key factor when models break down.
The \SETOL approach formalizes the empirical \HTSR \Phenomenology,
but, in doing so, assumes that the
underlying layer effective correlation matrix $\XECS$, is~\Typical, meaning that it can describe out-of-sample / test data.
Conversely, if the underlying weight matrix $\mathbf{W}$ is~\ATypical, then it is in some sense
overfit to the  training data and can not fully represent out-of-sample / test data.
Consequently, when $\mathbf{W}$ is \ATypical,  we argue that we can observe this, either in the ESD $\rho^{emp}(\lambda)$ directly
(i.e., when $\alpha< 2$),
and/or having $1$ or more unusually large matrix elements $W_{ij}$.

We conjecture such sub-optimal results, and particularly those occurring from overfitting, actually
arise when the underlying layer weight matrix $\mathbf{W}$ is \ATypical in some way,
in analogy to the results from the classic \SMOG theory (see Section~\ref{sxn:trad_smog}),
and, importantly, that we can use the \SETOL approach to detect when $\mathbf{W}$ is~\ATypical
and therefore a layer is overfit in some way.


Here, we identify two specific cases of \ATypical weight matrices---\CorrelationTraps and \emph{\OverRegularization}.%
\footnote{Later, in Section~\ref{sxn:empirical}, we will show that we can systematically induce both phenomena and observe their effects on the \HTSR HT PL metric $\alpha$ and the \SETOL \TRACELOG condition.}
\begin{enumerate}[label=3.3.\arabic*]
  \item 
  \textbf{\CorrelationTraps.} 
  $\mathbf{W}$ is \ATypical in that $\mathbf{W}$
 exhibits an anomalously large mean $(\bar{\WMAT})$.  
  We can observe these by randomizing the layer weight matrix, $\mathbf{W}\rightarrow\mathbf{W}^{rand}$, and then looking
  for eigenvalues that extend significantly beyond the MP edge of the random bulk (i.e., Spikes).  We call such random spikes
  \emph{\CorrelationTraps},  denoted as $\lambda_{trap}$, because they appear, in some extreme cases, to be associated with distorted ESDs
  and, importantly, lower test accuracies.  Examples of \CorrelationTraps are shown in Section~\ref{sxn:empirical-correlation_trap}.
  \item 
  \textbf{\OverRegularization.} 
  $\mathbf{W}$ is \ATypical in that $\mathbf{W}$ exhibits an anomalously large variance $(\sigma^2(\WMAT))$. 
  We can observe this when the layer  $\alpha < 2$.  Also, since \ALPHA a measure of implicit regularization, we say the layer with $\alpha<2$ is \emph{Over-Regularized}.
  In particular, when one layer is undertrained, having $\alpha>6$, it appears that other layers may become overtrained to compensate, and this can be seen with having $\alpha < 2$.
  These effects are studied in Section~\ref{sxn:empirical}.
  Additionally, we also observe that when evaluating the \SETOL \TRACELOG condition, when $\alpha < 2$, then $\Delta \lambda_{min}< 0$ (see Section~\ref{sxn:empirical-trace_log}).

\end{enumerate}


\subsubsection{Correlation Traps}
\label{sxn:Traps}

The first way we identify $\mathbf{W}$ as \ATypical is when it has an anomalously large mean $(\bar{\WMAT})$;
detecting this in general, however, requires more than just examining which elements $W_{i,j}$ are
anomalously large but, insteading, looking for what can be multi-element rank-1 perturbations in $\WMAT$.  Here, we can apply elementary RMT, as in the \HTSR approach.

The \HTSR \Phenomenology states that NNs generalize better when their layers ESDs are more HT---precisely because the 
tail eigenvalues arise from correlations in the weight matrices.
So one way is to identify \emph{atypicality} is to look for  large eigenvalues that
do not arise from correlations in $\mathbf{X}$,
but, rather, from one or a few spuriously large matrix elements $W_{i,j}$ and/or rank-1 perturbations in $\mathbf{W}$.
We call these eigenvalues  \CorrelationTraps, denoted by $\lambda_{trap}$
(i.e., see Section~\ref{sxn:empirical-correlation_trap}).

Indeed, if we randomize $\mathbf{W}$ element-wise, i.e $\mathbf{W}\rightarrow\mathbf{W}^{rand}$, we
expect the $W^{rand}_{i,j}$ matrix elements to be i.i.d and with a small mean
(unless something odd happens during SGD training).
Likewise, we expect the singular values of $\WMAT^{rand}$ to follow the MP distribution, to within
finite-size / TW fluctuations.
If we observe an eigenvalue $\lambda_{trap}$ extending beyond the MP bulk region, $\lambda_{trap}>\lambda^{+}_{bulk}$,
then the mean $W_{i,j}$ matrix element will also be anomalously large,
and we can identify $\mathbf{W}$ as \ATypical.
We must be careful, however, as we do not fully understand the origin of these atypicalities
and do not claim that every one is associated with suboptimal generalization.

By a \emph{\CorrelationTrap}, we mean that some anomaly in the training of $\mathbf{W}$ resulted
in one or more spuriously large eigenvalues $\lambda_{trap}$ in $W^{rand}$,
and that  whatever caused them also may, in some pronounced cases, tend
to ``trap the correlations in $\mathbf{X}$ itself,
preventing them from coalescing into a well defined PL Heavy Tail,
or otherwise distorting the ESD.
Whether they are a signature of training gone wrong, or whether they distort the dynamics of the tail correlations 
simply by being there, \CorrelationTraps can be expected to alter the shape of the ESD,
reducing the quality of the PL fits, and sometimes producing spurious $\alpha$ values.

Why would such anomalies occur in a NN?
It is conceivable that SGD will, when it fails to find usable correlations, instead
produce spurious
correlations in the form of large elements and/or rank-1 perturbations.
Also, the matrix itself may simply undergo an innocuous  mean-shift because the
mean is not explicitly controlled during training. Here,  mean-recentering may be beneficial.
\footnote{Similarly, when training NNs, frequently the weight matrices~\cite{baskin2021} or activations~\cite{choi2018_TR}
may need to be clipped during training to ensure good results.}

We will see, below in Section~\ref{sxn:empirical}, that we can induce a \CorrelationTrap both by shrinking the batch 
size, or, equivalently, increasing the learning rate, and that this is associated with degraded model performance
and small $\alpha$.
We seek to identify specific ways of identifying such traps because we 
reason that the presence of foreign large eigenvalues may disrupt the self-organization of correlations – or that failed learning may produce them as a by-product.

\paragraph{Detecting Correlation Traps with \RMT.}

\begin{figure}[ht]
    \centering
    \subfigure[Well-formed ESD]{ 
      \includegraphics[width=7cm]{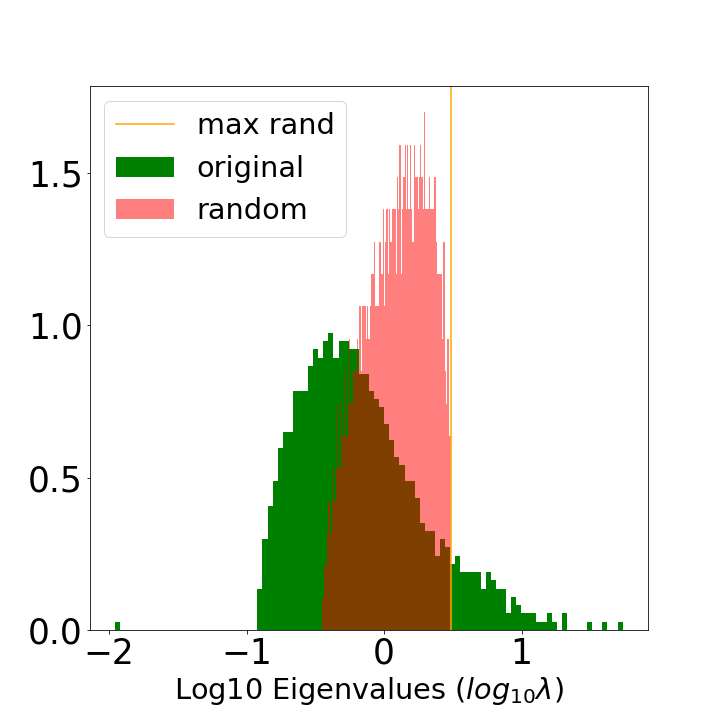}
      \label{fig:scale-shape-a}
    }                               
    \subfigure[ESD with \CorrelationTrap]{                   
      \includegraphics[width=7cm]{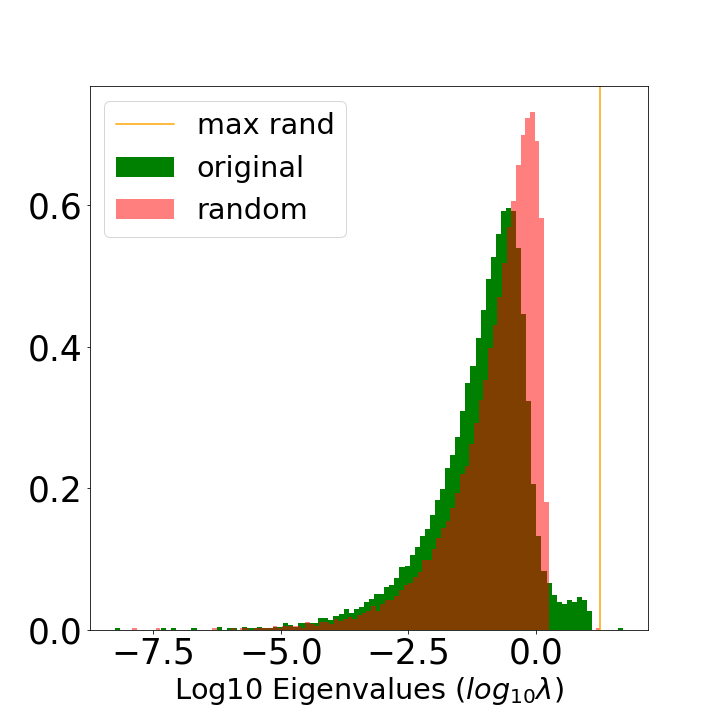}
      \label{fig:scale-shape-b}                           
    }                                                                                                                            
    \caption{Comparison of a well-formed, Heavy-Tailed ESD (a) to one with a Correlation Trap (b), in the VGG16 model (FC2 layer)}
  \label{fig:scale-shape}                                                                                                      
\end{figure}

RMT suggests that when a matrix $\mathbf{W}$ has unusually large elements $W_{ij}$, then the ESD will have one or more large eigenvalues $\lambda_{trap}$ lying outside the bulk edge $\lambda^{+}_{bulk}$ of the ESD, as predicted by MP theory. 
One can detect these so-called \CorrelationTraps in a weight matrix $\mathbf{W}$ by performing the following:
\begin{enumerate}
\item randomize $\mathbf{W}$ element-wise to obtain $\mathbf{W}^{rand}$;
\item compute the ESD for $\mathbf{W}^{rand}$; and
\item look for large eigenvalues $\lambda_{trap}\gg\lambda^{+}_{bulk}$.
\end{enumerate}
\WW~looks for \CorrelationTraps ($\lambda_{trap}$) in the ESD of the randomized $\mathbf{W}^{rand}$, that are larger than 
$ \lambda_{trap}>\lambda_{bulk}^{+}+\Delta_{TW} , $ 
where 
$\lambda^{+}_{bulk}$ is the MP bulk edge $\Delta_{TW}$ are the associated finite-size Tracy Widom (TW) fluctuations. 
This procedure detects \emph{any} anomaly in the matrix weights that produce spuriously large eigenvalues. 
It is implemented in \WW (using the randomize option), which was used to generate the plots in Figure~\ref{fig:scale-shape}.

See Figure~\ref{fig:scale-shape-a}, which displays the (log)-ESD of a \Typical SOTA NN layer $\mathbf{X}$ (green), i.e., on a log-linear scale, along with the (log)-ESD of that layer after randomizing it element-wise (red).
The two ESDs differ substantially:
the ESD of the original weight matrix  $\mathbf{W}$ (green) is very HT, whereas 
the ESD of the randomized weight matrix  $\mathbf{W}^{rand}$ (red) is an MP (and as predicted by the MP \RMT).
The orange line corresponds to the maximum eigenvalue of the randomized ESD.
Note that it is at the MP bulk edge of the red plot, indicating that this ESD is not affected by unusually large elements or other weight anomalies.
Here, we say that the ESD of $\mathbf{X}$ is HT, and that $\mathbf{W}$ is not HT element-wise.
\HTSR says this layer is well trained.

Contrast this with Figure~\ref{fig:scale-shape-b}, which displays the (log)-ESD of a NN layer with a \CorrelationTrap.
The ESD of $\mathbf{X}$ (green) is weakly HT, but it looks nothing like the ESD in Figure~\ref{fig:scale-shape-a}.
In fact, it looks very much like the ESD of the randomized weight matrix  $\mathbf{W}^{rand}$ (red), except for a small shelf on the right. 
The orange line again corresponds to the maximum eigenvalue of the randomized ESD, and this is just past this shelf.
Relative to the randomized ESD, this line depicts (an) unusually large element(s)---or, equivalently, a rank-1 perturbation of $\mathbf{W}^{rand}$.
By a \CorrelationTrap, we mean that some anomaly in the elements of $\mathbf{W}$ tends to ``trap'' the ESD of $\mathbf{X}$, concentrating the correlations in $\mathbf{X}$ into the small shelf of density around the orange line. 
\HTSR says this layer is not well-trained because it does not have a good PL fit.




\subsubsection{Over-Regularization}
\label{sxn:underfitting}

The second way we identify $\mathbf{W}$ as \ATypical is when it has an anomalously large variance $(\sigma^2(\WMAT))$.

The \SETOL theory -- a single-layer theory of learning -- casts the training 
of a NN layer in terms of how the correlations concentrate into the layer  \EffectiveCorrelationSpace (\ECS),
and becomes exact when the \TRACELOG condition is satisfied.
Analogously, the \HTSR theory -- a single layer \Phenomenology of learning -- casts training
of an N layer by fitting its ESD to a PL, and noting that the PL exponent $\alpha$ measures
the amount of implicit regularization in the layer.
Comparing the two approaches, we see that smaller $\alpha$ corresponds to the correlations
concentrating into a low-rank~\ECS.  In general, and likewise, the more the weight matrix
correlations concentrate  into a low-rank~\ECS, the better the layer has been regularized.
A natural question arises then, namely, can a layer be \emph{\OverRegularized} and
can we detect this?
and in large,
Empirically, we do indeed observe that over the course of training, $\alpha$ decreases, (See Figure~\ref{fig:mlp3-FC1-alpha-overloaded} 
(a), Section~\ref{sxn:hysteresis_effect},) and that the models predictions are concentrated into the~\ECS, 
(See Section~\ref{sxn:empirical-effective_corr_space}). Thus, we also interpret $\alpha$ and~\ECS concentration to be 
measures of learning itself, meaning that NNs are self-regularizing~\cite{MM18_TR_JMLRversion}.

Importantly, however, the \HTSR \Phenomenology indicates that \ALPHA usually lies in the Fat-Tailed Universality class,
such that $\alpha\in [2,6]$.  When $\alpha <2$, the ESD is Very Heavy Tailed (VHT), and, also,
this indicates that $\mathbf{W}$ has an anomalously large variance.  That is,  $\mathbf{W}$ is  \ATypical.
Occasionally, but very infrequently, we do observe $\alpha<2$, and in large, production quality models
(like Llama).
Interestingly, we also observe that, frequently, when the \HTSR $\alpha<2$, the \SETOL \TRACELOG
condition holds fairly well.  This is further explored in Section~\ref{sxn:empirical}

\begin{figure}[ht!]
    \centering
    \includegraphics[width=15cm]{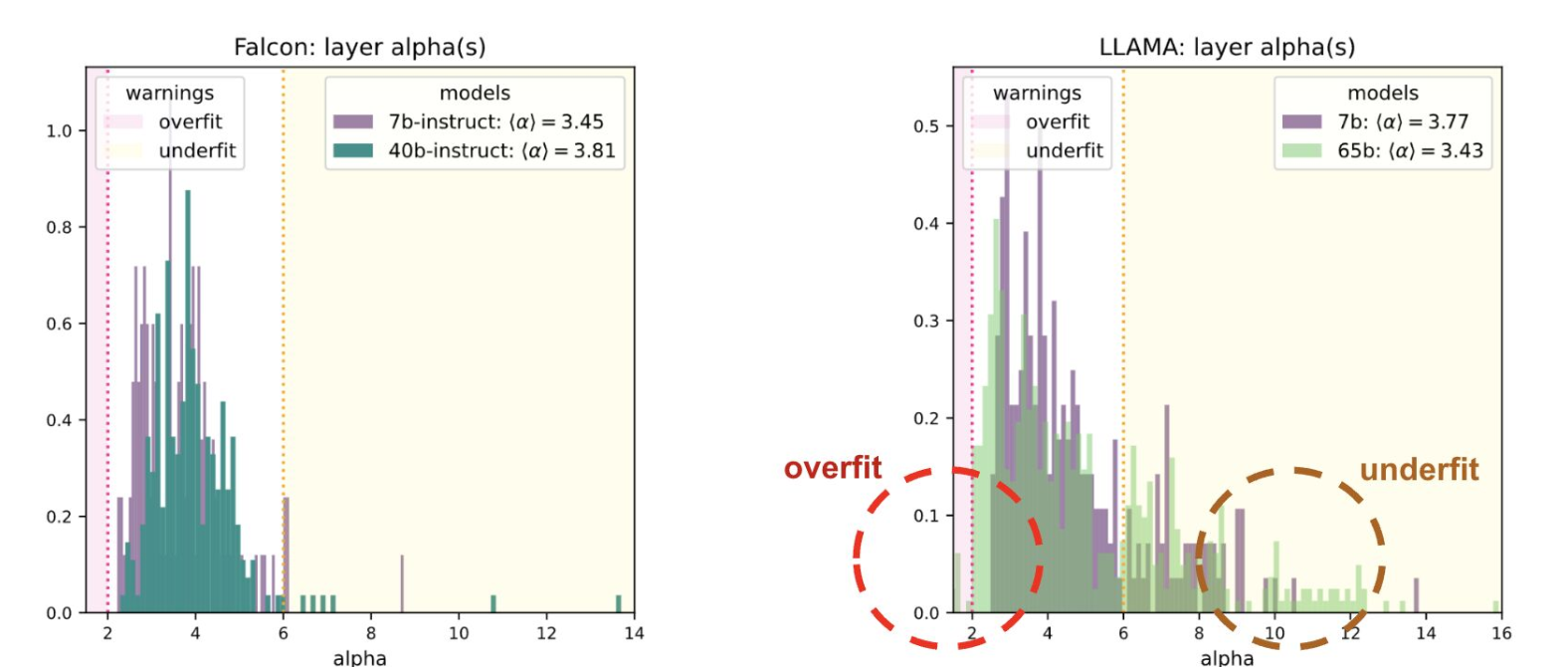}
    \caption{Falcon vs Llama}
   \label{fig:falcon_vs_llama}
\end{figure}

We have applied the \WW tool to have examined dozens
of modern, very large NNs; of particular interest are the so-called Large Language Models
(LLMs) that have revolutionized the field of AI.  To that end, in Figure.~\ref{fig:falcon_vs_llama},
we present the \WW layer \ALPHA metrics for the Falcon-40b and the Llama-65b LLMs.
\footnote{Similar results are found for the larger, more modern Llama models,
and can be found on the ~\WW website\cite{WW}}

For the Falcon-40b model, all of the layer \ALPHA range between $\alpha\in[2,6]$,
and therefore lie in the Fat-Tailed Universality class (in Table~\ref{tab:Uclass})
and are well-fit.
In contrast, looking at the Llama-40b layer \ALPHA, very many have $\alpha>6$,
indicating these layers are under-fit, and while a few have $\alpha <2$, suggesting
these are over-fit.  Finally, there are more layers with $\alpha~\sim2$ in Llama-65
vs Falcon-40b.

The observations on Llama-2 suggest that the layers with $\alpha\le 2$
are compensating for the layers with $\alpha>6$, and yielding suboptimal
performance for the Llama-65b architecture.
Based on these observations, we hypothesize that, in a multi-layer-perceptron (MLP),
when one layer does not or can not learn, then other layers will
have to compensate, and will be overloaded with the training data,
leading to $\alpha<2$, and even the \TRACELOG condition $\Delta \lambda_{min} < 0$.




In Section~\ref{sxn:hysteresis_effect}, we will test this hypothesis.
By reducing the trainable parameters in a small MLP, we can simulate the situation seen above in the Llama-65b model,
and observe the formation of a Very Heavy Tailed (VHT) ESD in the dominating layer weight matrix.
Overloading results from having too few parameters for the complexity of the task. Adding more data increases 
the load up to the total complexity of the task itself.

Moreover, we will also argue that in our experiments,
the model enters a kind of glassy meta-stable  phase, similar to the kinds of phases predicted by
classic \STATMECH theories of learning~\cite{SST92}
(described below).
Section~\ref{sxn:hysteresis_effect} will explore how far we can push the analogy of glassy systems in our experiments to 
stress test the \SETOL approach. In particular, we will see effects such as the slowing down of its dynamics, leading to a kind 
of hysteresis, specific to the under-parameterized regime.

\newpage
\section{Statistical Mechanics of Generalization (\SMOG)}
\label{sxn:SMOG_main}

In this section, we review the \STATMECH approach to learning: 
both to understand how it is usually applied in \StatisticalMechanicsOfGeneralization (\SMOG) theory; 
and to understand how our \SemiEmpirical approach in \SETOL is similar to and different from the traditional approach.
We will also obtain an expression for the \GeneralizationAccuracy (or \ModelQuality $\Q^{ST}$) for the
classic \StudentTeacher (ST) model of the Linear \Perceptron (in the AA, and at high-T),
as described in \cite{SST90,SST92}.
In Section~\ref{sxn:matgen}, we will generalize this to a \LayerQuality metric, $\Q$, for a layer in a general Multi-Layer Perceptron (MLP),
i.e., $\Q^{ST}\rightarrow\Q$, so that $\Q$ can then be expressed in terms of the ESD of the NN layer.

\paragraph{Outline.}
Here is an outline of this section.
\begin{itemize} 
\item
  \textbf{Approaches to the \SMOG.}
  In Section~\ref{sxn:trad_smog}, we explain the mapping from the \STATMECH theory of disordered systems
    to the \STATMECH theory of NN learning (\SMOG); and how our \SemiEmpirical approach (\SETOL)
    is similar to and different from the traditional approach.

  \item
      \textbf{Mathematical Preliminaries.}
    In Section~\ref{sxn:mathP}, we review the mathematical details of \STATMECH, providing definitions
      and detailed derivations of quantities and expressions necessary later.  

    \item
      \textbf{Student-Teacher Model.}
      In Section~\ref{sxn:SMOG_main-student_teacher}, we discuss the setup of the \StudentTeacher (ST) model
      as a general means to estimate the \AverageGeneralizationError empirically.
      First, in subsection~\ref{sxn:ST_OP_setup}, we describe the ST setup with an operational analogy.
       Then, in subsection~\ref{sxn:SMOG_main-st_av}, we derive the (new) result for the
    ST \ModelQuality, $\Q^{ST}$, using the setup of the classic (ST) model for the
    \GeneralizationError (and accuracy) of the \Perceptron (in the AA, and at high-T).
\end{itemize} 

\noindent
Additional information can be found in the Appendix.
\begin{itemize} 
\item
  \textbf{Symbols and Equations.}
  In Section~\ref{sxn:appendix_A}, we summarize the important symbols and key results,
         including the dimensions of the vectors and matrices, different notations for energies,
         and key equations.
   \item
   \textbf{Summary of the \SMOG.}
         In Section~\ref{sxn:summary_sst92}, we provide a more detailed analysis of the results we derive in Section~\ref{sxn:SMOG_main-st_av}.
         In particular, in Section~\ref{app:st-gen-err-annealed-ham}, we repeat the derivations of the ST \GeneralizationError $\AVGSTGE$ and related quantities (from Section~\ref{sxn:mathP}),
         using more concrete notation to align with \cite{SST90, SST92}; and
         in Section~\ref{sxn:appendix_Gan}, we use this to derive the matrix-generalization of the ST \EffectivePotential $\EPSLR$ (as well as the normalization for the weight matrices, necessary for later).
\end{itemize} 

\subsection{\STATMECH: the \SMOG approach and the \SETOL approach} 

\label{sxn:trad_smog}

In this subsection, we review the basic \STATMECH setup necessary to understand \SMOG theory as well as \SETOL.
This theory was developed in the 1980s and 1990s~\cite{SST90,SST92,Gardner_1985,Gardner_1988,engel2001statistical,EB01_BOOK}.

\begin{table}[t] 
\centering
\renewcommand{\arraystretch}{1.15} 
\begin{tabular}{c|c}
  \textbf{Statistical Physics} & \textbf{Neural Network Learning}                      \\ \hline
  Gaussian field variables     & Gaussian i.i.d (idealized) data  $\NDXI\in\mathcal{D}$            \\ \hline
  State Configuration          & Trained / Learned weights $\WVEC$                     \\ \hline
  State Energy Difference      & Training and Generalization Errors  $\AVGTE, \AVGGE$  \\ \hline
  Temperature                  & Amount of regularization present during training $T$       \\ \hline
  \AnnealedApproximation       & Average over data $\NDXI$ first, then weights $\WVEC$.                          \\ \hline
  \ThermalAverage              & Expectation w.r.t. the distribution of trained models \\ \hline
  \FreeEnergy                  & Generating function for the error(s) $F$             \\ \hline
\end{tabular}
\caption{Mapping between states and energies of a physical system and parameters of the learning process of a neural network.}
\label{table:statMech_to_NNs}
\end{table}

\begin{table}[t] 
\centering
\renewcommand{\arraystretch}{1.15} 
\begin{tabular}{c|c}
  \textbf{\SETOL Terminology} & \textbf{Explanation}                      \\ \hline
  \ModelQuality $\Q$          & \makecell{Generalization accuracy, \\in the AA and at high-T }      \\ \hline
  \LayerQuality $\Q$          & \makecell{Layer contribution to the accuracy, \\in the AA and at high-T}       \\ \hline
  \LayerQualitySquared $\QT$ &  \LayerQuality squared, used for technical reasons \\ \hline
  \Quality \GeneratingFunction $\Gamma_{\Q}, \Gamma_{\QT}$   & Generating function for \Quality    \\ \hline
  \AnnealedHamiltonian $H^{an}$                & \makecell{Energy function, \\for errors or accuracies}             \\ \hline
  \EffectiveHamiltonian $H^{eff}$     & \makecell{Exact energy function, \\but restricted to a low-rank subspace}      \\ \hline
\end{tabular}
\caption{Additional terminology introduced for the \SETOL.  
  Notice that the \Quality \GeneratingFunction $\Gamma$ is simply one minus the \FreeEnergy, $\Gamma:=1-F$,
but it is introduced because sign convention for the \FreeEnergy is always decreasing with the error.
  In contrast, we define the Hamiltonian in terms of the model error or accuracy, depending on the context.
}
\label{table:SETOL_terminology}
\end{table}

\paragraph{Traditional \SMOG theory.}
In traditional \SMOG theory, one maps the learning process of a NN to the states and energies of a physical system.
The mapping is given in Table~\ref{table:statMech_to_NNs}.
\SMOG theory models the SGD training of a \emph{\Perceptron} on the data, $\NDX$, to learn the optimal weights, $\WVEC$, as a Langevin process.%
\footnote{Typically, we have no guarantee of the true equilibrium in a high‐dim nonconvex landscape; so, when the \emph{\ThermodynamicLimit} exists,
the Langevin process converges or relaxes to the thermodynamic equilibrium.} The power of the \STATMECH approach comes from the fact that the core concept of \ThermalAverages corresponds to taking the expectation of a given quantity only \emph{over the set of trained models}, as opposed to uniformly over all possible models (or in a worst-case sense, over all possible models in a model class).
This capability is particularly compelling in light of the \STATMECH capacity to characterize disordered 
systems with complex non-convex energy landscape (which can even be \emph{\Glassy}, characterized by a
highly non-convex landscape~\cite{SST92, STS90, engel2001statistical},
and recognized classically by a slowing down of the training dynamics~\cite{gutfreund1985spin}).
Thus, concepts such as training and \GeneralizationError arise naturally from integrals that are familiar to \STATMECH; 
and theoretical quantities such as Load, Temperature, and \FreeEnergy also map onto useful and relevant concepts~\cite{MM17_TR}. 
The \FreeEnergy is of particular interest because it can be used as a generating function to obtain the desired \GeneralizationError and/or accuracy.
We wish to understand how to compute the associated thermodynamic quantities such as the expected value of the
various forms of the \AverageGeneralizationError $(\AVGGE)$, \PartitionFunction $(Z)$, and the
\FreeEnergy $(F)$ and other \GeneratingFunctions~$(\Gamma)$.

\paragraph{The \StudentTeacher model.}
We seek to compute and/or estimate the \AverageGeneralizationAccuracy for a \emph{fixed} \Teacher Perceptron $T$
by averaging over an ensemble of \Student $S$ \Perceptrons, in the \AnnealedApproximation (AA), and at
\HighTemperature (high-T); we call this ST \ModelQuality, and denote it $\Q^{ST}$.
In Section~\ref{sxn:matgen}, we generalize $\Q^{ST}$ to an
arbitrary layer in a \MultiLayerPerceptron, giving a \LayerQuality, i.e., $\Q^{ST}\rightarrow\Q$.
This formulation of the ST problem differs from the classic approach~\cite{SST92,engel2001statistical} in that
we treat the \Teacher as input to the theory rather than a random vector $\TVEC$
and that \Teacher provides the trained weights $\WVEC$ explicitly as opposed
to the labels $\Yt$.
In the simpler \Perceptron formulation, there are only a few degrees of freedom (i.e., the ST overlap $R$, the feature load, the inverse Temperature $\beta$, and the number of training examples $\ND$), whereas in the matrix generalization,
the structure of the empirical \Teacher weight matrix  $\TMAT=\WMAT$ provides $N\times M$ additional degrees of freedom,
allowing for a much richer theoretical analysis.
This is one of many ways that distinguishes the current approach from previous work.
Because of this, we present both a pedagogic derivation of $\Q^{ST}$
(for a general NN in Section~\ref{sxn:mathP}, and for the ST model specifically
in the Appendix, Section~\ref{sxn:summary_sst92}), whereas in
Section~\ref{sxn:SMOG_main-st_av} we provide a simple derivation, assuming
the \AnnealedApproximation and \HighTemperature at all times.

\paragraph{Towards a \SemiEmpirical Theory.}
In the \SETOL approach to \STATMECH, we want a matrix generalization of the \StudentTeacher \ModelQuality, $\Q^{ST}$, for a single \LayerQuality
$\Q\sim\Q^{NN}_{L}$ in an arbitrary \MultiLayerPerceptron (MLP).
This matrix generalization is a relatively straightforward extension of the classical (i.e., for a vector \Teacher) \SMOG ST \ModelQuality (but our \SETOL approach will use it in a conceptually new way).

In our matrix generalization, the \Teacher vector $\TVEC$ is replaced by a \Teacher matrix $\TMAT$ (i.e., $\TVEC\rightarrow\TMAT$); 
and, in our \SETOL approach, $\TMAT$ represents actual (pre-)trained NN weight matrix (i.e., $\TMAT\simeq\WMAT$) that has been trained on real-world, strongly correlated data $\ADD$.
As such, for our \SETOL theory, we seek an expression that can be parameterized by the \Teacher, and in particular by the ESD of the \Teacher.
This is what makes the basic method \SemiEmpirical: 
we make a modeling assumption that the \Teacher has the same limiting spectral distribution as the \Student, 
and hence the same PL fit parameter $\alpha$. 
This is all done with the understanding that later we will augment
(and hopefully ``correct'') our mathematical formulations with 
phenomenological parameters fit from experimental data.
To make the \SemiEmpirical method a \SemiEmpirical \emph{Theory}, 
we not only seek to parameterize our model; but we also try to understand 
how to derive \HTSR empirical metrics, such as $\ALPHA$ and $\ALPHAHAT$,
how they arise from this formalism, 
how they are related to the correlations in the data, and 
why they are transferable and exhibit \Universality.
Importantly, we do not just seek a method with adjustable parameters, but rather formulate the theory using
techniques similar to those to explain the origins of Quantum Chemisty \SemiEmpirical methods, resulting in formal
expressions that resemble a renormalized Self-Energy
and/or \EffectiveHamiltonian from such approaches
~\cite{Martin1994,Martin1996_CPL,MartinFreed1996, Brandow1965, freed1977, Freed1983, PhysRevLett.69.800}.
This gives what we call a \SemiEmpirical \emph{Theory}.

\subsection{Mathematical Preliminaries of Statistical Mechanics}
\label{sxn:mathP}


\paragraph{SubSection Roadmap}
Briefly, in the following subsection,
we start by defining an arbitrary NN model, with weights $(\WVEC)$.
Then, we explain the difference between using real-world $(\XVEC)$ and random data $(\XI)$.
This lets us define an energy error function, $\DETOP$,
the error the NN makes on the data.
We then explain how to take different kinds of \emph{\Thermodynamic} averages of the data,
including \emph{Sample} and \emph{\ThermalAverages} and the implications,
and the difference between computing errors and accuracy.
Next, we define the \emph{\FreeEnergy} $(F)$ for the error(s), and the \emph{\GeneratingFunction}  $(\Gamma)$
for the accuracy and/or quality.
From here, we explain the \emph{\AnnealedApproximation} (AA) and
how to define the  \emph{\AnnealedHamiltonian}, $\GAN$, a crucial expression
that will be the starting point later for our matrix model.
In the AA, $\GAN$ simplies to  $\HANHT=\EPSLw$, where $\EPSLw$ is an \EffectivePotential
that depends only on the weights $\WVEC$.
Likewise, we can define the \SelfOverlap, $\ETAw:=1-\EPSLw$, which is useful for
obtaining the \Quality.
We show how to obtain the \emph{Average Training and Generalization Errors} $\AVGTE$, $\AVGGE$
using the \STATMECH formalism, which defines them in terms of partial derivatives of the \FreeEnergy $(F)$.
Doing this, we show that in the AA and at high-T they are equivalent,
$[\AVGSTTE]^{an,hT}=[\AVGSTGE]^{an,hT}$,
and can both be expressed as a \ThermalAverage over all Students, as a function 
of the \Teacher, as $[\AVGSTGE]^{an,hT}=\THRMAVG{\GANHTR}=\THRMAVG{\EPSL(R)}$.
Note that these averages are obtained by using the \FreeEnergy as a \GeneratingFunction.
We then explain how to obtain the \ModelQuality as partial derivatives of a
\emph{\GeneratingFunction} $(\Gamma_{\Q})$.
We then discuss the more advanced techniques, the
\emph{\LargeN Approximation} and the \emph{\SaddlePointApproximation} (SPA),
which will be used extensively later.
Finally, we introduce  HCIZ integrals, which will be necessary to evaluate the matrix-generalized
form of $\Gamma_{\Q}$ to obtain the final result.

In this subsection, we will compare and contrast several types of averages, energies and other \Thermodynamic concepts we will encounter. 
\begin{enumerate}[label=4.2.\arabic*]
\item
  \textbf{Setup.}
  In Section~\ref{sxn:mathP_setup}, 
we will start by describing the basic setup of the problem, including the distinction between the actual training process and how we model the training process.

 \item
 \textbf{BraKets, Expected Values, and Thermal Averages.}
In Section~\ref{sxn:mathP_averages}, 
we will describe our use of physics BraKet notation for both Sample Averages (over the data $\XVEC$) and Thermal Averages (over the weights $\WVEC$) and ---in particular, under the \AnnealedApproximation (AA) and in the \HighTemperature (High-T) limit---showing how they relate to each other and to the notion of \Replica Averages.
\item
  \textbf{Free Energies and \GeneratingFunctions.} 
In Section~\ref{sxn:mathP_free_energies}, 
we will make a connection between these different averaging notions and \FreeEnergies and \GeneratingFunctions, showing how they relate to each other.

\item
  \textbf{The Annealed Approximation (AA) and the High-Temperature Approximation (high-T).}
  In Section~\ref{sxn:mathP_annealed}, we explain the \AnnealedApproximation, the \HighTemperature approximation,
  and the Thermodynamic \LargeN limit and the Saddle Point Approximation (SPA).
  We also introduce the \Quality \GeneratingFunction $\Gamma_{\Q}$%
  \item
    \textbf{Average Training and Generalization Errors and their \GeneratingFunctions.}
  In Section~\ref{sxn:mathP_errors}, we will show how to compute the Average Training and Test/Generalization Errors $\AVGTE, \AVGGE$
using the \FreeEnergy as a \GeneratingFunction, and how these errors are related to each other in the AA and High-T limit. 
\item 
  \textbf{The Thermodynamic limit.}
  In Section~\ref{sxn:largeN_and_SPA}, we discuss the \LargeN \ThermodynamicLimit and the \SaddlePointApproximation (SPA),
  along with the concepts of \SelfAveraging and \WickRotations.
\item 
  \textbf{From the ST Perceptron to a Matrix Model.} 
  Finally, in Section~\ref{sxn:from_vectors}, we introduce the physical concepts necessary to understand the \SETOL matrix model, including the \WideLayer \LargeN limit and the notion of \SizeConsistency.  We then describe how to obtain the \LayerQuality $\Q$ from a matrix-generalized \ThermalAverage over random matrices, called an HCIZ integral.
\end{enumerate}
The various symbols and other important results are summarized in the Appendix~\ref{sxn:appendix_A}

\subsubsection{Setup}
\label{sxn:mathP_setup}

In formulating \SETOL, 
we want a methodology to estimate quantities such as the generalization error and/or accuracy that does not rely on traditional methods that split the data into training and test sets because \SETOL makes \emph{data independent} predictions.
To accomplish this, we idealize the empirical data distribution as Gaussian fields, and we will use \STATMECH to construct quantities (basically, free energies or generating functions) so that we can compute training/testing errors by taking appropriate derivatives of these quantities.

In more detail, we imagine training a NN on $\ND$ training data instances, $\XVEC_{\mu}$, which are $m$-dimensional vectors,
with labels $y_{\mu}$, chosen from a large but finite-size training data set $\ADD$.
The goal of training a perceptron (or, later, a NN) is to learn the $m$ weights of the vector $\WVEC$ (or, later, a weight matrix $\WMAT$) by running a form of stochastic gradient descent (SGD) to minimizing a loss function $\mathcal{L}$ ($\ell_2$, cross-entropy, etc.). 
We want to approximate the actual network’s learning dynamics by an analytically tractable ensemble
so that we can then obtain 
analytic expressions for the \emph{\FreeEnergy} and \emph{\GeneratingFunction} we then use to compute
Thermodynamic averages such as the \emph{\AverageGeneralizationError} $(\AVGGE)$ and
\emph{\ModelQuality} $(\Q)$ (which is our approximation to the \emph{\AverageGeneralizationAccuracy}).

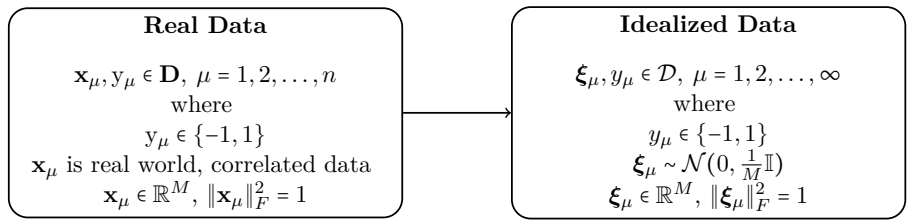
\begin{figure}[t] 
    \centering
\resizebox{0.75\textwidth}{!}{
\begin{tikzpicture}[
     thick, 
    rectnode/.style={rectangle, draw=black, thick, minimum width=6cm, minimum height=2.5cm, rounded corners=0.3cm}, 
    -> 
]

\node[rectnode] (realdata) at (0,0) {%
    \begin{minipage}{6cm}
        \centering
        \textbf{Real Data} \\
        \vspace{0.3cm}
        $\mathbf{x_\mu}, \mathrm{y}_\mu \in \mathbf{D},\;\mu=1, 2, \ldots, n$\\
        $\text{where }$ \\
        $\mathrm{y}_\mu \in \{-1, 1\}$ \\
        $\mathbf{x_\mu}\text{ is real world, correlated data }$ \\
        $\mathbf{x_\mu} \in \mathbb{R}^{M}$, $\Vert\mathbf{x_\mu}\Vert^{2}_{F}=1$
    \end{minipage}
};

\node[rectnode] (modeldata) at (8,0) {%
    \begin{minipage}{6cm}
        \centering
        \textbf{Idealized Data} \\
        \vspace{0.3cm}
        $\boldsymbol{\xi}_\mu, y_\mu \in \mathbf{\mathcal{D}},\;\mu=1, 2, \ldots, \infty$ \\
        $\text{where }$ \\
        $y_\mu \in \{-1, 1\}$ \\
        $\boldsymbol{\xi}_{\mu} \sim \mathcal{N}(0, \tfrac{1}{M} \mathbb{I})$ \\
        $\boldsymbol{\xi}_{\mu}\in \mathbb{R}^{M}$, $\Vert\boldsymbol{\xi}_{\mu}\Vert^{2}_{F}=1$
    \end{minipage}
};

\draw[->] (realdata) -- (modeldata);

\end{tikzpicture}
}
\caption{Mapping from a fixed set of $n$
  real-world, correlated data instances $[\mathbf{x},\mathrm{y}]\in\mathbf{D}$
  to an uncorrelated, random model of idealized data   $[\boldsymbol{\xi}, y]\in\mathbf{\mathcal{D}}$, drawn from a Gaussian i.i.d. distribution.
}
    \label{fig:data_mapping}
\end{figure}

\paragraph{Counting Samples and Features.}
We let the number of training samples be $\ND$ and the dimension 
(i.e., number of features) for each sample be $m$.
When we move to \SETOL matrix-model, there will still be $\ND$ training examples, but treated implicitly.
The $m$-dimensional weight vector $\WVEC$ will become an $N \times M$ weight matrix, $\WMAT$, $(N\ge M)$
described as having $N$ number of $M$-dimensional (input) feature vectors.   Table~\ref{tab:dim_notation} summarizes these conventions.
\begin{table}[H]
\centering
\begin{tabular}{l|l|l}
\toprule
 \textbf{Definition} & \textbf{Vector} & \textbf{Matrix} \\
\midrule
 Total number of data samples used in training. & $\ND$ & $\ND$ \\
 Number of features per training sample (Input dimension). & $m$ & $M$ \\
 Dimension of layer output (Output dimension)  & $1$ & $N$ \\
 Number of free parameters (in $R$) & $1$ & $M(M-1)/2$ \\
 Energy scaling & $\ND$ & $\ND\times N\times M$ \\
\bottomrule
\end{tabular}
\caption{In the ST \Perceptron \emph{vector} model, lowercase $m$ is the dimension of the weight vector (total parameters), which is also the number of features per sample. In the vector case, there is one free parameter -- the overlap $R$ (or angle $\theta$) between student and teacher. In the \SETOL \emph{matrix} model,  uppercase $N$ and $M$ are the input and output dimensions of the weight matrix, and the
matrix the overlap $\OVERLAP$, being an $M\times M$ symmetric matrix, has $M(M-1)/2$ free parameters.
The Energy scales as $\ND$ in the ST \Perceptron model, and as $\ND\times N\times M$ in the \SETOL matrix model.
}
\label{tab:dim_notation}
\end{table}

\paragraph{Actual and Idealized Data and Energies.}

Consider a large set $\ND$ of actual, real-world data, 
\begin{align}
  \label{eqn:x}
  \left( \XVEC_{\mu}, \MY_{\mu} \right) \in\ADD,\;\mu=1, \cdots, \ND,
\end{align}
where 
$\XVEC_{\mu}\in\mathbb{R}^{m}$ is an $m$-dimensional real vector, 
$\MY_{\mu}$ is a binary label taking values $\{-1,1\}$, and 
$\ADD$ denotes the finite-size dataset.
WLOG, we assume that $\XVEC_{\mu}$ is normalized such that the norm squared is unity:
\begin{equation}
  \label{eqn:XVEC_norm}
  \Vert\XVEC_{\mu}\Vert^{2} := \sum_{i=1}^{m} \XVEC_{\mu,i}^{2}=1
\end{equation}
We call $\XVEC^{\ND}$ an $\ND$-sized sample (of the training data instances $\XVEC$) from $\ADD$. 

We define model errors
as an energy $\DEL$, the difference squared between the \Student's and \Teacher's output.
Smaller energies correspond to smaller errors and therefore better models.
For example, for the mean-squared-error (MSE) loss, one has
\begin{align}
  \label{eqn:DEy}
  \DEL(\WVEC,\XVEC_{\mu},\MY_{\mu}):= (\MY_{\mu}-\NNOUT(\WVEC,\XVEC_{\mu}))^{2}  ,
\end{align}
where $\NNOUT(\WVEC,\XVEC_{\mu})$ is output prediction of the NN, as in \EQN~\ref{eqn:dnn_energy}.
\footnote{Note that the Energy Landscape function $\NNOUT$ is nonlinearly related to $\DEL$. We treat $\NNOUT$ and $\DEL$ as being the energy states of two separate, but interacting, thermodynamic systems, and the loss function $\mathcal{L}$ is a transfer function that relates them.}

We assume that there is a real-world training process that generates a \Teacher model $(T)$, trained on a particular dataset $\ADD$, and we seek a theory for the \Quality (or generalization accuracy) of this model.
 To estimate quantities such as the generalization error or generalization accuracy, we will adopt an approach that starts off by assuming idealized Gaussian data $\MDD$, 
 but ends with  the NN with a parametric model that we will fit with a \SemiEmpirical procedure (described later).
 To that end, in the theoretical setup,
the replacement scheme is,
\begin{align}
\label{eqn:model_real_world_expt}
  \ADD \rightarrow \MDD,\;\;\XVEC_{\mu} \rightarrow \XImu,\;\;  \Ymu \rightarrow y_{\mu}  ,
\end{align}
where we denote the model training and/or test data instances as $(\XI,y)$ 
such that
\begin{align}
    \label{eqn:xi}
  \left(\XI_{\mu}, y_{\mu} \right) \sim \MDD,\;\mu=1, \cdots, \infty  .
\end{align}
Here, $\XImu\in\mathbb{R}^{m}$ is a random vector (i.e., an $m$-dimensional random variable), sampled from an i.i.d $m$-dimensional joint distribution $\MDD$ of both the features $\XI_{\mu}$, which are Gaussian, and the labels $y_{\mu}$, which are binary NN outputs.

\subsubsection{BraKets, Expected Values, and Thermal Averages}
\label{sxn:mathP_averages}
Given the setup from Section~\ref{sxn:mathP_setup},
we will treat the total energy $\DETOPX$ as a sum over some $\ND$-size data set $\NDX$.
We can write the \TotalDataSampleError, using an overloaded operator notation, as
\begin{align}
  \label{eqn:detopxy}
  \DETOPXY :=\sum_{\mu=1}^{\ND}\DEL(\WVEC,\XVEC_{\mu}, \MY_{\mu})  ,
\end{align}
where the superscript $\ND$ in $\DETOPXY$ indicates this is a sum over the entire set of $\ND$ pairs $[\NDX, \MY^{\ND}]$.
 We should keep in mind that this depends on the specific set of $\ND$ data pairs $[(\XVEC_{\mu},\MY_{\mu})\in\ADD\;|\;\mu=1,\cdots,n]$.
Here, we treat 
the labels $\MY$ as
\emph{implicit} in $\DETOPX$.
We will therefore drop the $\MY_{\mu}$ and $y^{\ND}$ symbols, 
and simply write this total error / energy difference as
\begin{align}
  \label{eqn:detox_FIXLATER}
  \DETOPX :=\sum_{\mu=1}^{\ND}\DEL(\WVEC,\XVEC_{\mu})  ,
\end{align}
which is now a function of the entire set of $\ND$ vectors $[\NDX]$.%
\footnote{In the classic Student Teacher model, the labels  $\MY^{\ND}$ represent the Teacher outputs and are effectively treated as either uniform random variables to be averaged over later, or as the outputs of an optimal Teacher. In this work, the Teacher weights $\WVEC$, not labels, are input (i.e., a fixed ESD, later) so we can drop the labels.}
This operator notation will prove useful later in Section~\ref{sxn:SMOG_main-st_av}
(see \EQN~\ref{eqn:DE_L}) and in Appendix~\ref{sxn:summary_sst92}.

We will not, however, work directly with samples and sample averages.
Instead, we will model them.
To that end, we need to estimate them with a theoretical approach. For example, we can write the \TotalDataSampleError 
in terms of our random data variables $\NDXI$, written formally as
\begin{align}
\label{eqn:tdse}
\DETOPXI := \sum_{\mu=1}^{\ND}\DETmu ,
\end{align}
but to evaluate this we need to take an integral and/or \ExpectedValue over the data sample $\NDXIn$.

\paragraph{Expected Values.}

We need to compute various sums and integrals, sampling from a \emph{idealized} $\MDD$ for the \emph{actual} (i.e, real-world) data distribution $\ADD$,  over $\ND$-sized data samples (or data sets), and also over distributions of weights ($\WVEC, \SVEC$) and weight matrices ($\WMAT, \SMAT, \AMAT, \cdots$).
This will frequently (but not always) be defined as more familiar \ExpectedValues.
We will denote \ExpectedValues using physics \BraKet notion.
Importantly, we use the term \ExpectedValue in the physics sense, and BraKets will denote an un-normalized sum or integral;
when the quantity is to be normalized, we will denote the normalization explicitly.
For example, given a function $f(\XI)$, we write the BraKet integral as:
\begin{align}
 \label{eqn:EuT}
 \langle f(\XI) \rangle_{\XI}:=\int d\mu(\XI) f(\XI)  .
\end{align}
We would express an $\ND$-sized sample average over $f()$ as 
\begin{align}
    \label{eqn:EuT_normalized}
    \langle f(\NDXIn) \rangle_{\AVGNDXIn} :=& \frac{1}{\ND} \int d\mu(\NDXIn) f(\NDXIn) \nonumber \\
    =& \frac{1}{\ND} \left[\prod_{\mu=1}^{\ND} \int d\XI_{\mu}P(\XI_{\mu}) \right] \left[ \sum_{\mu=1}^{\ND}f(\XI_{\mu}) \right].
\end{align}
The BraKet $\langle\cdots\rangle_{\AVGNDXIn}$ denotes an integral over an $\ND$-sized sample of idealized
Gaussian-field data $\NDXIn$, with the convention that summation over $\ND$ points and normalization $\tfrac{1}{\ND}$
appears inside the \BraKet implicitly.

For example, we treat an \ExpectedValue of the \DataSampleError , which is correlated when using real-world data $\ADD$, using the uncorrelated idealized data $\MDD$; this is specified with the following mapping:
\begin{align}
  \label{eqn:Emap}
  \frac{1}{\ND}\DETOPX \xrightarrow{\text{Expected Value}} \langle \DETOPXI \rangle_{\AVGNDXIn}  , 
\end{align}
where the BraKet $\langle\cdots\rangle_{\AVGNDXIn}$ denotes the integral over the idealized dataset $\NDXIn$ of a normalized average over the sample.
In this case, we obtain:
\begin{align}
  \langle \DETOPXI\rangle_{\AVGNDXIn}
  :=  &\frac{1}{\ND}\int d\mu(\NDXIn) \DETOPXI \nonumber \\
  = &
  \frac{1}{\ND}\int \left[ \prod_{\mu=1}^{\ND}d\XI_{\mu}P(\XI_{\mu})\right] \left[ \sum_{\mu=1}^{\ND}\DETmu \right] , \nonumber \\
  = &
  \frac{1}{\ND}\sum_{\mu=1}^{\ND}\int d\XI_{\mu}P(\XI_{\mu})\DETmu  , 
    \label{eqn:avg_data_sample_error}
\end{align}
where $P(\NDXIn)$ is a product of $\ND$ i.i.d. $m$-dimensional Gaussian distributions.
The subscript $\AVGNDXIn$ indicates this is an
\ExpectedValue of an average of an $\ND$-size \emph{sample} of ideal data, where the \ExpectedValue is taken over datasets, and the average is over data points within each sample. The third line follows because the $\ND$ samples are i.i.d.
(This is used in both Sections~\ref{sxn:SMOG_main} and~\ref{sxn:matgen}.)
(The normalization $\tfrac{1}{\ND}$ ensures the \BraKet is a proper \ExpectedValue of a sample average.)
The measure $d\mu(\NDXIn)$ 
denotes the probability measure associated with a single 
\emph{realization} of a random data sample of size $\ND$ drawn from an $m$-dimensional idealized Gaussian distribution.
A subscript $\XI$ on the Ket as $\langle\cdots\rangle_{\XI}$ indicates that the integral is over potential data points, not an average of a data sample (i.e., there would be no $1/\ND$ prefactor).

\paragraph{\SizeExtensivity and \SizeIntensivity.}
A key requirement for the \ThermodynamicLimit in \STATMECH is \emph{\SizeExtensivity}:
that physically meaningful quantities (i.e, total energies and free energies)
scale linearly with the system size or number of parameters, i.e. $m$ (or $N\times M$, below).
Extensive quantities scale with system size, intensive ones do not.
Along with this, Thermodynamic average quantities should be \emph{\SizeIntensive},
meaning that they remain independent of $\ND$ (or $N\times M$, below) as the system size increases.
In our setting, \SizeExtensivity and \SizeIntensivity underpin the so-called \LargeN limits we employ,
ensuring that macroscopic observables become independent of
microscopic fluctuations so that the system approaches equilibrium as expected.

In the theory of the \StatisticalMechanics of Generalization (\SMOG), however, one also holds the (feature) load $=\ND/m$ fixed when taking the \ThermodynamicLimit, and we will do the same~\cite{Gardner_1985, SST92, engel2001statistical,MM17_TR}. In fact, we will take load to be unity, and not treat it explicitly. \footnote{For our purposes, we only require that $n/m$ is \emph{some fixed quantity} as we take this limit. We note that $n/m = 1$ is the Double Descent Worst Case, however assuming that $n / m$ is some other constant addresses this without changing any of the results.}
Moreover,  we will use $\ND$ as the primary variable below as we are interested in how the Free Energy and other quantities change with $\ND$. That is, we use the notational convention in ~\cite{Solla2023} (as opposed to \cite{SST92}). \textbf{Consequently, for our purposes here, this means that the \ThermodynamicLimit and the `\LargeN limit in $\ND$' are effectively the same.}

As an example of \SizeExtensivity and \SizeIntensivity, 
we write the \ExpectedValue (i.e., the data-average) of \DataSampleError $\DETOPXI$ (\EQN~\ref{eqn:avg_data_sample_error})
in the \LargeN limit in $n$ as
\begin{align}
\label{eqn:longEn}
  \lim_{n\gg 1} 
  \langle \DETOPXI\rangle_{\AVGNDXIn}=
  \lim_{n\gg 1}\frac{1}{\ND}
\int \left[ \prod_{\mu=1}^{\ND}d\XI_{\mu}P(\NDXIn) \right]
\left[ \sum_{\mu=1}^{\ND}\DEL(\WVEC,\XI_{\mu}) \right].
\end{align}
Here, the notation $(n \gg 1)$ means $\ND$ grows arbitrarily large, but is not necessarily
at the limit point $(n=\infty)$. And, again, the load is fixed, so $n\gg 1 \leftrightarrow m \gg 1$.
The \TotalDataSampleError $\DETOPXI$ is \SizeExtensive, whereas the
average $\langle\DETOPXI\rangle_{\AVGNDXIn}$ is \SizeIntensive.
This limit will be implicit later when taking a \SaddlePointApproximation (see below).
\footnote{As we are working within a ``physics-level of rigor'', we take some liberties in evaluating these \LargeN limits; and we leave the formal proofs for future work.  }

The data-averaged error  $\langle \DETOPXI \rangle_{\AVGNDXIn}$ will appear frequently below.
For convenience and for compatibility with \cite{SST92}, we denote it using the symbol $\EPSLw$:
\begin{align}
 \label{eqn:epsl}
 \EPSLw:=\lim_{n\gg 1}  \langle \DETOPXI \rangle_{\AVGNDXIn} \quad \text{(\SizeIntensive)}.
\end{align}
where, by our normalization here, $\EPSLw \in [0,1]$.
The symbol $\EPSLw$ is our theoretical estimate of the sample average $\DETOPXI$ (\EQN~\ref{eqn:longEn}),
well-defined for any $\ND$.
We also call $\EPSLw$ the \emph{\EffectivePotential}, which will be made clear below.

It is also convenient to write \emph{\TotalEffectivePotential} as an Energy, 
\begin{align}
 \label{eqn:detox}
 \DETOT := \ND\EPSLw\quad \text{(\SizeExtensive)}.
\end{align}
This will only be useful when the \ThermodynamicLimit exists, and this
can be reasonably expected for the \AnnealedApproximation (AA),
which is the regime in which \SETOL will be developed.%
\footnote{We should note that, while our model training and generalization errors are always expressed as energies, an energy is not necessarily a model error. }

\paragraph{From Errors to Accuracies: The \AverageGeneralizationAccuracy, the \Quality, and the \SelfOverlap.}
We have been discussing various forms of errors.
In \SETOL, we will, however, primarily be concerned with approximating the \emph{\AverageGeneralizationAccuracy},
or, more generally, the \Quality of a NN model and/or its layers.\footnote{Technically, the \Quality will estimate the average \emph{Precision} rather than the Accuracy.
This will distinction will be clarified in the Section~\ref{sxn:SMOG_main-student_teacher}.}
The average accuracy is simply one minus the error.
To represent this,
we introduce the \emph{\SelfOverlap} $\ETA(\WVEC)$, which is defined generally as
\begin{align}
 \label{eqn:def_eta}
 \ETA(\WVEC) := 1 -\EPSLw ,\;\;\ETA(\WVEC)\in[0,1] ,
\end{align}
and which 
describes the ``overlap'' between the true and the predicted labels.
Unlike here, however, in later sections
(\ref{sxn:SMOG_main-st_av}, \ref{sxn:matgen_mlp3}, and Appendix~\ref{sxn:quality})
we will first define a data-dependent \SelfOverlap, so that we may obtain
 $\ETA(\WVEC):=\langle\ETA(\WVEC,\XI)\rangle_{\AVGNDXIn}$ directly.

\paragraph{Braket Notation.}
We will use physics \BraKet notation, $\langle\cdots\rangle$,
to denote different kinds of sums and integrals, with superscripts and subscripts,
and for \ExpectedValues (estimated theoretical averages).
We use superscripts to denote the kind of integral or average:
\begin{center}
Thermal $\langle\cdots\rangle^{\beta}$,
\Annealed $\langle\cdots\rangle^{an}$,
high-T $\langle\cdots\rangle^{hT}$,
HCIZ $\langle\cdots\rangle^{IZ}$, etc.
\end{center}
We use subscripts to emphasize the dependent variables:
\begin{center}
  weights $\langle\cdots\rangle_{\WVEC}$, $\langle\cdots\rangle_{\SVEC}$, $\langle\cdots\rangle_{\SMAT}$ \nonumber \\
    \vspace{0.33cm}  
data $\langle\cdots\rangle_{\XI},
\langle\cdots\rangle_{\NDXI},
\langle\cdots\rangle_{\AVGNDXI}$
\end{center}
When averaging over whole datasets $\NDXI$, the subscript will appear with a bar (i.e. $\AVGNDXIn$), but when just integrating over a particular dataset, no bar will appear (i.e., $\NDXIn$). 
We also reuse these symbols for other quantities, such as the $\ZANHT$, $\AVGGE^{an,hT}$, $\GAN$, etc,
but may mix-and-match subscripts and superscripts for visual clarity.

\paragraph{Sign Conventions.}
Finally, we discuss the sign conventions used.  Since errors decrease with better models,
Energies $(\DET, \DETOT, \EPSLw, \cdots)$ and Free Energies $(F)$ are minimized to obtain better models.
Likewise, since accuracies increase with better models, Qualities $(\Q, \QT, \cdots)$,
\SelfOverlap $(\ETA)$, and \Quality \GeneratingFunction $(\Gamma)$ would be maximized to obtain better models.
An exception will be Hamiltonians $(H,\mathbf{H})$, where the sign convention will depend on context.

\paragraph{Thermal Averages (over weights).}

To evaluate the expectation value of some equilibrium quantity that depends on the weights $\WVEC$ (say $\mathbb{E}^{\beta}_{\WVEC}[f(\WVEC)]$), one uses a \ThermalAverage.
By this, we mean a \emph{\BoltzmannWeightedAverage}: given a function $f(\WVEC)$,
we define the \ThermalAverage over $\WVEC$ as
\begin{align}
\label{eqn:thrmavg}
\langle f(\WVEC)\rangle_{\WVEC}^{{\beta}}:=\dfrac{1}{Z_{\ND}}\int d\mu(\WVEC) f(\WVEC)e^{-\beta \DETOT}  ,
\end{align}
where the superscript $\beta$ denotes \ThermalAverage,
$\beta=\frac{1}{T}$ is an inverse temperature, and 
$Z_{\ND}$ is the normalization term (or Partition function), defined as
\begin{align}
\label{eqn:Zwn}
Z_{\ND}:=\int d\mu(\WVEC) e^{-\beta \DETOT},
\end{align}
defined for the $\ND$-size \emph{Data Sample} $\NDXIn$.
In particular, when we want to compute the \ThermalAverage of the \emph{Total Energy} difference or Error
$\DETOT$ over $\WVEC$, we could write
\begin{align}
\label{eqn:Detot}
\langle \DETOT \rangle^{\beta}_{\WVEC}:=\dfrac{1}{Z_{\ND}}\int d\mu(\WVEC) \DETOT e^{-\beta \DETOT} .
\end{align}
Importantly, we will never calculate the average errors directly like this.
Instead, we will calculate them from partial derivatives of the \FreeEnergy $F_{\ND}$ (as shown below).
Also, we may use $\langle \cdots \rangle^{\beta}_{\NDXIn}$ to denote what looks like a \ThermalAverage over the data;
this is not essential and only used once below and can be ignored for this section.

\paragraph{Other Notation: Overbars, Superscripts and Subscripts.}
As above, we may also occasionally denote averages using the common notation for expected values, $\mathbb{E}[\cdots]$.
See Table~\ref{tab:dimensions} and~\ref{tab:symbols} in Appendix~\ref{sxn:appendix} for a list of these and other notational conventions and symbols we use.

When discussing quantities such as the \FreeEnergy $(F)$, 
training and test errors/eneries $(\mathcal{E})$, 
the \LayerQuality $(\mathcal{Q})$, etc.,
we will place a bar over the symbol (i.e., $\bar{F}$, $\bar{\mathcal{E}}$, $\Q$, etc.) when referring to
an average over the data $\ND$.
Otherwise, we will refer to these quantities as the total (averaged) energy, error, quality, etc.

\charles{Finally, in this preliminary Section, we represent the dimensions with lower case $n,m$, but elsewhere
  (and sometimes below), we will use capital $N,M$.}

Finally, when referring to the model (i.e., theoretical)
training and generalization errors, we will use the superscript $ST$ for
the average \StudentTeacher training and generalization errors, $\AVGSTTE$ and $\AVGSTGE$, respectively, and
the superscript $NN$ for the matrix-generalized NN layer average
training and generalization errors, $\AVGNNTE$ and $\AVGNNGE$, respectively.
When referring to empirical errors, we denoted these as $\AVGEMPTE$ and $\AVGEMPGE$, respectively.

\subsubsection{Free Energies and \GeneratingFunctions} 
\label{sxn:mathP_free_energies}

If one needs an average energy (or error), 
it is often easier to calculate the associated \FreeEnergy and take corresponding partial derivatives
than it is to compute that quantity directly via an expected value or \ThermalAverage.
Generally speaking, a \FreeEnergy, $F_{\ND}$, is defined in terms of a partition function $Z_{\ND}$ as
\begin{align}
\label{eqn:F}
\beta F_{\ND}:=-\ln Z_{\ND}.
\end{align}
Keep in mind that $Z_{\ND}$ may actually be a function of the idealized data $\XI$ (or some other variables),
i.e., $Z(\XI)$, but we usually don't write this explicitly.
Likewise, while  both $F_{\ND}$ and $Z_{\ND}$ depend explicitly on the system size $\ND$,
we will only include these subscripts when emphasizing this.
Also, $F_{\ND}$ has units of Energy or Temperature, so $\beta F_{\ND}=-\ln Z_{\ND}$ is a ``unitless'' \FreeEnergy.
Each model (in single-layer models) and/or layer (in multi-layer models) will have its own \PartitionFunction and associated \GeneratingFunctions.
We call $F_{\ND}$ and $Z_{\ND}$ \emph{\GeneratingFunctions} because they can be used to generate the model errors. 
That is, given an $F_{\ND}$ and/or $Z_{\ND}$, we can ``generate'' the training and generalization errors with the appropriate partial derivatives w/r.t $\beta$ and $\ND$~\cite{LTS90, Solla2023}.

From this generating function perspective, i.e., when using a generating function to compute quantities of interest, we can work with other transformations of $F_{\ND}$.
Most notably, we will consider 
\begin{equation}
    \Gamma_{\ND} = n-F_{\ND} .
\end{equation}
where we note that these Energies scale as $\ND$ for the ST vector model, but for a  matrix model they scale as $\ND\times N\times M$; see Appendix~\ref{app:st-gen-err-annealed-ham}).
The quantity $\Gamma_{\ND}$ decreases as the error increases (as opposed to $F_{\ND}$, which increases), i.e., it increases as the accuracy of quality of the model increases.
Thus, we will use it as a generating function for the model quality/accuracy.  
For average Qualities, one has
\begin{equation} 
\label{eqn:GammaBar}
 \bar{\Gamma}_{\ND}:=1-\bar{F}_{\ND}
\end{equation} 
for the model or layer under consideration (see below).

\subsubsection{The Annealed Approximation (AA) and the High-Temperature Approximation (high-T)}
\label{sxn:mathP_annealed}

In the traditional \SMOG approach, one models the (\Typical) generalization behavior of a NN
by defining and computing the \ExpectedValue of the \FreeEnergy of the model.
The full expected value  of the \FreeEnergy, $\beta F_{\ND}=-\ln Z_{\ND}$, with respect to the (idealized) data $\NDXIn$, is:
\begin{align}
\label{eqn:mm_f_bar}
  \mathbb{E}_{\NDXIn}[\beta F_{\ND}]=\beta\bar{F}_{\ND}:=-\langle \ln Z_{\ND}\rangle_{\AVGNDXIn},
\end{align}
where $\langle \ln Z_{\ND}\rangle_{\AVGNDXIn}$ means where we average over  the $\ND$ samples of the data ($\NDXIn$, of size $\ND$).  This is also called the \Quenched \FreeEnergy.
This is, however,  frequently too difficult to analyze, and doing so typically
requires a so-called Replica calculation. 

The \AnnealedApproximation (AA) is a way of taking the data-average \emph{first} and greatly
simplifies the model under study and its analysis.  The standard way to move forward is to follow the methods used
in disordered systems theory \cite{SST92, EB01_BOOK}.
The mapping is:
\begin{align*}
  &\mbox{Average over the Data }&\leftrightarrow& \quad\mbox{\AnnealedApproximation}    &\leftrightarrow& \quad\mbox{Disorder Average} \\
  &\mbox{Learning the Weights }      &\leftrightarrow& \quad\mbox{NN Optimization Algorithm} &\leftrightarrow& \quad\mbox{\ThermalAverage}  .
\end{align*}
  through the NN optimization algorithm,
  and then evaluates the training or test accuracy
  by averaging over the (actual) data $\NDX$.
  In the theoretical analysis, however, the steps are reversed.
  One first averages over the (model) data $\NDXIn$ (i.e., a multi-variate Gaussian distribution)
  so that the disorder (variability in the data) is averaged out.
  Then, a \ThermalAverage is used to model the final state of NN learning process, the learned weights.

\paragraph{The Annealed Approximation (AA).} 
Formally, the AA makes the substitution
\begin{align}
\label{eqn:AA}
-\langle\ln Z_{\ND}\rangle_{\AVGNDXIn}\approx-\ln \langle Z_{\ND}\rangle_{\AVGNDXIn}.
\end{align}
Here, we are \emph{averaging over the disorder}.
We may associate: 
\begin{eqnarray*}
    -\beta\langle\ln Z_{\ND}(\NDXIn)\rangle_{\AVGNDXIn} &: \mbox{the \Quenched \FreeEnergy} \\
    -\beta\ln \langle Z_{\ND}(\NDXIn)\rangle_{\AVGNDXIn} &: \mbox{the \Annealed \FreeEnergy}.
\end{eqnarray*}
Applying the AA amounts to applying Jensens inequality \emph{as an equality}, 
and it allows will let us interchange integrals and logarithms when computing the data average:
\begin{align}
\label{eqn:Jensens}
\tfrac{1}{\ND}\int d\mu(\NDXIn)\ln(\cdots)   
\xrightarrow {\text{AA}}
\ln\tfrac{1}{\ND}\int d\mu(\NDXIn)(\cdots)
\end{align}
This will allow us to switch the order of the data and the thermal averages, i.e.,
\begin{align}
\label{eqn:switch}
\left\langle \THRMAVGw{\cdots} \right\rangle_{\AVGNDXIn}
\xrightarrow{\text{AA}}
\THRMAVGw{\left\langle \cdots \right\rangle_{\AVGNDXIn}},
\end{align}
greatly simplifying the analysis.

The use of the AA is common in \STATMECH, as it simplifies computations considerably; and 
it is chosen when it holds exactly (if, say, $x$ is a \Typical sample from $\mathcal{D}$ and $Z_w(\XI)$ has a well-defined mean).
In contrast, there are situations in \STATMECH when the average is \ATypical, and then it one can get different results for the \Quenched versus \Annealed cases.  In a practical sense, one imagines this may occur when the data is very
noisy and/or mislabeled, and this requires a special treatment~\cite{SST92}.

\paragraph{Annealed Hamiltonian $\GAN$ and Annealed Parition Function $Z^{an}$.}

When we apply the AA (as in \EQN~\ref{eqn:Jensens}), 
we average over the data $\NDXIn$ first. 
Doing this will allow us to develop a theory in terms a (Temperature dependent) \EffectivePotential. 

Following~\cite{SST92} (see their \EQN(2.30)), we will call this average the \emph{\Annealed Hamiltonian}, $\GAN$, 
which we define as  
  \begin{align}
   \label{eqn:Gan_def}
   \beta\GAN:= - \frac{1}{\ND}\ln  \int d\mu(\NDXIn)e^{-\beta\DETOPXI}.
  \end{align}
The \AnnealedHamiltonian is a simple ``mean-field-like'' Hamiltonian for the problem.
This can be seen by noting that we can express \EQN~\ref{eqn:Gan_def} as an integral over a single data example $\XI$:
 \begin{align}
   \nonumber
   \beta\GAN &=  - \frac{1}{\ND}\ln  \int d\mu(\NDXIn)e^{-\beta\sum_{\mu=1}^{\ND}\DEL(\WVEC,\XI_{\mu})} \nonumber \\
   &=  - \ln \left[\int d\mu(\NDXIn)e^{-\beta\sum_{\mu=1}^{\ND}\DEL(\WVEC,\XI_{\mu})}\right]^{\tfrac{1}{\ND}} \nonumber \\
   &=  - \ln \left[\prod_{\mu=1}^{\ND}\int d\mu(\XI_{\mu})e^{-\beta\DEL(\WVEC,\XI_{\mu})}\right]^{\tfrac{1}{\ND}} \\ 
   \label{eqn:Gan_simplified}
   &=  - \ln  \int d\mu(\XI)e^{-\beta \DEL(\WVEC,\XI)}
 \end{align}
This will be a critical piece needed to generalize the vector-based ST \Perceptron model to the matrix-generalized ST MLP model.
In BraKet notation, \EQN~\ref{eqn:Gan_simplified} can be expressed as
\begin{eqnarray*}
    \beta\GAN:=  -\tfrac{1}{\ND}\ln\left\langle e^{-\beta\DETOPXI}\right\rangle_{\NDXIn}= 
    -\ln \left\langle e^{-\beta\DEL(\WVEC,\XI)}\right\rangle_{\XI}.
\end{eqnarray*}

Using $\GAN$, we can define the \emph{Annealed Partition Function}, $Z^{an}_{\ND}$, as
\begin{align}
  \label{eqn:Zan_def}
  \ZAN 
  &:=  \int d\mu(\WVEC) e^{-\ND\beta \GAN} \nonumber \\
  &=\int d\mu(\WVEC) \exp\left[-\ND \left(- \frac{1}{\ND}\ln  \int d\mu(\NDXIn)e^{-\beta\DETOPXI}\right)\right] \nonumber \\
  &=  \int d\mu(\WVEC) \exp\left[\ln  \int d\mu(\NDXIn)e^{-\beta\DETOPXI}\right] \\ 
  \label{eqn:Zan_simple}
  &=  \int d\mu(\WVEC)  \int d\mu(\NDXIn)e^{-\beta\DETOPXI} .
\end{align}
where the lines after the first line follow by substituting \EQN~\ref{eqn:Gan_def} into \EQN~\ref{eqn:Zan_def}.
Note also that the order of the integrals in \EQN~\ref{eqn:Zan_simple} is exactly what we expect using the AA, as in \EQN~\ref{eqn:switch}.
Also, analogously to \EQN~\ref{eqn:Gan_simplified}, we can write $\ZAN$ as the product of the $n=1$ case, $\ZAN=[Z^{an}_{1}]^{\ND}$.
Finally, we will only need the high-T version, $\HANHT$, of $\GAN$, and this will take a very simple form.
\footnote{We will derive expressions for $\EPSL(R)$ and $\AVGSTGE(R)$ in \EQN~\ref{eqn:epslR} and \EQN~\ref{eqn:AVGSTGE_R}, respectively, using relatively simple arguments.
In Appendix~\ref{app:st-gen-err-annealed-ham} we present
a more detailed derivation of $\GANR$ and $\GANHTR=\EPSL(R)$
in Appendix~\ref{sxn:appendix_Gan} we show that these derivations generalize to the matrix-generalized case, $\GANMAT$.}


\paragraph{The High-Temperature (High-T) Annealed Hamiltonian $(\HANHT(\WVEC)=\EPSLw)$ and Partition Function $(\ZANHT)$. }
In addition to the AA, we will be evaluating our models at at high-T.
Notably, the \AnnealedHamiltonian $\HAN(\WVEC)$ in \EQN~\ref{eqn:Gan_def} is a non-linear function of $\beta$; by
taking the high-T approximation, we can remove this dependence and obtain
the simpler expression that $\HAN(\WVEC)=\EPSLw$.  This greatly simplifes both
the \PartitionFunction, i.e.,  $\ZANHT$, and subsequent results (below).

To obtain the high-T result, we can write the Taylor expansion for $e^{\beta\DETOPXI}$ and keep
the first two terms:
\begin{align}
  e^{-\ND\beta\DETOPXI} \simeq \underbrace{1 - \beta\DETOPXI}_{high-T}+ (\beta\DETOPXI)^{2} + \cdots  .
\end{align}

Let us now express $\GAN$ directly in terms of $\EPSLw=\langle\DETOPXI \rangle_{\AVGNDXIn}$ (see \EQN~\ref{eqn:epsl}) as a \ThermalAverage at high-T.
To do so, let's take a high-T expansion of \EQN~\ref{eqn:Gan_def} 
by expanding the exponential to first order in $\beta$, to obtain
\begin{align}
\beta\GANHT =&  - \tfrac{1}{\ND}\ln \int d\mu(\NDXIn) [1-\beta \DETOPXI] \nonumber \\
\approx&\;   - \frac{1}{\ND} \int d\mu(\NDXIn)\beta \DETOPXI \nonumber \\
=&\;\langle \beta\DETOPXI \rangle_{\AVGNDXIn} \nonumber \\
=&\;\frac{1}{\ND}\beta(\DETOT)  \nonumber \\
=&\;\beta \EPSLw  .
\label{eqn:Gan_highT}
\end{align}
Here, we have used the AA, and the property that $\ln(1 + y) \approx y$, for $|y| \ll 1$, and the fact that $\EPSLw$ takes the form given in \EQN~\ref{eqn:epsl}.
This gives $\GANHT:=\EPSLw$, which is no longer a non-linear function of $\beta$.
Moving forward, we will assume we are taking the high-T limit like this.

Given \EQN~\ref{eqn:Gan_highT}, we can now express the \Annealed \PartitionFunction at high-T directly in terms of
the Annelaed (i.e., data-averaged) error $\EPSLw$:
\begin{align}
  \nonumber
  \label{eqn:Zanht_def}
\ZANHT :=  &\int d\mu(\WVEC) e^{-\ND\beta\GANHT} \\ 
  =  &\int d\mu(\WVEC) e^{-\ND\beta\EPSLw} 
\end{align}
If we assume that at high-T we can make this approximation, 
since we only care about the small $\beta$ results. 
This will prove very useful later when working with HCIZ integrals.

\subsubsection{Average Training and Generalization Errors and their \GeneratingFunctions}
\label{sxn:mathP_errors}
Here, we show how to derive the Average \TrainingError $\AVGTE$ and  the \AverageGeneralizationError $\AVGGE$,
in the \AnnealedApproximation (AA), and at high-Temperature (high-T), using the \FreeEnergy $F_{\ND}$ and/or the \PartitionFunction $Z_{\ND}$ as
a generating function.  
In particular, we show that, in the AA and at high-T, these errors are (approximately) equal
and equal to the \ThermalAverage of the \EffectivePotential,
$\AVGTE^{an, hT} \approx \AVGGE^{an, hT} \approx \THRMAVGw{\EPSLw}$.

\paragraph{\GeneratingFunctions for the Errors: the \STATMECH way.}
In our theory, after applying the AA, we obtain expressions where the random model data $\NDXIn$ has been integrated out. This leaves formal quantities that depend only on the weights $\WVEC$, which are the variables being learned during training.
Since we are left with a distribution over $\WVEC$, we define the training error not explicitly as an average over the training data itself, but instead in terms of how the Free Energy, $\beta F :=-\ln Z_{\ND}$, varies with $\beta$, i.e., the amount of stochasticity in the model weights.

Following ~\cite{LTS90, Solla2023},
we define the \emph{\AverageTrainingError}, in the AA,
by differentiating $\ln Z^{an}_{\ND}$ with respect to $\beta$:
\begin{align}
  \label{eqn:avgte_def}
  \AVGTE^{an}
  := -\frac{1}{\ND}\dfrac{\partial (\ln Z^{an}_{\ND})}{\partial \beta} 
  = -\frac{1}{\ND}\frac{1}{Z_{\ND}}\dfrac{\partial Z^{an}_{\ND}}{\partial \beta} .
\end{align}
This error captures how the model predictions will vary with changes in the learned
weights $\WVEC$, which implicitly describes how the changes will vary with the
training data $\NDXIn$.
Similarly, 
we define the \emph{\AverageGeneralizationError}, in the AA,
by differentiating $\ln Z^{an}_{\ND}$ with respect to $\ND$, the number of data points.
Following 
\begin{align}
  \label{eqn:avgge_def}
  \AVGGE^{an}
  := -\frac{1}{\beta}\dfrac{\partial (\ln Z^{an}_{\ND})}{\partial n}    +\frac{1}{\beta}\ln z(\beta)  
  =  -\frac{1}{\beta}\frac{1}{Z_{\ND}}\dfrac{\partial Z^{an}_{\ND}}{\partial n}
  +\frac{1}{\beta}\ln z(\beta) ,
\end{align}
where $z(\beta)$ is a constant normalization term that depends only on $\beta$ (which, moving forward, we ignore, as it only shifts the scale).
This error captures how the model’s predictions will change as more data points are introduced.

In the \ThermodynamicLimit $(n \gg 1)$, these two definitions of the error become equivalent at High-T,
and they are equal to the \ThermalAverage of the \EffectivePotential:
\begin{align}
  \AVGTE^{an, hT} = \AVGGE^{an, hT} = \THRMAVGw{\EPSLw}\;,n\gg 1 .
\end{align}
To see this, substitute \EQN~\ref{eqn:Zanht_def} into \EQN~\ref{eqn:avgte_def}, and take the partial derivative w.r.t $\beta$, to obtain
\begin{align}
  \label{eqn:avgge_anhT}
  \AVGGE^{an, hT} :=&\frac{1}{\ND}\dfrac{\partial (-\ln \ZANHT)}{\partial \beta}  \nonumber \\
   =& - \dfrac{1}{\ND}\dfrac{\partial}{\partial \beta}\ln\int d\mu(\WVEC)e^{-\ND\beta\EPSLw} \\  \nonumber
   =&  \dfrac{
              \tfrac{1}{\ND}  \int  d\mu(\WVEC) n\EPSLw e^{-\ND\beta\EPSLw} 
             }{
              \int  d\mu(\WVEC) e^{-\ND\beta\EPSLw} 
   } \nonumber \\
   =&\langle\EPSLw \rangle_{\WVEC}^{\beta} .
  \end{align}
Likeswise, if we substitute \EQN~\ref{eqn:Zanht_def} into \EQN~\ref{eqn:avgge_def}, and take the partial derivative w.r.t. $\ND$, we obtain
\begin{align}
  \label{eqn:avgte_anhT}
    \AVGGE^{an, hT}  :=& \frac{1}{\beta}\dfrac{\partial (-\ln \ZANHT)}{\partial n} \nonumber \\
    =& - \dfrac{1}{\ND}\dfrac{\partial}{\partial \beta}\ln\int d\mu(\WVEC)e^{-\ND\beta\EPSLw} \\  \nonumber
   =&  \dfrac{
              \tfrac{1}{\beta}  \int  d\mu(\WVEC) n\EPSLw e^{-\ND\beta\EPSLw} 
             }{
     \int  d\mu(\WVEC) e^{-\ND\beta\EPSLw} 
   } \nonumber \\
   =&\langle\EPSLw \rangle_{\WVEC}^{\beta} .
\end{align}
Notice that both of these results arise because of the simple expression that appears in the exponent of $\ZANHT$, namely because $-\ND\beta\HANHT(\WVEC)=\ND\beta\EPSLw$.

This equivalence reflects the fact that when the system is large enough, adding a new data example to the
training distribution is formally equivalent to adding noise, making the two errors indistinguishable.
This approach allows us to define both training and generalization errors in terms of fundamental thermodynamic quantities,
providing a simplified formal framework suitable for empirical adjustment later.

Also, note that the model data variables $\XI$ do not enter the calculation because we 
integrated them out before the calculation of \ThermalAverage.
(This illustrates the difference between taking an annealed versus a quenched average.)
Also, our sign convention is consistent with a model of NN training that \emph{minimizes} the total loss
$(\mathcal{L})$ and/or ST error, and, therefore minimizes \FreeEnergies as well.

More generally, we see that we can use the \PartitionFunction, $Z_{\ND}$,
and/or the \FreeEnergy, $\beta F_{\ND}:=-\ln Z_{\ND}$, as a \GeneratingFunction to obtain any
desired Thermodynamic average by taking the appropriate partial derivative
of the corresponding form of $\ln Z$.
In Appendix~\ref{sxn:mathP} we show how to obtain 
$\AVGTE^{an}$ and $\AVGGE^{an}$ obtain explicitly in this way using $Z^{an}$.
\subsubsection{The Quality \texorpdfstring{$(\Q)$}{Q} and its Generating Function \texorpdfstring{$(\Gamma_{\Q})$}{(Gamma Q)}}
Here, we explain how to define what we call the \Quality $\Q$, which is
defined as the \ThermalAverage of the \SelfOverlap,  $\Q:=\THRMAVGw{\ETAw}$,
and which can be obtained from the associated \emph{\Quality~\GeneratingFunction} $\Gamma_{\Q}$.
See Table~\ref{tab:intensive_quantities} for a quick summary.
\begin{table}[H]
\centering
  \begin{tabular}{@{} l  p{0.50\linewidth}  l  c @{}}
  \toprule
\textbf{Symbol} 
  & \textbf{Definition \& Formula} 
  & & \textbf{Eq.\ \#} \\
\midrule
$\epsilon$  
  & \EffectivePotential (\LargeN, high-T)
  & 
    $\epsilon(\WVEC) = \langle \DETOPXI\rangle_{\AVGNDXI}$
  & \ref{eqn:epsl} \\[1ex]

$\eta$
  & \SelfOverlap\ (average accuracy):
  & $\eta(\WVEC) = 1 - \epsilon(\WVEC)$
  & \ref{eqn:def_eta} \\[1ex]

$\bar{F}_{\ND}$
  & Average Free Energy:
  & $\beta\,\bar{F}_{\ND} = -\langle\ln Z_{\ND}\rangle_{\AVGNDXI}$
  & \ref{eqn:mm_f_bar} \\[1ex]
$\Gamma_{\bar Q}$
  & Quality‐Generating Function:
  & $\Gamma_{\bar Q} = 1 - \bar{F}_{\ND}$
  & \ref{eqn:GammaBar} \\
$\bar Q$
  & Model \Quality\ (average self‐overlap):
  & $\bar Q = \THRMAVGw{\eta(\WVEC)}$
  & \ref{eqn:model_qualities} \\[1ex]
\bottomrule
\end{tabular}
\caption{Summary of the main intensive (average, per‐parameter) quantities.  Here $n$ is the number of free parameters.  The average model
  \Quality~$\bar Q$ is the model’s average accuracy (one minus the error), and the Quality‐Generating function $\Gamma_{\bar Q}$ plays the same role as the free energy $\bar F$ but with an opposite sign convention.}
\label{tab:intensive_quantities}
\end{table}

For our purposes below, we define the \ModelQuality (as in Eqn.~\ref{eqn:ProductNorm}) as our approximation to the
\AverageGeneralizationAccuracy for our model. 
We denote the \ModelQuality for the ST \Perceptron model, $\Q^{ST}$,
and for a general NN, $\Q^{ST}$, such that
\begin{align}
  \label{eqn:model_pre_qualities}
\Q^{ST}:=1-\AVGGE^{ST}  \\ 
\Q^{NN}:=1-\AVGGE^{NN}  
\end{align}
In this work, however, the \Quality will always be defined at high-T, and so we may write
\begin{align}
  \label{eqn:model_qualities}
  \Q^{ST}=1-[\AVGTE^{ST}]^{an,hT}=1-[\AVGGE^{ST}]^{an,hT} \\
  \Q^{NN}=1-[\AVGTE^{NN}]^{an,hT}=1-[\AVGGE^{NN}]^{an,hT} 
\end{align}
We also define a \LayerQuality, simply denoted $\Q$,
which will describe the contributions an individual layer makes to the overall \ModelQuality $\Q^{NN}$.
To obtain the \LayerQuality, we define an accuracy--or \Quality--\GeneratingFunction, denoted $\Gamma_{\Q}$, which
is analogous to a layer \FreeEnergy, but with the opposite sign convention.

Generally speaking, the \Quality \GeneratingFunction $\Gamma_{\Q}$ is defined in the AA, and at high-T and is given as
\begin{align}
  \label{ref:defGammaQ}
  \beta\Gamma_{\Q} := \ln\int d\mu(\WVEC) e^{\ND\beta\ETAw}
\end{align}

For example, for the single-layer ST \Perceptron, $\Gamma_{\Q^{ST}}:=n-F^{ST}_{\ND}$
(in units of energy or error).
The term $\Gamma_{\Q^{ST}}$ is directly related to the \emph{Total} Free Energy $F^{ST}_{\ND}$, which can be seen by substituting \EQN~\ref{eqn:def_eta}
for $\ETAw$ in \EQN~\ref{ref:defGammaQ}:
\begin{align}
  \label{ref:defGammaQtoF}
  \beta\Gamma_{\Q^{ST}}
  &= \ln\int d\mu(\WVEC) e^{\ND\beta(1-\EPSLw)} \nonumber \\
    &= \ln\int d\mu(\WVEC) e^{\ND\beta}e^{-\ND\beta\EPSLw} \nonumber \\
    &= \ln\left(\int d\mu(\WVEC) e^{\ND\beta}\right)+\ln\left(\int d\mu(\WVEC)e^{-\ND\beta\EPSLw}\right) \nonumber \\
   &= \ln e^{\ND\beta}+\ln\left(\int d\mu(\WVEC)e^{-\ND\beta\EPSLw}\right) \nonumber \\
  &= \ND\beta -\beta F^{ST}_{\ND}
\end{align}
Dividing by $\ND$, we can also recover the more general relation for the \emph{Average} Free Energy,
$\bar{\Gamma}_{\Q}=1-\bar{F}_{\ND}$. (\EQN~\ref{eqn:GammaBar}). For the matrix case we do something similar; see Appendix~\ref{sxn:quality}.

Likewise, one can show that the \Quality $\Q$
(again, always in the AA and at high-T) can be identified as the \ThermalAverage of the (data-averaged or Annealed)
\SelfOverlap, 
\begin{align}
  \label{eqn:Q_def_eta}
  \Q = \THRMAVGw{\langle\ETA(\WVEC,\XI)\rangle_{\AVGNDXIn}} =\THRMAVGw{\ETA(\WVEC)}
\end{align}
We can then obtain $\Q$ by taking the appropriate partial derivative of its \GeneratingFunction, $\Gamma_{Q}$.

For technical reasons, however, we will actually define and use a
\GeneratingFunction for the \AverageLayerQualitySquared $\QT$, denoted $\IZG$.
In analogy with Eqns.~\ref{eqn:avgte_def} and ~\ref{eqn:avgge_def}, and at high-T and \LargeN (explained below),
we can obtain $\QT$ (see Section~\ref{sxn:matgen} Eqn.~\ref{eqn:IZG_generate_Q2}) as
\begin{align}
  \label{eqn:QT_def}
  \QT := \frac{1}{\beta}\frac{\partial }{\partial \ND}\IZGINF
  \underset{\text{high-}T}{\approx}
\frac{1}{\ND}\frac{\partial }{\partial \beta}\IZGINF
\end{align}
See Section~\ref{sxn:matgen} and Appendix~\ref{sxn:quality} for more details.

\subsubsection{The Thermodynamic limit}
\label{sxn:largeN_and_SPA}
In the \SMOG, the \ThermodynamicLimit is expressed as the limit of the number of free parameters $m$ grows arbitrarily large, $m \gg 1$, but also holding the feature load $n/m$ fixed (i.e., see \cite{SST92}).  In these older approaches, one is interested in how the thermodynamic properties depend on the feature load. Here, however, we are interested in estimating the generalization accuracy, so we are interested how the thermodynamic properties vary with $n$, the number of training examples.  For that reason, we will express the \ThermodynamicLimit as the \LargeN limit in $\ND$. i.e., while $\ND$ grows arbitrarily large,  $n \gg 1$, while implicitly fixing the corresponding feature load ($n/m$ for the ST Percepton,  and $n/N$ for our ST matrix model, below).

To express the average \FreeEnergy $\bar{F}_{\ND}$ in the \LargeN limit in $\ND$, we can write
\begin{align}
  -\beta\bar{F}_{\ND} = \lim_{n\gg 1}\frac{1}{\ND}\ln \int d\mu(\WVEC) e^{-\ND\beta\EPSL(\WVEC)}  .
\end{align}
When this \LargeN approximation is well behaved,
then the total energy $\DETOT$ is extensive, i.e., when $\DETOT=n\EPSL(\WVEC)$;
and, consequently, the total \FreeEnergy is also extensive, i.e., $F_{\ND}=\ND\bar{F}_{\ND}$.
This property is a cornerstone of statistical mechanics, as it allows for meaningful macroscopic predictions from microscopic interactions.

\paragraph{\SelfAveraging.}
The existence of the limit signifies that the system is \emph{\SelfAveraging}, meaning that the macroscopic properties are independent of fluctuations, etc.
This also implies that the relevant averages
(i.e., training and generalization errors) are the same for almost all samples, or \emph{realizations of the disorder}.
Additionally, the Annealed and the Quenched averages,
$\ln \langle Z_{\ND} \rangle_{\AVGNDXIn}$ and $\langle \ln Z_{\ND} \rangle_{\AVGNDXIn}$, respectively,
become sharply peaked, and
\begin{align}
\langle \ln Z_{\ND} \rangle_{\AVGNDXIn} \approx \ln \langle Z_{\ND} \rangle_{\AVGNDXIn}, \quad \text{as } \ND \gg 1\;\; \text{(grows arbitrarily large)}.
\end{align}
For a NN, \SelfAveraging implies that the weights $\WVEC$ are \emph{\Typical} of the distribution,
and therefore the NN can generalize to similar but unseen test examples.

\paragraph{The Saddle Point Approximation (SPA).}In \STATMECH, one is frequently faced with intractable integrals, and one common solution is to apply \SaddlePointApproximation (SPA). The SPA is a \LargeN approximation and is used to describe the system interactions at \LargeN, forming the average or mean interaction over the data.
For example, when considering the \LargeN in $\ND$ case, we could write

\begin{align}
  \label{eqn:SPA}
  \int d\mu(\WVEC) e^{-\ND\beta\EPSL(\WVEC)}\approx \sqrt{\tfrac{2\pi}{\ND\EPSL(\WVEC^{*})}} e^{-\ND\beta\EPSL(\WVEC^{*})}  
  , \quad \text{as } \ND \gg 1
\end{align}
where $\WVEC^{*}$ satisfies the saddle point equations:
\begin{align}
  \EPSL(\WVEC^{*}):=\tfrac{\partial}{\partial \WVEC}\EPSL(\WVEC)\vert_{\WVEC=\WVEC^{*}}=0 \\
  \EPSL(\WVEC^{*}):=\tfrac{\partial^{2}}{\partial^{2} \WVEC}\EPSL(\WVEC)\vert_{\WVEC=\WVEC^{*}}>0  .
\end{align}
These equations guarantee that the solution $\EPSL(\WVEC^{*})$ is at a (local) minimum in $\WVEC$.
The SPA can be applied to many \LargeN integrals, as long as the integrand decays exponentially,  

\paragraph{Wick Rotation.} When working with complex oscillatory integrands, $i.e., e^{iNf(x)}$,   
the SPA is frequently combined with a \emph{\WickRotation}, 
which is used to convert an oscillatory integral to an exponentially decaying one.
The \WickRotation is an analytic continuation that rotates the integration variable by 90 degrees, i.e., $x \to \pm i x$, deforming the integration contour to pass through a saddle point of the integrand:
\begin{equation}
x\rightarrow ix,\;\;e^{iNf(x)}\rightarrow e^{-Nf(ix)},
\end{equation}
This is typically used in a formal sense; that is, one assumes the resulting complex integrand is both analytic and well-behaved.
One would then apply the SPA in the \LargeN in $N$ case.
Below, we will do just this, and  use the SPA with a Wick Rotation in both Appendices \ref{sxn:TraceLogDerivation} and \ref{sxn:tanaka}.

\paragraph{When the \Thermodynamic limit fails: \ATypical behavior.}
When the \ThermodynamicLimit fails to exist, the \FreeEnergy will contain additional, non-extensive terms, i.e.,
$F = n\bar{F}_{ex} + n^{1+x}\bar{F}_{non-ex},\;x>0$.
\footnote{When dealing with matrix integrals (below), $F\sim nNMF_{0}+\cdots$, when there are $n \times N \times M$ degrees
of freedom~\cite{PP95}.}
In this case, the AA may fail, the SPA may not apply, and the system may fail to be \SelfAveraging.
This causes the system to behave in an \ATypical way, 
possibly converging to a meta-stable and/or glassy phase.
Indeed, when the weights $\WVEC$ are \emph{\ATypical}, they may describe the training data well, 
but they would fail to describe the test data well; in this sense, we say the model is \emph{overfit} to the training data.
We will not explicitly consider a model that is non-extensive; however, we will
present empirical results where we suspect the model is overfit
(in Section~\ref{sxn:empirical-correlation_trap})
and, additionally, where we observe glassy behavior, which we refer to as a \emph{Hysteresis Effect}
(in Section~\ref{sxn:hysteresis_effect}).

\subsubsection{From the ST Perceptron to a Matrix Model.} 
\label{sxn:from_vectors}
In Section~\ref{sxn:matgen}, we will move from a single $m$-dimensional \Perceptron vector $\WVEC$ 
to a matrix model of $N \times M$ weights, which can be thought of
as $N$ interacting (feature) vectors of dimension $M$.
Moreover, in the Student-Teacher (ST) model used below,
in the vector case, the $m$ free parameters (o.e., degrees of freedom) of the feature vectors are integrated out,
whereas for the matrix case, we will retain \emph{a reduced set} of the $M$ degrees of freedom, $\MECS$. To do this, we need to ensure the \FreeEnergies and related quantities like the \LayerQualitySquared scale properly.

\SETOL uses two \LargeN approximations, the \Thermodynamic limit, which is a \LargeN limit in $\ND$, and the \emph{\WideLayer limit}, which is a \LargeN limit in $N$:
\begin{itemize}
    \item \Thermodynamic limit: $m \gg 1$ (the number of parameters or features, fixed load: $\ND/m = 1$)
    \item \LargeN limit in n: $n\gg 1$ (the number of training examples, fixed load: $\ND/m = 1$)
    \item \WideLayer limit: $N\gg 1$ (the number of feature vectors,  $n/N$ fixed)
\end{itemize}
Note that by ``limit'' we mean $\ND$ or $N$ simply grows arbitrarily large. In taking the \WideLayer \LargeN limit in $N$, we assume implicitly that  the \emph{layer load}, $\ND/N$, the number data points per feature vector or output dimension $N$, remains constant. \footnote{Although we do not enforce this explicitly here, we note that the so-called ``Chinchilla Scaling Law'' demonstrates an explicit compute-optimal frontier~\cite{hoffmann2022}, which we can expect practitioners will likely adhere to.}  
In doing this, we expect the matrix-generalized total Energies $(F, H^{an}, \cdots)$ to be \SizeExtensive in $\ND$ and $N$, the average quantities to be \SizeIntensive in both, 
and, as discussed below, \emph{\SizeConsistent} in $\MECS$.

\paragraph{\SizeConsistency.}
The notion of \emph{\SizeConsistency} is a notion commonly 
used Quantum Chemistry to describe interacting, correlated systems. It is often introduced through the Linked Cluster Theorem~\cite{Hubbard1959,Brandow1963},
which states that the \emph{average} energies and/or free energies $(\bar{F})$ of, in particular, a correlated system scale with $\MECS$,
the number of \emph{independent interacting components}:
\begin{align}
  \label{eqn:LCT}
  -\beta \bar{F}=\langle \ln Z \rangle_{\AVGNDXIn} &= \sum_{\mu=1}^{\MECS}\text{(Connected Components)} \nonumber \\
  &= \sum_{\mu=1}^{\MECS}\text{Cumulants($\mu$)+higher order terms} 
\end{align}
For \SETOL, below, these connected components will be the matrix-generalized cumulants from RMT.
In Quantum Chemistry, this is analogous to ensuring the renormalized Self-Energy and/or Effective Hamiltonian is \SizeConsistent.
For NNs, \SizeConsistency appears when scaling the number of feature vectors $N$ in our matrix model,
and it ensures that our layer Free Energy, etc., remain well-behaved as we increase $\MECS$.
For a simple example, see Appendix~\ref{sxn:summary_sst92}
 where we derive the expression for the matrix-generalized
\AnnealedHamiltonian $\HAN$.  
Both \SizeExtensivity (in $\ND$ and $N$) and \SizeConsistency (in $\MECS$ )
are crucial in our \SETOL analysis: they justify taking the \LargeN approximation for matrix integrals, and they ensure
our resulting HCIZ integral--a sum of integrated RMT cumulants (below)--
scales with the dimension of the \EffectiveCorrelationSpace (\ECS), $\MECS$.

\paragraph{HCIZ Integrals}
To generalize the Linear ST \Perceptron (in the AA, and at high-T) from \Perceptron vectors to MLP matrices,
we need to generalize the thermal average over the $m$-dimensional \Perceptron weight vectors $(\WVEC=\SVEC)$
to an integral over NN \Student $N \times M$ weight matrices ($\WMAT=\mathbf{S}$).
The resulting \PartitionFunction and \FreeEnergy will be expressed with what is called an HCIZ integral.
(See Tanaka~\cite{Tanaka2007, Tanaka2008}, Gallucio et al.~\cite{Bouchaud1998} and/or Parisi~\cite{PP95}, and also \EQN~\ref{eqn:tanaka_result} in Section~\ref{sxn:matgen}.)
As in \EQN~\ref{eqn:IZGINF_HCIZ}, we will define a \emph{\LayerQuality \GeneratingFunction}, $\IZG$, for the \LayerQuality (squared).
This will take the form of an HCIZ integral,
\begin{align}
\label{eqn:hciz_prelim}
\IZGINF := \lim_{N\gg 1}\frac{1}{N}\ln \underbrace{\int d\mu(\AMAT) \exp[\ND\beta N\Trace{\AMAT_N\XMAT_N}] }_{\mbox{HCIZ Integral}} 
  \approx \ND\beta\Trace{\mathcal{G}_{A}(\lambda)}  ,
\end{align}
that is evaluated in the \WideLayer\emph{\LargeN limit in $N$} (here, meaning as $N \gg 1$, but not at the limit $N=\infty$).
Here, $\AMAT_N$ and $\XMAT_N$ are $N \times N$ Hermetian matrices, (See Appendix \ref{sxn:appendix_A} for notation,) $d\mu(\AMAT)$ is a measure
over all random (Orthogonal) matrices (see \EQN~\ref{eqn:dmuA}),
and $\mathcal{G}_{A}()$ is called a \GEN~
(which are different from $\Gamma_{\ND}$, above, and will be made clear in Section~\ref{sxn:r_transforms} ), defined in \EQN~\ref{eqn:generating_function_A} in Section~\ref{sxn:matgen}. 
In applying this, we will actually write $\Trace{\AMAT\XMAT}=\tfrac{1}{N}\Trace{\AMAT\WMAT\WMAT^{\top}}=\tfrac{1}{N}\Trace{\WMAT^{\top}\AMAT\WMAT}$,
where $\mathbf{W}$ is an $N \times M$ layer weight matrix, and $\AMAT=\AMAT_N:=\tfrac{1}{\ND}\SMAT\SMAT^{\top}$ is an (Outer) layer
Correlation matrix, and here,  $\XMAT=\XMAT_N=\tfrac{1}{N}\WMAT\WMAT^{\top}$).
Moreover, to evaluate this, we will need to restrict $\AMAT$ (and $\XMAT$)
to the lower-rank \EffectiveCorrelationSpace (\ECS),  i.e., $\AMAT\rightarrow\AECS$. ($\XMAT$ is already restricted to the span of $\XECS$ by \ECS assumption.)

Notice that this looks similar to a \SaddlePointApproximation (SPA), but where the more complicated function
$\GNORM()$ now appears.
Also, $\GNORM()$ here depends only on the limiting form of the ESD of $\AMAT$, $\rho^{\infty}_{A}(\lambda)$,
and depends on the $M$ normalized eigenvalues $\lambda_{i}$ of $\XMAT$.

To evaluate this, we will form the \LargeN limit in $N$, using a result from Tanaka~\cite{Tanaka2007,Tanaka2008},
but slightly extended (in Appendix~\ref{sxn:tanaka}) to include the term $\ND\beta$ explicitly.
We can write
\begin{align}
  \label{eqn:izgin_def}
  \IZGINF:=\lim_{N\gg 1}\IZG ,
\end{align}
with the final result
  \begin{align}
    \label{eqn:hciz_tanaka}
    \IZGINF=\ND\beta\sum_{i=1}^{\MECS}\int^{\LambdaECS_{i}}_{\LambdaECSmin} dz R(z) ,
\end{align}
where $R(z)$ is a complex function, the \RTransform of the ESD $\rho(\lambda)$ of the \Teacher, and $\LambdaECS_{i}$ are the eigenvalues of \Teacher \CorrelationMatrix $\XECS$, restricted to the~\ECS.
For more details, see Section~\ref{sxn:matgen} and Appendices~\ref{sxn:TraceLogDerivation} and~\ref{sxn:tanaka}.

\paragraph{Branch Cuts and Phase Behavior.}
Free energies ($F,\Gamma$) often exhibit \emph{branch cuts} when expressed as analytic functions 
of complex parameters (i.e., temperature, coupling constants, or eigenvalue cutoffs),
and arise from singularities in the underlying integral representations of the partition function $Z_{\ND}$,
When a branch cut occurs, this demarcates non-analytic behavior
and this indicates the onset of a \emph{phase transition} where
macroscopic observables such as correlation lengths and/or variance-like quantities
may diverge or change character abruptly.  

In the context of our HCIZ-based construction, integrating a complex function like the $R(z)$-transform of
a \HeavyTailed ESD can produce precisely this phenomenon.
For example, if $R(z)$ has a square-root term, i.e., $(\sqrt{z-c})$, then it will have a branch cut at $z=c$,
and the \GeneratingFunction (i.e.,  negative Effective Free Energy) $\IZGINF$
will be non-analytic and we must choose the appropriate, physically meaningful branch, i.e., $(\Re[z]>\LambdaECSmin)$,
corresponding to the \ECS (where $\Re[z]$ is the Real-part of $z$)
We argue that this cut signifies a \emph{phase boundary}—an abrupt change
in the system’s correlation structure and corresponds to an emerging singularity in the \LayerQuality.
Even though we perform only a single exact RG step, 
(rather than fully iterating a renormalization flow), the appearance of a branch cut in $\IZGINF$ will encode
nontrivial \emph{phase-like} behavior in the \SETOL \HeavyTailed matrix model.

\subsection{Student-Teacher Perceptron}
\label{sxn:SMOG_main-student_teacher}

In this subsection, we present a unified view of the \emph{\StudentTeacher} (ST) \Perceptron 
model from both a practical and a theoretical perspective.  
The common sense modern understanding of Student-Teacher learning typically involves knowledge distillation: a larger, pretrained neural network (the \emph{Teacher}) is used to train a smaller network (the \emph{Student}), which learns to imitate the Teacher’s outputs $\Yt$.
In practical terms, this often helps the smaller Student achieve comparable performance more efficiently~\cite{hu2023TS}.
And it remains an open question as to why this works so well.
In contrast, classical theoretical models from the Statistical Mechanics of Generalization (\SMOG) interpret the Student-Teacher relationship differently. Here, the Teacher is an idealized theoretical model $T$, a Perceptron (vector) $\TVEC$, that explicitly generates (binary) labels $\Yt$ from some idealized data $\xi$, and the Students
$S$ are Perceptrons (vectors) of the same size as the Teacher, $\Vert\SVEC\Vert=\Vert\TVEC\Vert$, selected from a Boltzmann distribution.
The ST theory seeks simple analytic derivations of the generalization errors $\AVGSTGE$ under different simplifying assumptions (AA, high-T, linear activations, etc.),
in order to understand phenomena such as what is now called Double Descent\cite{Vallet1989}, NN storage capacity, learning with noisy labels,  memorization vs. generalization, and other 
fundamental learning phenomena\cite{Opper01,OK96_CHAPT,Eng01,EngelAndVanDenBroeck,SST90,SST92}.
A key result is that $\AVGSTGE$ is a simple analytic function of the ST \emph{Overlap}, $R=\SVEC^{\top}\TVEC$.
Our \SETOL approach is motivated by both these perspectives, and introduces a novel, \SemiEmpirical twist. 
We seek a matrix-generalization of the classical vector ST model.
And while inspired by the ST Perceptron, 
rather than an idealized theoretical Teacher $T$, we take the Teacher as an empirical input—a layer weight matrix, $\TMAT=\WMAT$,
taken from a real model--a model you might train yourself, download, or acquire elsewhere.
We aim to estimate the layer accuracy (or precision),
which we call \LayerQuality $\Q$, using an ensembles of similar Students, $S\sim T$, i.e. $N \times M$ matrices, $\SMAT,\TMAT$.
$\Q$ is also a function of the ST Overlap, now a matrix,
$\OVERLAP=\tfrac{1}{N}\SMAT^{\top}\TMAT$,
but now also critically depends on the ESD of $\TMAT$.
And while our Students are the same size as the Teacher, 
we will find later that the optimal Teacher is actually much smaller than the one we started out with. 

\paragraph{Practical framing.}
Imagine being a real-world practitioner who has trained a large-scale neural network (NN), or downloaded it from the internet, or received it by other means --- an increasingly common scenario.
 We call this NN the \Teacher, because it is the primary empirical input to the theory. The term ``\Teacher'' means that we consider this object to be the original source of supervision -- this trained model replicates the training labels $\Yt$, often to perfection (interpolation) or nearly so.\footnote{If nearly so, then we consider this label noise to be the contribution of the \Teacher, which \SETOL can characterize.}
You also hold a dataset $\ADD$ that approximates the distribution on which $T$ was trained, yet you lack a reliable yardstick for evaluating $T$’s outputs: perhaps $\ADD$ is contaminated with test data, or $T$ is an LLM with highly arbitrary outputs, or you just
don't have access to test data.  In any case, you can escape this bind by applying a \emph{Student–Teacher} protocol.  
You retrain one or more \Students\ $S_{1},S_{2},\dots$ on $\ADD$ using ordinary tools (SGD), that resemble the Teacher $T$
(same size, amount of regularization, etc),
but now slightly varying the training procedures (i.e., initial conditions, optimization hyperparameters, etc.)--
tasking them solely with \emph{imitating} $T$’s predictions. 
Averaging over many such Students, you can obtain an \emph{indirect but quantitative} estimate of $T$’s performance.

\paragraph{Why this works.}  
We view $T$ as an \textit{inevitable} outcome of the training pipeline: if you reran the same recipe many times, you would likely converge on a narrow family of similar models.  
The tighter that distribution, the more \emph{Precise} it is, and—assuming the ground-truth concepts $\Yt$
are at least \emph{nearly} realizable— the closer any single draw (your Teacher) must sit to the truth.  
By comparing Student and Teacher outputs, you can approximate the Teacher’s generalization performance even without an explicit hold-out set.  
Notably, if the Teacher perfectly interpolates its training data, the Student’s error directly estimates the Teacher’s \emph{true} \GeneralizationAccuracy.
Otherwise, it captures the Teacher’s \emph{Precision} in reproducing its own noisy or biased labels.

\paragraph{An extended theory.} We seek a succinct, analytic formulation of the ST \AverageGeneralizationError, denoted $\AVGSTGE$.  
We work in the \AnnealedApproximation (AA) -- a simplification often valid when networks are sufficiently large and can nearly interpolate their training data. 
Under this approximation, one obtains closed-form expressions for the \StudentTeacher \emph{Overlap} $(R)$
and thus for the \Teacher overall error or accuracy.  
Take note also that we do not assume that the labels $\Yt$ are realizable by the \Teacher, but, rather,
take the Teacher as empirical input to the theory.  
This point is both subtle -- since the derivations  proceed closely at times to those found in~\cite{engel2001statistical,EngelAndVanDenBroeck,SST90,SST92}
-- and also radical, because our derived quality metrics $\Q$ now
depend directly on empirical properties of the \Teacher.
These results lay the foundation for our \emph{\SemiEmpirical} approach, in which we supplement this theoretical form with empirical measurements (e.g., from the trained weight matrices) to account for real-world correlations in the data and the model’s internal structure.  
\textbf{As far as we know, this is completely new \SemiEmpirical approach for modeling NNs.}

\begin{enumerate}[label=4.3.\arabic*]
\item
  \textbf{Operational Setup.}
  In subsection~\ref{sxn:ST_OP_setup} we explain how to set up our formulation\StudentTeacher
  model in an \emph{operational} manner. 
  In particular, we emphasize the difference between \emph{true accuracy} (vs.\ ground-truth labels)
  and \Precision (vs. the Teacher’s own labels). We also the discuss the difference between
  our \Quality metrics and the \emph{Generalization Gap}.

  \item
    \textbf{Theoretical Student-Teacher Average Generalization Error $(\AVGSTGE)$.}
    In subsection~\ref{sxn:SMOG_main-st_av}  we outline how to derive $\AVGSTGE$ using the AA.  
    We introduce the key expressions that will serve as the starting point for our extended \SemiEmpirical theory in subsequent subsubsection.
\end{enumerate}

\subsubsection{Operational Setup}
\label{sxn:ST_OP_setup}
\begin{figure}
  \begin{center}
    \begin{tikzpicture}[scale=1.4, every node/.style={transform shape}]
    \node[draw, circle, minimum size=0.15cm, inner sep=0.05cm] (A) at (0, 3) {};
    \node[draw, circle, minimum size=0.15cm, inner sep=0.05cm] (B) at (0, 2) {};
    \node[draw, circle, minimum size=0.15cm, inner sep=0.05cm] (C) at (0, 1) {};
    \node[draw, circle, minimum size=0.15cm, inner sep=0.05cm] (D) at (0, 0) {};

    \node[left] at (A.west) {$\XI_1$};
    \node[left] at (B.west) {$\XI_2$};
    \node[left] at (C.west) {$\XI_3$};
    \node[left] at (D.west) {$\XI_4$};

    \node[draw, circle, minimum size=0.15cm, inner sep=0.05cm] (Y) at (2, 2.5) {};
    \node[draw, circle, minimum size=0.15cm, inner sep=0.05cm] (Z) at (2, 0.5) {};

    \node[right] at (Y.east) {$\Yt$};
    \node[right] at (Z.east) {$\Ys$};
  
    \draw[->, orange] (A) -- (Y) node[pos=0.2, above] {$S_1$};
    \draw[->, orange] (B) -- (Y) node[pos=0.2, above] {$S_2$};
    \draw[->, orange] (C) -- (Y) node[pos=0.2, above] {$S_3$};
    \draw[->, orange] (D) -- (Y) node[pos=0.2, above] {$S_4$};

    \draw[->, blue] (A) -- (Z) node[pos=0.5, above] {$T_1$};
    \draw[->, blue] (B) -- (Z) node[pos=0.5, above] {$T_2$};
    \draw[->, blue] (C) -- (Z) node[pos=0.5, above] {$T_3$};
    \draw[->, blue] (D) -- (Z) node[pos=0.5, above] {$T_4$};

  \end{tikzpicture}
  \end{center}
  \caption{Pictorial representation Student and Teacher Perceptrons.}
\end{figure}
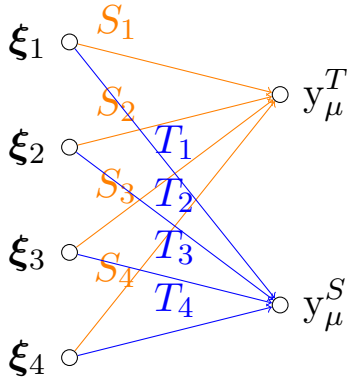

Here, describe the basic setup of the classic \StudentTeacher model, taking an operational view from the perspective of a practitioner training real-world \Student and \Teacher models. Specifically, we present the \AnnealedApproximation (AA) in a practical light,
and use it to explain the difference between computing the \emph{Empirical \GeneralizationError}, $\AVGEMPGE$, for the \emph{\TrueAccuracy}
and for the \emph{\Precision} of a \Teacher model.
Table~\ref{tab:st_notation} summarizes some of the key notation we will be using in this section, and also the next two sections for reference.
\begin{table}[h]
\centering
\begin{tabular}{@{}ll@{}}
\toprule
\textbf{Symbol} & \textbf{Meaning} \\
\midrule
$T$                      & Teacher model (fixed network) \\
$S$                      & Student model (trained to mimic $T$) \\
$\ADD$             & True dataset of $(\DATA,\Ytrue)$ pairs \\
${\mathcal D}$ & Gaussian‐field idealization of $\mathcal D$ \\
$ \AVGGE^{emp}$    & Average empirical test‐set error estimate \\
 $\AVGGE^{T}$          & Average, true generalization error of $T$ (unknown) \\
$\AVGGE^{ST}$       & Student–Teacher average generalization error \\
$\SVEC,\;\TVEC$                      & Student and Teacher weight vectors \\
$\Ys,\;\Yt$                      & Student and Teacher labels $(+1|-1)$  for example $\mu$\\
$\AVGR$                      & Student-Teacher  Overlap $\displaystyle \AVGR=\SVEC^\top \TVEC$ \\
$\epsilon(R)$            & \EffectivePotential (i.e.,\ $1-\AVGR$) \\
$\eta(\SVEC,\TVEC)$            & \SelfOverlap (i.e., $\AVGR$) \\
$\Q^{ST}$            & Student-Teacher Perceptron \Quality \\

\bottomrule
\end{tabular}
\caption{Key notation for our formulation of the Student–Teacher model.}
\label{tab:st_notation}
\end{table}

\paragraph{Test Error of the Teacher.}
We start by describing how to obtain a simple formal expression for the empirical test errors of the \Teacher $T$.
Let us say we have a model, called \Teacher $(T)$, which maps some \emph{actual} (i.e., correlated,  meaning the features are not i.i.d.,) data
$\DATA\in\ADD$ to some known or \emph{true} labels $(\Ytrue)$
(where,  $\Ytrue$ is, say, an $N$-dimensional vector of binary labels).
We might say that $\Ytrue$ represents the \emph{\GroundTruth} for the problem.
Operationally, we train the \Teacher $T$ to reproduce or at least approximate the true labels $\Ytrue$.
\begin{align}
 T:\DATA\rightarrow \Yt \approx \Ytrue.
\end{align}
If $T$ reproduces the true labels exactly, we might say that the \Teacher has been
trained to \emph{\Interpolation}, and therefore, $\Yt = \Ytrue$.
Indeed, most models today are trained to \emph{\Interpolation}, and we don't need to
necessarily worry about the difference between the true and the predicted \Teacher labels.
Formally, however, and to better understand the AA, it is beneficial to discuss the implications
of this distinction.

Following \EQN~\ref{eqn:dnn_energy}, let's say the \Teacher outputs are specified by
the NN output (or inference energy) function $\NNOUT$
\footnote{
We refer to outputs of $\NNOUT(\TVEC,\DATA )$, when applied to a data point $\DATA$, as energies because they are effectively un-normalized probabilities for the class outputs (for labels $\Ymu=1$ or $-1$).  }
\begin{align}
\label{eqn:T_ENN}
\Yt=\NNOUT(T,\DATA) 
\end{align}
so that we may write the \emph{Empirical} \AverageTrainingError
$\AVGEMPTE$
as 
\begin{align}
\label{eqn:Eg_train}
\AVGEMPTE:= \frac{1}{N^{train}}\sum_{\mu=1}^{N^{train}}\mathcal{L}[\Ytrain,\NNOUT(T,\DATAtrain)]  .
\end{align}
Ideally, we seek the \emph{True} \AverageGeneralizationError of the \Teacher, denoted  $\AVGGE^{T}$. 
Of course, this is unknowable, but in practice, we estimate $\AVGGE^{T}$ 
by measuring the error of the \Teacher predictions on some test (or hold-out) set $(\DATA^{test}, \MY^{test})$.
We call this the \emph{Empirical \AverageGeneralizationError}, $\AVGEMPGE$, and write
\begin{align}
\label{eqn:Eg_test}
 \AVGGE^{T}\approx \AVGEMPGE:= \frac{1}{\Ntest}\sum_{\mu=1}^{\Ntest}\mathcal{L}[\Ytest,\NNOUT(T,\DATAtest)] .
\end{align}
To measure the error, the loss function $\mathcal{L}$ may be an L1 $(\ell_1)$ or L2 $(\ell_2)$ loss;
whereas for training a NN model, it is usually something like a cross-entropy loss.
\footnote{Note that $\AVGGE^{T}$ will be treated as an energy function, as is $\NNOUT$, yet they are related nonlinearly. This is because they are energies in \emph{two different systems} -- training, and inference -- and $\mathcal{L}$ is a transfer function that bridges them.}

If we don't have a hold-out set, however, we can still estimate $\AVGGE^{T}$ using the \StudentTeacher approach.

\paragraph{Estimating the Student-Teacher Error: Accuracy vs. Precision.}
Imagine we train a Student model $S$ (with the same architecture as the Teacher $T$) on the real-world dataset $\ADD$, using $T$'s outputs as targets:
\begin{equation*}
S: \ADD \to y^{S}_\mu \approx y^{T}_\mu,
\label{eqn:S_ENN}
\end{equation*}
where $y^T_\mu = \NNOUT(T, x_\mu)$ are the Teacher’s predicted labels or energies.

\begin{figure}[ht] 
  \centering
  \resizebox{0.75\textwidth}{!}{

\begin{tikzpicture}

    \draw[->] (0,0) -- (6,0) node[below] {Value};
    
    \draw[thick, red] plot [smooth] coordinates {(1,0) (2,1) (3,3) (4,1) (5,0)};
    
    \draw[thick, blue] (3,0) -- (3,3.2);
    \node[blue, above] at (2.5,3.2) {\textbf{Accuracy} };
    \draw[->] (2.5,3.) -- (3,3.);
    \node[black, below] at (0.75,3.2) {\textbf{Teacher output} $\Yt$};

    \draw[<->, blue, thick] (3,3.2) -- (4,3.2);
    
    \fill[darkgreen] (4,3) circle (0.15);
    \node[right, below] at (5.5,2.9) {\textbf{Ground truth} $\Ytrue$};
    
    \draw[->] (2.5,1) -- (3.5,2);
    \node at (0.75,0.75) {\textbf{Student outputs} $\Ys$};

    \draw[<->, red, thick] (1,-0.5) -- (5,-0.5);
    \node[red, below] at (3,-0.5) {\textbf{Precision}};
    
    \begin{scope}[xshift=9cm, yshift=2cm]
        \draw[thick] (0,0) circle (1.8);
        \draw[thick, fill=white] (0,0) circle (1.2);
        \draw[thick, fill=white] (0,0) circle (0.6);

        \fill[blue] (-0.5, 0.0) circle (0.10);
        \foreach \x/\y in {-0.6/0.3, 0.1/-0.7, -0.15/0.5, -0.3/-0.5, -0.75/0.75, -0.9/-0.2} {
            \fill[red] (\x,\y) circle (0.08);
        }
        
        \fill[darkgreen] (0,0) circle (0.15);
        
        \node[left] at (2.,-2.5) {\textbf{Bullseye Example}};
    \end{scope}

\end{tikzpicture}
  }
  
\caption{Precision vs. Accuracy}
 \label{fig:precision}
\end{figure}
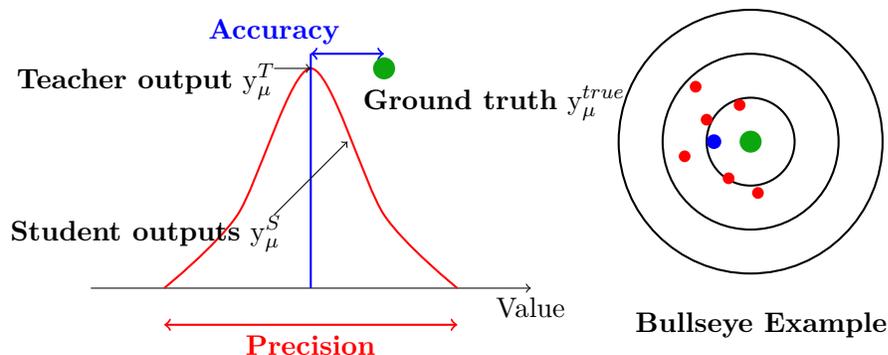

Figure~\ref{fig:precision} illustrates the following two scenarios side by side:
\begin{itemize}
  \item \textbf{Accuracy (Interpolation) regime:}
    If $T$ perfectly interpolates the true labels ($y^T_\mu = y^\text{true}_\mu$), then
    \( \|y^S_\mu - y^T_\mu\| = \|y^S_\mu - y^\text{true}_\mu\| \)
    measures how well the student recovers the Ground Truth labels—i.e., the Student's \emph{training accuracy}.

  \item \textbf{Precision (Noisy teacher) regime:}
    If $T$ itself makes mistakes ($y^T_\mu \neq y^\text{true}_\mu$), then
    \( \|y^S_\mu - y^T_\mu\| \)
    estimates how faithfully $S$ reproduces $T$ (its precision), rather than true accuracy.
\end{itemize}

By analyzing the Student–Teacher overlap $R = \sum_\mu s^{\top} t$, one can show that the average generalization error of $T$ depends simply on $1-R$ under the annealed approximation—-even when $T$ was trained on noisy labels. In practice, we exploit this fact by training a large ensemble of students to estimate $R$ and hence recover the teacher’s true error.
%

%
The \StudentTeacher model also explains why NNs can generalize even when trained to \Interpolation on noisy data (which has been a source of confusion \cite{Understanding16_TR}). In this model, the \GeneralizationError $\AVGGE^{T}$ is a simple function of the overlap $R$ between the \Teacher $T$ and the \Students $S$, i.e., $\AVGGE^{T}\sim \THRMAVG{1-\EPSLR}$. So even if the \Teacher $T$ is trained on noisy data, as long as there are \Students $S$ with significant overlap $R$ with the \Teacher, the \Teacher \GeneralizationError $\AVGGE^{T}$ can be considerably small. For more details, see \cite{MM17_TR_v1}.

\paragraph{Learning the Student.}
Moving forward, we will always assume the \Teacher is trained to \Interpolation because this
actually corresponds to the \AnnealedApproximation, whereas if the \Teacher makes
errors, we may need to consider \Quenched averages, explained below.  

Imagine now that to estimate the empirical \AverageGeneralizationError, $\AVGEMPGE$,
by training a very large number of Students, and computing the average ST error on some test set.
Let us break the data set into training $\DATAtrain$ and test $\DATAtest$ examples, 
train models on the training data (that is, find the optimal model weights), 
and evaluate the $S$ and $T$ models on the test data.

The \Student learning task can be written as in \EQN~(\ref{eqn:dnn_opt})
as the following optimization problem over the training data:
\begin{align}
\underset{\{S'\}}{\argmin}\sum_{\mu=1}^{N^{train}}\mathcal{L}[\NNOUT(S',\DATAtrain),\NNOUT(T,\DATAtrain)]   ,
\label{eqn:ST-learning-task}
\end{align}

If the \Teacher is trained to \Interpolation, then the NN optimization problem 
training a \Student to reproduce the \GroundTruth labels, so that $\Ys\sim y_{\mu}^{true}$
for both the training and test sets.
\begin{align}
\underset{\{S'\}}{\argmin}\sum_{\mu=1}^{N^{train}}\mathcal{L}[\NNOUT(S',\DATAtrain),\Ytrue]   ,
\label{eqn:ST-learning-intepolation}
\end{align}

If not, then the \Student is reproducing the possibly
incorrect \Teacher labels, and, importantly, the \Student $S$ \emph{now depends explicitly
on how the \Teacher was trained.}    That is, we should denote that the learned
\Student \emph{explicitly depending} on $T$, i.e. $S\rightarrow S[T]$.
This will be important below.

\paragraph{The \ThermalAverage.} 
In \STATMECH, the average over Student weights $\langle\cdots\rangle_{\SVEC}$ is a \ThermalAverage $\THRMAVG{\cdots}$.  This means we should restrict the choice of $S$ to the same Boltzmann distribution we assume $T$ is drawn from.  
In a practical sense, we would expect our Students $S$ to be trained with the roughly same amount of regularization (same weight decay, etc.). In principle, we should be able to admit many kinds of $S$, including those arising from
different optimization paths which lead to the same equilibrium sample. 
\footnote{See~\cite{WILSON20031429}, which shows that for NNs trained with SGD, nearly all learning rates, if they are not too large, converge to the same minima given enough time. Additionally, ~\cite{tsukada2024} shows that one can effectively train with any batch size and learning rate—as long as you schedule them correctly (i.e., growing them)—and still guarantee convergence to the same solution.}
While the parameter $\beta$ is like a fixed regularization setting,
it actually modulates the trade-off between energy and entropy—
sometimes called \emph{entropic regularization}—and therefore how strongly the
data constrains the model, and not specifically how the model is trained.
Moreover, in the high-T (small-$\beta$) limit the
measure becomes broader, so we can admit more Student weights that fluctuate 
around the mean.
In practice, however, we find
that different optimizer settings (batch sizes, learning rates, stopping criteria, etc.)  can give very different results.   
Therefore, we must be careful not to admit any arbitrary Student into the average
because this would amount to averaging over \emph{several} Boltzmann distributions and this should
be treated explicitly.

\paragraph{The \AverageGeneralizationError.}
We may seek to estimate the Empirical \AverageGeneralizationError
by replacing the test predictions in \EQN~\ref{eqn:Eg_test} with the student predictions
$y_{\mu}^{test}\rightarrow \Ys$, and then averaging directly over the test data $\DATAtest$
for all possible or available test examples.

If we have a very large number of suitable Students
(say, drawn from some random distribution), then we can try to estimate the 
\AverageGeneralizationError of the \Teacher, i.e., $\TGE^{T}\approx\AVGEMPGE$.
$\AVGEMPGE$ is given by an average loss, the average is 
over all possible Students $N_S$,  and then  over all  $\Ntest$ test data points $\DATAtest\in\mathbf{D}$ 
\begin{align}
  \AVGEMPGE
  &=
  \frac{1}{\Ntest}\sum_{\mu=1}^{\Ntest}
  \frac{1}{N_S}\sum_{S}
  \mathcal{L}[\NNOUT(S,\DATAtest),\NNOUT(T,\DATAtest)]  \\ \nonumber
    &=
  \frac{1}{N_S}\sum_{S}
    \frac{1}{\Ntest}\sum_{\mu=1}^{\Ntest}
    \mathcal{L}[\NNOUT(S,\DATAtest),\NNOUT(T,\DATAtest)] ,
\label{eqn:emp_gen_error}
\end{align}
where (ideally) $\Ntest$ is extremely large.

In Bra-Ket notation, we may also write
\begin{align}
  \AVGEMPGE
  &= \langle \langle \DETOPSTx \rangle_{S} \rangle_{\DXtest}\\ \nonumber
  &= \langle \langle \DETOPSTx \rangle_{\DXtest} \rangle_{S}
\end{align}
where $\DETOPSTx:=\mathcal{L}[\NNOUT(S,\DATAtest),\NNOUT(T,\DATAtest)]$.
For the empirical estimate, it does not matter what order we take the sums in,
but we are not estimating the
the True \AverageGeneralizationError  of the \Teacher, $\AVGGE^{T}$,
unless $T$ is trained to \Interpolation.
For the theoretical estimate, however, the order can be important, and this also depends on
if $T$ is trained to \Interpolation or not.
\footnote{This approach can be likened to the Bootstrap method~\cite{efron1993bootstrap} used for error estimation.  However, the Bootstrap method predominantly emphasizes variations in the input data $\NDX\in\mathbf{D}$, while in this context, we are essentially bootstrapping over the students $S$.}

\paragraph{Annealed vs. Quenched Averages.}
Recall that in the \STATMECH approach to computing errors, we do not break the data into
training and test, but, instead, to obtain the \AverageGeneralizationError, $\AVGGE$, use
the \GeneratingFunction approach. In doing this, we need to compute both the \ThermalAverage
over the model weights ($S,T$), and also take the data average over the entire available idealized data set $\MDD$.
And the order can matter.

In the case where the \Teacher is trained to \Interpolation, may train the \Student 
independently of \emph{when} the \Teacher.  
But if \Teacher is \emph{not} trained to \Interpolation, then formally we must train the \Teacher
first to obtain target predictions for the \Student.  That is, the \Student formally
depends on the \Teacher, $S[T]$.
The empirical errors in $T$ would then formally depend on the specific instantiation of the data  $\NDXIn\in\MDD$,
and therefore, conceptually imagine that we must first average over the data
before averaging over the weights.
\emph{Training to \Interpolation} corresponds conceptually to working with a model
in the \AnnealedApproximation, whereas not doing so corresponds to \Quenched case.

Practically, when the \Teacher is not trained to \Interpolation, 
we may need to resample the training data and training an ensemble of models to compensate for anomalies in training data (bad labels, noise, etc.) that may cause the underlying model to overfit to the training data.
Theoretically, within \SMOG, this is equivalent to \emph{quenching} the system to the data (a term analogous to quickly cooling a physical system, frezing in any defects).
In contrast, when one \emph{anneals} a physical system, one heats up and cools it down slowly, and repeatedly, thereby removing any defects (of data anomalies for NNs, or material defects in a physical system).

In \STATMECH, one can perform a so-called quenched average using a replica calculation,
effectively removing the dependence on test and/or training data
from the final estimate for $\AVGEMPGE$.
However, the theoretical quenched result may differ significantly from the annealed case when the underlying model is unrealizable~\cite{SST92}. 
This may occur when the training data is very noisy and/or the model architecture is such that it can not correctly predict all the training labels.
In such cases, the model will always have some finite, non-zero Average \TrainingError, $\AVGTE > 0$,
even in the \LargeN limit of infinite data $\ND=\infty$. In such a case, this indicates
a highly complex error landscape with many local minima separated by extremely high barriers,
and a slowing down of the dynamics.%
\footnote{In modern ML parlance, one might say the model can not be evaluated at interpolation, although 
in practice such a model might have a zero empirical \TrainingError since it may overfit the specific training data.}

While it is commonplace to train ensembles and/or use cross-validation when training small models (as the above discussion assumes),
this could be extremely expensive and impractical in modern ML, e.g., for very large models like Large Language Models (LLMs).
For such massive NNs, one needs a theory that can detect anomalies in training directly from observations during and/or after training.
This is a hallmark of the \SETOL approach, and it distinguishes \SETOL from the classic \STATMECH approach.

\
\paragraph{Generalization Gap vs. Model Quality.}
\label{sxn:SMOG_main-model_quality}
We should distinguish between what is typically done in the \SLT literature versus the \STATMECH approaches.
In \SLT, one is typically interested in modeling the \emph{\GeneralizationGap}.
The \GeneralizationGap quantifies the difference between a models performance on training data versus unseen test data:
\begin{align}
  \label{eqn:gen_gap}
  \mathcal{E}^{emp}_{gap}:= \TTE[\DXtrain]- \TGE[\DXtest]
\end{align}
In contrast, in \STATMECH approaches, one considers the \ModelGeneralizationError directly,
which is sometimes called the \ModelQuality in the \SLT literature.
Model quality is an indication of the models accuracy, precision, recall, or any other relevant metric based on the task at hand.

While related, in developing an analytic theory, the \GeneralizationGap and
the \ModelQuality (or \ModelGeneralizationError) require conceptually different approaches.
This is because the \GeneralizationGap depends on a specific realization of the training data,
whereas our \ModelGeneralizationError will be formulated on a random training data set
(and then corrected later with empirical data).
In this sense, any theory of the \GeneralizationGap  requires a formalism where the
predicted model error is \Quenched to the training data, which is not what we want.
In contrast, the \ModelGeneralizationError  will be formulated using the \AnnealedApproximation (AA),
and is therefore both conceptually and mathematically simpler.
Notably, the \HTSR model \Quality metric trend well when compared to the \ModelGeneralizationError ,
whereas methods from the \SLT literature tend to only describe the \GeneralizationGap well.\cite{YTHx23_KDD}

\paragraph{What about Grokking?} There could be a scenario where the Teacher $T$ is trained to intepolation, achieving perfect training accuracy, and, yet, the test accuracy remains low;
This scenario happens in Grokking,  in both the \emph{pre-Grokking} and \emph{anti-Grokking} 
phases, and has been analyzed using the \HTSR theory.\cite{prakash2025grokking}  
In the pre-Grokking case,  the \Teacher has not properly converged, and has one or more
layer $\alpha\gg 2$, whereas in the anti-Grokking phase, the layers have may be overfit
or overcorrecting, with $\alpha\ll 2$. In either case, we might argue that \Teacher has not 
achieved \Thermodynamic equilibrium, and a naive ST protocol may fail!  \emph{This and other complex situations motivate why it is necessary to have a robust theory that
can characterize the internal states of the NN layers, and without needing access to test or training data.}

\subsubsection{Theoretical Student-Teacher Average Generalization Error \texorpdfstring{$(\AVGSTGE)$}{(AVGSTGE)}}
\label{sxn:SMOG_main-st_av}

Here, we seek a simple, formal expression for the
\StudentTeacher Average \GeneralizationError, $\AVGSTGE$,
that can be used as the starting point for our extended \SemiEmpirical theory.

\paragraph{The Idealized Data.}
To develop a \SemiEmpirical theory of the \Teacher \GeneralizationError, $\TGE^{T}$, 
instead of training and evaluating a NN model using real data $(\DX)$,
we seek a simple, analytical expression with parameters that can be fit to empirical measurements.
So in addition to using a model for our NN, we must specify a idealized model for the data.
In a real NN, the data $\DX$ is correlated, and, in fact, very strongly correlated;
and this is reflected in the layer weight matrices.
However, to be tractable, our starting theoretical expressions use uncorrelated (i.i.d) data.
Formally, we must replace the correlated data 
with some uncorrelated, random model of the data, i.e., $\XVEC\rightarrow\XI$.
As described in Figure~\ref{fig:data_mapping},
our \DataModel is a standard Gaussian $N(0,\sigma^{2}\mathbb{I})$ model for the input data
\begin{align}
\DATA\rightarrow\XImu,\;\;\XImu\in N(0,\sigma^{2}\IMm) ,
\label{eqn:mwm_replace_1}
\end{align}
where $N(0,\sigma^{2}\IMm)$ denotes a Gaussian distribution with zero mean and variance $\sigma^{2}=\tfrac{1}{m}$,
\and $\XImu$ is normalized such that $\Vert\XImu\Vert^{2}:=\sum_{i=1}^{m}(\XImu)^{2}_{i}=1$ for all $\ND$ data vectors.  The full idealized sample is denoted $\NDXI$, and is an $n\times m$ matrix, or, equivalently, a $n$-dimensional vector  of $m$-dimensional vectors $\XImu$. 

We make this distinction between actual $\ADD$ and idealized data $\MDD$ to emphasize that,
later, we will use our so-called \SemiEmpirical procedure to
account for the real correlations in the actual data phenomenologically
by taking some analytical parameter of the theory and fitting it to the real world observations,
here, on the ESD of the NN weight matrices.

\paragraph{The ST Error Model and the Annealed Potential $\EPSLSTx$.}
We now model \Teacher error $\AVGGE^{T}$ with the
\emph{\AverageSTGeneralizationError} $\AVGSTGE$, which is obtained
 by \emph{first} computing the ST error function
$\DETOPSTL$
over the set of \emph{all} possible $\ND$ input examples $\XI$.  Define the data-dependent ST test error function--or Energy--as 
\begin{align}
\label{eqn:DE_L}
\DETOPSTL:=\sum_{\mu=1}^{\ND}\mathcal{L}[\NNOUT(\SVEC,\XI_{\mu}),\NNOUT(\TVEC,\XI_{\mu})]  .
\end{align}
where $\mathcal{L}(\SVEC,\TVEC, \XImu)$ is simply the $\ell_2$ loss.  This measures the error
between the \Student and the \Teacher; it is zero when their predictions are identical,
$(\Ys=\Yt)$,  and is nonzero otherwise.

We aim to derive a simple expression for the  \AverageSTGeneralizationError, $\AVGSTGE$, and to do this, 
we define the  \EffectivePotential for the data-averaged ST \GeneralizationError $\EPSL(\SVEC,\TVEC)$, as in \EQN~\ref{eqn:epsl}, as:
\begin{align}
\label{eqn:STerror}
\EPSLSTx = \langle\DETOPSTL\rangle_{\AVGNDXI}:=\frac{1}{\ND}\int d\mu(\NDXI)\DETOPSTL
\end{align}
The measure $d\mu(\NDXI)$ will end up being a Gaussian measure over $\ND$ samples
(see Appendix~\ref{sxn:summary_sst92}), and the intent is to evaluate it
in the \LargeN limit in $\ND$, thereby sampling all possible inputs in the idealized data space, $\NDXI\in\MDD$.

As in Section~\ref{sxn:mathP}, by applying the AA, we can rewrite the \AverageSTGeneralizationError, $\AVGSTGE$:
first, a simple average over all the possible inputs $\NDXI$; and, 
second, then as a Thermal average over all Students $S$, in the AA, and at high-T 
\begin{align}
\label{eqn:MGE}
\AVGSTGE:=\THRMAVG{\EPSLSTx} .
\end{align}
(Recall that in this regime, $\AVGSTGE=\AVGGE^{an,hT}$.)

In the classic \STATMECH approach, the average $\THRMAVG{\cdots}$ is
a \ThermalAverage in the canonical ensemble with $\beta$ fixed,
as explained in Section~\ref{sxn:mathP}.  Here, we will do something similar, as the \Student
average $\THRMAVG{\cdots}$ will be computed from the associated
generating function $\IZG$ for the matrix-generalized case  (an HCIZ integral defined over all students,
and in both the \LargeN \Thermodynamic limit in $\ND$, and the \WideLayer \LargeN  limit in $N$).

Recall that above, the empirical estimate for $\AVGEMPGE$ depended on a
specific instantiation of the model for the training data $\DATAtrain$,
i.e  $\AVGSTGE$ is \Quenched to the training data.
For that reason, for the final result, we needed to take a second,
quenched average over all possible data sets.
Here, we do not need to consider this and always work in the \AnnealedApproximation(AA).
This is because we incorporate
the specific effects of the real-world training data $(\NDX)$ after we derive our formal expressions
by fitting the model parameters to empirical data.
The final expression for $\AVGSTGE$, derived below,
will be generalized to $\AVGNNGE$, matrix-generalization of  the classic \STATMECH formula
for the \LinearPerceptron, in the \Annealed and High-T approximations.
(see Appendix~\ref{sxn:summary_sst92}). 

\paragraph{The Annealed Potential as a function of the overlap $(\EPSL(\AVGR))$.}

We want an expression for the data average of the ST test error, from \EQN~\ref{eqn:STerror}, generalized from \Perceptron vectors to NN layer weight matrices.
For the \Perceptron, one obtains different expressions for the ST error function, depending on 
the type of activation function $h(x)$ in \EQN~\ref{eqn:dnn_energy};
The simplest are the Linear and Boolean \Perceptrons, and
for both (and with $\ell_2$ loss),
 $\EPSL(\SVEC,\TVEC)$ is simply a function of the ST overlap $\AVGR$~\cite{SST92}.
This gives $\EPSLSTx\rightarrow\EPSL(\AVGR)$, where
\begin{align}
  \label{eqn:Rdef}
\AVGR=\SVEC^{\top}\TVEC=\sum_{i=1}^{m}s_{i}t_{i}, \;\;\AVGR\in[0,1].
\end{align}
which is simply the dot product between the $m$-dimensional \Student $\SVEC$ and \Teacher $\TVEC$ weight vectors.
 Notice that the data vectors are normalized such that $\Vert\SVEC\Vert^{2}=\Vert
\TVEC\Vert^{2}=1$ and 
we assume there are \emph{no inversely-correlated} Students--Students that predict more than half the Teacher labels incorrectly. And, importantly, the number of free parameters becomes $1$.

For a \LinearPerceptron~\cite{SST92},%
\footnote{In the classic approach for the ST model, the theory examined different expressions $\EPSL(\AVGR)$.
For example, one can consider the  Boolean \Perceptron~\cite{SST92,Ros62}, with activation function $h(x)=\mbox{sgn}(x)$, 
i.e., the Heaviside step function. Then, the error is
$
\EPSL(\AVGR)=1 - \dfrac{1}{\pi}\arccos(\AVGR).
$
In both cases, perfect learning occurs when $R=1$~\cite{SST92}.
}
with activation function $h(x)=x$,  we can obtain the error function as
\begin{align}
\EPSL(\AVGR)=1-\AVGR,\;\;\AVGR\in[0,1].
\label{eqn:LinearPerceptronError}
\end{align}

\begin{figure}[ht]
\centering

\begin{minipage}{0.48\linewidth}
\centering
\begin{tikzpicture}[scale=1.5]

    \draw (0, 0) circle (1.5cm);

    \draw[->,thick,red]  (0,0) -- (60:1.5cm) node[pos=0.7, above right] {\(\mathbf{s}\)};
    \draw[->,thick,blue] (0,0) -- (90:1.5cm) node[pos=0.75, above left] {\(\mathbf{t}\)};

    \draw[<-] (0,1.125) arc (90:60:1.125cm) node[pos=0.7, below left] {\(\theta\)};

    \draw[thick,blue] (0,0) -- (-1.5,0);
    \draw[thick,blue] (0,0) -- ( 1.5,0);

    \shade[left color=gray!50, right color=gray!10]
      (0,0) -- (-1.49,0) arc (180:150:1.5cm) -- cycle;
    \shade[left color=gray!10, right color=gray!50]
      (0,0) -- (1.49,0) arc (0:-30:1.5cm) -- cycle;

    \draw[thick,red]  (0,0) -- (150:1.5);
    \draw[thick,red]  (0,0) -- (330:1.5); 

    \node at (0,-1.75) {\( \AVGR =\mathbf{s}^\top \mathbf{t} \)};

\end{tikzpicture}
\end{minipage}%
\hfill
\begin{minipage}{0.48\linewidth}
\centering
\begin{tikzpicture}[scale=1.5]

    \shade[ball color=gray!5, opacity=0.3] (0,0) circle (1.5cm);

    \begin{scope}
      \clip (0,0) circle (1.5cm);
      \fill[red!50, opacity=0.5]
        [rotate around={-15:(0,0)}, xscale=1.5, yscale=0.4]
        (0,0) circle (1.0);
    \end{scope}
    \draw[thick, red]
      [rotate around={-15:(0,0)}, xscale=1.5, yscale=0.4]
      (0,0) circle (1.0);

    \begin{scope}
      \clip (0,0) circle (1.5cm);
      \fill[blue!50, opacity=0.5]
        [rotate around={0:(0,0)},    
         xscale=1.5, yscale=0.5]     
        (0,0) circle (1.0);
    \end{scope}
    \draw[thick, blue]
      [rotate around={0:(0,0)}, xscale=1.5, yscale=0.5]
      (0,0) circle (1.0);

    \draw[->,thick,red]  (0,0) -- (1.0,1.2)
      node[pos=0.8, above left]  {\(\mathbf{S}\)};
    \draw[->,thick,blue] (0,0) -- (0,1.8)
      node[pos=0.7, left] {\(\mathbf{T}\)};


    \begin{scope}
      \clip (0,0) circle (1.5cm);
      \fill[purple!30, opacity=0.4]
        (0,0)
         -- (15:1.5cm)
         arc [start angle=15, end angle=40, radius=1.5cm]
         -- cycle;
    \end{scope}

    \draw[thick, dashed, purple!70!black] (0,0) -- (15:1.5cm);
    \draw[thick, dashed, purple!70!black] (0,0) -- (40:1.5cm);

    \draw[->, thick, purple!80!black]
      (20:0.4) arc[start angle=20, end angle=35, radius=0.4];
    \node[purple!80!black] at (27:0.65) {\(\theta\)};

    \node at (0,-1.75) {\( \AVGR = \frac{1}{N} \,\mathrm{Tr}\bigl(\mathbf{S}^\top \mathbf{T}\bigr) \)};

\end{tikzpicture}
\end{minipage}

\caption{Comparison of 2D and 3D representations of the vector and matrix Student--Teacher overlap \(R\).
\textbf{Left:} \(\AVGR = \mathbf{s}^\top \mathbf{t}\) (and  normalized s.t. $\Vert\SVEC\Vert^{2}=\Vert\TVEC\Vert^{2}=1$).
\textbf{Right:} \(R = \tfrac{1}{N}\,\mathrm{Tr}\bigl(\mathbf{S}^\top\mathbf{T}\bigr)\) with conic sections on the sphere (red \(\mathbf{S}\), blue \(\mathbf{T}\)), plus a purple wedge for the angle.  Averaged over matrix dimension $N$ (and implicitly normalized over to $1/M$).
Note that $R\in[0,1]$, where $R=1$ indicates Student perfect learning (and $R\ge0$ because there no Students that predict more than half the Teacher labels incorrectly).
}
\label{fig:overlaps}
\end{figure}
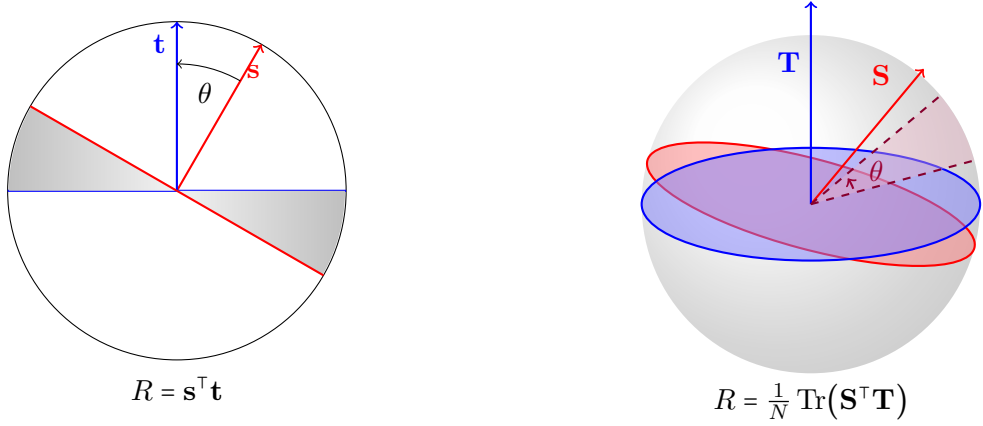


\paragraph{Derivation of the ST error $(\EPSL(\AVGR))$ for the Linear Perceptron.}
To derive \EQN~\ref{eqn:LinearPerceptronError},
define the data-dependent ST error (\EQN~\ref{eqn:DE_L}) in terms of an $\ell_2$ loss function.

Write the data-averaged ST error as
\begin{align}
\DETOPSTLL= & \frac{1}{2} \sum_{\mu=1}^{\ND}(\Ys - \Yt)^{\top} (\Ys - \Yt)
\label{eqn:deriveSTerror}
\end{align}
Define the $\ND$ label vectors  
$\YsVEC:=[\YsIS{1}, \YsIS{2}, \cdots, \YsIS{\ND}]$ and 
$\YtVEC:=[\YtIS{1}, \YtIS{2}, \cdots, \YtIS{\ND}]$.
This lets us write the total Energy as
\begin{align}
\nonumber
\DETOPSTLL= & \frac{1}{2} \Trace{(\YsVEC - \YtVEC)^{\top} (\YsVEC - \YtVEC)} \\
\nonumber
=& \frac{1}{2} \Trace{(\YsVEC)^{\top} \YsVEC - 2 (\YsVEC)^{\top} \YtVEC + (\YtVEC)^{\top} \YtVEC } \\
\nonumber
=& \ND - \Trace{(\YsVEC)^{\top} (\YtVEC)}\\
=& \ND- \Trace{\ETA(\SVEC,\TVEC,\NDXI)},
\label{eqn:deriveSTerror2}
\end{align}
where we define the \emph{data-dependent \SelfOverlap}:
\begin{equation}
\ETA(\SVEC,\TVEC,\NDXI):=(\YsVEC)^{\top}(\YtVEC)
\end{equation}

The expression $\ETA(\SVEC,\TVEC,\NDXI)$ is analogous to the ST overlap $R$, but before the data has been integrated out.
It is convenient to work directly with
the \SelfOverlap $\ETA(\SVEC,\TVEC,\NDXI)$ because it will appear later in \EQN~\ref{eqn:eta_mat_avg_def} (in Section~\ref{sxn:matgen}), 
when formulating the matrix-generalized overlap operator~$\OVERLAP$.

In defining $\ETA(\SVEC,\TVEC,\NDXI)$, we replace the individual labels $(\Ys,\Yt)$ with the Energy functions $\NNOUT$ that generate them, giving an expression in terms of the weights $(\SVEC,\TVEC)$ and the Gaussian data variables $(\XI)$. We will then integrate out the data variables, leaving an expression just in terms of the weights.  
Using the $\NNOUT$ Energy generating or output function (\EQN~\ref{eqn:dnn_energy}, \ref{eqn:S_ENN}, \ref{eqn:T_ENN}), we can write the labels as
\begin{align}
\Ys=\SVEC^{\top}\XI_{\mu},\;\;
\Yt=\TVEC^{\top}\XI_{\mu}  .
\end{align}
This now gives the data-dependent \SelfOverlap explicitly as
\begin{align}
  \label{eqn:eta_vec_xi_def}
\ETA(\SVEC,\TVEC,\NDXI) =\sum_{\mu=1}^{\ND} (\SVEC^{\top}\XI_{\mu})^{\top} (\TVEC^{\top}\XI_{\mu}) = \sum_{\mu=1}^{\ND}\XI^{\top}_{\mu}  (\SVEC^{\top} \TVEC )\XI_{\mu} 
\end{align}
After integrating over the data, we have the \emph{data-independent \SelfOverlap}, $\ETA(\SVEC,\TVEC)$:
\begin{align}
\ETA(\SVEC,\TVEC) := \langle\ETA(\SVEC,\TVEC,\NDXI)\rangle_{\AVGNDXI}
   = & \frac{1}{\ND}\int d\mu(\NDXI)\ETA(\SVEC,\TVEC,\NDXI) \nonumber \\
= & \frac{1}{\ND}\int d\mu(\NDXI)\sum_{\mu=1}^{\ND}\XI_{\mu}^{\top}\SVEC^{\top}\TVEC\XI_{\mu} \nonumber \\
= & \frac{1}{\ND}\int d\mu(\NDXI)\sum_{\mu=1}^{\ND}\XI_{\mu}^{\top}\AVGR\XI_{\mu} \nonumber \\
= & \AVGR\;\frac{1}{\ND}\sum_{\mu=1}^{\ND}\int d\mu(\XI_{\mu})\XI_{\mu}^{\top}\XI_{\mu} \nonumber \\
   = &\AVGR. 
      \label{eqn:eta_vec_avg_def}
\end{align}

where the third equality holds because $\AVGR$ is a scalar constant, and the fourth holds because the elements of $\XI$ are i.i.d. and normalized to unit variance.
(See Section~\ref{app:st-gen-err-annealed-ham}) .

We can now obtain the \EffectivePotential $\EPSL(\AVGR)$ for the data-averaged ST test error (see \EQN~\ref{eqn:epsl}),  as in \EQN~\ref{eqn:LinearPerceptronError},
\begin{equation}
\label{eqn:epslR}
\EPSL(\AVGR)=\langle\DETOPSTLL\rangle_{\AVGNDXI} =  1 - \AVGR.
\end{equation}

In traditional \STATMECH (e.g., \cite{SST92}), one is interested in how the \emph{Total 
\GeneralizationError} $\TGE(\AVGR)$ depends on $\AVGR$.
With these simple error functions, \EQN~\ref{eqn:MGE} reduces to a function over $\AVGR$,
and the \AverageSTGeneralizationError $\STGE(\AVGR)$ is then obtained by taking a \emph{\ThermalAverage} over the Students 
\begin{align}
\label{eqn:AVGSTGE_R}
\AVGSTGE(\AVGR)=\THRMAVG{\EPSL(\AVGR)}=
\THRMAVG{1-\langle\ETA(\SVEC,\TVEC,\NDXI)\rangle_{\AVGNDXI}}=
\THRMAVG{1-\ETA(\SVEC,\TVEC)}=
\THRMAVG{1-\SVEC^{\top}\TVEC}=
\THRMAVG{(1-\AVGR)}  ,
\end{align}
where $\THRMAVG{\cdots}$ is a \ThermalAverage over the \Student weight vector $\SVEC$.
Recall that even at high-T, this restricts the Students to ones trained with a similar
amount of regularization as the Teacher (i.e. the measure  $d\mu(\SVEC)=d\SVEC$ is independent 
of $\beta$).

The average \Quality for the ST \Perceptron, $\Q^{ST}$,
is just the \AverageGeneralizationAccuracy, so we can write
\begin{align}
\label{eqn:QST_final}
\Q^{ST} := 1 - \AVGSTGE(\AVGR) 
       = \THRMAVG{1 - \EPSL(\AVGR)} 
       = \THRMAVG{\langle\ETA(\SVEC, \TVEC, \NDXI)\rangle_{\AVGNDXI}} 
       = \THRMAVG{\ETA(\SVEC, \TVEC)} 
       = \THRMAVG{\SVEC^{\top}\TVEC} 
       = \THRMAVG{\AVGR}.
\end{align}
\EQN~\ref{eqn:QST_final} is the starting point for deriving a \SEMIEMP theory for the \WW quality metrics (\ALPHA,\ALPHAHAT);
see Section~\ref{sxn:matgen_mlp3}.
To generalize this expression, we will start with the \SelfOverlap $\ETA(\SMAT,\TMAT,\NDXI)$ for a
\MultiLayerPerceptron (MLP3) in Section~\ref{sxn:matgen}.

Before doing this, however, we note that we can obtain this expression for $\AVGSTGE$ by defining the
\AnnealedHamiltonian $\HANHT(\AVGR)$, at high-Temperature, as in Section~\ref{sxn:mathP}, \EQN~\ref{eqn:Gan_highT}.
Indeed, it is really $\HANHT(\AVGR)=\EPSL(\AVGR)$ that we must generalize to the matrix case 
in the next section, which we do (using a technique
similar to a Replica calculation, but still in the AA).

\paragraph{Final Remarks.}
Additional results are provided in Appendix~\ref{sxn:summary_sst92}. In particular, 
In Appendix~\ref{app:st-gen-err-annealed-ham}, we use the ideas from this section  and the previous one to derive the full non-linear vector form of the \AnnealedHamiltonian $\GAN$ (\EQN~\ref{eqn:Gan_def})
for the \LinearPerceptron, in the AA.
Then, in Appendix~\ref{app:st-gen-err-annealed-ham}, we derive the matrix generalization
$\GANR\rightarrow\GANMAT$ of these quantities using
a \LargeN in $n$ expansion (as opposed to a full replica calculation).
This derivation explains how to obtain the implicit $1/M$ normalization for
the weight matrices $\WMAT$ that is necessary for the final results in Section~\ref{sxn:matgen}.
(Notice this is similar to the implicit normalization $1/m$ on the ST vectors)
Then, taking the case $N=1$, we can recover the original result for $\GANR$.
We can then express the Annealed Hamiltonian at High-T, i.e.,
$\GANHTR=\EPSLR$ (\EQN~\ref{eqn:Gan_highT}).
This shows that  our new derivation is consistent with the full  and the High-T results for
classic ST \Perceptron model, for a Linear Perceptron in the AA.~\cite{SST92}

\begin{table}[t]
  \raggedright
\hspace*{-1.5cm}
\renewcommand{\arraystretch}{1.25} 
\begin{tabular}{|c|c|c|c|}
  \hline
  Quantity & Traditional \SMOG & \makecell{\LinearPerceptron \\ in Traditional \SMOG} & \makecell{Matrix Generalization \\ for \SETOL} \\ \hline

  Total (Idealized) Data Error 
    & $\DETOPXI$ (\ref{eqn:detox})
    & $\DETOPSTL$ (\ref{eqn:deriveSTerror}) 
    & $\DETOPNN$ (\ref{eqn:DE2}) \\ \hline

   Annealed Hamiltonian
    & $\HANHT=\EPSLw$ (\ref{eqn:epsl}) 
    & $\GANHTR=\EPSLSTx=1-\AVGR$ (\ref{eqn:epslR}) 
  & $\GANMATHT = N(\IM-\OVERLAP)$ (\ref{eqn:GANHTmatR}) \\

  (Data-Averaged Error) 
    & (AA, at high-T) 
    & (and at \LargeN) 
    & (only for a layer)  \\ \hline

    \SelfOverlap 
    & $\ETAw = 1-\EPSLw$~(\ref{eqn:def_eta})

    & $\ETA(\SVEC,\TVEC)=\SVEC^{\top}\TVEC$ (\ref{eqn:eta_vec_avg_def})
    & $\ETA(\SMAT,\TMAT)=\tfrac{1}{N}\SMAT^{\top}\TMAT$ (\ref{eqn:eta_mat_avg_def})  \\ \hline
    \hline

  \ModelQuality 
    & $\Q:=1-\AVGGE$ 
    & $\Q^{ST}:=1-\AVGGE^{ST}$ (\ref{eqn:model_qualities})
   & $\Q^{NN}:=1-\AVGGE^{NN}$  (\ref{eqn:model_qualities})\\ 

  in terms of \LayerQuality
    & 
    & 
   & $\Q^{NN}:=\prod_{L} \Q^{NN}_{L}$ \\ \hline
\end{tabular}
\caption{Summary of key quantities compared across traditional \SMOG models,  the \Student-\Teacher (ST) \LinearPerceptron--in the \AnnealedApproximation
(AA) and at high-Temperature (high-T) and at \LargeN in $\ND$, and the matrix-generalized forms as the starting point to frame \SETOL.
The total ST Error of Energy, $\DELBF$, represents the difference (squared) between the model and its labels for the ST model between
the \Student and \Teacher predictions.
The \AnnealedHamiltonian is the Energy function for this Error after it is averaged over the model for the training data
(an $\ND$-dimensional i.i.d. idealized Gaussian dataset,  $\NDXIn$).
In the AA, the \AnnealedHamiltonian is equal to the \EffectivePotential.  For the ST model,  this is one minus the average overlap, $\HANHT(R)=(1-\AVGR)$;
for the \SETOL, this is  the ($M$-dimensional) identity minus the overlap operator/matrix, $\HANHT(\OVERLAP)=N(\IM-\OVERLAP)$. 
The \SelfOverlap $\eta(\cdots)$ is used to describe the Accuracy (as opposed to the Error) for both the ST model and
its matrix-generalized form.
Finally, the different forms of the \Quality are defined.  Generally speaking, the \Quality $\Q$ is an approximation to some measure
of $1$ minus the \AverageGeneralizationError, $\Q:=1-\AVGGE$ (in the AA, at high-T, at \LargeN, and with whatever else
approximations are applied).
For the ST model, having just 1 layer, the \ModelQuality and the \LayerQuality are the same, and denoted $\Q^{ST}$.
For \SETOL, the \ModelQuality $\Q^{NN}$ is a product of individual \LayerQualities $\Q^{NN}_{L}$.
(Note that the  final \SETOL \LayerQuality $\Q$ is defined in terms of the \LayerQualitySquared $\QT$,
and the starting point for this is expressed with the \LayerQualitySquared Hamiltonian $\HBARE=\OLAPTOLAP$.
}
\label{table:quantities_general_vect_matrix}
\end{table}

\clearpage

\newpage
\section{\SemiEmpirical Theory of the \HTSR Phenomenology}
\label{sxn:matgen}

In this section, we present the main technical elements of our \SemiEmpirical Theory of Deep Learning (\SETOL).
Our goal is to explain and, where possible, derive the \HTSR PL metrics \ALPHA $(\alpha)$ and \ALPHAHAT $\hat{\alpha}$
from first principles, and, in doing so, also present the \TRACELOG condition and
newly proposed \WW\DETX metric.
To do this, we introduce a Matrix Generalization of the Student-Teacher model for a Linear \Perceptron
(See Section~\ref{sxn:SMOG_main-st_av}), 
adapted here for a (3-layer) \MultiLayerPerceptron  (MLP3).
We seek a theory for the \LayerQuality $\Q=\Q^{NN}_{L}$ of a NN, where
this \LayerQuality now corresponds to the (approximate) contribution each layer makes to the total
generalization accuracy, or total \Quality $\Q^{NN}$.
For technical reasons, we actually seek a formal expression(s) for the \LayerQualitySquared,
$\QT\approx(\Q^{NN}_{L})^2$. 
We say that the \SETOL is \SemiEmpirical because the final result $\QT$ is expressed directly in terms
of the empirically observable spectral properties of the Teacher layer weight matrix $\TMAT=\WMAT$.


\begin{enumerate}[label=5.\arabic*]
\item
\textbf{Matrix Generalization of the ST Model.}
Section~\ref{sxn:matgen_mlp3} generalizes
classical \STATMECH vector-based ST model of Section~\ref{sxn:SMOG_main-student_teacher}
to obtain a \LayerQuality for a single layer in an NN.
It starts by first formulating the learning problem for
the NN generalization accuracy or quality, $\Q^{NN}$,
of a 3-layer MLP (MLP3).
We then replace vectors with $N \times M$ matrices $\SVEC,\TVEC\rightarrow\SMAT,\TMAT$,
and obtain and expression for the NN \SelfOverlap $\ETA(\SMAT,\TMAT,\NDXI)$,which then gives a matrix-generalized overlap operator
$\OVERLAP:=\langle\ETA(\SMAT,\TMAT,\NDXI)\rangle_{\AVGNDXI}=\tfrac{1}{N}\SMAT^{\top}\TMAT$.
This can be related to a single-layer matrix-generalization of the ST \AnnealedHamiltonian, 
presented in Appendix~\ref{sxn:summary_sst92},
$\HANHT(\OVERLAP):=N(\IM-\OVERLAP$),
where, importantly, the scalar overlap $R$ is now a matrix $\OVERLAP$ of $M\times M$ adjustable parameters.
Importantly, being empirical quantities, the weight matrices are implicitly normalized by $1/\ND$, and, like the ST vectors, also implicitly normalized by $1/M$. (See Appendix~\ref{sxn:appendix_Gan}).

\item
\textbf{The \LayerQualitySquared $\QT$.}
Section~\ref{sxn:matgen_quality_hciz} presents the expression for NN \LayerQualitySquared $\QT$.
Following the ST analogy, we define a \ThermalAverage over possible \Student weight matrices $\SMAT$
for the matrix overlap, giving $\Q^{NN}_{L}:=\THRMAVGIZ{\HANHT} =\THRMAVGIZ{\OVERLAP}$.
For technical reasons, however, we actually seek the (approximate)  \LayerQualitySquared, $\QT\approx (\Q^{NN}_{L})^2$,
defined as $\QT:=\THRMAVGIZ{\OLAPTOLAP}$.
To evaluate $\QT$, rather than sampling all random \Student matrices $\SMAT$ directly,
we switch measures to the (Outer) \Student Correlation matrices\footnote{
We adopt the convention ``Inner'' for smaller, $M\times M$ full rank \Student correlation matrices, and ``Outer'' for larger, $N\times N$ rank-deficient \Student Correlation matrices.
Note that both are scaled as $1 / N$, meaning that their nonzero eigenvalues are the same.
} 
$\AMATN = \tfrac{1}{N}\SMAT\SMAT^{\top}$ along with their Inner counterparts
$\AMATM = \tfrac{1}{N}\SMAT^{\top}\SMAT$.
Importantly, we argue that the measures $d\mu(\AMATM)\leftrightarrow d\mu(\AMATN)$,
can be interchanged for our purposes, making them effectively equivalent.
This reparameterization leads us to an integral of the HCIZ type (as in \EQN~\ref{eqn:hciz_prelim})
which, as shown by Tanaka \cite{Tanaka2007, Tanaka2008}, is expressed in terms of the \RTransform $R_{\AMATM}(z)$, defined for the tail of the limiting form of the ESD of $\AMATM$,  $\rho^{\infty}_{\AMATM}(\lambda$).

Then, we introduce the \EffectiveCorrelationSpace (\ECS), and two key approximations,
the \IndependentFluctuationApproximation (\IFA) and
the Exact Renormalization Group (\TRACELOG) Condition.
We impose the \IFA (described below) because it is necessary for the final result.
The \TRACELOG condition states that the determinant of the (effective) \Student correlation matrix is unity, $\Det{\AECS}=1$.
Critically, this condition can be tested empirically by assuming the (effective) \Teacher correlation matrix
also follows the testable \TRACELOG condition, $\Det{\XECS}=1$. \textbf{The empirically testable \ERG condition is a key result of this work}.

\item
\textbf{The \WideLayer \LargeN limit in $N$.}
Section~\ref{sxn:matgen_evaluation_hciz} presents the core result,
(as in \EQN~\ref{eqn:QT_result}),
closed-form or semi-analytic expression(s) for the \LayerQualitySquared $\QT$
formed in the \WideLayer \LargeN limit in $N$.
Restricted to the \ECS, and under the \TRACELOG condition and the \IFA, our
HCIZ integral for $\QT$ becomes tractable at large-$N$, giving an expression that can be parameterized
in terms of $\MECS$ eigenvalues $\LambdaECS$ of the \Teacher correlation matrix 
restricted to the \ECS $\XECS$.
In doing this, the $\MECS$ \Teacher eigenvalues are treated as experimental observables, and 
become the effective \SemiEmpirical parameters (i.e., $\alpha$, $\LambdaMax$) of the \SETOL approach.

\item
\textbf{Selecting the \HeavyTailed \RTransform.}
Section~\ref{sxn:r_transforms} presents several models of different \RTransforms.
Evaluating $\QT$ requires evaluating selecting an \RTransform $R(z)$ for the Teacher Empirical Spectral Density (ESD),
and also ensure that it is analytic and single-valued on the domain of interest-- the \ECS and/or tail of the ESD.
We examine four possible models for $R(z)$: \emph{(i)} the \emph{Bulk$+$Spikes} (BS),
\emph{(ii)} the \emph{Free Cauchy} (FC) model, 
\emph{(iii)} the \emph{\InverseMP} (IMP) model, 
and \emph{(iv)} the  \LevyWigner (LW) model.
First, as a trivial case, the tail of ESD can be treated as a collection of spikes,
and the ESD is simply a sum of Dirac delta functions; in this case,
$\QT$ becomes a ``Tail Norm'', the Frobenius Norm of the PL tail.
When the layer is \Ideal, i.e., $\alpha\sim 2$ and $\Det{\XECS}\sim 1$,
one can use  \emph{Free Cauchy} (FC) model.  and the resulting \LayerQuality $\Q$
is approximately the Spectral Norm $\LambdaMax$.
Since $\LambdaMax$ increases with decreasing $\alpha$, the FC model yields the \HTSR \ALPHA metric.
One can also use  \emph{\InverseMP} (IMP) model. 
Being a model for the full \Teacher ESD (not just the tail) the IMP \RTransform contains a \emph{branch cut} in the complex plane
which aligns with the start of the \ECS \PowerLaw tail.
Finally, Using the \LevyWigner (LW) model, one can model cases where
$\alpha\le 2$ and derive the \HTSR \ALPHAHAT metric.
\end{enumerate}

\vspace*{1em}

These core elements form a bridge between well-established empirical properties of large-scale NNs 
and a tractable ST-based theory. In the subsequent sections, we formalize the key steps: 
\emph{(i)} setting up the matrix-based ST problem, \emph{(ii)} defining our HCIZ integrals 
over restricted correlation matrices (\ECS), and \emph{(iii)} analyzing the resulting 
\emph{Layer Quality} (or \emph{Quality-Squared}) expressions in the \LargeN limit in $N$.

\subsection{\MultiLayer Setup: MLP3}
\label{sxn:matgen_mlp3}

In this section, we describe the matrix generalization of the ST  model of the \LinearPerceptron; and, 
in particular, a matrix-generalized version of the key quantities we derived in Section~\ref{sxn:SMOG_main-st_av}.

\paragraph{A simple model.}

Consider a simple NN with three layers (two hidden and an output), i.e., a three-layer \MultiLayerPerceptron, denoted as the \emph{MLP3} model.
(This is a \emph{very} simple model of a modern NN with hundreds of layers and complex internal structure.)

Ignoring the bias terms, \emph{Without Loss of Generality}, (WLOG), the NN outputs
 $E_{1\mu},E_{2\mu},E_{3\mu}$ for each layer, as defined in \EQN~\ref{eqn:dnn_energy}, are given by:
\begin{align}
\nonumber
  \EVEC_{1\mu} &:= \frac{1}{\sqrt{N_1}} h (\WMAT_1^\top \boldsymbol{\xi}_{\mu}) , \\
\nonumber
  \EVEC_{2\mu} &:= \frac{1}{\sqrt{N_2}} h (\WMAT_2^\top \EVEC_{1\mu})     , \\
\label{eqn:nflow}
             y_{\mu} := \EVEC_{3\mu} &= \frac{1}{\sqrt{N_3}} h (\WMAT_{3}^\top \EVEC_{2\mu})     ,
\end{align}
where $\EVEC_{\#\mu}$ denotes a vector of energies or NN outputs for that layer, and $h$ is a general function or functional, denoting either a non-linear activation or a more complex layer structure, such as a CNN or an RNN. Note that for binary classification, $\EVEC_{3\mu}=1|-1$ is just a binary number.
We can consider $h(\cdot)$ to be an (unspecified) activation function, but here we simply take $h(x)=x$.

As in \EQN~\ref{eqn:deriveSTerror}, let us specify the ST error, or total Energy, specifically in terms of the $\ell_2$ or MSE loss:
\begin{equation}
\label{eqn:DE}
\DETOPNN = \frac{1}{2}\sum_{\mu=1}^{\ND} (\Ys - \Yt)^2.
\end{equation}

We now develop the matrix generalized form of the self-overlap $\eta$:

\subsubsection{Data-Dependent Multi-Layer ST Self-Overlap \texorpdfstring{$(\ETA(\SMAT,\TMAT))$}{ETA(S, T)}}

Following the same approach in Section~\ref{sxn:SMOG_main-st_av}, it is convenient to rewrite the total Energy $\DELBF$ in \EQN~\ref{eqn:DE} as:
\begin{align}
\label{eqn:DE2}
\DETOPNN
   := \frac{1}{2} \Trace{ (\YsVEC - \YtVEC)^\top (\YsVEC - \YtVEC) }
   = \ND - \Trace{ (\YsVEC)^{\top} \, \YtVEC  }
   = \ND - \ETAMLPXI
\end{align}
where the \SelfOverlap $\ETAMLPXI$
is of the same form as the (vector) \LinearPerceptron (in \EQN~\ref{eqn:deriveSTerror}).
Note that $\ETAMLPXI$ depends on the (idealized) data $\NDXI$
because we have not evaluated the expected value $\langle \cdots \rangle_{\NDXI}$ yet.

Using the general expression from \EQN~\ref{eqn:nflow} for the action of the NN on the input data $\boldsymbol{\xi}$,
we can write the formal expression of the ST error for the simple MLP3 model as:
\begin{align}
\label{eqn:overlap1}
\ETAMLPAVG  
& =\langle\ETA(\SMAT_1,\SMAT_2,\SMAT_3,\TMAT_1,\TMAT_2,\TMAT_3,\NDXI)\rangle_{\AVGNDXI}  \nonumber \\ 
& :=  \frac{1}{\ND}\operatorname{Tr}\left[
    \left\{
    \frac{1}{\sqrt{N_3}} h\left(
        \SMAT_3^{\top} 
        \frac{1}{\sqrt{N_2}} h\left(
            \SMAT_2^{\top} 
            \frac{1}{\sqrt{N_1}} h\left(
                \SMAT_1^{\top} \NDXI 
            \right)
        \right)
    \right)
    \right\}^{\top} \right. \nonumber \\  
& \quad\quad\quad\quad \left. \times
    \frac{1}{\sqrt{N_3}} h\left(
        \TMAT_3^{\top} 
        \frac{1}{\sqrt{N_2}} h\left(
            \TMAT_2^{\top} 
            \frac{1}{\sqrt{N_1}} h\left(
                \TMAT_1^{\top} \NDXI 
            \right)
        \right)
    \right)
  \right]
\end{align}

So far, we have not used any particular assumption on the form of the NN or the data, 
other than that the layer structure used to write the explicit expression for the form eventually needed,
$\ETA(\SMAT,\TMAT)$, a single layer \SelfOverlap.
As a next step, we show which assumptions are needed in order to reformulate the setup as
an effectively a single layer linear model for a NN.

\subsubsection{A Single Layer Matrix Model}
Following others in the literature~\cite{SMG2013_TR}, and for simplicity, one can restrict to the simplifying case that the function $h(x)$ is the identity function.
To evaluate \EQN~\ref{eqn:overlap1}, there are three possibilities.
First, we can multiply all the matrices together, and treat a multi-layer NN effectively as a single layer.
Under this assumption, \EQN \ref{eqn:overlap1} simplifies to
\begin{align}
\label{eqn:overlap2}
  \ETAMLPAVG &= \frac{1}{\ND}\tfrac{1}{N_3 \, N_2 \, N_1} 
  \Trace{ (\NDXI)^\top 
    \SMAT_1 \SMAT_2 \SMAT_3 
    \TMAT_3^{\top} \TMAT_2^{\top} \TMAT_1^{\top} 
    (\NDXI) }.
\end{align}
While this is possible, it would not lead to layer-by-layer insights (as \HTSR-based approaches do).
Second, we could attempt to expand \EQN~\ref{eqn:overlap2} into inter- and intra-layer terms, 
which we could readily do if the $S$ and $T$ matrices were square and the same shape, and then apply Wick's theorem:
\begin{align}
\ETAMLP\approx\frac{1}{\ND}\prod_{l=1}^{L}\tfrac{1}{N_l}\ETAMATAVG=
\frac{1}{\ND}\prod_{l=1}^{L}\tfrac{1}{N_l}\Trace{ (\NDXI)^{\top} \SMAT^{\top}_{l} \TMAT_{l} (\NDXI)) } +\text{intra-layer cross terms}.
\end{align}
However, these matrices are not square, and we don't know how to express the intra-layer cross terms.
Finally, we can simply assume that the individual layers are statistically independent, in which case we can treat each layer independently.
By ignoring the intra-layer cross-terms, let us write the single-layer self-overlap $\ETAMAT$ as:
\begin{align}
  \label{eqn:eta}
  \ETAMAT =
        \ETAMATAVG\rightarrow \
        \tfrac{1}{\ND}\tfrac{1}{N} \Trace{ (\NDXI)^{\top} \SMAT^{\top}_{l} \TMAT_{l} (\NDXI) }  .
\end{align}
This third approach is the one we will adopt.
Moving forward, we will drop the layer subscript, $l$, and we will consider a \SETOL as a single-layer theory.

\subsubsection{The Matrix-Generalized ST Overlap 
\texorpdfstring{$(\ETA(\SMAT,\TMAT)$)}{ETA(S, T)}.}
We can now relate the \SelfOverlap for the NN layer $\ETA(\XI)_{l}$ in \EQN~\ref{eqn:eta}
as a matrix-generalized form of the ST \EffectivePotential $\EPSL(R)=1-R$.
Since we can interpret the Trace as an expected value over the model data $\NDXI$, this gives the desired
\begin{align}
  \label{eqn:eta_mat_avg_def}
  \ETA(\SMAT,\TMAT)
  :=\tfrac{1}{N} \Trace{\SMAT^{\top}\TMAT} 
\end{align}
We have dropped the $1/n$ because (as noted above), the $1/\ND$ will be implicit in the $\TMAT$ and $\SMAT$ matrices because they represent empirical observables. That is, the weights are the learned parameters, effectively averaged over the entire real-world training data set $\ADD$.
This is the matrix generalized form of \EQN~\ref{eqn:eta_vec_avg_def}.

\subsection{Quality Metrics of an Individual Layer as an HCIZ Integral}
\label{sxn:matgen_quality_hciz}

In this subsection, we describe how to generalize the (Thermal) average over the Students $\THRMAVG{\cdots}$ to an 
integral over random \Student matrices, $\THRMAVGIZ{\cdots}$, called an HCIZ integral.

\subsubsection{A Generating Function Approach to Average \QualitySquared of a Layer}
\label{sxn:matgen_quality_hciz_A}

For our matrix generalization, we need to express the \LayerQuality $\Q$ in terms of the data-averaged
\SelfOverlap in \EQN~\ref{eqn:eta_mat_avg_def} for a individual layer.
\begin{itemize}
\item
\textbf{Student-Teacher Overlap  $\mathbf{R}$}
For the vector-based \Perceptron ST model, the data-averaged \SelfOverlap appears in the expression for the
\LayerQuality  in \EQN~\ref{eqn:QST_final}, and is just the average ST vector overlap $R=\SVEC^{\top}\TVEC$. 
We define 
\begin{equation}
\mathbf{R} =  \tfrac{1}{N}\SMAT^{T} \TMAT  .
\end{equation}
The overlap matrix $\OVERLAP$ describes the average interactions between $N$ interacting $M$-dimensional layer (i.e., feature) vectors.  Also, notice that we have dropped $1/\ND$ term.
For the vector-based ST model, the ST \Quality $\Q^{ST}$ in \EQN~\ref{eqn:QST_final} is expressed as
the \ThermalAverage $\Q^{ST}(\AVGR)=\THRMAVG{\AVGR}$. We seek a similar formal expression for the individual matrx-generalized \LayerQuality $\Q^{NN}_{L}$, this time in terms of $\OVERLAP$ and a matrix-generalied \ThermalAverage.
\item
\textbf{Model and Layer Qualities $\Q^{NN}$, $\Q^{NN}_{L}$}
We define the \ModelQuality $\Q^{NN}$ ,
as explained in Subsection~\ref{sxn:htsr-metics}, \EQN~\ref{eqn:ProductNorm},
to be a product of individual NN Layer Qualities $\Q^{NN}_{L}$,
and, as in \EQN~\ref{eqn:model_qualities},
approximates the total NN \AverageGeneralizationAccuracy $(1-\AVGGE^{NN})$:
\begin{equation}
 \Q^{NN}:=\prod_{L}\Q^{NN}_{L}\approx 1-\AVGGE^{NN} 
\end{equation}
The individual $\Q^{NN}_{L}$ expresses the contribution that layer makes
to the approximate total NN \AverageGeneralizationAccuracy.
\item
 \textbf{\LayerQualitySquared $\QT$}
For  technical convenience, however, rather than compute
the NN \LayerQuality $\Q^{NN}_{L}$ directly, we will work with the \emph{\AverageLayerQualitySquared}, 
defined as
\begin{align}
  \label{eqn:QT_1}
  \QT:=\THRMAVGIZ{\OLAPTOLAP}
\end{align}
where  $\THRMAVGIZ{\cdots}$ is now a \ThermalAverage over Student weight matrices $\SMAT$--
an HCIZ integral.

This choice means that the final \LayerQuality $\Q$ we use approximates what would be the
matrix-generalized NN \LayerQuality $\Q^{NN}_{L}$ (above) as
\begin{align}
  \label{eqn:QT_2}
  \Q:=\sqrt{\QT}
  =\sqrt{\THRMAVGIZ{\OLAPTOLAP}} \\ \nonumber
  \approx\THRMAVGIZ{\sqrt{\OLAPTOLAP}} \\ \nonumber
  \approx\THRMAVGIZ{\OVERLAP} \\ \nonumber
  = \Q^{NN}_{L}
\end{align}

\item
 \textbf{Overlap Squared}
 The Overlap operator (squared)  $\OLAPSQD$ is defined  in terms of \EQN~\ref{eqn:A2} such
 that we can 
\begin{align}
  \label{eqn:TraceR2withA2}
  \QT :=\OLAPSQD
  =\tfrac{1}{N^{2}}\Trace{\TMAT^{\top}\SMAT\SMAT^{\top}\TMAT}
  =\tfrac{1}{N}\Trace{\TMAT^{\top}\AMATN\TMAT}  .
\end{align}
This choice, (as opposed to $\Trace{\OVERLAP\OVERLAP^{\top}}$, which would correspond with \EQN~\ref{eqn:A1},) places $\QT$ in the form of the HCIZ integral, as in \EQN~\ref{eqn:hciz_prelim}.
See Appendix~\ref{sxn:tanaka} for a detailed discussion of why we define $\QT$ using the \emph{Outer} Student Correlation matrix $\AMATN$ (defined in Section~\ref{sxn:setol}, and  \EQN~\ref{eqn:A2} below).
\item
 \textbf{\GeneratingFunction}
For the vector-based ST model, we could compute $\Q^{ST}$ using a generating function, $\STG$.
For our matrix generalization, we compute $\Q$ from a \emph{\LayerQualitySquared \GeneratingFunction} $\IZG$, given as
\begin{align}
  \label{eqn:betaIZG_S}
  \IZG:=  \tfrac{1}{N}\ln \int d\mu(\SMAT) \exp\left[\ND\beta N\OLAPSQD\right] .
\end{align}
We normalize the $\IZG$ because we are going to take the \LargeN limit in $N$, and we need to keep the \LayerQuality finite.  This allows us to implicitly fix the effective load of the layer, $\ND/N$, as $\IZG$ is also in the \ThermodynamicLimit and therefore at \LargeN in $\ND$, although we do not explicitly set the layer (feature vector) load $\ND/N$.
See Appendix~\ref{sxn:quality} for the derivation of \EQN~\ref{eqn:betaIZG_S} (and recall the discussion in Section~\ref{sxn:mathP}). 
\end{itemize}

We cannot evaluate \EQN~\ref{eqn:betaIZG_S} directly; but we will be able to evaluate it if we transform it into an HCIZ integral (as in \EQN~\ref{eqn:hciz_prelim}). To do this, however, requires a trick which will allow us to work with different forms of the Student $\AMAT$ (and Teacher $\XMAT$) Correlation matrices.

\paragraph{From weight matrices to Correlation matrices.}
To evaluate $\QT$ in terms of derivatives of \EQN~\ref{eqn:betaIZG_S}, we need to introduce the change of measure:
\begin{align}
  \label{eqn:changeOfMeasure}
  d\mu(\SMAT)\rightarrow d\mu(\AMAT) ,
\end{align}
and then restrict $d\mu(\AMAT)$ to resemble just the generalizing eigencomponents of the \Teacher correlation matrix $\XMAT$.
%
We have two choices for the \Student \CorrelationMatrix $\AMAT$, call them $\AMATM$
(Inner) and $\AMATN$ (Outer), defined as:
\begin{eqnarray}
  \label{eqn:A1}
 \textrm{(Inner) }\;\AMATM&:=& \tfrac{1}{N}\SMAT^{\top}\SMAT \quad\mbox{(which is $M \times M$)} \\
  \label{eqn:A2}
 \textrm{(Outer) }\;\AMATN &:=& \tfrac{1}{N}\SMAT\SMAT^{\top} \quad\mbox{(which is $N \times N$)} .
\end{eqnarray}
Note that $\frac{1}{N}$ is the correct scaling on each of these.
\EQN~\ref{eqn:A1} is consistent with our definition of the layer \CorrelationMatrix, and we use it as the starting point
below to derive the \VolumePreserving \TRACELOG condition (Appendix~\ref{sxn:TraceLogDerivation}).
\EQN~\ref{eqn:A2} is consistent with Tanaka, which requires $\AMAT$ be $N\times N$, but
we still need a \emph{Duality of Measures} to rederive this (Appendix~\ref{sxn:tanaka})~\cite{Tanaka2007, Tanaka2008}.

\paragraph{Duality of measures.}
For either form of $\AMAT$, the measure $d\mu(\AMAT)$ is the same because we will restrict the measures to the~\ECS
space of non-zero eigenvalues $(\lambda_{i}\gg 0)$.
We note that $\AMATM$ and $\AMATN$ have the same eigenvalues $\lambda_{i}$, or ESD,
up to the additional zero eigenvalues $(\lambda_{i}=0)$ in the null space of $\AMATN$.  
Consequently, both forms of $\AMAT$ have 
the same non-zero part
of the ESD $(\rho(\lambda)\gg 0)$, and the same Trace $(\Trace{\AMATM}=\Trace{\AMATN})$.
In the large-N approximation, the ESD of (either form of)
$\AMAT$, $\rho^{\infty}_{\AMAT}(\lambda)$,
becomes continuous (and bounded), but remains zero in the null space.
(Notice that following the physics approach, in the limit, $N$ remains finite but we nevertheless assume a continuous spectrum)
Consequently,
when integrating over the eigenvalues  $\lambda$, one can interchange
$\AMATM$  with $\AMATN$, such that
\begin{align}
 \int d\mu(\AMAT)\;[\cdots]\;\leftrightarrow  \int d\lambda\;[\cdots]\; \rho^{\infty}_{\AMATM}(\lambda) \leftrightarrow  \int d\lambda\;[\cdots]\; \rho^{\infty}_{\AMATN}(\lambda)
\end{align}
This equivalence will be essential both to derive the \TRACELOG condition (Section ~\ref{sxn:TraceLogDerivation}),
and to (re)derive the core result by Tanaka (Section ~\ref{sxn:tanaka}).
Additionally, and WLOG, we may occasionally denote the Student Correlation matrix as
$\AECS$ instead of explicitly using  $\AMATM$ and $\AMATN$.

%

\subsubsection{Evaluating the Average Quality (Squared) Generating Function }
\label{sxn:matgen_quality_hciz_B}

We can write the generating function $\IZG$ 
in \EQN~\ref{eqn:betaIZG_S} 
in terms of $\AMATN$, giving, as in Eqn.~\ref{eqn:QT_dS}, 
\begin{align}
  \label{eqn:IZG_dmuS}
  \IZG = \tfrac{1}{N}\ln \int d\mu(\SMAT)  e^{ \ND\beta N Tr[\tfrac{1}{N}\TMAT^T\AMATN\TMAT]) }  .
\end{align}
To recast \EQN~\ref{eqn:IZG_dmuS} as an HCIZ integral, as in \EQN~\ref{eqn:hciz_prelim},
we must perform a change of measure,
 from $N\times M$ \Student weight matrices $\SMAT$ to $N\times N$ Outer \Student Correlation matrices $\AMATN$, as in \EQN~\ref{eqn:changeOfMeasure}.

When we perform the change of measure on
\EQN~\ref{eqn:IZG_dmuS},
we obtain the following expression: 
\begin{align}
  \label{eqn:IFA2_integral}
  \nonumber 
  \IZG 
  &\approx 
    \tfrac{1}{N}
  \ln \int d\mu(\AMAT)
  e^{\ND\beta N Tr[\tfrac{1}{N}\TMAT^{\top}\AMATN\TMAT] }
  e^{\tfrac{N}{2}\ln(\Det{\AMATM})} \\
  & = 
  \tfrac{1}{N}
  \ln
  \left\langle
  e^{\ND\beta N Tr[\tfrac{1}{N}\TMAT^{\top}\AMATN\TMAT] }
  e^{\tfrac{N}{2}\ln(\Det{\AMATM})}
    \right\rangle_{\AMAT}
\end{align}
where the latter expresses the former in \BraKet notation. Since the measures of $\AMATM$ and $\AMATN$ are equivalent, this evaluates to the same quantity regardless of which is used.
This expression is derived in Appendix~\ref{sxn:TraceLogDerivation} (see \EQN~\ref{eqn:IZG_integral}):
it contains the original overlap term that depends $\AMATN$ as well as a new term from
the transformation that depends on $\Det{\AMATM}$ that is not yet defined.

\subsubsection{The Effective Correlation Space (\ECS)}

Currently, \EQN~\ref{eqn:IFA2_integral} is a ``formal'' expression, 
and we have not identified and/or justified its realm of applicability.
Fortunately, we have empirical evidence to suggest that this can be
made ``physically'' meaningful, by ``restricting'' the integral
to the tail of the Empirical Spectral Density (ESD). Since the ESD is an empirical object, this is where the theory becomes Semi-Empirical.

Prior work on HTSR theory indicates that during training the generalizing parts of a layer weight matrix concentrate
into $\rho_{tail}^{emp}(\lambda)$, the tail of the ESD, as the ESD becomes more \PowerLaw (PL),
and layer PL exponent  $\alpha\rightarrow\ 2+$ (from above)~\cite{MM19_HTSR_ICML,MM20_SDM,MM18_TR_JMLRversion,MM20a_trends_NatComm,YTHx23_KDD}. 
This suggests the integral in \EQN~\ref{eqn:QT}
should instead average over a low rank subspace spanned \emph{only by} the generalizing eigen-components of the Inner Correlation matrix, $\AMATM=\tfrac{1}{N}\SMAT^{\top}\SMAT$
(or equivalently, the Outer Correlation matrix $\AMATN$).
\textbf{We call this subspace the \EffectiveCorrelationSpace (\ECS).}

Let $\AECS$ be the matrix spanned by the largest $\MECS$ eigen-components $\AMAT$
in either form, $\AMATM$ (Inner) or $\AMATN$ (Outer), and using physics BraKet notation, giving
\begin{align}
  \label{sqn:AecsDefined}
  \AECS:=\mathbf{P}_{\EFF}\AMAT,\;\;  \mathbf{P}_{\EFF}:= \sum_{i=1}^{\MECS}|\LambdaECS_{i}\rangle\langle\LambdaECS_{i}|
\end{align}
where $\mathbf{P}_{\EFF}$ is the projection operator onto the subspace spanned by the eigenvector
$|\LambdaECS_{i}\rangle$  associated with the eigenvalue $\LambdaECS_{i}$ of $\AMATM$ or, equivalently, $\AMATN$.
We denote the corresponding projected \Student correlation matrices with a tilde, as follows:
\begin{align}
  \AMATM \rightarrow \AECSM\;\; \text{such that}\;\; d\mu(\AMATM) \rightarrow d\mu(\AECSM) \\ \nonumber
  \AMATN \rightarrow \AECSN\;\; \text{such that}\;\; d\mu(\AMATN) \rightarrow d\mu(\AECSN)  \nonumber
\end{align}
where now the matrices $\AECSM$ and $\AECSN$ are restricted to the \ECS, and the measure $d\mu(\AECSM)$ is similarly restricted.  We denote the eigenvalue $\lambda$ with the tilde, $\LambdaECS$, when we want to emphasize it is in the \ECS--\emph{but may drop the tilde} if it is clear from the context.
 See Appendix~\ref{sxn:TraceLogDerivation} for more..
   
When an ESD is \emph{\FatTailed} the~\ECS is at least as large if not larger than the PL tail of the ESD.
For an ESD with $\alpha\ge 2$, the generalizing eigen-components, i.e.$|\lambda_{i}\rangle$, will mostly
concentrate into the tail of the ESD, but some will remain in the bulk, so
the~\ECS will be larger than the tail.
In this case that $\MECS\ge M^{tail}$,
and
$\rho_{ECS}(\lambda)\supseteq\rho_{tail}(\lambda)$, as depicted in Figure~\ref{fig:ECS_space}.
When $\alpha=2$,  the layer is \Ideal (as in Section~\ref{sxn:setol_overview}),
in that all of the generalizing eigen-components to have now concentrated completely into the tail, so that
 $\MECS=M^{tail}$,
and
$\rho_{ECS}(\lambda)=\rho_{tail}(\lambda)$.
(When $\alpha< 2$, this suggests that the layer is overfit, and the layer may have a \CorrelationTrap and/or
frequently also has many near-zero eigenvalues.)
\\
\\
This leads to the following Model Selection Rule (MSR) for the \ECS:
\begin{quote}
  When transforming the measure $d\mu(\SMAT) \rightarrow d\mu(\AECS)$, we invoke an eigenvalue cutoff rule that
  prescribes how to replace $\AMAT$ with a low-rank effective matrix $\AMAT\rightarrow\AECS$,
  where the cutoff $\LambdaECS\ge\LambdaECSmin$ is chosen so that the~\ECS at least contains the PL tail
  and, importantly, such that $\det(\AMAT)=\det(\AMATM)=\det(\AMATN)$ is well defined.  
\end{quote}
Formally, this means that when we evaluate the \Quality (squared) $\QT$, \GeneratingFunction $\IZG$
or other relevant averages, we restrict the measure (i.e., integral or sum) to the eigencomponents in the tail of the ESD of
$\AMAT$ (or $\XMAT$, when appropriate)
starting with $\LambdaECSmin$.  
To our knowledge, this proposed MSR is completely novel.%

Restricted to the~\ECS, we now replace \EQN~\ref{eqn:IFA2_integral} with:
\begin{align}
  \nonumber 
  \IZG 
  &=
    \tfrac{1}{N}
  \ln \int d\mu(\AECS)
  e^{\ND\beta N Tr[\tfrac{1}{N}\TMAT^{\top}\AECSN\TMAT] }
  e^{\tfrac{N}{2}\ln(\Det{\AECSM})} \\
  \label{eqn:IFA2_braket}
  & =
    \tfrac{1}{N}
  \ln
  \left\langle
  e^{\ND\beta N Tr[\tfrac{1}{N}\TMAT^{\top}\AECSN\TMAT] }
  e^{\tfrac{N}{2}\ln(\Det{\AECSM})}
    \right\rangle_{\AECS}
\end{align}
where we have used the formal Duality of Measures, $d\mu(\AECS)=d\mu(\AECSM)=d\mu(\AECSN)$.
\footnote{Also, we could replace $\TMAT\rightarrow\TECS$, but to simplify the notation, we do not do this.}

\subsubsection{Two Simplifying Assumptions: the \IFA and \TRACELOG Condition}
\label{sxn:matgen_quality_hciz_C}

It now remains how to define the cutoff for the~\ECS space. To accomplish this, we make the following assumptions.
\begin{itemize}
\item
\textbf{The Independent Fluctuation Assumption (\IFA).}
This condition states that the two terms appearing in the exponential of \EQN~\ref{eqn:IFA2_braket} are statistically independent:
\begin{align}
  \label{eqn:IFA2}
  \IZG \approx 
    \tfrac{1}{N}
\ln
  \left\langle
  e^{\ND\beta N Tr[ \tfrac{1}{N}\TMAT^{\top}\AECSN\TMAT] }
  \right\rangle_{\AECS}
  \left\langle
  e^{\tfrac{N}{2}\ln(\Det{\AECSM})}
    \right\rangle_{\AECS}.
\end{align}
\item
\textbf{The Exact Renormalization Group Condition (\TRACELOG).}
This condition states that the determinant of the \Student (and \Teacher) Correlation matrix is unity, such that:
\begin{align}
\label{eqn:tlc_mm}
\Det{\AECS}=1 
\quad\mbox{or}\quad
\Trace{\ln \AECS}=0  ,
\end{align}
(and with $\AMAT$ additionally normalized to $M$, as explained in the Appendix, Section~\ref{app:st-gen-err-annealed-ham})
so that when we replace the measure over \Student layer weight matrices $d\mu(\SMAT)$ with a measure over all \Student Correlation matrices $d\mu(\AMAT)$,
\emph{restricted to the~\ECS}, the second term in \EQN~\ref{eqn:IFA2} becomes unity, $\langle \exp\left[\tfrac{N}{2} \Trace{ \ln[det\;\AMATM] } \right]\rangle_{\AECS}=1$,
and vanishes in the final expression for $\IZG$.
\end{itemize}

\noindent
At this point, the \IFA is made purely for mathematical convenience, i.e., 
we have not demonstrated it empirically, but it is not implausible as a statistical modeling assumption. 
On the other hand, we can test the \TRACELOG condition empirically;
this is a critical part of our \SETOL approach.

Notably, when taking the large-$N$ approximation of $\IZG$, we effectively do this independently in two steps.
First, we apply a \SaddlePointApproximation (SPA) to the second term in \EQN~\ref{eqn:IFA2},
leading to the \TRACELOG condition (see Appendix~\ref{sxn:TraceLogDerivation}).
Most importantly, we can test the \TRACELOG condition empirically,
and this is another critical part and justification of our \SETOL approach.
Second, when applying the result by Tanaka (\EQN~\ref{eqn:hciz_tanaka}),
we are applying an SPA to the result, but assuming we are
restricted to the~\ECS.

\paragraph{Empirical Tests of the \TRACELOG Condition and the~\ECS.}

Since we require that the students resemble the fixed \Teacher, a reasonable estimate for the average of the Student
Correlation matrix $\AECS$
(either $\AECSM$ or $\AECSN$)  (in the~\ECS)  is the point estimate provided by the actual (known) fixed \Teacher correlation matrix $\XECS$, 
\begin{align} 
\label{eqn:IFA-ppint}
\Big\langle \det\AECSM\Big\rangle_{\AECS}=
\Big\langle \det\AECSN\Big\rangle_{\AECS}
\simeq\det\XECS=\prod_{t}\lambda_{t}=1\;\;\forall\LambdaECS_{t}\in\rho^{emp}_{tail}(\lambda) ,
\end{align}
for all eigenvalues $\LambdaECS_{t}$ in $\rho^{emp}_{tail}(\lambda)$ the tail of the
ESD of $\XECS$, i.e., $\LambdaECS_{t}\ge\LambdaECSmin$.
How should one choose $\LambdaECSmin$ in this expression?
We already know that most NNs layers have \FatTailed ESDs, but
if the \Teacher ESD is simply \emph{\RandomLike} or even Bulk+Spikes, then the expected value
$\langle\Det{\AMATM}\rangle$ itself of the determinant
of a random full rank \Student Correlation matrix $\AMATM$
might be well defined and easy to estimate, and we might not even need to define the lower rank~\ECS.
Indeed, in these cases, the correlated eigencomponents, if they exist, may be buried in the bulk
region of the ESD and not readily identifiable.
Because $\rho_{tail}^{emp}(\lambda)$ is \FatTailed
and \PowerLaw (PL), this poses some difficulty, we need to first define the \EffectiveCorrelationSpace (\ECS).

\begin{figure}[t]
  \begin{center}
  \includegraphics[width=10cm]{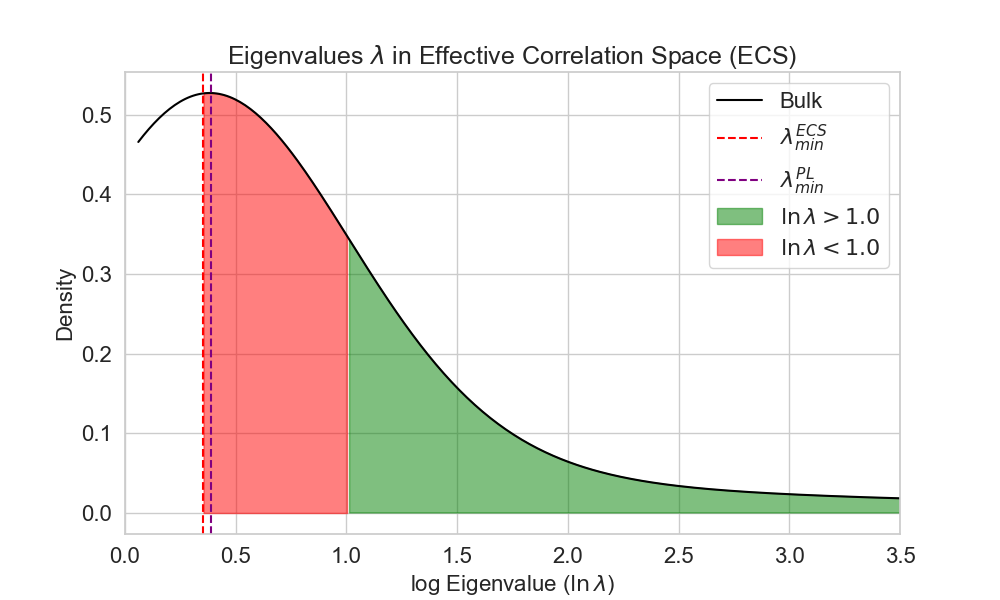}
  \caption{The image depicts a typical
    \EmpiricalSpectralDensity (ESD) of a layer correlation matrix $\XMAT$, with \WW Power-Law (PL) exponent
    $\alpha=2.0$.
    The green and red shaded regions depict
    eigenvalues $\lambda$ in the \EffectiveCorrelationSpace (\ECS) of $\XECS$, defined by $\lambda>\lambda_{min}^{ECS}$.
    The x-axis displays the eigenvalues on the log scale, $\ln\lambda$.
    The vertical red line is at the start of the PL tail  $(\lambda_{min}^{PL})$.
    The purple, vertical line is at the start of the~\ECS tail  $(\lambda_{min}^{ECS})$.
    The green shaded region depicts those eigenvalues where $\ln\lambda>1.0$,
    whereas the red shaded region depicts those eigenvalues where $\ln\lambda<1.0$.
    The~\ECS is defined such that the Exact Renornmalization Group (\TRACELOG) condition is best satisfied, i.e $\sum\ln\lambda= 0$ for $\lambda\ge\lambda_{min}^{ECS}$.
    }
  \end{center}
  \label{fig:ECS_space}
\end{figure}

Because the \Teacher ESD is most likely MHT and PL, 
if we choose $\LambdaECSmin:=\lambda_{min}^{ECS}$ too small, and the tail extends too far into the bulk region of the ESD,
then for all practical purposes $\Det{\XECS}\ll 1$.
On the other hand, if we choose $\LambdaECSmin$ too large,
then we only capture very large eigenvalues, and for all practical purposes $\Det{\XECS} \gg 1$.
Therefore, if we set the scale of $\XECS$ appropriately, we can choose a $\LambdaECSmin$ such that $\Det{\XECS}=1$.
In this case, by choosing $\LambdaECSmin$ appropriately, we can estimate the expected value of
$\langle\Det{\AMATM}\rangle$  with an empirical point estimate over the \Teacher Correlation matrix, which is unity.
\begin{equation}
\label{eqn:detX}
\vert\det\XECS\vert \simeq 1 ; \quad \Trace{ \ln\XECS } = \ln\vert\det\XECS\vert \simeq 0  .
\end{equation} 

This expression can now be used in a practical calculation to define a low-rank subspace that both allows us to evaluate the HCIZ integral,
and to identify, in principle, the generalizing components of the layer.
We also refer to \EQN~\ref{eqn:detX} as the \TRACELOG condition, which is technically its empirical form.
For example, Figure~\ref{fig:ECS_space} depicts the eigenvalues in the~\ECS for a \Typical ESD with PL $\alpha=2.0$..
Notice that the start of the PL tail, $(\lambda_{min}^{PL})$, is very close to the start of the~\ECS tail. $(\LambdaECSmin)$,
i.e. $\Delta \lambda_{min}:=\LambdaECSmin-\LambdaPLmin\approx 0$.
Also, notice that while there are many large eigenvalues, $\ln\lambda>1.0$, there are numerous small eigenvalues as well,
$\ln\lambda<1.0$, such that the $\sum\ln\lambda\approx 0$ for $\lambda\ge\LambdaECSmin$. In other words, the red and green shaded areas have the same measure.
Additional plots like Figure~\ref{fig:ECS_space}, generated with \WW,  can be found in Section~\ref{sxn:empirical},
in Figure~\ref{fig:mlp3-detx-gap}
as well as plots of $\Delta \lambda_{min}$ vs. the \WW PL $\alpha$ for several real-world examples,
in Figures~\ref{fig:CV_ESD_trends} and ~\ref{fig:LLM_ESD_trends}.

\subsection{Evaluating the Layer Quality \texorpdfstring{$(\Q)$}{Q} in the Large-\texorpdfstring{$N$}{N} Limit}
\label{sxn:matgen_evaluation_hciz}

To generate the Average \Quality, $\QT$, we first take the \LargeN limit of $\IZG$ in $N$,
\begin{equation}
  \label{eqn:IZG_limit}
\IZGINF := \lim_{N \gg 1} \IZG 
= \lim_{N \gg 1}  \tfrac{1}{N}
\ln \; 
  \Expected[\AECS]{ 
    \exp\,
      N \beta \Trace{\tfrac{1}{N}\TMAT^{\top}\,\AECSN\,\TMAT}
  } 
\end{equation}
and then take the appropriate partial derivative,
analogously to how we did for $\AVGSTGE$; see Section~\ref{sxn:quality} for more details.
This gives (as in \EQN~\ref{eqn:QT_def})
\begin{align}
\label{eqn:IZG_generate_Q2}
\QT := \frac{1}{\beta}\frac{\partial }{\partial \ND}\IZGINF  
\underset{\text{high-}T}{\approx}\;
\frac{1}{\ND}\frac{\partial }{\partial \beta}\IZGINF 
\end{align}
Notice that since we are at high-Temperature, it doesn't matter which partial derivative we take,
and we expect both results to be yield the same expression.

This HCIZ integral in \EQN~\ref{eqn:IZG_limit} can be evaluated
(i.e in the \LargeN limit in $N$) using a result by Tanaka ---provided
the matrix $\AECSN$ is low rank, which holds when the \TRACELOG condition is satisfied.
Thus, moving forward, we will assume an
\EffectiveCorrelationSpace (\ECS) of rank $\MECS$, where $\LambdaECSmin$ is the $M^{th}$-largest eigenvalue of $\XECS$,
and defines the start of the~\ECS (and whatever branch-cut is necessary to integrate $R(z)$).

Tanaka's result for the~\ECS can be expressed as:
\begin{equation}
  \label{eqn:tanaka_result}
  \underset{N\gg 1}{\lim}\frac{1}{N}\ln
\Expected[\AECS]{\exp\left(\ND\beta \Trace{\TMAT^{\top}\AECSN\TMAT}\right)}
  =\ND\beta\sum\limits_{i=1}^{\MECS}\;\GNECSI,
\end{equation}
where the sum now only includes the eigenvalues of $\XECS$ (in the~\ECS), $\beta=\tfrac{1}{T}$
is the Inverse-Temperature, $\ND$ is the size of the training dataset, and $\LambdaECS$ is an eigenvalue of $\XECS$, the Teacher
Correlation matrix projected into the \ECS space.
$\AECSN$ is the $N \times N$ form of the \Student Correlation matrix,
with $N-M$ non-zero eigenvalues, and $\TMAT$ is the $N\times M$ \Teacher  weight matrix
(also effectively projected into the \ECS, i.e. $\TMAT\rightarrow\TECS$ here).
$\GNI$ is the \GEN, defined below.
\footnote{We use the notation $\Expected[\AECS]{\cdots}$ for expected value and placed $\tfrac{1}{N}$ on the L.H.S.
to help the reader compare this to the original expressions in~\cite{Tanaka2007, Tanaka2008}.
Also,  in \cite{Tanaka2007, Tanaka2008},$\beta=1|2$, but, in fact, if one inserts $-\beta$ as an inverse temperature into the final expression, it simply factors out.}
This gives
\begin{equation}
\label{eqn:tanaka_result2}
\IZGINF = \ND\beta\sum\limits_{i=1}^{\MECS}\;\GNECSI,
\end{equation}

This gives a final expression for the Average \LayerQuality (Squared) $\QT$ as
\begin{equation}
\label{eqn:Q2_result}
\QT = \sum\limits_{i=1}^{\MECS}\;\GNECSI,
\end{equation}
Note that $\QT$ is independent of $N$ and $\beta$,
and, indeed, \EQN~\ref{eqn:IZG_generate_Q2} is an equality.

The average \Quality (squared) can be expressed as a sum over
\GeneratingFunctions $\GN$, which depend only the statistical properties of the
actual \Teacher Correlation  matrix  $\XECS$ (projected into the~\ECS).
Each term in the sum, $\GNECSI$, takes the form
\begin{equation}
\label{eqn:generating_function_A}
 \GN:=\int\limits_{0}^{\lambda}R_{\AMAT}(z)dz \xrightarrow{\text{\ECS}}\int\limits_{\LambdaECSmin}^{\lambda}R_{\AECS}(z)dz
\end{equation}
where $R_{\AECS}(\LambdaECS)$ is the \RTransform from RMT,
and $\LambdaECSmin$ is the lower bound of the~\ECS spectrum.
Importantly, the \RTransform for a Heavy-Tailed ESD may have a branchcut at or near the
start of the~\ECS (as explained in Section~\ref{sxn:r_transforms}), so restricting the integrand
to start at $\LambdaECSmin$ is critical.

\charles{removed section below for space reasons; see tex file}

%
%

Since we expect the best Student matrices to resemble the actual \Teacher matrices, we expect the \Student correlation matrix $\AECS$ to have similar spectral properties to our actual empirical correlation matrices $\XECS$.
That is, from the perspective of \HTSR theory and the classification into \PhasesOfTraining~\cite{MM18_TR_JMLRversion}, we expect all the $\AECS$ to be in the same phase as $\XECS$ (and, in addition, to have the same PL exponent value).   That is, 
\\
\\
\emph{We expect the \RTransform of $\AECS$ to have the same functional form as the $R$-transform of $\XECS$.}
\\
\\
If our (\Teacher) NN weight matrix exhibits a HT PL, then the tail the ESD ($\rho_{tail}(\lambda)$) of the \Student and \Teacher will both take the limiting form of a PL, with the same empirical variance $\sigma^{2}$ and (critically) the same PL exponent $\alpha$:
\begin{equation}
\label{eqn:R_PL}
  \rho_{tail}[\AECS](\lambda)\sim\rho_{tail}[\XECS](\lambda)\sim\lambda^{-\alpha}.
\end{equation}

Up until this point, our derivation of $\QT$ only depends on the \TRACELOG condition, irrespective of the exact functional form of $R(z)$,
therefore the \SETOL approach is tested by examining how well the \TRACELOG condition and the \ECS holds for the layers in very well-performing models.
We do this in Section~\ref{sxn:empirical-trace_log}.

\subsection{Modeling the R-Transform}
\label{sxn:r_transforms}

To apply \SETOL, the model must satisfy the \TRACELOG condition--which occurs during the case of \IdealLearning.
For most cases of NN models, the ESD are HT; and in practice, one usually would select $R(x)$ that reflects this.
But to be more general, we formally extend the theory to allow the practitioners to
both model the ESD as just a collection of discrete spikes, and to even correct for
\CorrelationTraps and/or \VeryHeavyTailed (VHT) ESDs.
Most importantly,  we derive expressions for both the~\WW~\ALPHA and~\ALPHAHAT metrics, valid
for the case \IdealLearning where $\alpha=2$, and formally extended for other cases.

The goal of this section is to obtain formal expressions for the \LayerQuality, $\Q$ which is
given in terms of what we somewhat imprecisely refer to as a \GEN.
In many cases, however, the resulting approximate expressions for $\Q$ do take the
form well-known norms, including the Frobenius norm, the Spectral norm, and the Shatten norm.
The models for $R(z)$  we use are presented in Table~\ref{tab:known_r_transforms}.
The inal expression for the \LayerQuality $\Q$ extended \emph{Semi-Empirically} to the non-ideal layers that can then be used to obtain the \HTSR \ALPHA and \ALPHAHAT metrics; these are given in Table~\ref{tab:htsr_layer_quality}.

\subsubsection{Elementary Random Matrix Theory}
\label{sxn:r_transforms:elementary_rmt}

We begin with some useful notions definitions from \RandomMatrixTheory.
Using the ESD $\rho(\lambda)$, defined as
\begin{equation}
\label{eqn:rgo}
\rho(\lambda):=\frac{1}{M}\sum_{i=1}^{M}\delta(\lambda-\lambda_{i})  ,
\end{equation}
we can express the \emph{\GreensFunction} (or \emph{\CauchyStieltjes} transform $C(z)$) by%
\footnote{Notice our naming and sign convention in \EQN~\ref{eqn:Cz}.
We equate the \GreensFunction $G(z)$ with
the (positive) \CauchyStieltjes transform, $G(z)=C(z)$.   Other works may use the opposite sign convention, $G(z)=-C(z)$.}
\begin{equation}
\label{eqn:Cz}
G(z)=C(z):=\int \mathrm{d}\lambda \frac{\rho(\lambda)}{z-\lambda} .
\end{equation}
From $G(z)$, we can recover the ESD, $\rho(\lambda)$, using the inversion relation
\begin{equation}
\label{eqn:GzInverse}
\rho(\lambda)=\lim_{\epsilon\rightarrow 0+}\frac{1}{\pi}\mathrm{IM}(G(\lambda+i\epsilon))  ,
\end{equation}
where $\mathrm{IM}$ is the imaginary part of $G(z)$, and where the $\lim_{\epsilon\rightarrow 0+}$ means to take the limit approaching from the upper half of the complex plane.
The \RTransform, $R(z)$, can be defined using the Blue function $B(z)$ 
\begin{equation}
\label{eqn:Rz}
R(z):=B(z)-\frac{1}{z}  ,
\end{equation}
where the Blue function $B(z)$~\cite{Zee1996} is the functional inverse of the Greens Function $G(z)$,%
\footnote{The Blue function was first introduced by Zee~\cite{Zee1996} to model, among other things, spectral broadening in quantum systems.
Briefly, given a deterministic Hamiltonian matrix $\mathbf{H}_{0}$, with eigenvalues $\lambda^{0}_{i}$,
one can model the spectral broadening of $\lambda^{0}_{i}$ by adding a random matrix $\mathbf{H}_{1}$ to $\mathbf{H}_{0}$:
$\mathbf{H}=\mathbf{H}_{0}+\mathbf{H}_{1}$.  
The resulting eigenvalues of $\mathbf{H}$ now contain some level of randomness, $\sigma$, i.e., $\lambda=\lambda^{0}+\sigma$.  
To model the ESD of $\mathbf{H}$, one then specifies the \RTransforms for $\mathbf{H}_{0}$ and $\mathbf{H}_{1}$; the full ESD of $\mathbf{H}$
can then be reconstructed by adding the two \RTransforms together $R(z)=R_{0}(z)+R_{1}(z)$.
Zee notes that $R(z)$  is  the self-energy $\Sigma(z)$ from quantum many body theory~\cite{Zee1996}.}
satisfying 
\begin{equation}
\label{eqn:GzRelation}
B[G(z)]=z  .
\end{equation}
By specifying the $R(z)$ transform, we specify the complete ESD, $\rho(\lambda)$.
Here, we are actually only interested in the tail of $\rho(\lambda)$.
That is, we can given $R(z)=R(z)_{tail}+R(z)_{bulk}$, we only need $R(z)\approx R(z)_{tail}$.

Before moving forward, it is first necessary to demonstrate the \RTransform $R(z)$ exists for the power-law ESDs that arise in \SETOL; this is shown in Appendix~\ref{sxn:RTransformExists}.  Specifically, it is shown that while the $R(z)$ does not formally exist for a pure power-law (PL) tail with $\alpha=2$, it does generally exist for \emph{Truncated} power-law (TPL) tails, defined such that the domain is compact, i.e., $\lambda\in[\LambdaECSmin, \LambdaECS_{max}]$. 

\subsubsection{Known \RTransforms and Analytic (Formal) Models}
\label{sxn:r_transforms:known_r_transforms}

There are only a few known analytic results for the explicit \RTransform $R(z)$.
The ones we need are in Table~\ref{tab:known_r_transforms}.
Below, we review some of them, explaining what ESD they correspond to,
and what the resulting \GEN~$G(\lambda)$ would be if applied
as a model $R(x)$ here.
\begin{table}[ht!]
  \centering
  \renewcommand{\arraystretch}{1.25} 
\begin{tabular}{|c|c|c|}
  \hline
  Model & \textbf{HTSR Universality Class} & \textbf{$R(z)$}\\  \hline
  \hline
  Discrete & Bulk$+$Spikes, MHT, HT & $\tfrac{1}{\MECS}\sum_{i=1}^{\MECS}\lambda_{i}$   \\ \hline
  \hline
  Wishart Models & &\\ \hline
  Multiplicative-Wishart & HT/VHT& $\dfrac{\epsilon\phi z^2}{2 - \epsilon\phi^2 z^2}$ \\  \hline
  Inverse Marchenko-Pastur (IMP) & HT/VHT &  $\dfrac{\kappa-\sqrt{\kappa(\kappa-2z)}}{z}$   \\  \hline
  \hline
  L\'evy Wigner (LW) &   & \\  \hline
  Free Cauchy (FC) ($\alpha_{l}=1$) & HT $\alpha=2$ & $a+i\gamma$ \\ \hline
  General L\'evy  ($\alpha_{l}\ne 1$) & VHT $\alpha<2$   & $a+bz^{\alpha-2}$ \\  \hline
\end{tabular}
  \caption{Known \RTransforms for random matrix ensembles relevant to modeling heavy-tailed spectral densities (eigenvalues or singular values squared).
    The \emph{Multiplicative-Wishart} model has two real, non-zero parameters, $\epsilon$ and $\phi$; for more details,
    see \cite{PW16_NIPS}.
  For the \emph{\InverseMP}, as given by Bun~\cite{BunThesis}, $\kappa=\frac{1}{2}(Q-1)$ where, $q=\frac{1}{Q}=\frac{M}{N}\le 1$.
  The \emph{L\'evy-Wigner} (LW) model describes Wigner-like square random matrices
  (as opposed to Wishart-like or Correlation Matrices), where the elements are drawn from a L\'evy-Stable distribution.
  The resulting LW ESD is Heavy-Tailed Power Law, and characterized by the L\'evy exponent $\alpha_{l}$.
  The LW $R(z)$ is parameterized by a (real) shift parameter $a$,
  a complex phase factor $b$ (that depends on 3 real parameters   $\alpha_{l}, \beta$, and $\gamma$),
  and, of course, $\alpha_{l}$.
  The Free Cauchy (FC) model is a special case of the IW model, corresponding to the L\'evy $\alpha_{l}=1$, and the \HTSR $\alpha=2$. 
  We will extend the LW models to the rectangular case for our modeling purposes here by making
  the association   $\alpha = \alpha_{l}+1$ for $\alpha\le 2$.
   (Also, we take thew variance $\sigma=1$ for all models.)
   Generally speaking, the L\'evy $R(z)$ are more complicated; 
   for more details, see~\cite{BJNx01_TR,BJNx06_TR,BJ09_TR}.
}
\label{tab:known_r_transforms}
\end{table}

\charles{Discuss the Table briefly, and then each model in its own subsubsection.
We can create plots to show how these models can treat heavy tails, and explain the parameter fitting
}

\subsubsection{Discrete Model: Bulk\texorpdfstring{$+$}{+}Spikes, MHT, HT}

Here, we consider modeling the tail ESD, $\rho_{tail}(\lambda)$, as a collection
of discrete spikes $\lambda_{spike}$, 
where $\lambda_{spike}\ge \LambdaECSmin$.
This could be for modeling an ESD in the Bulk$+$Spikes (BS) \HTSR Universality class,
or it could be applied to an MHT or HT ESD as a discrete distribution with
no inherent structure; its a modeling choice.
Here, \RTransform for the \ECS is sum of Dirac delta functions.
This lets us compute the \LayerQuality $\Q$ in closed form in terms of the Teacher weight matrix
$\TMAT=\WMAT$ as a Tail norm, the Frobenius norm of the tail eigenvectors.

Let the tail of the ESD have $\MECS = M^{tail}$ eigenvalues that define the \ECS, i.e.,  
\begin{equation}
\label{eqn:spikes_model_0}
\rho_{tail}(\lambda)=\sum_{i=1}^{\MECS}\delta(\lambda-\LambdaECS_{i}) .
\end{equation}
where the $\LambdaECS_{i}$ are normalized by $\tfrac{1}{M}$.

The \GreensFunction $G(z)$ is then
\begin{equation}
\label{eqn:spikes_model_1}
G(z) = \int d\lambda \dfrac{ \rho_{tail}(\lambda) }{z - \lambda} =
\sum_{i=1}^{\MECS} \int d\lambda \dfrac{\delta(\lambda-\LambdaECS_{i}) }{z - \lambda} =
\sum_{i=1}^{\MECS} \dfrac{1}{z-\LambdaECS_{i}}  ,
\end{equation}
and the Blue function for each individual term $i$ is $\frac{1}{z-\LambdaECS_{i}}$, i.e., $B(z)=\LambdaECS_{i}+\frac{1}{z}$. 
Now, using the additive property of the \RTransform, we can express the total $R(z)$ as the sum of the \RTransforms for the individual terms $i$, giving
\begin{equation}
\label{eqn:spikes_model_2}
R(z) =\sum_{i=1}^{\MECS}\left(\left(\LambdaECS_{i}+ \dfrac{1}{z}\right) - \dfrac{1}{z}\right)=\sum_{i=1}^{\MECS}\LambdaECS_{i}  .
\end{equation}
This gives the \GEN~$\GN$ as
\begin{align}
\label{eqn:spikes_model_3} 
\GN 
&= \int_{\LambdaECSmin}^{\lambda}\sum_{i=1}^{\MECS}\LambdaECS_{i} d\lambda \\ \nonumber
&= \sum_{i=1}^{\MECS}\LambdaECS_{i} \int_{\LambdaECSmin}^{\lambda} 1 d\lambda \\ \nonumber
&= \left(\sum_{i=1}^{\MECS}\LambdaECS_{i}\right)\left(\lambda-\LambdaECSmin\right)
\end{align}
Seeing that $\LambdaECSmin$ is usually quite small, we make the approximation
\begin{align}
\GN  \approx \left(\sum_{i=1}^{\MECS}\LambdaECS_{i}\right)\left(\lambda\right)  ,
\end{align}
which gives the \QualitySquared approximately as
\begin{align}
  \label{eqn:spikes_model_4}
  \QT = \sum_{i=1}^{\MECS}\GNECSI 
 \approx\left(\sum_{i=1}^{\MECS}\LambdaECS_{i}\right)^{2}  .
\end{align}
We now see that we can define $\Q:=\sqrt{\QT}=\sum_{i=1}^{\MECS}\LambdaECS_{i}$ is what we call a Tail Norm,
the Frobenius norm of the tail eigenvalues.

\subsubsection{Free Cauchy Model (\texorpdfstring{$\alpha = 2$}{alpha = 2})}
\label{sxn:r_transforms:free_cauchy}

For the \emph{Free Cauchy} (FC) model the \RTransform is a constant
\begin{equation}
\label{eqn:free_cauchy_R}
R(z)[\mathrm{FC}] = a + i\gamma ,
\end{equation}
where the shift parameter $a>0$ translates the FC spectrum (i.e. so it overlaps with the the Power Law tail of the ESD),
and the scale parameter $\gamma>0$ sets the tail width.
We do not attempt to fit $a$ and $\gamma$ to a real-world ESD here but rather simply use this model formally,

Because $R(z)$ is independent of $z$, the integral is straightforward:
\begin{equation}
\label{eqn:free_cauchy_G}
\GN[\mathrm{FC}] = \int_{\LambdaECSmin}^{\lambda} R(z)[\mathrm{FC}]dz = (a + i\gamma)(\lambda - \LambdaECSmin).
\end{equation}

As explained in Appendix~\ref{sxn:tanaka}, when $R(z)$ is complex,  we keep the \emph{Real} part $\GNECSI$, giving,
\begin{equation}
\label{eqn:free_cauchy_G_abs}
\Re[\GN[\mathrm{FC}]] = a(\lambda - \LambdaECSmin).
\end{equation}

To obtain a formal expression for $\ALPHA$, we make two approximations.
First, we take the lower cut-off to be near zero $(\LambdaECSmin \approx 0)$.
\begin{equation}
\label{eqn:free_cauchy_G_max}
\bigl|\GN[\mathrm{FC}]\bigr|_{\LambdaECSmin \approx 0} \approx a\lambda
\end{equation}
Second, we assume the \LayerQualitySquared is dominated by the largest term in the sum:
\begin{equation}
\label{eqn:free_cauchy_G_max2}
\QT_{\mathrm{FC}} = \sum_{i=1}^{\MECS}\bigl|\GN[\mathrm{FC}]\bigr|_{\LambdaECSmin \approx 0}
= a\sum_{i}^{\MECS} \LambdaECS_{i} \approx a\lambda_{max}
\end{equation}

For the Heavy-Tailed ESDs, we generally find that as $\lambda_{\max}$ increases, the \HTSR $\alpha$ decreases.
Juat for illustrative purposes,
If we take $\log_{10}\lambda_{\max}\sim\frac{1}{\alpha}$ (assuming $\log_{10}\lambda_{\max}>1$), and take the square root of both sides,
we obtain the desired result for the log \Quality, namely, that it scales inversely with $\alpha$
\begin{equation}
\label{eqn:free_cauchy_Q}
\log_{10}\Q_{\mathrm{FC}} \sim \frac{1}{\alpha}
\end{equation}

If we extend this formally beyond the Ideal  case to $\alpha\ge 2$, we now can explain why the \HTSR $\alpha$
makes a good \LayerQuality metric in deep networks,
as smaller $\alpha$ suggests higher quality layers and therefore a higher overall model quality and/or accuracy.

\subsubsection{Inverse Marchenko-Pastur Model of \IdealLearning}
For a more realistic model of a layer ESD, 
here, we consider the \InverseMP (IMP) model~\cite{BunThesis}.
The IMP model treats the ESD of $\XMAT^{-1}$ when the ESD of $\XMAT$ itself is MP.
As a parametric model, it can be quite effective at treating VHT and HT (or \FatTailed) ESDs,
$\alpha\le 4$, but works best when $\alpha=2.0$.
To do this, one simply considers  $\kappa$,  as an adjustable parameter which effectively captures the rank or soft-rank of an ECS-like space in these non-ideal layers.
It is an excellent model for the ESD when $\alpha=2.0$ (and $Q=2$).
Using this model, we can derive an expression for the \HTSR \ALPHAHAT \LayerQuality metric,
$\ALPHAHATEQN :=\ALPHAHATLONG)$ as a leading order term in the final expression for $\log_{10}\QT$.

\begin{figure}[ht]
    \centering
    \subfigure[IMP Distribution, $\kappa=0.5$]{ 
      \includegraphics[width=7.5cm]{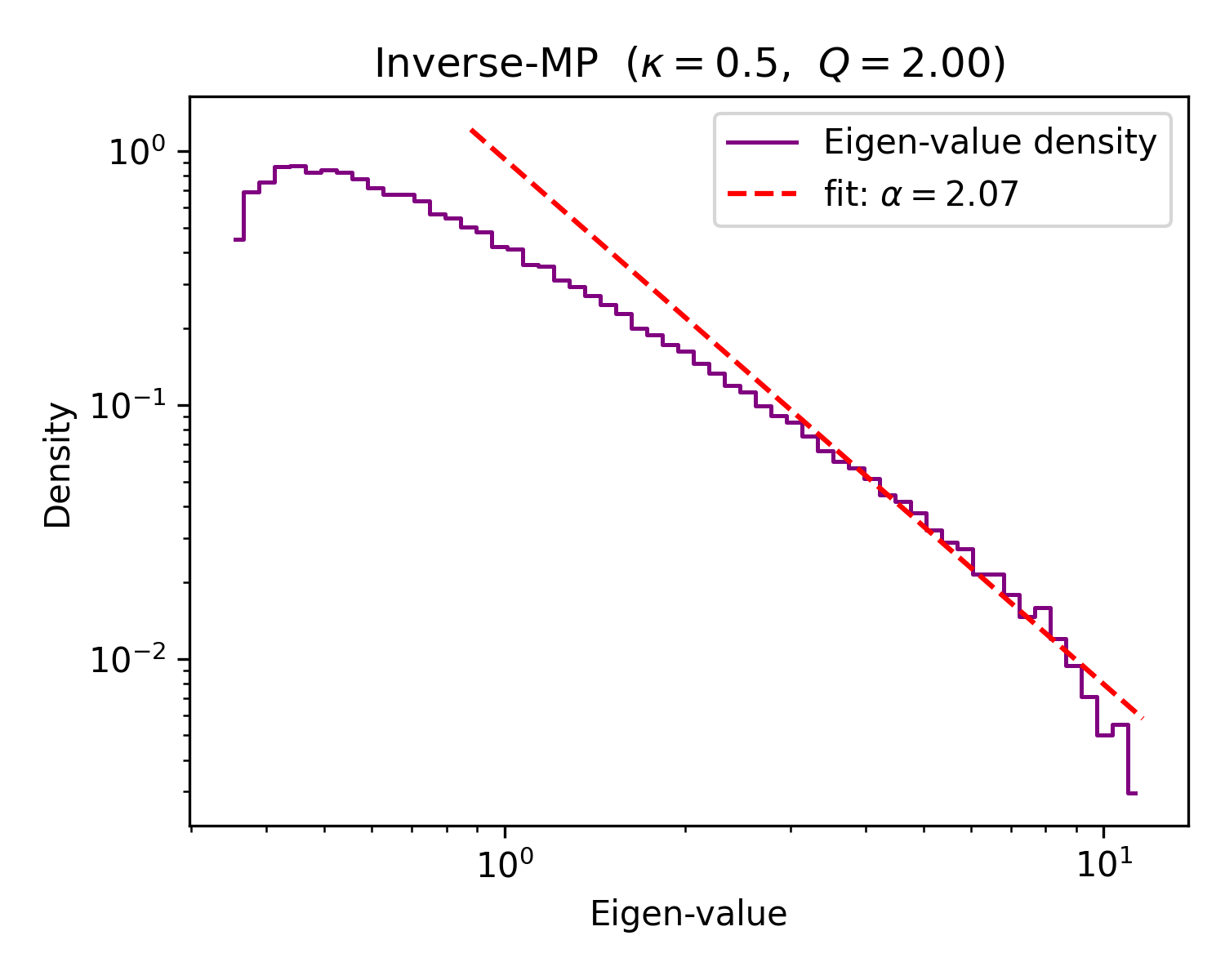}
      \label{fig:IMPplotESD}                               
    }                               
    \subfigure[IMP $\kappa$ vs. \HTSR $\alpha$]{ 
      \includegraphics[width=7.5cm]{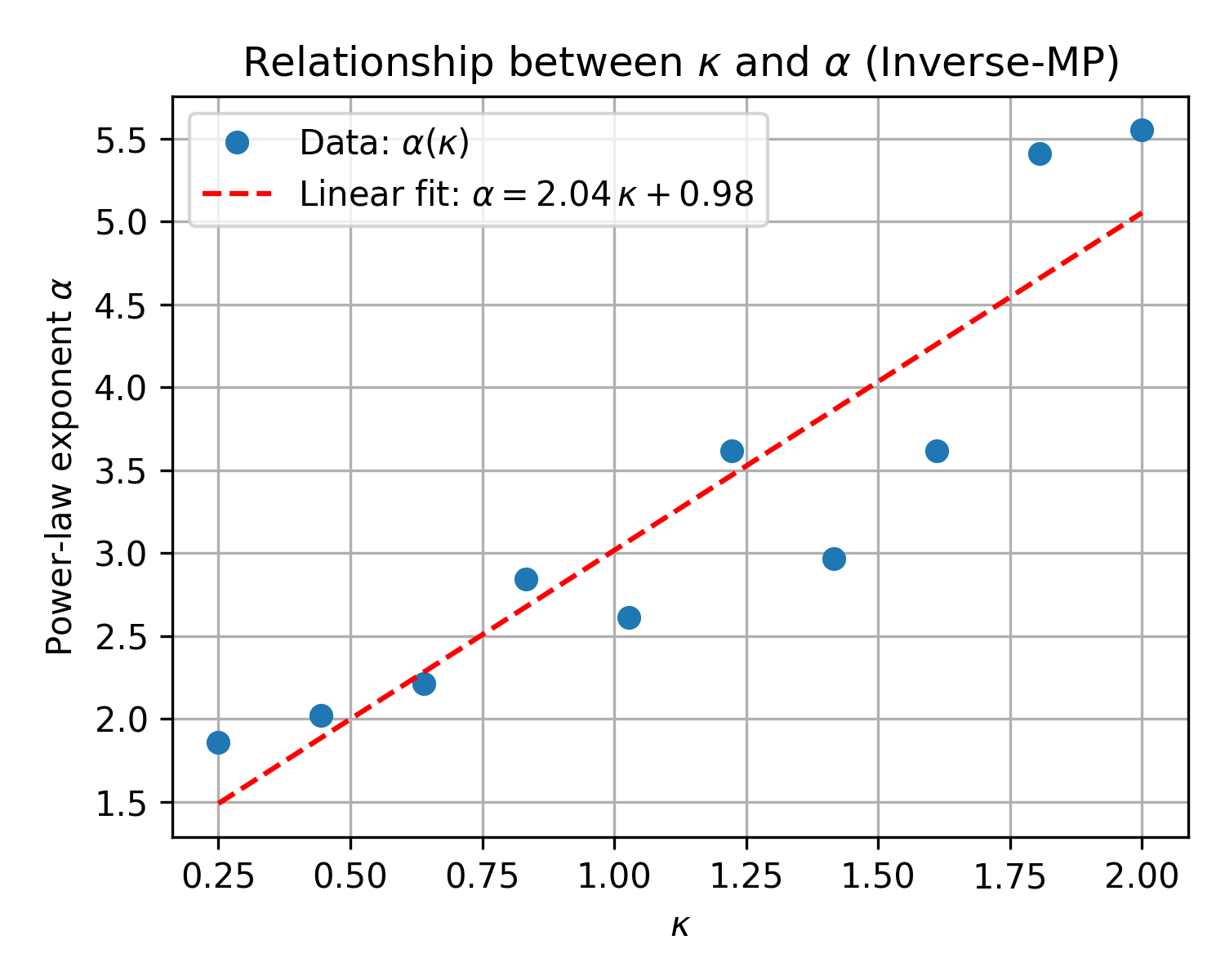}
      \label{fig:IMPPlotKappaVsAlpha}                                
    }
    \caption{
      (\ref{fig:IMPplotESD}) plots a typical \InverseMP (IMP) ESD  $\kappa=0.5$,  along with Power Law (PL) fits, with PL exponent $\alpha=1.86$.
      (\ref{fig:IMPPlotKappaVsAlpha}) depicts the linear relationship between $\kappa$ and $\alpha$ for a few randomly generated examples.
    }
  \label{fig:IMPplots}                                                                                                      
\end{figure}   

In Figure~\ref{fig:IMPplots}, we generate a random IMP model ESD with $\kappa=0.5$, $Q=2.0$, fit the ESD
to a PL, and find $\alpha=1.08$; this is a reasonably accurate model of an PL ESD for $\alpha\approx 2$.
For larger $\kappa$, the fits are not as good, but the model is still useful to suggest a heuristic model for the
\LayerQuality in terms of the \HTSR $\alpha$.
In Figure~\ref{fig:IMPPlotKappaVsAlpha}, we show that there is a simple relationship between $\kappa$ and $\alpha$,
which allows us to formally state $\alpha=2\;\kappa$.

\begin{figure}[t]
    \centering
    \subfigure[IMP $R(z)$, real $z$.]{
        \includegraphics[width=7.5cm]{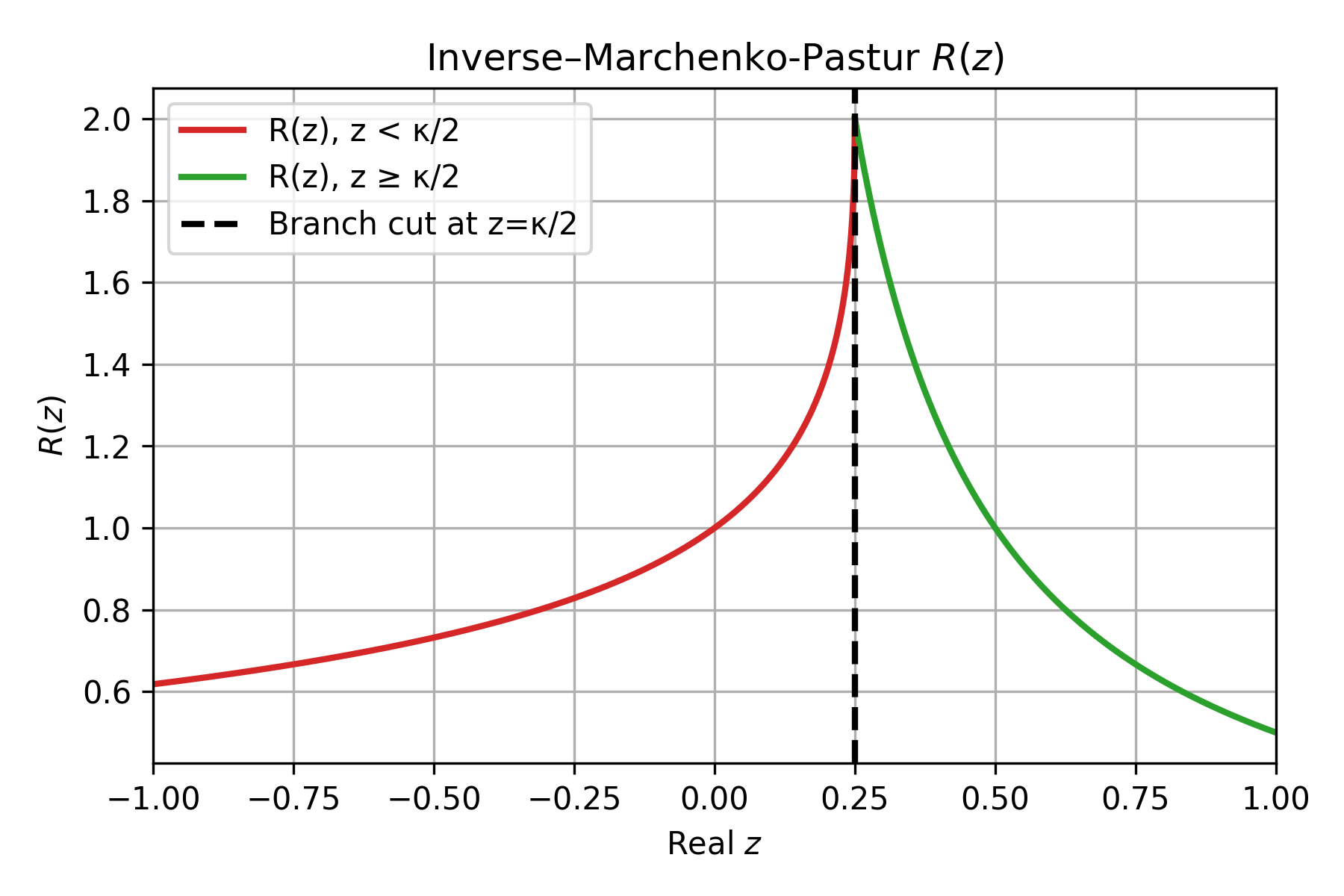}
        \label{fig:IMP_R}
    }
    \subfigure[Branch cut at $z=\kappa/2$]{
        \includegraphics[width=7.5cm]{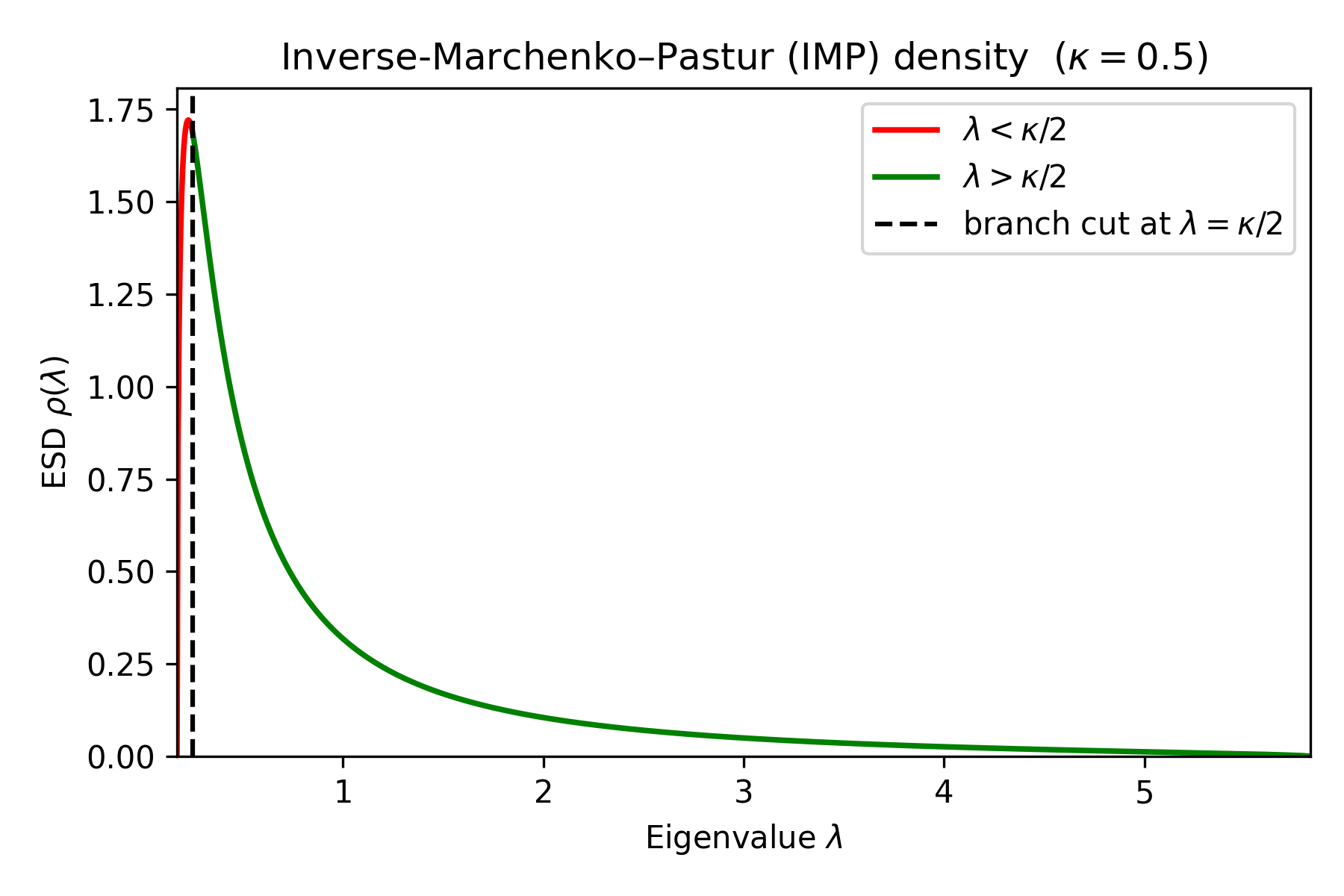}
        \label{fig:IMP_branch_cut}
    }
    \caption{(a) The function $R(z)$ of the \InverseMP (IMP) model, with a singularity at $z = \kappa/2$. (b) The branch cut in the Empirical Spectral Density (ESD) at $\kappa = 0.5$.}
    \label{fig:R_branch_cut_combined}
\end{figure}

Lets consider $R(z)$ for the \InverseMP model, denoted $R(z)[IMP]$.
To integrate this function, we require that it be analytic.
At first glance, it may seem that that $R(z)[IMP]$ is not analytic because it
has a pole at $z=0$ and because the square-root term $\sqrt{\kappa(\kappa-2z)}$  creates branch
cut at and $z=\kappa/2$ (and $z=\infty$).
Figure~\ref{fig:R_branch_cut_combined} presents this in two ways:
Figure~\ref{fig:IMP_R} shows the R-transform $R(z)[IZ]$ for real $z$, highlighting its singular behavior and the location of the branch cut at $z = \kappa/2$; and
Figure~\ref{fig:IMP_branch_cut} shows the corresponding branch cut in the ESD of the Inverse MP model (for $\kappa = 0.5$).
We select the branch cut starting at $z=\kappa/2$ and ending at $z=\infty$,
which allows us to at least formally defined the integral along the physically meaningful part of the ESD:
\begin{equation}
\label{eqn:IMP_model_1} 
\GN[\mathrm{IMP}] := \int_{\LambdaECSmin}^{\lambda} \Re[R(z)][\mathrm{IMP}] dz  ,
\end{equation}
noting that we expect $\LambdaECSmin\ge\kappa/2$ and we take the \emph{Real} part of $R(z)$, $\Re[R(z)]$.

It turns out, however, that due to the branch cut in $R(z)[\mathrm{IMP}]$,
the function $\GN[\mathrm{IMP}]$ is not analytic in the domain we need. 
To correct for this, we will instead model the \LayerQualitySquared using the \emph{Real Part} $\Re[\GN[\mathrm{IMP}]]$,
which yields (See Appendix~\ref{sxn:IMP}):
\begin{equation}
  \label{eqn:IMP_model_2}
G(\lambda)[IMP] = \int_{\LambdaECSmin}^\lambda \frac{\kappa}{z}\;dz = \kappa \left[ \ln z \right]_{\LambdaECSmin}^\lambda = \kappa \left( \ln \lambda - \ln \LambdaECSmin\right).
\end{equation}
Notice this expression is very similar to the expression for the Free Cauchy model above (\EQN~\ref{eqn:free_cauchy_G_max2}), except
now it is terms of the logarithm of $\lambda$, and rescaled by $\kappa$.
By associating $\alpha=2\;\kappa$, we recover the \ALPHAHAT formular, albeit with a different normalization.

\subsubsection{The Multiplicative-Wishart (MW) model}
The Multiplicative-Wishart (MW) model  has two real, adjustable parameters $\epsilon,\phi$.
It treats $\XMAT$ as resulting from a product of random matrices, and is good for modeling an ESD with a Very Heavy Tail (VHT).
It has been used previously to model the heavy tail of the Hessian matrix in NNs~\cite{PW16_NIPS}.
This model would work better for fitting HT ESDs that decay slower than a PL or TPL.
We note that unlike the IMP, the MW model does not have a branch cut at the ECS, but it does
have $2$ poles, complicating the integration of the \RTransform.
We will not consider this model further here and instead leave this for future work.

\subsubsection{Levy-Wigner Models and the \ALPHAHAT Metric}
Here, we consider General Levy-Wigner (LW) model..
We show how to obtain the \WW \ALPHAHAT metric by modeling VHT ESDs with an approximation to a Levy distribution, suitable for cases where the \HTSR $\alpha\le 2$.  

The~\ALPHAHAT metric has been developed to adjust for \SCALE anomalies that arise from issues like \CorrelationTraps,
rank collapse, and overfitting, which can in many cases make \ALPHA smaller than expected, and even smaller than $2$.
We model these VHT ESDs \emph{as if} they follow a Levy-Stable distribution, generated from a \LevyWigner random matrix.
The \LevyWigner (LW) model treats $\XMAT$ as if it were a \Wigner matrix (and not actually a rectnalgyle correlation  matrix),
s we must decide how to map the \HTSR $\alpha$ to the L\'evy $\alpha_{l}$; for this, we select $\alpha_{l}=\alpha-1$.
(See in Table~\ref{tab:known_r_transforms}.)
The LW model is defined in terms of 3 parameters:
$a$ is a shift parameter, and $b$ is a complex phase factor depending on 2 real factors, $\beta$ and $\gamma$.
Strictly the ESD for an LW model, $\rho_{LW}(\lambda)$, is defined by its characteristic function (i.e., the Fourier Transform of $\rho_{LW}(\lambda)$), but when the ESD is VHT, $\rho_{LW}(\lambda)\sim\lambda^{-\alpha_{l}-1}$, and when $\alpha_{l}\approx 1$, the ESD resembles a PL HT ESD with the \HTSR $\alpha\approx 2$.

Let us model the \RTransform of our \VeryHeavyTailed (VHT) ESDs as
\begin{align}
\label{eqn:LW_model_0} 
R(z)[\mathrm{LW}] & = bz^{\alpha_{l}-1},\;\alpha_{l}\in(0,1) \\ \nonumber
  & = bz^{\alpha-2},\;\alpha\in(0,2),
\end{align}
where $b$ is an unspecified constant (possibly negative and/or complex).
This is not a particularly good model for VHT ESDs in practice; it is simply an approximate model, and
we choose to obtain a formal expression.

Integrating $R(z)[\mathrm{LW}]$, and (as above) taking the approximation $\LambdaECSmin\approx 0$, we obtain 
\begin{equation}
\label{eqn:LW_model_1} 
\GN[\mathrm{LW}] \approx \tfrac{b}{\alpha-2} \lambda^{\alpha-2}  .
\end{equation}
For simplicity, if we now choose $b=\alpha_{l}=\alpha-2$, then $\QT_{\mathrm{LW}}$ takes the form of a Shatten Norm
\begin{equation}
  \label{eqn:LW_model_2}
  \QT_{\mathrm{LW}} = \frac{1}{\MECS}\sum_{i}^{\MECS}\lambda^{\alpha-1}  .
\end{equation}
Selecting the largest eigenvalue as the dominant term in the sum gives
\begin{equation}
\label{eqn:LW_model_2max}
  \QT_{\mathrm{LW}} \approx \lambda_{max}^{\alpha-1}  .
\end{equation}
Taking the logarithm of $\QT_{\mathrm{HT}} $, we obtain 
\begin{equation}
\label{eqn:LW_model_3} 
\log \QT[\mathrm{LW}] \approx (\alpha-1)\log\lambda_{max}
\end{equation}

As with the Free Cauchy (FC), let us approximate $\QT$ by the largest term in the sum over $\GNECSI$, and then let $\lambda=\lambda_{max}$, giving
\begin{equation} 
\label{eqn:LW_model_4} 
\ALPHAHATEQN = \log_{10} \QT_{\mathrm{LW}} \approx (\alpha-1)\log\lambda_{max}   .
\end{equation}
We present this as a formal example, noting that it is slightly different from the result for the IMP model.
(See Table~\ref{tab:htsr_layer_quality}). 
We do not claim this is a valid empirical model, as we have not attempted to fit a real-world ESD to Levy-stable distribution.  
We leave this to a future study, noting however, there has been some work doing such fits~\cite{li2024exploring}.

Ideally, we would like to have a rigorous expression for $R(z)$ not just
in the case of \IdealLearning but also for the entire \FatTailed Universality class.
This is nontrivial to obtain and we will attempt this in a future work.
For now, we will take a different approach and evaluate $R(z)$ explicitly using numerical techniques.

\paragraph{Scaling insight.}
For the VHT Universality class, we may expect $\lambda_{\max}$ to be fairly large such that
$\lambda_{max}>1$ and therefore $\log_{10}\lambda_{max}>0$
This implies that  $\log_{10}\QT$ decreases  as $\alpha$ (and thus $\alpha_l$) decreases, even though
$\log_{10}\lambda_{max}$ simultaneously \emph{increases} in the VHT class.
This opposing interplay explains why correlation traps produce small $\alpha$ yet yield deceptively moderate quality scores.

\paragraph{Historical remark.}
An earlier study on \ALPHAHAT, reported the opposite trend, with $\ALPHAHAT$ decreasing with increasing model quality.
This is because we employed an un–normalized covariance, suggesting that $\lambda_{max}<1$ and $\log_{10}\lambda_{max}<0$
(although we have not rigorously checked this).
The present $M$ normalization clarifies that the VHT regime is meaningful \emph{only} when $\alpha<2$,
indicating the layer and therefore the model is overfit (in some unspecified way).

\subsubsection{Summary of heuristic models.}
Using the results for different \RTransforms, we can construct several different, related heuristic models for the \LayerQuality
that resemble the very successful \WW \HTSR \ALPHA and \ALPHAHAT metrics.  These heuristic models allow us extend
the \SETOL results for \Ideal learning (i.e., $\alpha=2$) to a wider range of NN layers, covering the \HTSR
HT and VHT Universality classes and correspondingly, NN layers that are both underfit $(\alpha>2)$
and overfit $(\alpha<2)$.

\begin{table}[ht]
  \centering
  \renewcommand{\arraystretch}{2}
  \begin{tabular}{|c|p{3cm}|c|c|} 
    \hline
    \textbf{Model} & \textbf{Tail Norm} & \textbf{WW Metric} & \textbf{$\log\mathcal{Q}$ (Log Quality)} \\ \hline\hline

    Bulk$+$Spikes (BS)
    & Frobenius Norm
    & N/A
    & $\displaystyle \log\left(\sum_{i=1}^{\MECS}\LambdaECS_{i}\right)$ \\ \hline

    Free Cauchy (FC)
    & Spectral Norm
    & \ALPHA $\;\;\alpha$
    & $\displaystyle \log\lambda_{max}\sim 1/\alpha$ \\ \hline
    
    L\'evy Wigner (LW)
    & Shatten Norm
    & \ALPHAHAT $\;\;\hat{\alpha}$
    & $\displaystyle (\alpha-1)\,\log\lambda_{\max}$ \\ \hline
     \hline
    \InverseMP (IMP)
    & ECS Boundary
    & \ALPHAHAT $\;\;\hat{\alpha}$
    & $\displaystyle 2\alpha[\log\lambda_{\max}-\log\lambda_{\min}]$ \\ \hline

  \end{tabular}
  \caption{Closed-form or leading-order expressions for the log \LayerQuality
            $\log\mathcal{Q}$ derived from the integrated $R$–transform for each core
            tail-model, simplified to show the relation to the ~\WW~\ALPHA and ~\ALPHAHAT metrics.
            For each model, we interpret the final result.
            Bulk$+$Space (BS): Sums the $\LambdaECS_{i}$ in the \ECS, giving a Frobenius Tail norm.
            Free Cauchy (FC): Yields the Spectral Norm, $\lambda_{max}$. Since this scales as $1/\alpha$, this explaines the~\HTSR~\ALPHA metric as it shows  why smaller $\alpha$ yields higher $\mathcal{Q}$.
            L\'evy Wigner (LW): Yields a Shatten Norm, which can be approximated by~\ALPHAHAT, $\ALPHAHATEQN$,
            Being for the VHT Universality class, it implies that heavier tails ($\alpha\in(1,2)$) depress \LayerQuality.
            \InverseMP (IMP): Also gives ~\ALPHAHAT, and, importably, a branch cut that defines the \ECS Boundary
             }
  \label{tab:htsr_layer_quality}
\end{table}

Table~\ref{tab:htsr_layer_quality} consolidates the derived expressions for the log \LayerQuality, \(\log\Q\), across multiple \(R\)–transform models, each capturing distinct heavy-tail characteristics of the ESD within the \SETOL framework. The Discrete (Spikes) model yields a Frobenius-like tail norm by summing the effective eigenvalues in the \ECS, directly tying quality to tail magnitude (cf. \EQN~\ref{eqn:spikes_model_4}). The Free Cauchy (FC) model links \(\log\Q\) inversely to the \HTSR metric \ALPHA, demonstrating that smaller \(\alpha\) values enhance quality (cf. \EQN~\ref{eqn:free_cauchy_Q}). The L\'evy-Wigner (LW) model connects \(\log\Q\) to \ALPHAHAT through the term \((\alpha-1)\log\lambda_{\max}\), revealing how very heavy tails depress layer quality (cf.\ \EQN~\ref{eqn:LW_model_4}). The Inverse MP (IMP) model defines the \ECS boundary via a branch cut at \(\kappa/2\) and approximates \(\log\Q\) as \(0.5\log\lambda_{\max}\), providing a robust metric for \IdealLearning scenarios. Each model thus offers a unique lens on layer quality, enabling direct mapping between theoretical \(R\)–transform assumptions and empirical \WW metrics.

\paragraph{Practical Implications for Neural Network Analysis}
These models empower researchers to tailor layer-quality assessments to specific \HTSR universality classes, boosting the predictive power of \WW metrics like \ALPHA and \ALPHAHAT. The Discrete model excels when the ESD exhibits clear eigenvalue spikes, delivering a sharp quality signal for Bulk$+$Spikes layers. The FC model supports \IdealLearning by linking quality directly to \(1/\alpha\), which simplifies cross-layer comparisons. The LW model helps detect overfitting by adjusting for \CorrelationTraps via \ALPHAHAT in very heavy-tailed regimes. Finally, the IMP model offers a parametric fit for both HT and VHT tails, with \(\kappa\) tuning alignment to real-world ESDs (cf.\ Figure~\ref{fig:IMPplots}). By integrating these models, practitioners can compute precise layer-quality metrics that optimize both performance and interpretability.

\subsection{Computational Random Matrix Analysis}
\label{sxn:comp_rmt}
The \RTransform is the generating function for the \emph{\FreeCumulants} of RMT.
Formally--and if we assume a model without a branch-cut or troublesome poles--
one can define $R(z)$ as a series expansion in $z$,
\begin{equation}
  \label{eqn:Rz_expansion}
  R(z) := \kappa_1 + \kappa_2 z + \kappa_3 z^2 + \ldots 
\end{equation}
where the coefficients $\kappa_{k}$ are the free cumulants, which can be expressed
in terms of the matrix moments $m_{k}$~\cite{potters_bouchaud_2020}, defined (here) as
\begin{equation}
  \label{eqn:mk_defn}
  m_{k}:=\Trace{\XECS^{k}}=\sum_{i=1}^{\MECS}(\LambdaECS)^{k}
\end{equation}
where $\LambdaECS_{k}$ is the k-th eigenvalue of the effective correlation matrix $\mathbf\XECS$,
which has been mean-centered and normalized by its standard deviation.

The free cumulants are defined recursively as
\begin{equation}
  \label{eqn:kappa_defn}
  \kappa_k := m_k - \sum_{\text{partitions of } n} \prod_{\text{blocks } B} m_{|B|} 
\end{equation}
\charles{finish explanation}

The first 5 \emph{\Cumulants} are, explicitly,
\begin{align}
  \label{eqn:kappa_defn_2}
  k_1 = m_1 \\ \nonumber
  k_2 = m_2 - m_1^2 \\ \nonumber
  k_3 = m_3 - 3 m_2 m_1 + 2 m_1^3 \\ \nonumber
  k_4 = m_4 - 4 m_3 m_1 - 2 m_2^2 + 10 m_2 m_1^2 - 5 m_1^4 \\ \nonumber 
  k_5 = m_5 - 5 m_4 m_1 + 15 m_3 m_1^2 + 15 m_2^2 m_1 - 35 m_2 m_1^3 - 5 m_3 m_2 + 14 m_1^5
\end{align}

Using these definitions, we can estimate the \LayerQuality matrix $\QT$ for our experimental models
(in Section~\ref{sxn:empirical} by computing $\GN$ for the effective correlation space
(i.e., the tail of the layer ESD,) however it may be defined. That is, we use
\begin{align}
  \label{eqn:G_lambda_series}
\GNI=\kappa_{1}\frac{\LambdaECS}{\MECS}+\frac{\kappa_{2}}{2}\left(\dfrac{\LambdaECS}{\MECS}\right)^{2}+\cdots
\end{align}
This is implemented in the ~\WW package.
In Section~\Ref{sxn:empirical_comp_r_transforms}, we examine some examples of this approach to computing \LayerQualities directly.

\newpage
\section{Empirical Studies}
\label{sxn:empirical}

In this section, we present empirical results. 
Our goals are to justify key technical claims, including key assumptions underlying our \SETOL approach, and to illustrate the behavior of \SETOL with respect to various parameters and hyperparameters.
Importantly, it is \emph{not} our goal to demonstrate that layer PL exponent~\ALPHA~and~\ALPHAHAT~perform well for diagnostics and predicting model quality for SOTA NN models, as that has been demonstrated previously~\cite{MM20a_trends_NatComm,MM21a_simpsons_TR,YTHx22_TR}.

Since the \SETOL theory presented in Section~\ref{sxn:matgen} is (effectively) a single layer theory,
in order to carefully test (as opposed to simply use) \SETOL, we need to limit the degree of inter-layer interactions present in the model.
To do so, we consider a three-layer \MultiLayerPerceptron (MLP3), trained on MNIST~\cite{MNIST1998}. 
We refer to the hidden layers as ``FC1 and ``FC2. Their output sizes and parameter counts are shown in 
Table~\ref{tab:mlp3}.

\begin{table}[h]
\begin{center}
        \begin{tabular}{| c | c | c | c |}
                \hline
                Layer & Units & Weight Parameters           & $\%$ of total \\ \hline \hline
                FC1   &   300 & $768 \times 300 = 230,700$  & $88.2\%$      \\ \hline
                FC2   &   100 & $300 \times 100 = 30,000$   & $11.4\%$      \\ \hline
                out   &    10 &  $10 \times 100 = 1000$     & $0.38\%$      \\ \hline
        \end{tabular}
\end{center}
\caption{Dimensions of each FC layer in the MLP3 model, along with weight matrix parameter count and fraction of the 
total.}
\label{tab:mlp3}
\end{table}

The following are the main topics we consider.

\begin{enumerate}[label=6.\arabic*]
\item
\textbf{\ModelQuality: \HTSR \Phenomenology.}
The \HTSR \Phenomenology provides a metric of model quality in the form of the PL exponent $\alpha$.%
\footnote{Prior work has shown that the \ALPHAHAT metric $(\hat{\alpha})$ accurately describes variations in model 
quality as a function of architecture changes~\cite{MM21a_simpsons_TR}. Since we do not vary the depth of the model in 
our evaluations, the \ALPHA metric ($\alpha$) is of interest in this setting.} 
In particular, smaller values of $\alpha$ (e.g., 
values of $\alpha$ closer to $2$ than $3$ or $4$) should correspond to better models, e.g., having smaller test errors; and
a minimal error should be obtained when $\alpha=2$.
See Section~\ref{sxn:empirical-test_acc}.
\item 
\textbf{\EffectiveCorrelationSpace.}
The \SETOL theory is based on the notion of an \EffectiveCorrelationSpace, in which the learning and generalization occurs. 
This is the low-rank subspace $\mathbf{W}^{\EFF}$ of each layer $\mathbf{W}$ that approximates the teacher $\mathbf{T}$.
In particular, 
our measure of model quality should be restricted to $\mathbf{W}^{\EFF}$.
The \EffectiveCorrelationSpace can be identified from the tail of the ESD, $\rho_{tail}(\lambda)$, and it can be chosen according to one of several related Model Selection Rules.
See Section~\ref{sxn:empirical-effective_corr_space}.
\item 
\textbf{Evaluating the \TRACELOG  Condition.}
In the \HTSR \Phenomenology, when a model is very well-trained, the layer PL exponent $\alpha\simeq 2$.
In the \SETOL theory, when a model is very well-trained, the eigenvalues in the tail will satisfy the Empirical \TRACELOG  Condition, given in \EQN~(\ref{eqn:detX}).
In Section~\ref{sxn:empirical-trace_log}, we provide a detailed analysis of this effect.
\item
\textbf{Computational Model Qualities.}
In Section~\ref{sxn:empirical_comp_r_transforms}, we empirically compare the HTSR layer-quality exponent $\alpha$ against the computational \RTransform-derived metric $\QT$ on fully trained MLP3 models. We show that for the FC1 layer, $\QT$ closely tracks $\alpha$ with small error bars and the expected batch-size trend, whereas for FC2, $\QT$ exhibits much larger variability—demonstrating that $\alpha$ provides a more stable and robust measure of layer quality.  
\item
\textbf{\CorrelationTraps.}
Recall, from Section~\ref{sxn:Traps}, that if a layer weight matrix $\mathbf{W}$ has a \CorrelationTrap
(and, in particular, those arising from SGD training with very large learning rates)
then it is likely that the test (and train) accuracy will be degraded, and $\alpha$ will drop below its optimal value. 
See Section~\ref{sxn:empirical-correlation_trap} for an empirical demonstration of this.
\item 
\textbf{Overloading and Hysteresis Effects.}
The experiments described so far validate that \SETOL makes valid predictions in the $\alpha \gtrsim 2$ range.
Beyond that point, \SETOL only predicts ``atypicality, in the sense of spin glass theory~\cite{nishimori01}. 
See Section~\ref{sxn:hysteresis_effect} for an examination of how the MLP3 behaves when it is pushed further out of that range of validity, e.g., by training only one layer, while keeping the others frozen. In particular, we compare results when a single layer is either over- vs under-parameterized.
\end{enumerate}

\noindent
We trained the MLP3 model independently, using both the Tensorflow 2.0 framework (using the Keras api, and on Google Colab) and pytorch, with the goal of consistent, reproducible results.
Each setting of batch size, learning rate, and trainable layer was trained with $5$ separate starting random seeds, and error bars shown in plots below represent one standard deviation taken over these $5$ random seeds. 
Each training run used the same early stopping criteria on the train loss: training was halted when train loss did not decrease by more than $\Delta E_{train}=0.0001$, over a period of $3$ epochs. 
In doing so, each model was trained with a different number of epochs; and, at the end, the best weights were chosen for the model.
See Appendix~\ref{sxn:appendix_MLP3details} for more details on the MLP3 model and the training setup.
We provide a Google Colab notebook with the exact details, allowing the reader to reproduce the results as desired.

The dominant generalizing components of $\mathbf{W}$ reside in $\mathbf{W}^{\EFF}$ such that it captures the functional contribution of $\mathbf{W}$ to the NN; and thus
\charles{FINISH THOUGHTS HERE}

\subsection{\HTSR Phenomenology: Predicting Model Quality via the \ALPHA metric}
\label{sxn:empirical-test_acc}

\begin{figure}[t]
  \center
  \subfigure[Batch size experiment]{
    \includegraphics[width=6cm]{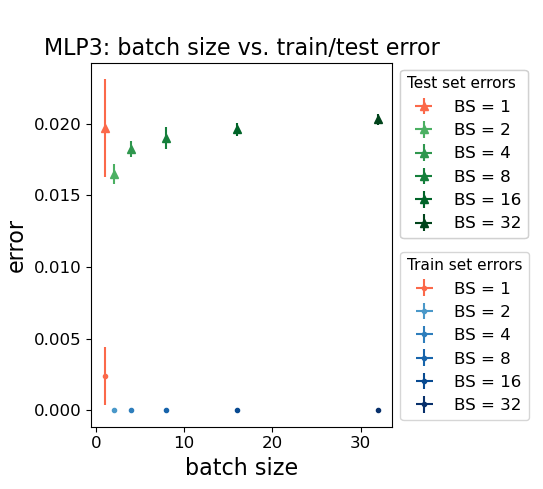}
    \label{fig:mlp3-accuracies-bs}
  }
  \subfigure[Learning rate experiment]{
    \includegraphics[width=6cm]{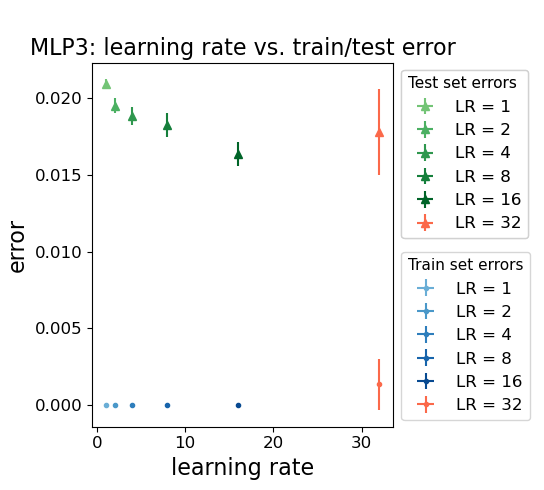}
    \label{fig:mlp3-accuracies-lr}
  }
  \caption{Train / test errors in the MLP3 model as a function of batch size, and learning rate. Observe the inverse relationship between batch size (a) and learning rate (b). As batch size decreases, test error generally decreases, until batch size reached $bs=1$. Similarly, as learning rate increases, test error decreases until $lr=32\times$ the SGD default value of $0.01$.
  }
  \label{fig:mlp3-accuracies}
\end{figure}

Here, we want to determine how the quality of our MLP3 model varies with the \ALPHA metric. 
From previous work~\cite{MM20a_trends_NatComm,MM21a_simpsons_TR,YTHx22_TR}, we expect that \ALPHA metrics for the FC1 and FC2 layers should be well-correlated with the test accuracy, while varying some suitable training knob, such as learning rate or batch size, that can modulate the test accuracy.%
\footnote{Since we do not change the depth of the model here, we expect the \ALPHA metric to follow the \ALPHAHAT metric, also predicting the test accuracies~\cite{MM21a_simpsons_TR}.}

In doing this, the goal is not to achieve \Thermodynamic equilibrium,
but, instead, by using a common stopping rule, to simulate 
in more realistic situations where models may not have fully converged.   This  allows us to test the theory outside of
its more limited apparent range of validity, 
thereby demonstrating the robustness of \SETOL.

We vary the batch size from small to large, i.e., $bs\in[1,2,4,8,16,32]$, following the setup of previous work on the \HTSR~\Phenomenology~\cite{MM18_TR_JMLRversion}. 
We expect similar effects by varying the learning rate, as it is known that small batch sizes correspond directly to large learning rates~\cite{SKYL17_TR,WT11}. 
Thus, we conducted a second set of experiments where the learning rate was varied by a factor of 
$[1\times,2\times,4\times,8\times,16\times,32\times]$, relative to the SGD default value of $0.01$. 
Adjusting the learning rate or batch size allows us to systematically vary the layer PL exponent $\alpha$ between roughly $2$ and $4$, i.e., within the range in which \SETOL should make the most reliable predictions. 
As an added benefit, it also allows us to use the very small batch size of $1$ to force the model into a state of over-regularization, which we also analyze below.


Consider Figure~\ref{fig:mlp3-accuracies}, which plots the final train and test accuracies as a function of the hyperparameter (batch size or learning rate) used during training for the MLP3 model.
Figure~\ref{fig:mlp3-accuracies-bs} varies batch size, and Figure~\ref{fig:mlp3-accuracies-lr} varies learning rate.
Recall that error bars represent one standard deviation taken over $5$ independent starting random seeds. 
In Figure~\ref{fig:mlp3-accuracies-bs}, we see that by decreasing the batch size ($bs$), and holding other knobs constant, we can systematically improve the train and test accuracy, up to a point. 
In particular, for $bs \ge 2$, both the test and train accuracies increase with decreasing batch size, consistent with previous work~\cite{MM18_TR_JMLRversion}.
Further decrease beyond $bs=2$ leads to \emph{lower} model quality, i.e., larger error and larger error variability.
In Figure~\ref{fig:mlp3-accuracies}, we see that increasing the learning rate ($lr$) by a factor $x$ has an exactly analogous effect as decreasing the batch size by $1/x$.%
\footnote{One could, of course, mitigate this by fiddling with other knobs of the training process, but that is not our goal.  Our goal here is not to use a toy model to demonstrate the properties and predictions of \SETOL.}

\begin{figure}[t]
    \centering
    \subfigure[$\alpha_{FC1}$]{
        \includegraphics[width=6cm]{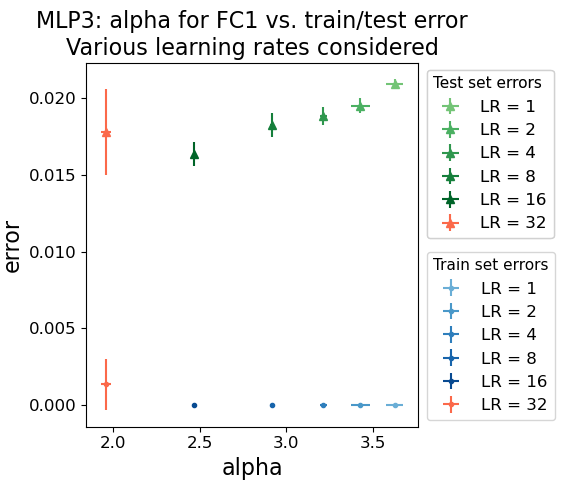}
        \label{fig:mlp3-alpha-fc1-by-lr}
    } 
    \subfigure[$\alpha_{FC2}$]{
        \includegraphics[width=6cm]{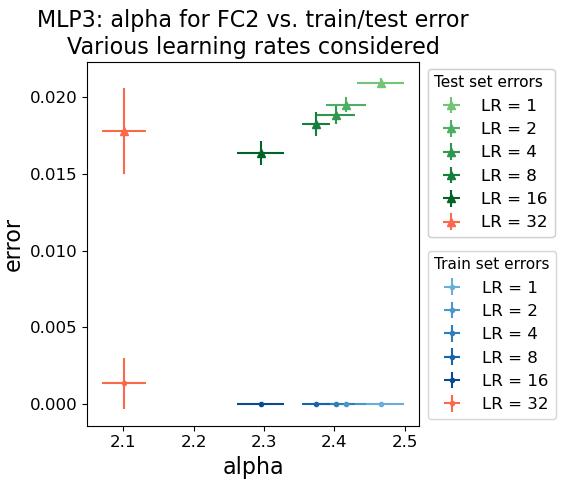}
        \label{fig:mlp3-alpha-fc2-by-lr}
    } 

    \caption{
            Train / test errors in the MLP3 model in the {\bf \emph{\LearningRate}} experiment as a function of $\alpha_{FC1}$ (a) and $\alpha_{FC2}$ (b). 
            Observe the regular downward progression of $\alpha$ and error as the learning 
            rate increased in both (a) and (b). When learning rate was $32\times$, (shown in red), $\alpha_{FC1}$ fell 
            below $2$, coinciding with a drastic increase in both train and test error. The results here almost 
            perfectly replicate those of the Batch Size experiment, shown in Figure~\ref{fig:mlp3-alphas-bs}.
              for Figure~\ref{fig:mlp3-alphas-bs}.}
 \label{fig:mlp3-alphas-lr}
\end{figure}

\begin{figure}[t] 
    \centering
    \subfigure[$\alpha_{FC1}$]{
        \includegraphics[width=6cm]{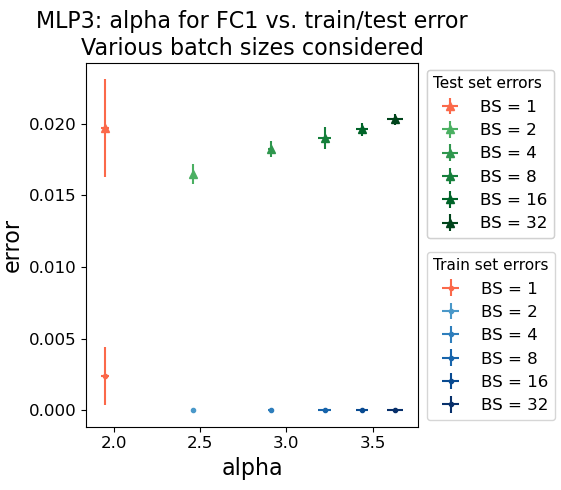}
        \label{fig:mlp3-alpha-fc1-by-bs}
    } 
    \subfigure[$\alpha_{FC2}$]{
        \includegraphics[width=6cm]{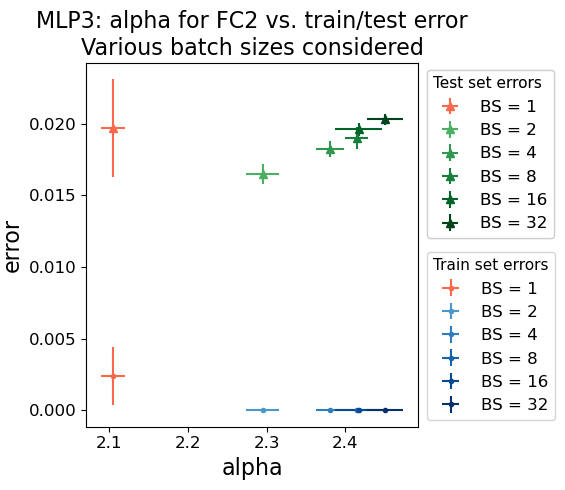}
        \label{fig:mlp3-alpha-fc2-by-bs}
    } 

    \caption{
            Train / test errors in the MLP3 model in the {\bf Batch Size} experiment as a function of $\alpha_{FC1}$ 
            (a) and $\alpha_{FC2}$ (b). Observe the regular downward progression of $\alpha$ and error as the batch size
            decreased in both (a) and (b). When batch size was $1$, (shown in red), $\alpha_{FC1}$ fell below $2$, 
            coinciding with a drastic increase in both train and test error. The results here almost perfectly 
            replicate those of the \LearningRate experiment, shown in Figure~\ref{fig:mlp3-alphas-lr}.
    }
 \label{fig:mlp3-alphas-bs}
\end{figure}

The transition between $lr=16\times$ (or $bs=2$), which is locally optimal for the setting of other hyperparameters, and $lr=32\times$ normal (or $bs=1$), which is not, provides a demonstration of a distinct change in the behavior of \ALPHA, concordant with the sudden increase in the error and error variability. 
To explore this in the context of \SETOL, consider Figure~\ref{fig:mlp3-alphas-lr} and Figure~\ref{fig:mlp3-alphas-bs}, which plot error as a function of \ALPHA, for different learning rates and batch sizes, respectively.

Figure~\ref{fig:mlp3-alphas-lr} plots the \ALPHA metrics $\alpha_{FC1}$ and $\alpha_{FC2}$, as learning rate is varied, demonstrating that both metrics are well-correlated with the test accuracies, for all learning rates less than $16\times$ normal. 
In particular, as we drive \ALPHA in FC1 down to an \Ideal value of $\alpha\simeq 2$, the test error decreases monotonically (Figure~\ref{fig:mlp3-alpha-fc1-by-lr}).
Beyond that point, further decrease of the batch size sees \ALPHA decrease below its \Ideal value of $2$ in the FC1 layer.
This corresponds not only with \emph{higher errors}, but also with \emph{larger error bars}, on both train and test error. 
The dramatic increase in train error and train error variability is particularly telling, because it suggests that when 
$\alpha_{FC1}$ passes below $2$, the model enters into a ``glassy state, and is unable to relax down to $0$ train error.

In Figure~\ref{fig:mlp3-alpha-fc2-by-lr}, we consider \ALPHA for FC2, and we see that $\alpha_{FC2}$ approaches $2$, but 
does not reach it, even for $lr=32\times$. This failure to achieve $\alpha_{FC2} \approx 2$, along with the much greater size 
of FC1, (See Table~\ref{tab:mlp3},) suggests that FC1 is the critical layer for the models performance. 
This also highlights some of the interplay between the layers, (which is not captured by a single layer theory) -- as $\alpha_{FC1}$ has narrow error bars throughout, $\alpha_{FC2}$ shows much more variation by way of its wider error bars. 
Thus, while model accuracy kept improving as learning rate increased up to $16\times$, this was likely driven by a better $\alpha_{FC1}$, more than by $\alpha_{FC2}$. 

In Figure~\ref{fig:mlp3-alphas-bs}, we consider batch size, and we see a near identical replication of these results, in terms of the relation of train error, test error and \ALPHA in the two layers. 
Consequently, the remainder of the experiments will focus on the learning rate experiment, as both produced substantially the same results.

\subsection{Testing the Effective Correlation Space}
\label{sxn:empirical-effective_corr_space}

Here, we will address the question: 
\begin{quote}
How shall we \emph{test the assumption} of the \EffectiveCorrelationSpace?
\end{quote}
Recall that the \SETOL theory estimates model quality by evaluating the ST \GeneralizationError as an integral over the theoretical training data $\boldsymbol{\xi}$.
This integral assumes each layer weight matrix can be replaced with an effectively lower rank form, i.e., $\mathbf{W}\rightarrow\mathbf{W}^{\EFF}$, corresponding to the span of the eigencomponents defined by the tail of the ESD, $\rho_{tail}(\lambda)$. 
In the \HTSR \Phenomenology, the tail is defined by the fact that $\rho_{tail}(\lambda)$ follows a PL distribution, above some minimal value $\lambda_{min}$.
In our \SETOL theory, the tail is defined by choosing the minimal value $\lambda_{min}$ to satisfy the Empirical \TRACELOG  condition.
These methods of realizing $\mathbf{W}^{\EFF}$ are essentially \emph{Model Selection Rules} (MSRs) for the \EffectiveCorrelationSpace. 
%
Importantly, in neither approach is $\lambda_{min}$ just some ``rank parameter to be chosen by yet some other MSR on 
the basis of rank, or magnitude alone, (that, in particular, does not know about \HTSR or \SETOL, which consider the 
\emph{shape} of the ESD).

Thus, to test the assumption of the \EffectiveCorrelationSpace, we want to show that the models predictions are in fact controlled predominantly by the tail, where the specific choice of the rank parameter depends on \HTSR or \SETOL as we expect.
We can emulate this theoretical construct and estimate (trends in the) test accuracies by evaluating the train and/or 
test accuracies of the trained MLP3 model -- after replacing the MLP3 layer weight matrices $\mathbf{W}_{FC1}$ and $\mathbf{W}_{FC2}$ with a low-rank approximation consisting of \emph{only} the tail:
\begin{align*}
 \mathbf{W}^{\EFF}_{FC1}:= P_{tail}\mathbf{W}_{FC1} \\
 \mathbf{W}^{\EFF}_{FC2}:= P_{tail}\mathbf{W}_{FC2}  ,
\end{align*}
where $P_{tail}$ is a projection operator selecting only the tail of the ESD with TrucnatedSVD.
(That is, we will use the  low-rank TruncatedSVD approximation at the inference step, not at the training step, as is more common.)
A Truncated model is one whose weight matrices $\mathbf{W}_*$ have been replaced by truncated matrices $\mathbf{W}^{\EFF}_*$. 
We denote the difference between the original models accuracy and the Truncated models train and test accuracy as $\Delta E_{train}$ and $\Delta E_{test}$, respectively:
\begin{align*}
  \Delta E_{train}:=  E_{train}(\mathcal{D}) - E^{\EFF}_{train}(\mathcal{D}) \\
  \Delta E_{test}:=  E_{test}(\mathcal{D}) - E^{\EFF}_{test}(\mathcal{D})  ,
\end{align*}
where $E^{\EFF}_{train}$ denotes the error of the TruncatedSVD model on the training portion of the dataset $\mathcal{D}$,a
and  $E^{\EFF}_{test}$ denotes corresponding test error for the TruncatedSVD model.

\paragraph{The \POWERLAW~and \TRACELOG~Model Selection Rules}
\charles{is MSR still a good theme here?  In the orginal paper we discussed the MSR up front. Now
  we seem to just casually mention it here.}
If we use good MSRs, then we expect that $\Delta E_{train}\to 0$ and $\Delta E_{test}\to 0$ as the models approach \IdealLearning. 
We consider the following MSRs,%
\footnote{We considered other MSRs that do not ``know about \HTSR or \SETOL, but they (expectedly) perform in trivial or uninteresting ways for testing the assumption of the \EffectiveCorrelationSpace.  Thus, we are not introducing just some arbitrary low-rank approximation, as is common, but instead that the specific \SETOL-based MSR matters.} 
which are associated with the \HTSR and \SETOL approaches.
\begin{itemize}
\item 
The \POWERLAW~MSR: 
All eigenvalues lying in the tail of the ESD, 
$\lambda_i \ge \LAMBDAPL$, where $\LAMBDAPL$ is the start of the PL tail, as determined by the \WW~PL fit, which is based on \cite{CSN09_powerlaw}.
\item 
The \TRACELOG~MSR: 
All eigenvalues lying in the tail of the ESD, 
such that they satisfy the \TRACELOG  Condition, i.e., $\lambda_{i}\ge \LAMBDADETX$, where $\prod\lambda_{i:\lambda_i \ge \LAMBDADETX}\simeq 1$.
\end{itemize}
\chris{CH TODO: Find a place to call out the need for Normalizing $W$ to Frobenius norm = $M$.}

%



\subsubsection{Train and test errors by epochs}
\label{sxn:trunc_err_epochs}
\begin{figure}[t]
    \centering
    \subfigure[$lr=1\times$]{
        \includegraphics[width=5cm]{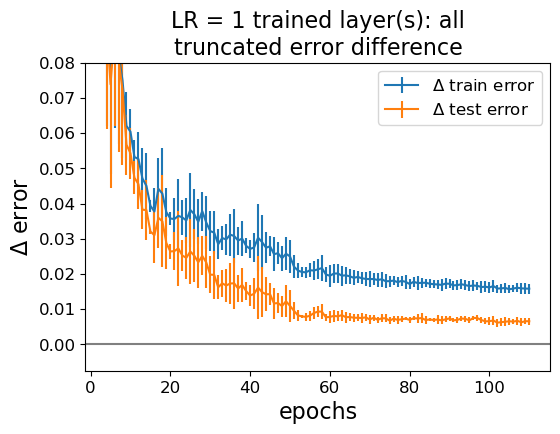}
        \label{fig:mlp3-trunc_err_epochs_xmin_1}
    }
    \subfigure[$lr=2\times$]{
        \includegraphics[width=5cm]{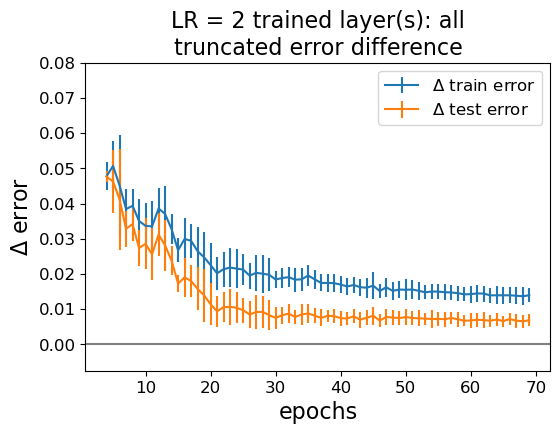}
        \label{fig:mlp3-trunc_err_epochs_xmin_2}
    }
    \subfigure[$lr=4\times$]{
        \includegraphics[width=5cm]{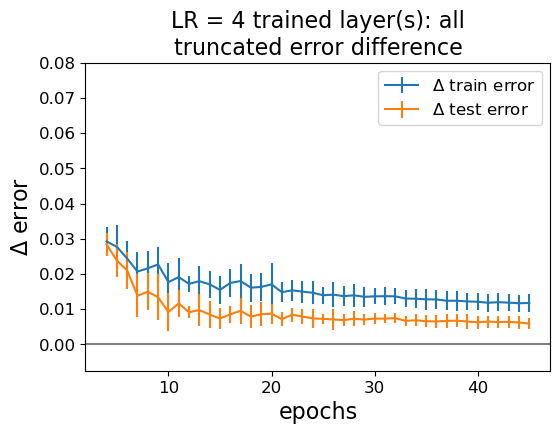}
        \label{fig:mlp3-trunc_err_epochs_xmin_4}
    }\\
    \subfigure[$lr=8\times$]{
        \includegraphics[width=5cm]{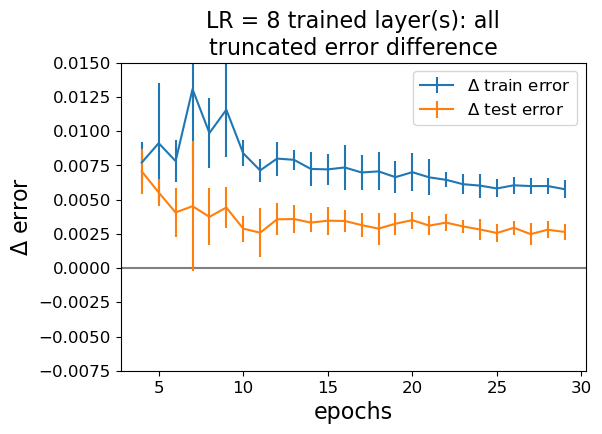}
        \label{fig:mlp3-trunc_err_epochs_xmin_8}
    }
    \subfigure[$lr=16\times$]{
        \includegraphics[width=5cm]{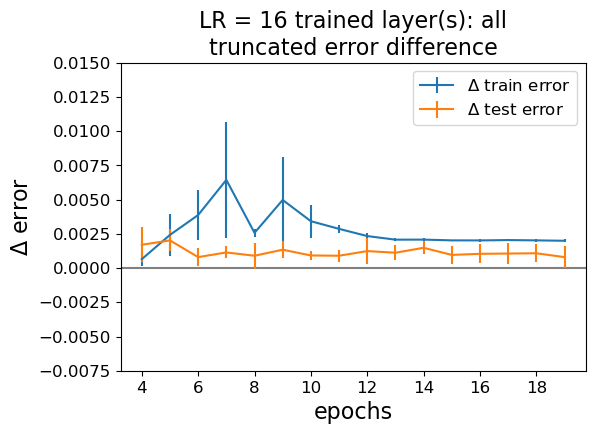}
        \label{fig:mlp3-trunc_err_epochs_xmin_16}
    }
    \subfigure[$lr=32\times$]{
        \includegraphics[width=5cm]{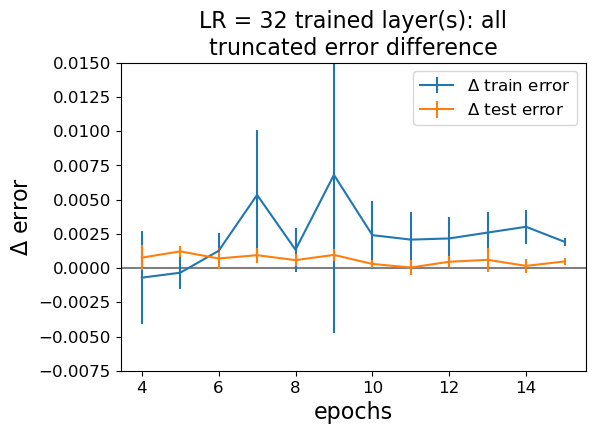}
        \label{fig:mlp3-trunc_err_epochs_xmin_32}
    }
    \caption{
            $\Delta E_{train}$ (blue) and $\Delta E_{test}$ (orange) for various learning rates, using the 
            \POWERLAW~MSR. As learning rate increases we can see that $\Delta E_{train}$ and $\Delta E_{test}$ both tend 
            towards lower asymptotic minima, which they reach after fewer epochs of training. We can also see that 
            (after the first few epochs,) $\Delta E_{train}$ (blue) is always higher than $\Delta E_{test}$ (orange). 
            Observe that in the bottom row (d--f) the yaxis is contracted to make the variation more visible. In (f) we can see 
            that as learning rate surpasses its optimal setting, the gap between $\Delta E_{train}$ and $\Delta 
            E_{test}$ begins to increase again, and has wider error bars, suggesting that the excessively large learning 
            rate is disrupting the MLP3s ability to lear the \EffectiveCorrelationSpace.
    }
    \label{fig:mlp3-msr-results-xmin}
\end{figure}

To see how the \EffectiveCorrelationSpace forms, we plot how $\Delta E_{train}$ and $\Delta E_{test}$ evolve over training, for each of the various learning rates considered.%
\footnote{When batch size was varied, results did not significantly differ, and so they are omitted. 
} 

We start with the effect of the \POWERLAW~MSR.
See Figure~\ref{fig:mlp3-msr-results-xmin}, where we see that $\Delta E_{train}$ and $\Delta E_{test}$ generally trend 
downwards as they approach minimum train error. 
When the learning rate is larger, the models converge more quickly, and $\Delta E_{train}$ and $\Delta E_{test}$ also converge to lower values. 
Recall from Figure~\ref{fig:mlp3-accuracies-lr} that $lr=16\times$ had the lowest test error. In 
Figure~\ref{fig:mlp3-trunc_err_epochs_xmin_16}, we see that it also has the lowest $\Delta E_{train}$ and $\Delta 
E_{test}$. A lower $\Delta E_{train}$ or $\Delta E_{test}$ means that more of the models correct predictions are due to 
the low rank tail, meaning that the tail generalizes better, and we see here that when the tail generalizes best, the 
model was the most accurate. 

In each plot, we also see that the error bars are wide early on, before suddenly becoming much narrower. This transition 
is more visible in the larger learning rates shown in~\ref{fig:mlp3-trunc_err_epochs_xmin_8}--\ref{fig:mlp3-trunc_err_epochs_xmin_32},
but can also be seen in \ref{fig:mlp3-trunc_err_epochs_xmin_1}--\ref{fig:mlp3-trunc_err_epochs_xmin_4}, albeit less 
clearly. Most interestingly of all, this transition is preceded by a brief period, sometimes a single epoch, in which the 
error bars are drastically wider, in a way that is reminiscent of a first-order phase transition. Again, this 
phenomenon can be seen most clearly in ~\ref{fig:mlp3-trunc_err_epochs_xmin_8}--\ref{fig:mlp3-trunc_err_epochs_xmin_32}.

\begin{figure}[t]
    \centering
    \subfigure[$lr=1\times$]{
        \includegraphics[width=5cm]{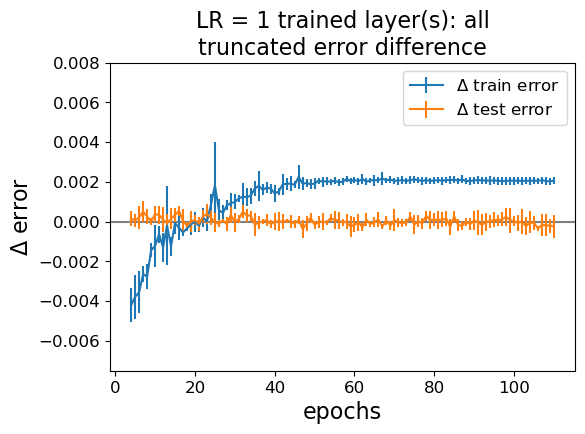}
        \label{fig:mlp3-trunc_err_epochs_detX_1}
    }
    \subfigure[$lr=2\times$]{
        \includegraphics[width=5cm]{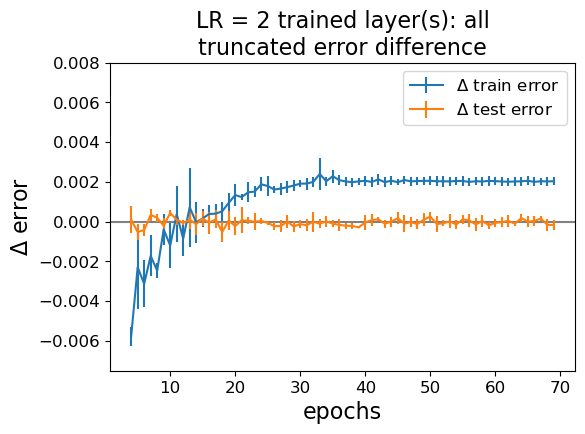}
        \label{fig:mlp3-trunc_err_epochs_detX_2}
    }
    \subfigure[$lr=4\times$]{
        \includegraphics[width=5cm]{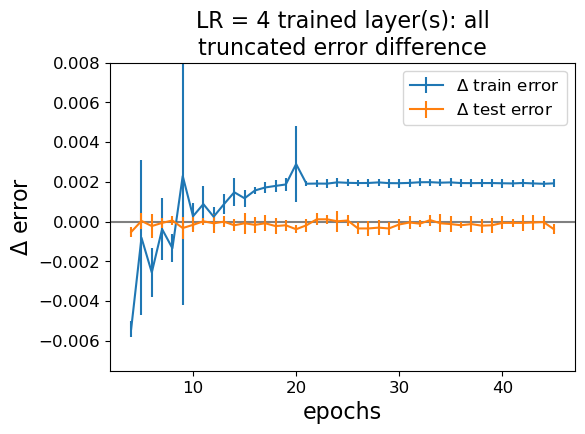}
        \label{fig:mlp3-trunc_err_epochs_detX_4}
    }\\
    \subfigure[$lr=8\times$]{
        \includegraphics[width=5cm]{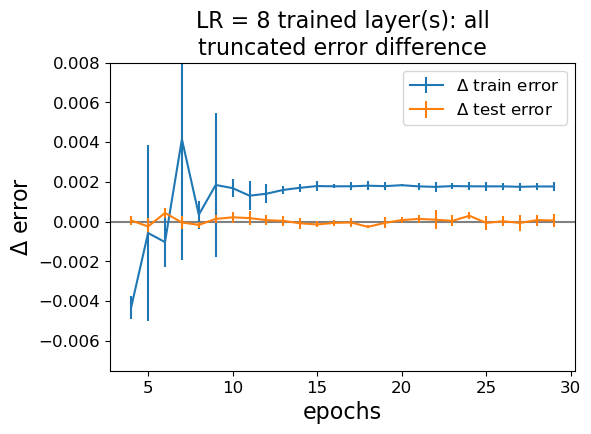}
        \label{fig:mlp3-trunc_err_epochs_detX_8}
    }
    \subfigure[$lr=16\times$]{
        \includegraphics[width=5cm]{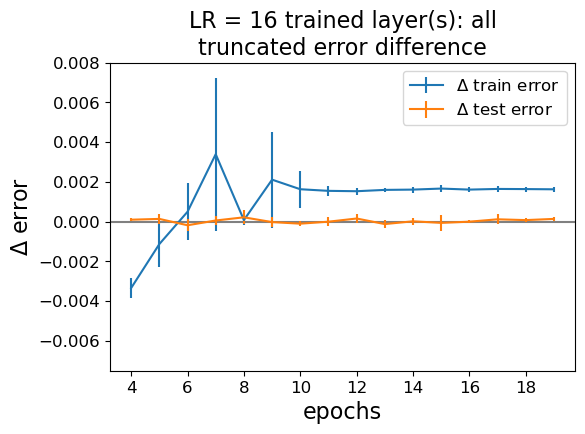}
        \label{fig:mlp3-trunc_err_epochs_detX_16}
    }
    \subfigure[$lr=32\times$]{
        \includegraphics[width=5cm]{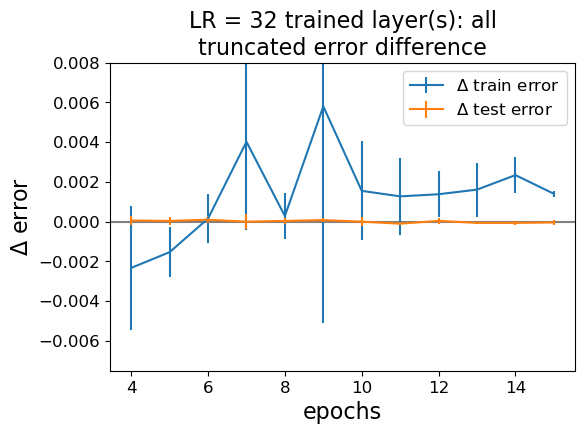}
        \label{fig:mlp3-trunc_err_epochs_detX_32}
    }
    \caption{
        $\Delta E_{train}$ (blue) and $\Delta E_{test}$ (orange) for selected learning rates, using the \TRACELOG~MSR. 
        For all learning rates, $\Delta E_{test}$ is centered on $0$, meaning that the \TRACELOG~Effective Correlation 
        Space explains almost all variation in out-of-sample predictions, but it does \emph{not} explain all of the 
        training set predictions (blue). NOTE: The y axis is the same in all plots, and is much narrower than in 
        Figure~\ref{fig:mlp3-msr-results-xmin}. In (a--e) we can see that $\Delta E_{train}$ converges to approximately 
        $0.002$. Compare with Figure~\ref{fig:mlp3-trunc_err_epochs_xmin_16}, which reaches a minimum of approximately 
        $0.0025$. As in Figure~\ref{fig:mlp3-trunc_err_epochs_xmin_32}, the learning rate of $32\times$ normal disrputs the 
        MLP3s ability to discover the \TRACELOG \EffectiveCorrelationSpace.
    }
    \label{fig:mlp3-msr-results-detX}
\end{figure}

We next consider the effect of the \TRACELOG~MSR.
See Figure~\ref{fig:mlp3-msr-results-detX}, which also shows the development of $\Delta E_{train}$ and $\Delta E_{test}$ 
over epochs, where we see a very different pattern in the train error and test error. 
The difference in the train error is because, as the model is untrained in the early epochs, the \TRACELOG~MSR 
\emph{over}-estimates the tail by choosing a $\lambda_{min}$ that is too small. 
Thus, $\Delta E_{train}$ actually increases to its asymptotic value at the final epoch. 
In the earliest epochs, the truncated train error is even \emph{less} than the full MLP3 models error, suggesting that 
signal is forming in the large eigenvalues in these early epochs, but is swamped by the randomness of the early initial 
weights, some of which is then removed by truncation. As epochs progress, this effect disappears.
Here again we can see the ``phase-transition-like behavior of the train error, as the error bars are wide early on, up to a transition having an abnormally large error bar, after which they stabilize.

Perhaps most interestingly of all, we see that under the \TRACELOG Model Selection Rule (MSR), $\Delta E_{test}$ is flat \emph{throughout 
training}, and for all learning rates. This implies that there is \emph{no} point in training where the 
\TRACELOG tail generalizes badly. In other words, almost none of the out-of-sample variance falls outside of the \EffectiveCorrelationSpace, as the \TRACELOG~MSR over-estimates the \EffectiveCorrelationSpace. It also bears noticing that under the 
\POWERLAW~MSR, (Figure~\ref{fig:mlp3-msr-results-xmin},) this is decidedly not the case, because the \POWERLAW~MSR \emph{under}-estimates the \EffectiveCorrelationSpace. As we will see in \ref{sxn:empirical-trace_log}, as \ALPHA approaches $2$, the gap between the two MSRs goes to $0$.

There are two final points of comparison between Figures~\ref{fig:mlp3-msr-results-xmin} and~\ref{fig:mlp3-msr-results-detX}.
First, 
although it appears in Figure~\ref{fig:mlp3-msr-results-detX} that $\Delta E_{train}$ converges to a larger value than in Figure~\ref{fig:mlp3-msr-results-xmin}, this is because the scale of the y-axis is $10\times$ smaller. 
That is, the \POWERLAW~MSR is biased towards over-estimating $\lambda_{min}$, which means it over-truncates, producing a 
larger $\Delta E_{train}$ or $\Delta E_{test}$ 
than the \TRACELOG~MSR, which is biased towards under-estimating $\lambda_{min}$. 
Second,
in both Figure~\ref{fig:mlp3-msr-results-xmin} and~\ref{fig:mlp3-msr-results-detX}, we can see that $\Delta E_{test}$ is consistently lower than $\Delta E_{train}$. 
Clearly, truncating has a larger effect on train predictions, meaning that no matter how long the model is trained, some 
of the train predictions are still derived from the bulk. The fact that they remain is evidence that the gradient does not degrade them, meaning that they were most likely contributing spuriously correct answers. Yet, the test predictions are far less affected, meaning that only the \EffectiveCorrelationSpace contributes to the models ability to generalize.

\subsubsection{Truncation and Generalization}

\begin{figure}[t]
  \centering
  \includegraphics[width=15cm]{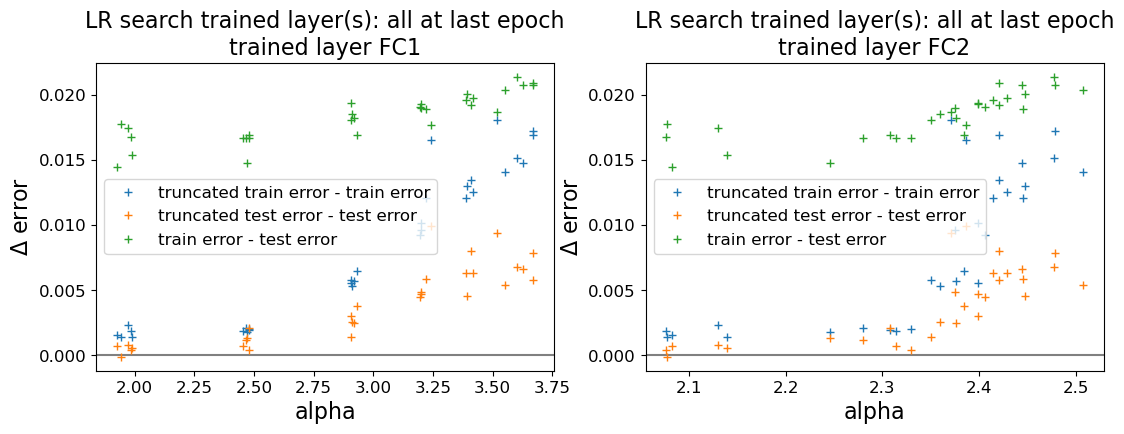}
  \caption{
        Train and test error gaps using the \POWERLAW~MSR, as a function of alpha in the FC1 and FC2 layers of MLP3 
        models, at the \emph{final epoch} of training. We can see that as alpha decreases towards $2$, (right to left), 
        $\Delta E_{train}$ and $\Delta E_{test}$ generally decrease as well, meaning that the closer $\ALPHA$ is to $2$, 
        the more the Effective Correllation Space explains the train and test predictions. The gap between un-truncated 
        train and test error, (\EQN \ref{eqn:gen_gap},) decreases as well until $\alpha_{FC1} < 2$.
  }
  \label{fig:mlp3-alpha-generalization-gap}
\end{figure}

Given that $\Delta E_{test}$ is always lower than $\Delta E_{train}$ for the \POWERLAW~MSR, and similarly for \TRACELOG~after a certain point in training, it is clear that the \EffectiveCorrelationSpace has something to do with generalization.
However, this leaves open the question of what role precisely \ALPHA plays. 
In Figure~\ref{fig:mlp3-alpha-generalization-gap}, we plot $\Delta E_{train}$ and $\Delta E_{test}$ with the \POWERLAW~MSR as a function of \ALPHA (rather than epochs) for layers FC1 and FC2, as well as the \GeneralizationGap -- that is, $E_{test} - E_{train}$. 
(Recall \EQN~\ref{eqn:gen_gap}.) 
Learning rate is not explicitly shown, but its effect can be seen in the clusters of points that each learning rate generates.

In both layers, $\Delta E_{train}$ and $\Delta E_{test}$ steadily decrease with $\alpha_{FC1}$, until it passes below 
$2$, after which the relation deteriorates somewhat.
This is especially prominent in FC2. 
Recall from Figure~\ref{fig:mlp3-alphas-lr}, Section~\ref{sxn:empirical-test_acc}, that when $\alpha_{FC1}$ passed below $2$, the train error and test error both increased and exhibited larger variability. 
From this we interpret \ALPHA as a measure of \emph{regularization} (which is consistent with its introduction as a measure of 
implicit self-regularization~\cite{MM18_TR_JMLRversion}).
Regularization has the effect of keeping train and test accuracy closer together, and generally, as \ALPHA in the dominant layer decreases towards $2$ from above, the train-test error gap decreases.

\subsection{Evaluating the \TRACELOG  Condition}
\label{sxn:empirical-trace_log}

Having established that the PL tail of the ESD, defined by eigenvalues above $\LAMBDAPL$, is a major factor in determining model quality in the MLP3 model, we now examine how well the \TRACELOG  Condition compares with it. 
In particular, we demonstrate that when the tail of a layer ESD is described well by the \HTSR \Phenomenology, i.e., when 
it is well-fit by a PL with $\rho(\lambda)_{tail}\sim\lambda^{-\alpha}$, with PL exponent $\alpha\simeq2$, then the 
eigenvalues in the tail defined by the PL fit, i.e., $\lambda\ge\LAMBDAPL$, \emph{also} satisfy the \TRACELOG  Condition 
of \EQN~\ref{eqn:detX}---\emph{a key assumption of the \SETOL theory}.
This is a rather remarkable empirical result that couples \HTSR and \SETOL; it has its basis in our \SETOL derivation; and it provides the basis for an inductive principle that is based on the product of eigenvalues rather than an eigenvalue gap.

We can denote the eigenvalue that best fits the \TRACELOG  Condition as $\LAMBDADETX$. 
Then, to measure how well this condition holds, we can compute
\begin{align}
        \label{eqn:D_lambda_min}
        \Delta \lambda_{min} = \LAMBDAPL - \LAMBDADETX  .
\end{align}
In Sections~\ref{sxn:detx-mlp3} and~\ref{sxn:detx-sota}, we will see the trend that as $\alpha$ approaches $2$, $\LAMBDAPL$ and $\LAMBDADETX$ also approach one another, and hence $\Delta \lambda_{min}$ goes to $0$, from above, both for our toy MLP3 model as for SOTA models.
In our MLP3 model, we will see that a crossing of the equality condition coincides with over-regularization and a degradation in model accuracy. 
In pre-trained ResNet\cite{ResNet15_TR}, VGG\cite{VGG14_TR} and ViT\cite{VIT20_TR} models, we will also see, empirically, that in general $\Delta \lambda_{min}$ remains positive, just as $\alpha$ remains above $2$.

\subsubsection{The MLP3 model}
\label{sxn:detx-mlp3}

\begin{figure}[t] 
    \centering
    \subfigure[$LR=8\times$]{
      \includegraphics[width=5cm]{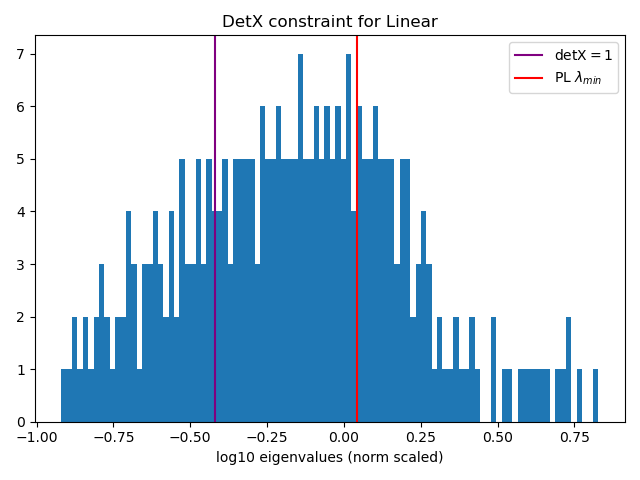}
        \label{fig:mlp3-randesd-8}
    }
    \subfigure[$LR=16\times$]{
      \includegraphics[width=5cm]{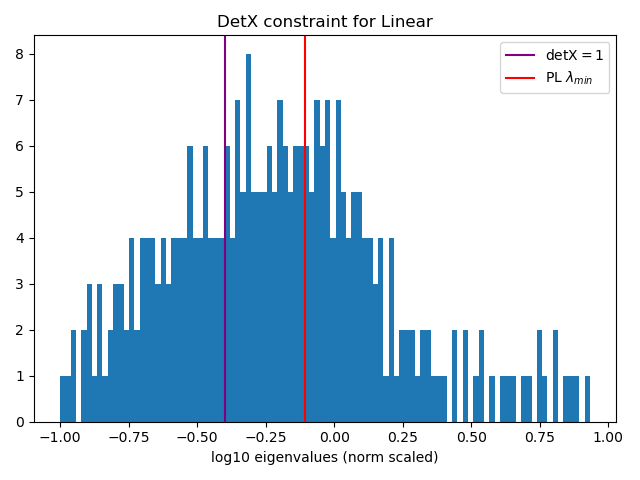}
        \label{fig:mlp3-randesd-16}
    }
    \subfigure[$LR=32\times$]{
      \includegraphics[width=5cm]{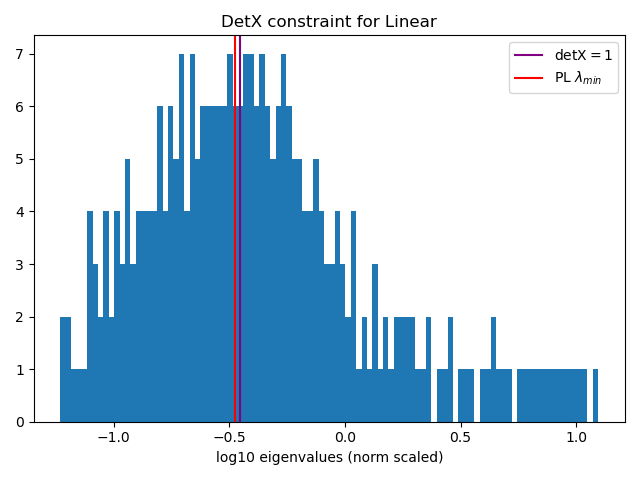}
        \label{fig:mlp3-randesd-32}
    }
    \caption{
        Log-Linear ESDs for three learning rates in the FC1 layer of MLP3. The red line shows $\LAMBDAPL$, and the 
        purple line shows $\LAMBDADETX$. Observe that the purple line is to the left of the red line, but as the LR 
        increases they move closer together. However, when LR is $32\times$, where both 
        train and test accuracy suffered (c), the red line is to the left of the purple line. This is often a signature 
        of an \OverRegularized layer, and indeed the FC1 layer in this model had $\alpha < 2$. (See Figure~\ref{fig:mlp3-alpha-fc1-by-lr} in 
        Section~\ref{sxn:empirical-test_acc}.)
    }
    \label{fig:mlp3-tracelognorm}
\end{figure}

Consider Figure~\ref{fig:mlp3-tracelognorm}, which shows $\LAMBDAPL$ and $\LAMBDADETX$ in the FC1 layer of three MLP3 models, each sharing a common starting random seed, that were trained with the largest learning rates.
The $\LAMBDAPL$ and $\LAMBDADETX$ eigenvalues are marked by red and purple vertical lines, respectively; and thus $\Delta \lambda_{min}$ is the distance between red and purple lines.
As learning rate increases, the red and purple lines draw closer, and they are closest for $lr=32\times$ (Figure~\ref{fig:mlp3-randesd-32}). 
(Compare this with Figure~\ref{fig:mlp3-alphas-lr}, Section~\ref{sxn:empirical-test_acc}, which shows that this corresponds with an increase in test accuracy, up to $lr=16\times$, but at $lr=32\times$ \ALPHA fell below $2$ and accuracy suffered.) 
In Figure~\ref{fig:mlp3-randesd-8}-\ref{fig:mlp3-randesd-16}, the purple line is left of the red line; but in Figure~\ref{fig:mlp3-randesd-32}, the red and purple lines cross, such that $\LAMBDAPL < \LAMBDADETX$.
This is analogous to the case where $\alpha$ crosses below $2$. 
This suggests that the absolute \TRACELOG  is minimized when $\alpha\simeq2$, which (remarkably) is exactly when the \HTSR \Phenomenology predicts the layer is \Ideal.

(Observe that \Ideal does not necessarily mean optimal under a finite sized training set, but rather that the finite-sized system behaves the most similarly to its infinite limit.)

\begin{figure}[t] 
    \centering
    \subfigure[Layer FC1 for various learning rates]{
        \includegraphics[width=7cm]{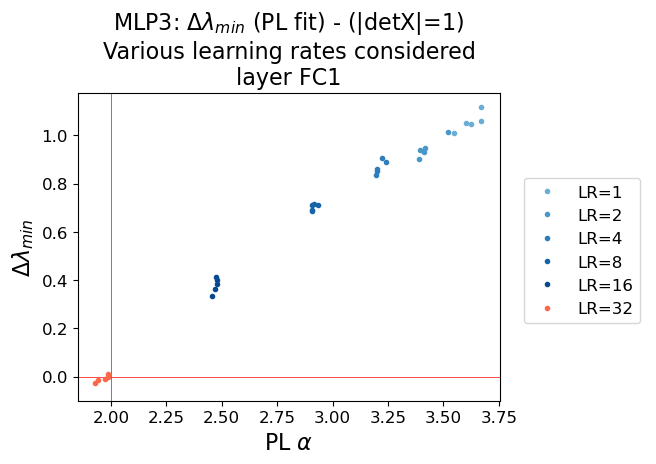}
        \label{fig:mlp3-detx-lr_search_fc1}
    }
    \subfigure[Layer FC2 for various learning rates]{
        \includegraphics[width=7cm]{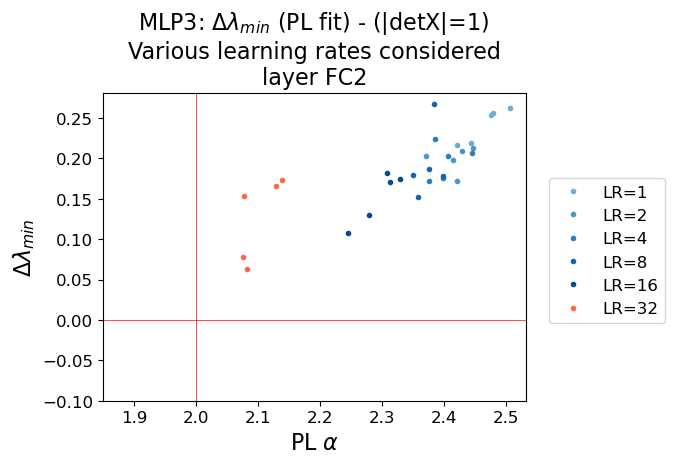}
        \label{fig:mlp3-detx-lr_search_fc2}
    }\\
    \subfigure[Layer FC1 for various batch sizes]{
        \includegraphics[width=7cm]{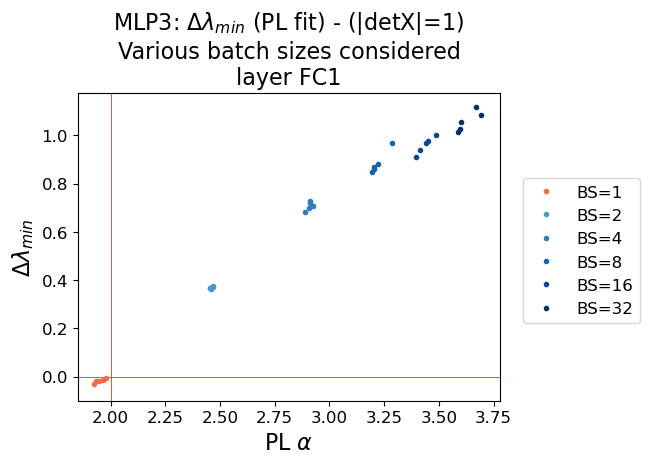}
        \label{fig:mlp3-detx-bs_search_fc1}
    }
    \subfigure[Layer FC2 for various batch sizes]{
        \includegraphics[width=7cm]{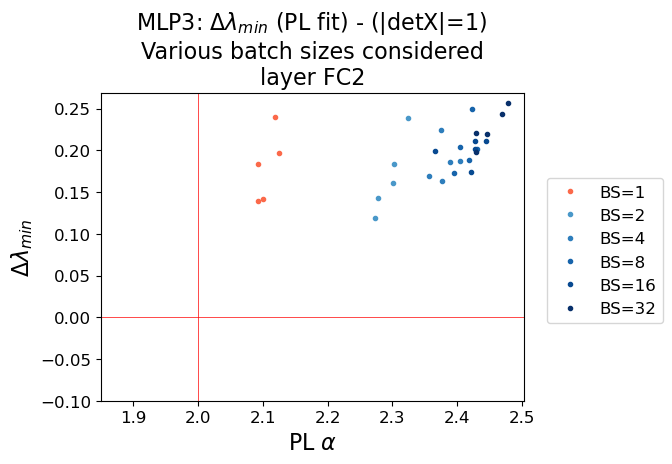}
        \label{fig:mlp3-detx-bs_search_fc2}
    }
    \caption{
            MLP3 Model: Comparison of the PL \ALPHA (x-axis), with the difference between $\LAMBDAPL$ and $\LAMBDADETX$ 
            (y-axis). The thin red lines indicate critical values of $\alpha=2$ and $\Delta \lambda_{min} = 0$. As 
            learning rate increases (a--b) or batch size decreases (c--d), we can see that in layer FC1, which dominates 
            the model, (See Table~\ref{tab:mlp3},) $\alpha$ goes to $2$, and $\Delta \lambda_{min}$ goes to $0$. Observe 
            that both critical values are crossed at the most extreme hyper-parameter selection, (red,) corresponding 
            with over-training. Layer FC2 shows a weaker tendency towards the critical values (b, d), and is disrupted 
            at the most extreme hyper-parameter values (red).
    }
 \label{fig:mlp3-detx-gap}
\end{figure}

We can compare  $\alpha$ and $\Delta \lambda_{min}$ more broadly by plotting $\Delta \lambda_{min}$ directly as a function of $\alpha$ in a single plot spanning all random seeds and learning rates or batch sizes.
This is shown in Figure~\ref{fig:mlp3-detx-gap}. 
Critical values of $\alpha=2$ and $\Delta \lambda_{min} = 0$ are shown as vertical and horizontal red lines, respectively. 
Values for various learning rates are plotted for layer FC1 
(Figure~\ref{fig:mlp3-detx-lr_search_fc1}) and FC2 (Figure~\ref{fig:mlp3-detx-lr_search_fc2}), as well as for various batch sizes in layer FC1 (Figure~\ref{fig:mlp3-detx-bs_search_fc1}) and FC2 (Figure~\ref{fig:mlp3-detx-bs_search_fc2}).

For layer FC1 (Figures~\ref{fig:mlp3-detx-lr_search_fc1} and~\ref{fig:mlp3-detx-bs_search_fc1}), in both cases we see near-linear march towards the critical tuple of $(\alpha, \Delta\lambda_{min}) = (2, 0)$.
In addition, passing this critical value coincides with diminished train and test accuracy, (recall 
Figure~\ref{fig:mlp3-accuracies}), suggesting that just as $\alpha=2$ is a threshold of over-regularization, $\LAMBDAPL < \LAMBDADETX$ may be as well. 
Since FC1 is the dominant layer, comprising roughly $8/9$ of the weights of the model, (Table~\ref{tab:mlp3},) we expect 
FC1 to most closely match the performance of the model as a whole.

For layer FC2, which comprises roughly the other $1/9$ of the models weights, there is a similar coevolution, but it is weaker. 
As learning rate or batch size exceeds their critical values, rather than going to $(2, 0)$ as in FC1, we instead see that the relationship simply breaks down, with the gap growing larger even as $\alpha$ decreases. 
Given that FC1 \emph{has} passed the critical $\alpha=2$ threshold, we conjecture that the breakdown of the relationship between $\alpha$ and $\Delta \lambda_{min}$ is due to FC1 becoming \ATypical.

\subsubsection{State-of-the-Art (SOTA) models}
\label{sxn:detx-sota}

Here, we consider SOTA models, 
in particular VGG pre-trained models~\cite{VGG14_TR}, the ResNet series~\cite{ResNet15_TR}, the ViT 
series~\cite{VIT20_TR}, and the DenseNet series~\cite{DenseNet17_TR}.
We show that as $\alpha$ approaches $2$, the Log-Trace Condition holds better and better, i.e., $\Delta\lambda_{min}$ 
approaches $0$.

\begin{figure}[t] 
    \centering
    \subfigure[VGG Series]{
      \includegraphics[width=7cm]{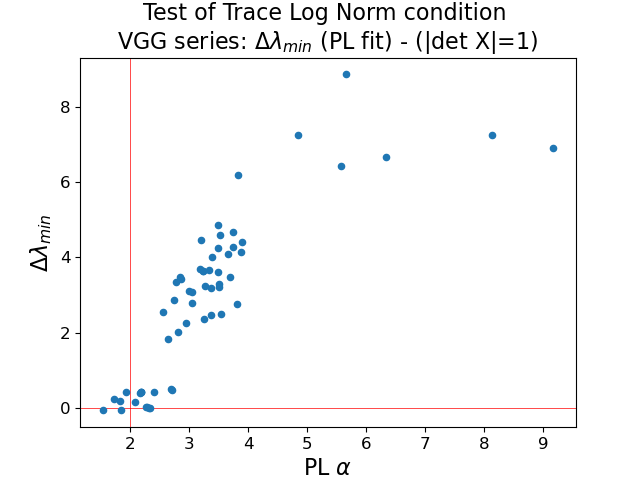}
      \label{fig:VGG_trend}
    }
    \subfigure[ResNet Series]{
      \includegraphics[width=7cm]{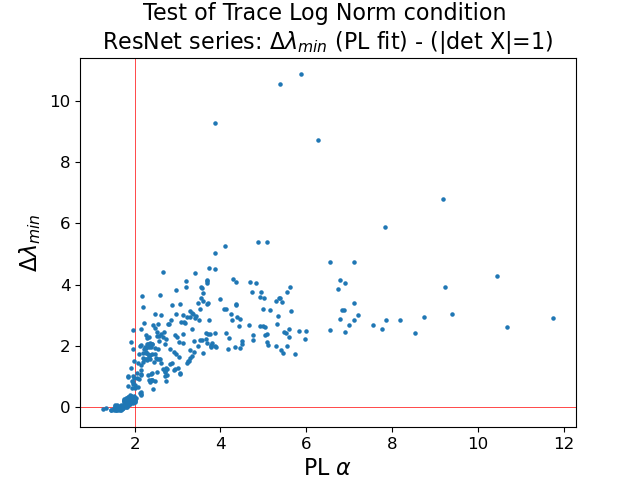}
      \label{fig:ResNet_trend}
    } \\
    \subfigure[ViT Series]{
      \includegraphics[width=7cm]{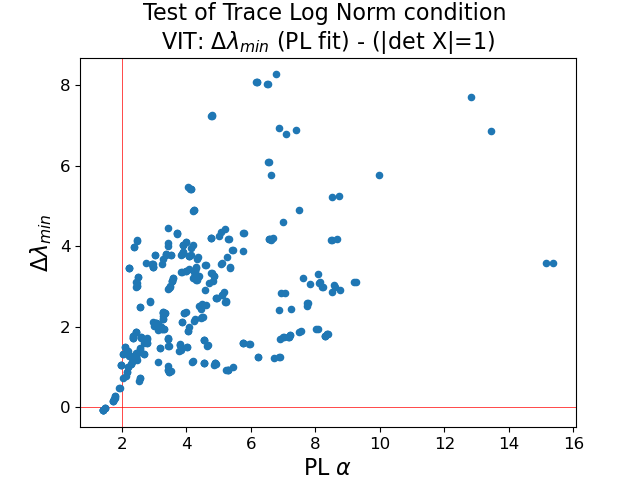}
      \label{fig:VIT_trend}
    }
    \subfigure[DenseNet Series]{
      \includegraphics[width=7cm]{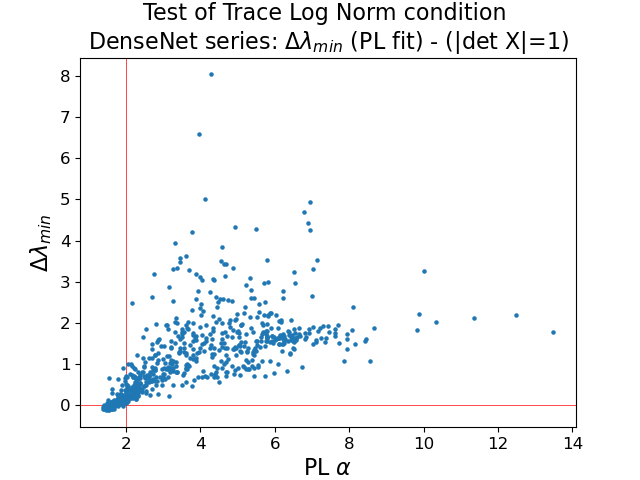}
      \label{fig:DenseNet_trend}
    }
    \caption{Difference between the two $\lambda_{min}$ estimates, $\Delta\lambda_{min}$, (\EQN~\ref{eqn:D_lambda_min}), as a 
        function or $\alpha$, for linear and convolutional layers in series of VGG~\cite{VGG14_TR}, ResNet~ 
        \cite{ResNet15_TR}, ViT models~\cite{VIT20_TR} and DenseNet models~\cite{DenseNet17_TR}. Layer matrices for all 
        models in the series were pooled to create each plot. In (a), (VGG,) we see three clusters of points -- those 
        with $\Delta \lambda_{min}$ close to $0$ and $\alpha$ close to $2$, those with $\Delta \lambda_{min}$ above $2$ 
        and $\alpha > 2.5$, and those with $\Delta \lambda_{min}$ above $6$ and $\alpha > 3.5$. In (b), (ResNet,) we see 
        that in general, as $\alpha$ shrinks towards $2$, $\Delta \lambda_{min}$ tends towards $0$, overshooting 
        slightly. We also see that the difference $\Delta\lambda_{min}$ is almost always positive, with few exceptions, 
        and even the layers that do not overshoot form a kind of ``funnel shape pointing towards the critical point 
        $(2, 0)$. In (c), (ViT,) we also see the same general relationship between $\alpha$ and $\Delta \lambda_{min}$ 
        across layers of several ViT models. Observe that ViT models do not have convolutional layers, and in spite of 
        this, the overall pattern is similar. In (d), (DenseNet,) we see a similar overall trend as in (b), except that 
        $\Delta\lambda_{min}$ tends to decrease sooner, but there are also more layers with $\alpha < 2$ and 
        $\Delta\lambda_{min}$ above $0$.
    }
  \label{fig:CV_ESD_trends}
\end{figure}

Figure~\ref{fig:CV_ESD_trends} plots $\alpha$ versus the difference $\Delta\lambda_{min}$, (\EQN~\ref{eqn:D_lambda_min}).
Layer matrices from all models in each series are pooled to generate the plots.%
\footnote{For convolutional layers, the \WW tool first computes eigenvalues for all channels-to-channels linear operators separately, and then pools them in order to compute $\alpha$, $\LAMBDAPL$ and $\LAMBDADETX$. For instance, in a $64\times 64\times 3\times 3$ weight tensor, there would be $9$ separate linear operators of $64\times 64$, giving $576$ eigenvalues, which would then be pooled to compute $\alpha$, $\LAMBDAPL$ and $\LAMBDADETX$.} 
Notice that in Figure~\ref{fig:VGG_trend} -- \ref{fig:DenseNet_trend}, $\Delta\lambda_{min}$ approaches zero as $\alpha\rightarrow 2$ from above. 
Individual points may have a large $\Delta\lambda_{min}$ for an $\alpha$ near to $2$, but the overall trend is apparent. 
The rapid decrease of $\Delta\lambda_{min}$ as $\alpha$ approaches $2$ from above implies that the PL tail rapidly 
takes on a unit \TRACELOG  (if it doesn't already have it.)

As we saw in comparison of Figures~\ref{fig:mlp3-msr-results-xmin} and~\ref{fig:mlp3-msr-results-detX}, 
(Section~\ref{sxn:trunc_err_epochs},) the \TRACELOG tail is generally larger, and always has highly generalizing 
components. Thus, it is plausible that as layers reach the limit of the amount of information that can be encoded in 
them, i.e. as $\alpha$ goes to $2$, the \POWERLAW~tail expands to fill the \TRACELOG tail. This effect can be seen 
clearly in Figure~\ref{fig:CV_ESD_trends} in the condensing of the ``funnel shape.

Recall from Figure~\ref{fig:mlp3-detx-gap} that layer FC1 dominated the model, (Table~\ref{tab:mlp3},) producing a clear 
progression of $\alpha$ and $\Delta\lambda_{min}$ towards $(2, 0)$, as a function of learning rate or batch size, whereas FC2 showed a slightly less clear relationship. 
In larger models having dozens or hundreds of layers, we would not expect any one layer to dominate as thoroughly.
Moreover, it is the architecture that varies between models in each series, not (necessarily) the hyperparameters, meaning that there would not be a straight line, as in Figures~\ref{fig:mlp3-detx-lr_search_fc1} and~\ref{fig:mlp3-detx-bs_search_fc1}. 
However, with all of the layers contributing to varying degrees, we nevertheless see a clear trend in all plots of Figure~\ref{fig:CV_ESD_trends}. 
These results show how the single-layer \SETOL theory extends from the MLP3 model, which is dominated by a single layer, to larger models where many layers interact in complex ways, but still reflect the same overall trend.

\begin{figure}[t] 
    \centering
    \subfigure[Falcon 7B]{
      \includegraphics[width=7cm]{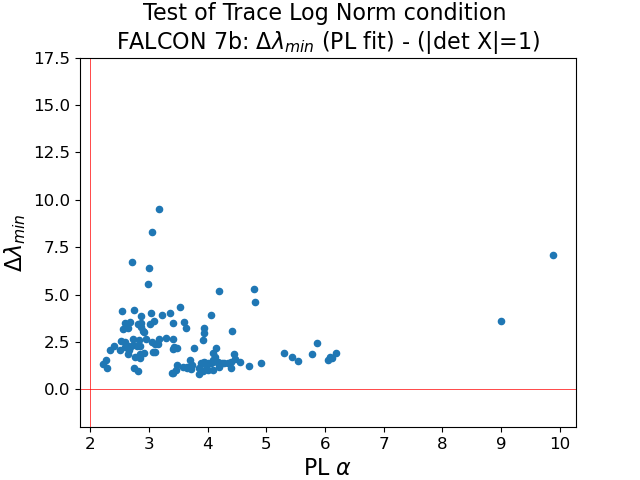}
      \label{fig:falcon7B_trend}
    }
    \subfigure[Falcon 40B]{
      \includegraphics[width=7cm]{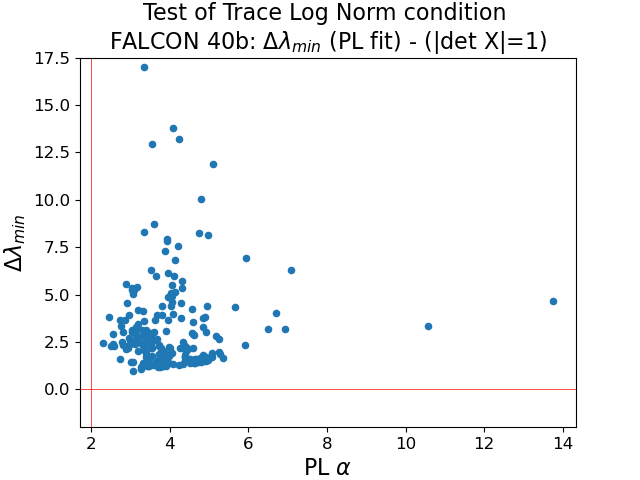}
      \label{fig:falcon40B_trend}
    }
    \subfigure[Llama 13B]{
      \includegraphics[width=7cm]{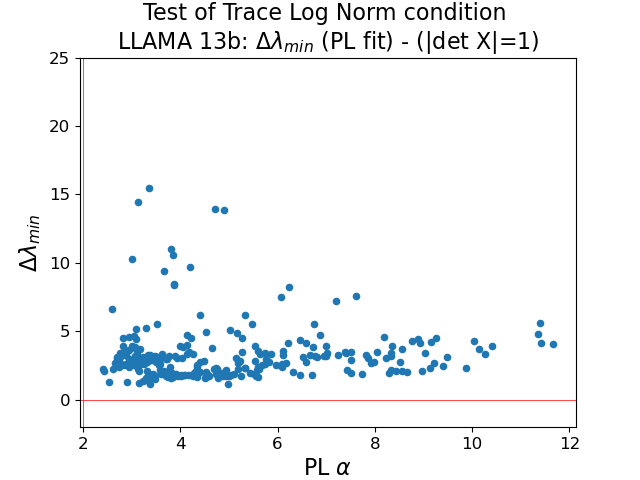}
      \label{fig:llama13B_trend}
    }
    \subfigure[Llama 65B]{
      \includegraphics[width=7cm]{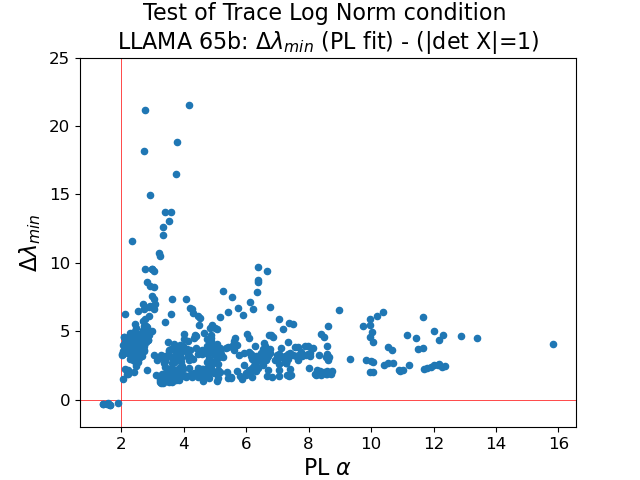}
      \label{fig:llama65B_trend}
    }
    \caption{Difference between the two $\lambda_{min}$ estimates, $\Delta\lambda_{min} = \LAMBDAPL - \LAMBDADETX$, 
        as a function of $\alpha$, for all linear layers in the FALCON\cite{falcon40b}(a-b) and 
        LLAMA \cite{touvron2023_TR}(c-d) language models for varying numbers of parameters. As in Figure 
        \ref{fig:CV_ESD_trends}, we see that in recent Large Language Models, $\Delta\lambda_{min}$ remains positive, 
        except where $\alpha < 2$ (d). Otherwise, a ``funnel shape can still be seen leading towards the critical 
        point $(2, 0)$ as in Figures~\ref{fig:ResNet_trend} and~\ref{fig:VIT_trend} Observe that the x- and y-axes are 
        different between sub-figures due to the differences in scale of each model.
    }
  \label{fig:LLM_ESD_trends}
\end{figure}

The overall pattern of relationship between $\Delta\lambda_{min}$ and $\alpha$ can also be seen in 
Figure~\ref{fig:LLM_ESD_trends}, which shows plots for Large Language Models (LLMs) of the Falcon~\cite{falcon40b} and 
LLAMA~\cite{touvron2023_TR} model families, for different numbers of parameters. Observe that each 
subfigure~\ref{fig:falcon7B_trend}--\ref{fig:llama65B_trend} shows a single 
model, rather than a collection of models in a family, as in Figure~\ref{fig:CV_ESD_trends}.
The y-axis is the same between models in the same family. 
As in Figure~\ref{fig:CV_ESD_trends}, there is a general outline of a ``funnel shape pointing towards the critical 
point $(2, 0)$, with the exception that it is only reached in the case of LLAMA-65b, (Figure~\ref{fig:llama65B_trend}). 
This suggests that these LLMs are larger than they necessarily need to be, consistent with prior work~\cite{YHTx21_TR}, 
but also that they are well guarded against \OverRegularized layers beyond the critical point
($\alpha=2$ and $\Delta\lambda_{min}=0$).

\subsection{Layer Qualities with Computational \RTransforms}
\label{sxn:empirical_comp_r_transforms}

\begin{figure}[ht]
    \centering
    \subfigure[FC1 batch size vs $\QT$]{
        \includegraphics[width=6cm]{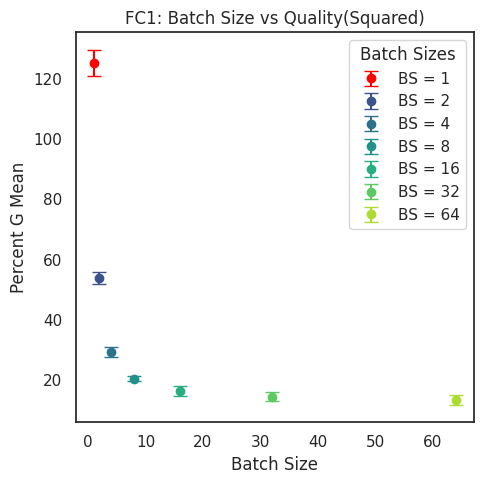}
        \label{fig:FC1_bs_vs_Q}
    }
    \hspace{1cm} 
    \subfigure[FC1 Layer $\alpha$ vs $\QT$]{
        \includegraphics[width=6cm]{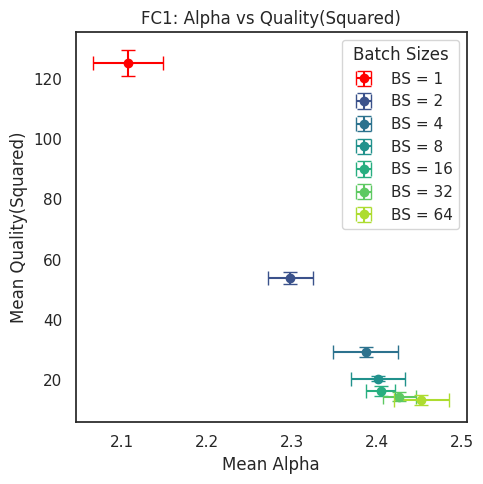}
        \label{fig:FC1_alpha_vs_Q}
    }
    \hspace{1cm} 
    \subfigure[FC2 batch size vs $\QT$]{
        \includegraphics[width=6cm]{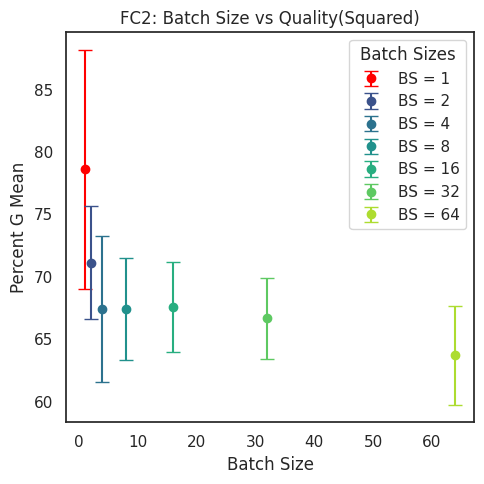}
        \label{fig:FC2_bs_vs_Q}
    }
    \hspace{1cm} 
    \subfigure[FC2 Layer $\alpha$ vs $\QT$]{
        \includegraphics[width=6cm]{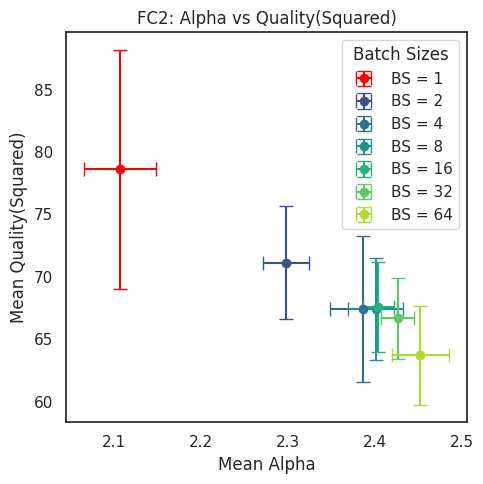}
        \label{fig:FC2_alpha_vs_Q}
    }
    \caption{
      Evaluation of the computational \RTransform \LayerQualitySquared metric $\QT$
      on the fully trained MLP3 model(s) for different batch sizes.
    }
    \label{fig:MLP3_qualities}
\end{figure}

In this Section, we look at how the \HTSR layer quality HT PL metric $\alpha$
compares to computing the \LayerQualitySquared $\QT$ using
the Computational \RTransform method proposed in Section~\ref{sxn:comp_rmt}
for fully trained MLP3 model(s).
Figure~\ref{fig:MLP3_qualities}
presents results for FC1 and FC2 layers, comparing
both the batch size and the mean $\alpha$ metric to the
mean $\QT$, and the results are quite different.

For FC1, as shown in Figure~\ref{fig:FC1_bs_vs_Q}, the quality metric \(\QT\) increases as the batch size decreases, following the expected trend. Notably, for batch size = 1, the quality metric exceeds $100\%$,
which suggests that the layer may be overfit in this scenario, which is similar to results obtained earlier. Furthermore, in Figure~\ref{fig:FC1_alpha_vs_Q}, the quality metric \(\QT\) shows a strong correlation with the $\alpha$ metric, consistent with the theoretical predictions of the \HTSR framework.
Earlier results suggest that the FC1 layer captures most of the correlation in the data,
so it is interesting that this layer also has much smaller error bars, indicating more consistent quality across different batch sizes.

In contrast, for FC2, while the general trends are similar (as shown in Figures~\ref{fig:FC2_bs_vs_Q} and~\ref{fig:FC2_alpha_vs_Q}), the error bars on the quality metric $\QT$ are much larger. This indicates significantly higher variability in computational quality compared to FC1. The large error bars for FC2 suggest that, while the $\QT$ metric is theoretically grounded, it is less effective than the existing \HTSR layer quality metric $\alpha$, which exhibits greater stability and reliability in capturing layer behavior. These differences highlight the distinct computational characteristics of the two layers, with FC1 demonstrating more predictable trends that align closely with theoretical expectations.

\subsection{Inducing a Correlation Trap}
\label{sxn:empirical-correlation_trap}

\begin{figure}[t]
    \centering
    \subfigure[MLP3 \LearningRate $16\times$]{
      \includegraphics[width=6cm]{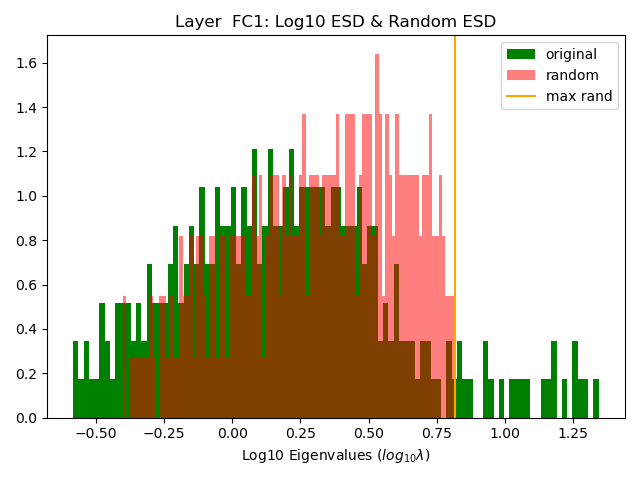}
        \label{fig:mlp3-accuracies3_lr16_trap}
    }
    \subfigure[MLP3 \LearningRate $32\times$]{
      \includegraphics[width=6cm]{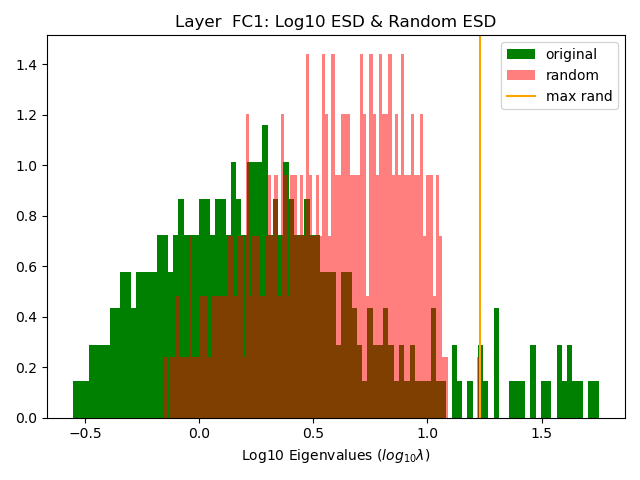}
        \label{fig:mlp3-accuracies3_lr32_trap}
    }
    \caption{ESD plots for learning rate $lr=16\times$ and $lr=32\times$ normal, shown on Log-Lin scale, as computed 
    using the \WW tool, for the FC1 weight matrix $\mathbf{W}$ (green) and an element-wise randomized $\mbox{rand}(\mathbf{W})$ (red).  
    This provides an example of inducing a \CorrelationTrap in the MLP3 model, simply by increasing the learning rate used during model training.  
        See Section~\ref{sxn:Traps}.
            }
    \label{fig:mlp3-accuracies3}
\end{figure}

Previous work on the \HTSR \Phenomenology \cite{MM20a_trends_NatComm,MM21a_simpsons_TR} has shown that one can look for 
quantitative deviations from the necessary pre-conditions of tradtitional \RMT (particularly that the weights are 
$0$-mean and finite variance) to detect when a model layer suffers from some other anomaly in the elements, which we 
call a ``\CorrelationTrap (see Section~\ref{sxn:Traps} and ~\cite{MM20a_trends_NatComm,MM21a_simpsons_TR}). A 
\CorrelationTrap may cause, or be caused by, the over-regularization leading \ALPHA to fall below $2$, (see 
Section~\ref{sxn:underfitting}). Here, we explore this in greater detail, in light of our \SETOL.

%
Lets look at the ESDs of the FC1 layer of the MLP3 model, for learning rates $lr=16\times$ and $lr=32\times$ normal. 
We will be interested in the general shape of the ESD of $\tfrac{1}{N}\mathbf{W}^T\mathbf{W}$.
For the purposes of detecting a \CorrelationTrap, we will randomize $\mathbf{W}^T\mathbf{W}$ element-wise, and then observe its largest eigenvalue $\lambda^{max}_{rand}$. 

Figure~\ref{fig:mlp3-accuracies3} shows the ESDs of the original matrix (green) and the element-wise 
randomized $\mbox{rand}(\tfrac{1}{N}\mathbf{W}^T\mathbf{W})$ (red). 
Observe in particular $\lambda^{max}_{rand}$ for each learning rate factor.
For $lr=16\times$, Figure~\ref{fig:mlp3-accuracies3_lr16_trap} shows 
that the ESD of $\tfrac{1}{N}\mathbf{W}^T\mathbf{W}$ is HT, whereas the ESD of the randomized matrix is essentially a distorted semi-circle---as expected from the well-known MP result; and 
that $\lambda^{max}_{rand}$ lies at the edge of the random \MPBulk ESD.
(A similar result is seen for smaller learning rates.)  
In contrast, for $lr=32\times$, Figure~\ref{fig:mlp3-accuracies3_lr32_trap} shows 
that while the original ESD is again HT, the ESD of the randomized has one large element, $\lambda^{max}_{rand}$, that pulls out from the MP bulk.
This is the signature of a \CorrelationTrap; and it co-occurrs with the exact learning rate setting that degraded the train and test accuracies, pushing \ALPHA below its optimal value of $\alpha\simeq 2$.
When this happens,  both the estimation of \ALPHA and the formation of a PL tail are potentially disrupted. 
%

\CorrelationTraps have been observed previously~\cite{MM20a_trends_NatComm,MM21a_simpsons_TR}, using the \HTSR \Phenomenology.
However, \SETOL provides an explanation for why this would be expected to occur --- non-standard element-wise distributions will tend to interfere with the properties of the spectrum which \SETOL analyzes. 
Our derivation in Section~\ref{sxn:empirical} suggests that in order to apply the \SETOL effectively one must avoid (or remove) such traps. 
This too has been observed previously~\cite{MM20a_trends_NatComm,MM21a_simpsons_TR}.


\subsection{Overloading and the Hysteresis Effect}
\label{sxn:hysteresis_effect}
Obtaining a value of $\alpha$ outside the $\alpha \gtrsim 2$ range is indicative of ``overloading~\cite{SST92} that layer. 
This can be accomplished, e.g., by training only one layer in the MLP3 model.
As the two layers have very different sizes, we see markedly different behaviors, that are nevertheless consistent with theory.
In particular, this lets us test the theory in both the \emph{Strongly Over-Parameterized Regime} $(\ND \gg N\times M)$ 
and the \emph{Under-Parameterized Regime} $(\ND \ll N \times M)$ 

The \SETOL theory is based on the idea that NNs undergoing training behave like Statistical Mechanic systems relaxing to an equilibrium. 
So far, we have tested the theory under conditions that are approaching \Ideal. 
However, for the theory to be useful in practice, we must also examine how it performs in non-\Ideal situations. 
Of particular interest, we would like to examine the theory under conditions where the training dynamics slows down, i.e., when it is in a ``glassy''
or meta-stable state. 
One way we can do this is to train only one layer, and freeze the rest. 
Doing so overloads the single trainable weight matrix, as a function of the ratio of examples to trainable
parameters~\cite{SST92,Gardner_1988,MM17_TR}, and we expect this to cause $\alpha_{FC1}$ or $\alpha_{FC2}$ to drop well below $2$. As in \ref{sxn:empirical-effective_corr_space}, we will be examining this effect over the entire course of training.

\subsubsection{Baseline: Loading onto both FC1 and FC2}
\label{sxn:hysteresis_baseline}
\begin{figure}[t] 
    \centering
        \includegraphics[width=15cm]{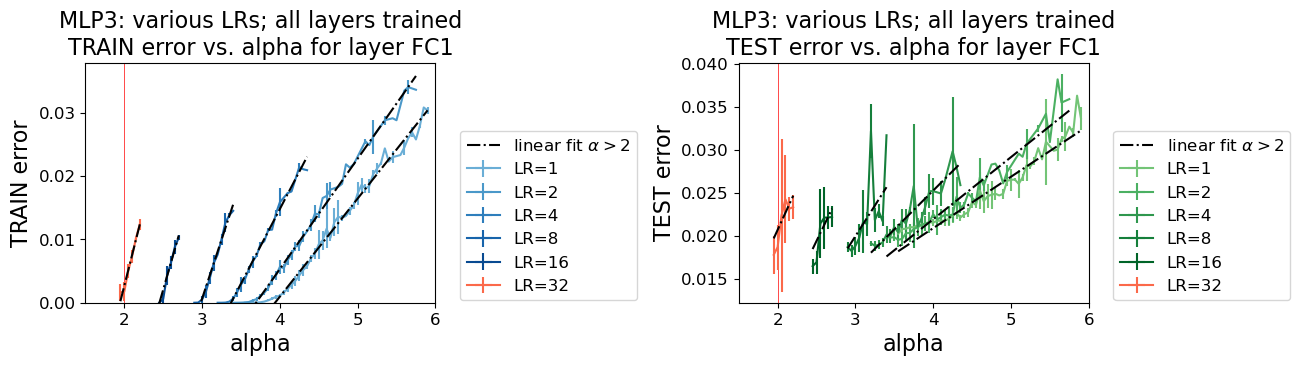} \\
    \begin{tabular}{ccc}
      (a)\hspace{5cm} & (b) 
    \end{tabular}
        \includegraphics[width=15cm]{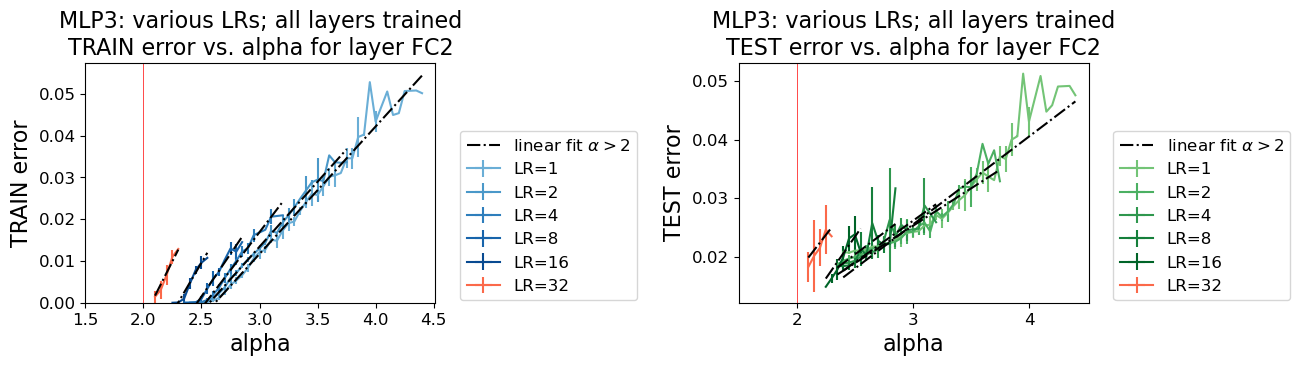}
    \begin{tabular}{ccc}
      (c)\hspace{5cm} & (d) \\
    \end{tabular}
    \caption{
        Train (a, c) and test (b, d) accuracy as a function of $\alpha_{FC1}$ (a, b) and $\alpha_{FC2}$ (c, d) when {\bf all 
        layers are trained}. Red vertical lines show the critical value of $\alpha = 2$, and dashed black lines show 
        linear fits of error, using only points where $\alpha > 2$ and train error $> 0.001$. 
        For FC1 (a, b), we can see that each learning rate 
        produces a different trajectory of train error (a) and test error (b) as a function of $\alpha_{FC1}$, 
        showing that even though FC1 dominates overall, (Table~\ref{tab:mlp3},) FC2 still plays a modulating role. 
        (Cf. Figure~\ref{fig:mlp3-FC1-alpha-overloaded} where there is only one trajectory.)
        In (c, d) we can see that $\alpha_{FC2}$ never goes below $2$. As in (a, b), each learning rate 
        produces a different trajectory, though there is greater overlap for lower learning rates. See discussion in 
        Section~\ref{sxn:hysteresis_baseline}.
    }
    \label{fig:mlp3-baseline-load}
\end{figure}

To start, Figure~\ref{fig:mlp3-baseline-load} shows $\alpha_{FC1}$ and $\alpha_{FC2}$, binned in units of $0.05$, versus 
train and test error over all epochs of training
\footnote{Excluding the first four epochs when the matrix is still essentially random.} 
for each of the different learning rates considered. 
(Cf. Figures~\ref{fig:mlp3-alphas-lr} and~\ref{fig:mlp3-alphas-bs}, Section~\ref{sxn:empirical-effective_corr_space},
which show only the final epoch.)
Binning was done so as to facilitate averaging over the $5$ starting random seeds; 
linear fits are shown separately for each learning rate; and
error bars represent one standard deviation within each bin. 
The critical value of $\alpha=2$ is shown as a vertical red line in all plots. 
For each learning rate, train error and $\alpha$ decrease together during training, which can be seen for both 
$\alpha_{FC1}$ (a) and $\alpha_{FC2}$ (c), reading each line from the top right to bottom left. Test errors, (b, d) show 
a similar trend, but with wider error bars. 
Observe that the range of the y-axis is narrower for test error to make detail more visible.
\chris{CH TODO: Double check these plots. The y-axis should be the same vertically.}

We see in Figure~\ref{fig:mlp3-baseline-load} (a,b) 
that the $32\times$ learning rate causes $\alpha_{FC1}$ to decrease faster than any other, putting it on course to 
fall below $2$ before train error reaches $\simeq 0$. 
(See Figure~\ref{fig:mlp3-accuracies}, at the beginning of this Section.) 
We also see that the slower learning rates cause train error to reach $\simeq 0$ well before $\alpha_{FC1}$ can reach $2$, and explains as to why their test error was higher.


\subsubsection{Overloading FC1: {Strongly Over-Parameterized Regime} \texorpdfstring{$(\ND\gg N\times M$)}{n>>N x M}}
\label{sxn:hysteresis_effect_FC1}

\begin{figure}[t] 
    \centering
        \includegraphics[width=15cm]{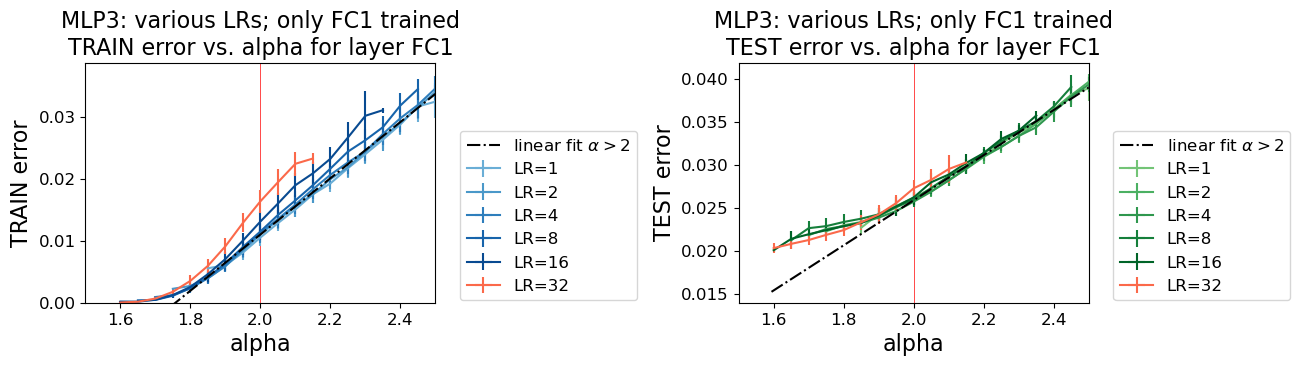}
    \begin{tabular}{ccc}
      (a)\hspace{5cm} & (b) \\
    \end{tabular}
    \caption{
            Train and test accuracy as a function of $\alpha_{FC1}$ when only FC1 is trained (a, b). 
            Red vertical lines show the critical value of $\alpha_{FC1} = 2$, and dashed black lines show 
            linear fits of error, using only points where $\alpha_{FC1} > 2$. 
            In contrast with Figure~\ref{fig:mlp3-baseline-load}, when only 
            FC1 is trained, we can see that no matter the learning rate, there is ony one trajectory, for both train 
            error (a) and test error (b). Hence, for visibility, only one linear fit, using all learning rates pooled, 
            is shown. Crucially, we can see in (b) that as $\alpha_{FC1}$ passes below $2$, the \emph{test error} 
            trajectory changes, for all learning rates, even as the \emph{train error} trajectory does not, until it 
            reaches $\sim 0$. This suggests that even though test accuracy can still decrease when $\alpha_{FC1} < 2$, it 
            does so at a decreased rate relative to $\alpha_{FC1}$. See discussion in 
            Section~\ref{sxn:hysteresis_effect_FC1}.
    }
    \label{fig:mlp3-FC1-alpha-overloaded}
\end{figure}

In contrast, when only FC1 is trained, Figure~\ref{fig:mlp3-FC1-alpha-overloaded} shows that, as expected, 
$\alpha_{FC1}$ decreases well below $2$. Training error (a) generally trends downward as $\alpha_{FC1}$ decreases, 
but no matter the learning rate, the relation is the same, or nearly so, because only one layer is being trained. 
Consequently, all learning rates were pooled to produce one linear fit for enhanced visibility.

In Figure~\ref{fig:mlp3-FC1-alpha-overloaded} we can see 
demonstration of a crucial claim of \SETOL theory: that for $\alpha_{FC1} > 2$, (vertical red line,) the test 
error declines linearly; (b) shows that test error is almost perfectly linear with decline in $\alpha_{FC1}$. However, when $\alpha_{FC1} < 2$, the curve bends 
upward. 
Furthermore, the precision with which the trajectory changes as $\alpha_{FC1}$ passes the threshold provides ample 
validation that the estimator of~\cite{CSN09_powerlaw} is indeed accurate. We 
observe that test error may continue to decrease after $\alpha_{FC1} < 2$, however the rate of decrease is significantly less. 
We can also see that in some sense, the model is ``doomed'' to always have train error reach $\simeq 0$ when 
$\alpha_{FC1} \simeq 1.7$, i.e., after $\alpha = 2$, because of the number of trainable parameters, and perhaps because 
of the lack of a modulating influence of FC2 seen in Figure~\ref{fig:mlp3-baseline-load}.

\chris{This might be a good place to cite Temperature Balancing, Layer-wise Weight Analysis, and Neural Network Training because this corroborates the idea that learning rates work differently on different layers.}


\subsubsection{Overloading FC2: {Under-Parameterized Regime} \texorpdfstring{$(\ND \ll N\times M$)}{n << NxM}}
\label{sxn:hysteresis_effect_FC2}

\begin{figure}[t] 
    \centering
        \includegraphics[width=15cm]{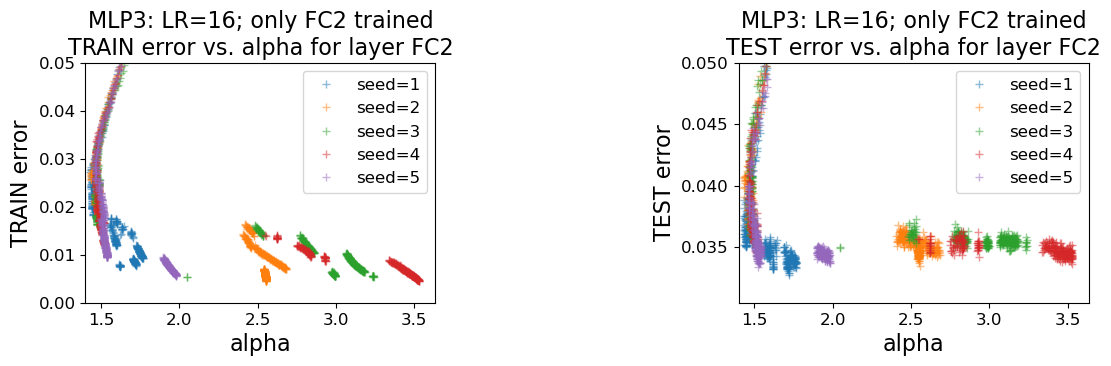}
    \begin{tabular}{ccc}
      (a)\hspace{5cm} & (b) \\
    \end{tabular}
        \includegraphics[width=15cm]{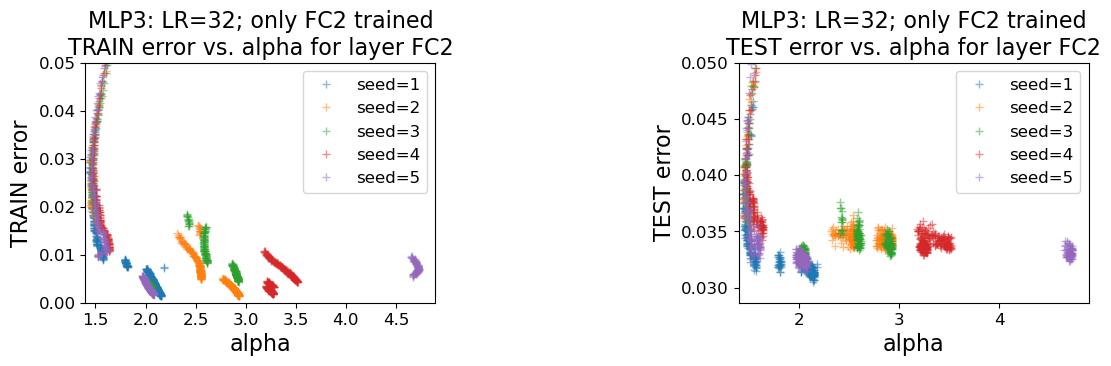}
    \begin{tabular}{ccc}
      (c)\hspace{5cm} & (d) \\
    \end{tabular}
        \caption{Train error (a, c) and test error (b, d) as a function of $\alpha_{FC2}$ when all other layers are 
        frozen, for the two largest learning rates, $lr=16\times$, (a, b) and $lr=32\times$(c, d). Cf. 
        Figure~\ref{fig:mlp3-baseline-load}, wherein all layers were trained. Due to the markedly different 
        behavior of each random seed, they cannot be plotted as means and error bars, and are instead shown separately. 
        The path taken by all seeds up to $\alpha_{FC2} = 1.5$ has a slight curvature characteristic of a 
        hysteresis-like behavior. Observe also that the fragmenting into separate paths, due to the breakdown of the 
        PL tail, coincides roughly with each seeds path reaching its minimum test error. The y-axis is scaled 
        differently for train and test error to make variation more visible. See discussion in 
        Section~\ref{sxn:hysteresis_effect_FC2}.
    }
    \label{fig:mlp3-FC2-alpha-hysteresis}
\end{figure}

The MNIST dataset has $60,000$ training examples, which means that an FC1-only model is over-parameterized, but FC2 is 
substantially \emph{under}-parameterized~\cite{DLM19_Exact_TR}. (Table~\ref{tab:mlp3}) This drastically changes the 
meaning of an experiment where only FC2 is trained. Figure~\ref{fig:mlp3-baseline-load} (Section 
\ref{sxn:hysteresis_baseline}) shows $\alpha_{FC2}$ vs. train error (c) and test error (d) when \emph{all} layers are 
trained. There we see a different relationship between $\alpha_{FC2}$ and train and test error for each learning rate. 
None of the training runs reached an $\alpha_{FC2}$ of $2$, as the load was split between both layers FC1 and FC2.

Figure~\ref{fig:mlp3-FC2-alpha-hysteresis}, however, shows a starkly different relationship. 
When only FC2 is trained, 
each random seed produces a different trajectory, meaning that they cannot be interpretably plotted as a single curve, even with error 
bars. Train (a, c) and test (b, d) error rates are shown as a function of $\alpha_{FC2}$ for the two highest learning 
rate factors, $16\times$ (a, b) and $32\times$ (c, d). Lower learning rate factors, (not shown,) showed the same trend, except 
that they did not progress as far as those shown in Figure~\ref{fig:mlp3-FC2-alpha-hysteresis}.

First we see, for all seeds, $\alpha_{FC2}$ decreases all the way down to $1.5$, after which it begins to rebound to the 
right. As it 
does, train error continues to decrease. We can also see that test error continues to decrease as well, down to its minimum 
value of slightly more than $0.03$, as $\alpha_{FC2}$ continues to increase for a short time. However, at the exact 
point where test error reaches a minimum, the PL tail itself begins to fracture, leading to different estimates 
of $\alpha_{FC2}$ for each seed. As each of the five starting random seeds are shown separately, we can see that each of 
them terminates at a different point, some of which are closer to $2$ than others.

Such a reversion to a more \Typical value of $\alpha_{FC2}$, prior to the fracturing of the PL tail, is 
reminiscent of a spin glass system relaxing towards its minimum energy configuration, in a way that retains some memory 
of the path taken on the way to its current state. 
\chris{The claim of hysteresis is based on the bending seen in the curve (left, top to bottom,) just before fracturing 
        happens. We can also see that though the curves fracture, they all continue bending strongly to the right.
        The theory predicts that alpha will decrease, but it doesnt predict that it will rebound as we see there. The 
        only mechanism we know of that would predict such bending, seen most clearly in (a, c), is hysteresis.
}
We conjecture that if the model were trained for sufficiently many 
epochs, (perhaps many thousands,) then the tail would re-form, and $\alpha_{FC2}$ would reform, and revert all the way back to its 
stable value of $2$.

\newpage
\section{Conclusion and Future Directions}
\label{sxn:conclusions}

In this work, we have introduced \SETOL, a \emph{Semi-Empirical Theory of (Deep) Learning}
that unifies concepts from \StatisticalMechanics (\STATMECH), \HeavyTailed (HT) Random Matrix Theory (\RMT), and quantum-chemistry-inspired approaches to strongly correlated systems~\cite{Martin1996,Martin1996_CPL,Martin1998}. \SETOL aims to provide a solid theoretical foundation for the Heavy-Tailed Self-Regularization (\HTSR) phenomenology, including the widely used \ALPHA and \ALPHAHAT HT Power Law (PL) \LayerQuality metrics, which are implemented in the open-source \WW toolkit. Specifically, \SETOL reformulates the Neural Network (NN) learning problem for a single layer as a matrix generalization of the \StudentTeacher (ST) model for \Perceptron learning, where the Teacher is now taken as 
empirical input, and the result is in \AnnealedApproximation (AA) at high temperatures (high-T).
This reformulation results in a Free Energy (i.e., Generating Function $\IZG$)  expressed as an integral over random \Student correlation matrices, commonly referred to as the Harish-Chandra–Itzykson–Zuber (HCIZ) integral. To evaluate this integral, we recast the solution using a technique derived from first principles, analogous to a taking single step of the \WilsonExactRenormalizationGroup (\ERG) theory~\cite{NobelPrizeRG,PhysRevLett.69.800}.
Leveraging more recent results~\cite{Tanaka2007,Tanaka2008}, we express the \LayerQuality as a sum of integrated matrix-cumulants from \RMT (i.e., \RTransforms). Finally, we conduct both direct and observational experiments to validate key assumptions of the \SETOL framework and empirically connect it to the \HTSR theory.

\vspace{1em}
\noindent
\textbf{Key Contributions and Observations.}
\begin{enumerate}[label=\Alph*.]
\item 
\emph{Rigor for the \HTSR Phenomenology.}
\SETOL explains \emph{why} power-law (PL) exponents in the layer spectral densities (e.g.\ $\ALPHA$ and 
\ALPHAHAT) act as robust diagnostics of generalization, even in large, complex architectures without access 
to training or test data. Our analysis ties these \HeavyTailed ESDs to a \ScaleInvariant  \VolumePreservingTransformation
associated with an \emph{Effective} \FreeEnergy, and
suggesting (in analogy with traditional \STATMECH phases in learning theory)
that the \HTSR condition $\alpha\approx 2$ marks a
phase boundary between optimal generalization and overfitting.

\item 
\emph{Matrix-Generalized \StudentTeacher (ST) Model.}
\SETOL is formulated as \SemiEmpirical matrix generalization of
the classical (vector-based) ST perceptron learning,  taking the Teacher now as empirical input, and incorporating $N\times M$ layer 
weight matrices, $\WVEC\rightarrow\WMAT$.
Key to this generalization is isolating the top eigenvalue/eigenvector directions---called the
\emph{Effective Correlation Space} (\ECS)---before evaluating the resulting partition function (or HCIZ 
integral). The ECS contains the \HTSR PL tail, validating that the tail captures the dominant layer generalizing components.

\item 
\emph{\TRACELOG Condition \& $\ALPHA=2$.}
A remarkable empirical observation, predicted by \SETOL, is that layers near \emph{ideal} training also satisfy a 
$\prod \LambdaECS_{i} \approx 1$ in their tail eigenvalues (i.e., $|detX|=1$); equivalently, $\sum \ln\LambdaECS_{i} = \ln\prod \LambdaECS_{i} \approx 0$.
We call this the \TRACELOG condition (i.e, the $detX$ option in \WW).
Empirically, this condition appears when the \HTSR $\ALPHA \approx 2$. \SETOL thereby \emph{unifies} two 
previously separate heuristics for “optimal” or so-called \Ideal  behavior.

\item 
  \emph{Empirical Validation of \SETOL.}
To validate the \TRACELOG and the \ECS conditions, we trained small (3-layer) \MultiLayerPerceptron (MLP3)
on MNIST and under varying batch sizes and learning rates.  Using this, we verified that
when the \HTSR $\alpha\approx 2$ the \SETOL \TRACELOG condition also (usually) holds,
and that one can reproduce the training accuracy by retaining only the \ECS.
This is further confirmed by using the \WW tool to examine the \ALPHA and \DETX metrics
common, open-source CV and NLP  models, including modern LLMs 
(ResNets, DenseNets,    ViTs, and LLMs like LLaMA and Falcon).

\item
  \emph{Correlation Traps \& OverFitting.}
  We  observe that when layer ESDs with $\alpha<2$, they exhibit behavior interpreted as
  \emph{over-regularization} and/or  \emph{overfitting}.
  For example, we observe what we call \emph{\CorrelationTraps}, large rank-one perturbations in the
  (randomized)  layer weight matrix $\mathbf{W}$ that can be induced by training with
  excessively small batch sizes (bs=$1$) and/or large learning rates, and are associated with degraded test accuracy,
  and which cause the \HTSR to drop to $\alpha<2$.
  These results are consistent with other results, applying \HTSR theory to understand Grokking\cite{prakash2025grokking}.
  
  Additionally, we can induce overfitting by freezing all but one layer and
  then training, which causes the layer $\alpha<2$.  By training in
  the underparameterized regime, we observe path-dependent, “glassy” behavior.

\item
\emph{OverParameterized and UnderParameterized regimes.}
  These above results indicate that the \HTSR and \SETOL approaches can be applied effectively in the overparameterized regime, even in conditions beyond the apparent range of validity of the theory ($\alpha\gg 2)$.  In the underparameterized regime, they are less effective.  It is noted that modern transformers like LLMs may appear to be underparameterized, but the multiplicative nature of their interactions suggests they are overparameterized.~\cite{hay2024}, making \HTSR and \SETOL very applicable for such AI models.
  
\item 
\emph{Connection to Semi-Empirical Methods.}
Conceptually, \SETOL parallels well-known \emph{\SemiEmpirical} methods in quantum
chemistry~\cite{Martin1996, Martin1996_CPL,Martin1998}, wherein complicated many-body Hamiltonians are approximated by 
effective theories, but fitted or validated using empirical data. By retaining only the largest spectral modes 
(\ECS) and imposing the \TRACELOG condition, we can describe crucial low-rank correlations while discarding less 
relevant interactions, much like in Freed--Martin Effective Hamiltonian  and/or Wilson’s  \ExactRenormalizationGroup 
(ERG) approach~\cite{Freed1983,Martin1994,Martin1996,Martin1996_CPL,  MartinFreed1996, Martin1998, PhysRevLett.69.800}.
\end{enumerate}

\vspace{1em}
\noindent
\textbf{ERG Analogy: A One-Step View.}
From a \RenormalizationGroup perspective, restricting to the measure on the partition function to
the \ECS is akin to performing a \emph{single step} of the Wilson Exact Renormalization Group (ERG).
In doing this, we are discarding bulk “uninteresting” degrees of freedom in favor of the strongly correlated 
HT \emph{long-ranged} modes. This leads to an effective model with fewer degrees of freedom but 
\emph{renormalized} interactions--interactions that are dominated by the largest eigenvalues.
This analogy with   \ERG theory suggests that the \HTSR phenomenology, where $\alpha\in[2,6]$ in the Fat-Tailed
\Universality Class, is essentially describing a near-critical phase  when
$\alpha \approx 2$ and satisfies $\Trace{\ln \LambdaECS}=\ln\prod \LambdaECS_i\approx 0$ in its ECS.
Departing from  this point ($\alpha<2$) leads to suboptimal results, consistent with the multi-phase 
pictures in \STATMECH spin glass theories of learning~\cite{SST92,Gardner_1988,Eng01, EB01_BOOK}.

\vspace{1em}
\noindent
\textbf{Toward Understanding “Why Deep Learning Works.”}
A key question in deep learning theory is why large neural networks achieve strong generalization despite operating in highly non-convex optimization landscapes. From the perspective of \ERG theory, this phenomenon can be partially understood through the concentration of generalization-relevant components. Specifically, models trained in regimes exhibiting \emph{Power-Law (PL)} correlations lead to the emergence of effective low-dimensional descriptions, where irrelevant modes are suppressed. The \WW \ALPHA metric quantitatively captures this concentration by characterizing how the optimization landscape stabilizes near criticality, with $\alpha \approx 2$ signaling optimal generalization.

\vspace{1em}
\noindent
\textbf{Student-Teacher Knowledge Distillation.}
\SETOL also provides an explanation of why Student-Teacher distillation works so well.
When training a Student DNN $S$ to emulate a larger Teacher $T$, the Student inherits the inductive biases and generalization behavior of Teacher, which \SETOL explains is concentrated in $T$ into smaller (lower rank) layer weight matrices.
Thus, the Student converges to a similar region of the optimization landscape, shaped by the same spectral and statistical constraints that govern the Teacher, allowing
for a smaller model to effectively mimic, if not reproduce, the Teacher's outputs.  

\vspace{1em}
\noindent
\textbf{A Novel Way to Characterize Generalization.}
Many results that use the AA / High-Temperature limit effectively state that ``with enough data and a wide enough NN, one can learn anything''. But how much is ``enough''? \SETOL takes this one step further by giving useful empirical measures that converge to the same Ideal, whose progress can be precisely tracked in finite settings. Our experimental results show that this occurs both in a simple model, as well as in large production models. These measures strongly indicate the point at which further learning in a particular layer will be counterproductive.

\vspace{1em}
\noindent
\textbf{Relation to Levy Spin Glasses and Heavy-Tailed Random Matrix Models.}
One longstanding puzzle is why large NNs avoid the worst of highly non-convex optimization, despite nominal 
exponential degeneracies. \SETOL offers a partial explanation: if the trained model has \emph{Levy-like} or VHT
PL correlations, then typical spin-glass degeneracies can be lifted, leaving the layer in a finite number 
of near-critical minima. In analogy with older work on \emph{Levy Spin Glasses}\cite{Bouchaud1998}, it is
proposed \WW \ALPHA metric effectively measures how \emph{rugged yet stable} the
effective energy landscape is for each layer, with $\alpha\approx 2$ signifying a sweet spot of \Ideal generalization.

\subsection{Future Directions}

\paragraph{1.\quad Multi-Layer   \ERG and Layer Interactions.}
While \SETOL is formulated per-layer, modern DNNs stack many layers, each potentially with different $\alpha$. 
An improved approach would treat layer-layer interactions (relax the \IFA). It is possible 
that certain layers (e.g.\ final fully-connected heads in LLMs) exhibit $\alpha$ far from 2, while others 
converge near 2---raising questions about how best to address or combine them.

\paragraph{2..\quad Practical Diagnostics and Fine-Tuning.}
The open-source \WW tool has already seen success diagnosing layer quality. Integrating \SETOL’s 
\TRACELOG condition may refine this further, helping users identify correlation traps or “under-exploited” 
layers. There is also strong potential for using \ALPHA and \TRACELOG-based signals during training or 
fine-tuning in the training of very large LLMs: e.g.\ automatically adapting learning rates to push each layer closer to $\alpha=2$\cite{YTHx22_TR},
for fine-tuning models with significantly less memory\cite{Qing2024AlphaLoRA},
for compressing large LLMs\cite{alphapruning_NEURIPS2024}, and other practical applications.

\paragraph{3.\quad Correlation Traps and Meta-Stable States.}
Although our experiments show how small batch sizes or large learning rates can induce correlation traps, 
a quantitative theory of \emph{where and why} traps occur remains open. Clarifying these states could 
enable \emph{trap-avoidance} strategies, e.g.\ partial re-initialization or specialized regularizers that 
favor lower-rank updates in the ECS. In large-language-model (LLM) contexts, correlation traps might 
manifest as \emph{hallucinations} or \emph{mode collapse}, motivating deeper analysis.

\paragraph{4.\quad Analyzing the Layer Null Space.}
One critical but often overlooked factor is the potential null space within model layers, which can emerge during overfitting. This null space represents parameter directions that fail to contribute meaningfully to generalization but instead encode redundant or overly specific patterns tied to the training data which might be ignored or forgotten.
Future work should examine if and when NN layers have components in their null space,
which contribute significantly to the performance of the model.

\paragraph{5.\quad Layer-Layer Cross-Terms.}
\SETOL is a single layer theory, however, as noted in Section~\ref{sxn:matgen_mlp3}, it would 
be desirable to extend the theory to including layer-layer cross terms.
While we don't have an exact expression for this, we can propose a phenomenological guess
that the leading order term would be the integrated \RTransform,
defined for the overlap between nearest-neighbor weight matrices
(i.e., $\WMAT_{1},\WMAT_{2}$) that can be algined along a common axis.
This term would take the form $\mathbb{G}_{\OVERLAP_{1,2}}(\lambda_{1,2})$
where $\OVERLAP_{1,2}\sim\WMAT_{1}^{\top}\WMAT_{2}$ and $\lambda_{1,2}$ is an
eigenvalue of $\OVERLAP_{1,2}$.  Note that the open-source~\WW tool can
identify and compute the intra-layer interactions.\cite{WW}


\vspace{1em}
\noindent
\textbf{Concluding Remarks.}
\SETOL as a \SemiEmpirical theory merges first-principles methods from \STATMECH and \RMT with empirical 
insights from \HTSR and the open-source \WW tool. It clarifies \emph{how} \HeavyTailed  layer weight 
matrices can emerge from training on realistic data, and \emph{why} their spectral exponents so reliably 
predict generalization quality without peeking at training/test sets. In so doing, \SETOL not only offers 
new insights into the “\emph{why does it work}” question of deep learning but also suggests a roadmap for 
improving DNN models by focusing attention on that near-critical subspace of their largest 
eigenvalues. We are optimistic that future developments along these lines---extending the single-step Wilson
\ExactRenormalizationGroup analogy, understanding layer-layer interactions expansions, and systematically diagnosing correlation 
traps, and developing better lagre-scale training methods---will yield more robust, data-free metrics for training, fine-tuning, and compressing next-generation 
neural networks.

\noindent
\paragraph{Acknowledgments.}
We would like to thank Matt Lee of Triaxiom Capital and Carl Page of the Anthropocene Institute.
We also thank Mirco Milletarì and Michael Mahoney for helpful conversations.


\bibliographystyle{unsrt}
{\small

}


\appendix
\break
%

\section{Appendix}
\label{sxn:appendix}


\subsection{Data Vectors, Weight Matrices, and Other Symbols}
\label{sxn:appendix_A}

See Table~\ref{tab:dimensions} for a summary of various vectors and matrices, including their dimensions;
see Table~\ref{tab:symbols} for a summary of various various symbols used throughout the text; and 
see Table~\ref{tab:energies} for a summary of types of ``Energies'' used throughout the text.


\renewcommand{\arraystretch}{1.2} 

\begin{center}
\begin{table}[ht]
  \begin{tabular}{| l | c | r |}
    \hline
    Number of NN Layers & index $L$ & $N_{L}$ \\ \hline
    Number of Data Examples & index $\mu$ & $\ND$ \\ \hline
    Number of (input) Features & index $i,j$ & $m$ \\ \hline
    Actual Data (Matrix) & $D$ & $n \times m$ \\ \hline
    Model Data (Matrix) & $\mathcal{D}$ & $n \times m$ \\ \hline    
    \Teacher \Perceptron Weight Vector & $\TVEC$ & $m$ \\ \hline    
    \Student \Perceptron Weight Vector & $\SVEC$ & $m$ \\ \hline        
    Actual Input Data Vector & $\DATA$ & $N_{f}\times 1$ \\ \hline
    Gaussian model of Input Data Vector & $\boldsymbol{{\xi}}_{\mu}$ & $N_{f}\times 1$ \\ \hline
    Actual Input Data Label & $\MY_{\mu}$ & $+1|-1$ \\ \hline
    Model Input Data Label & $y_{\mu}$ & $+1|-1$ \\ \hline      
    General Weight Matrix & $\mathbf{W}$ & $N\times M$ \\ \hline
    General \CorrelationMatrix & $\mathbf{X}=\frac{1}{N}\mathbf{W}^{\top}\mathbf{W}$ & $M\times M$ \\ \hline
    Input Layer Weight Matrix & $\mathbf{W}_{1}$ & $N \times M$ \\ \hline
    Hidden Layer Weight Matrix & $\mathbf{W}_{2}$ & $N\times M$ \\ \hline
    Output Layer Weight Matrix & $\mathbf{W}_{3}$ & $M\times 2$ \\ \hline
    \Teacher Weight Matrix & $\mathbf{T}$ & $N\times M$ \\ \hline
    \Student Weight Matrix & $\mathbf{S}$ & $N\times M$ \\ \hline
    \StudentTeacher Overlap Matrix & $\OVERLAP=\tfrac{1}{N}\mathbf{S}^T\mathbf{T}$ & $M\times M$ \\ \hline              
    Inner \Student \CorrelationMatrix & $\mathbf{A}_M=\tfrac{1}{N}\mathbf{S}^{\top}\mathbf{S}$ & $M\times M$  \\ \hline
    Outer \Student \CorrelationMatrix & $\mathbf{A}_N=\tfrac{1}{N}\mathbf{S}\mathbf{S}^{\top}$ & $N\times N$  \\ \hline
   Inner ~\ECS \Student \CorrelationMatrix & $\AECSM$ & $M\times M$  \\ \hline
   Outer ~\ECS \Student \CorrelationMatrix &$\AECSN$ & $N \times N$  \\ \hline
   ~\ECS \Teacher \CorrelationMatrix & $\XECS$ & $M\times M$  \\ \hline
    \hline
  \end{tabular}
  \caption{Summary of of various vectors and matrices, including their dimensions.}
\label{tab:dimensions}
\end{table}
\end{center}


\renewcommand{\arraystretch}{1.35} 

\begin{center}
\begin{table}[ht]
  \begin{tabular}{| l | c |}
    \hline
    \Perceptron \StudentTeacher (ST) Overlap & $R=\SVEC^{\top}\TVEC=\sum_{i}s_{i}t_{i}$ \\ \hline
    \StudentTeacher (ST) Overlap Operator& $\OVERLAP=\tfrac{1}{N}\SMAT^{\top}\TMAT$ \\ \hline       
    Matrix Generalized ST Overlap & $\tfrac{1}{N^2}\OLAPSQD$  \\ \hline
    \StudentTeacher \SelfOverlap & $\eta(\NDXI)=\mathbf{y}^{\top}_{T}\mathbf{y}_{S}$  \\ \hline
    $\ell_2$-Energy or $\ell_2$-Error & $\DETSTLL$ \\ \hline
    $\ell_2$-Energy Operator Form & $\DETOPSTLL:=\sum_{\XI}\DETSTLL$ \\ \hline
    $\ell_2$-Energy Matrix Operator Form & $\DETOPST:=N(\IM-\tfrac{1}{N}\SMAT^{\top}\TMAT)$\\ \hline
    \EffectivePotential  & $\epsilon(R)=\epsilon(S,T)=\langle\DETOPSTLL\rangle_{\AVGNDXI}$ \\ \hline
    \LinearPerceptron  $\epsilon(R)$ at high-T, large-$N$ & $\epsilon(R)=1-R$ \\ \hline
    \AnnealedApproximation (AA)& $\langle\ln Z\rangle_{\AVGNDXI}\approx\ln\langle Z\rangle_{\AVGNDXI}$ \\ \hline
    \AnnealedHamiltonian &  $\GAN$ \\ \hline
    \AnnealedHamiltonian at high-T &  $\GANHT=\EPSL(\WVEC)$ \\ \hline  
    Average \StudentTeacher \GeneralizationError & $\AVGGE^{ST}=\THRMAVG{\epsilon(R)}$ \\ \hline
    Average \StudentTeacher \GeneralizationAccuracy & $1-\AVGGE^{ST}=\THRMAVG{\eta(R)}$ \\ \hline
    Matrix Layer \QualitySquared & $\QT=\THRMAVGIZ{\OLAPTOLAP}$ \\ \hline
    Equivalent Notation for Averages & $\langle\cdots\rangle_{A}=\int\cdots d\mu(\mathbf{A})=\mathbb{E}_{\mathbf{A}}[\cdots]$\\ \hline
    Projection Operator onto ~\ECS & $\mathbf{P}^{ecs} := \sum |\LambdaECS_{i}\rangle\langle\LambdaECS_{i}|,\;i=1\cdots\MECS$ \\ \hline
    Average over~\ECS \Student Correlation Matrices & $ \langle \cdots \rangle^{\beta}_{\AECS}=\int \cdots d\mu(\AECS)=  \mathbb{E}_{\AECS}[\cdots]$ \\    \hline

    \TRACELOG Trace-Log-Determinant Relation & $\Trace{\ln\mathbf{A}}=\ln\det\mathbf{A}$ \\ \hline
    Effective Correlation Measure Transform & $d\mu(\mathbf{S})\rightarrow d\mu(\AECS)$ \\ \hline
    HCIZ Integral (Tanaka's Notation)&${\mathbb{E}_{\AECS}}[\exp(\ND\beta\Trace{\TMAT^{\top}\AMAT\TMAT)}$\\ \hline
      HCIZ Integral (\LargeN in $N$ explicit) & 
$\Expected[\AECS]{\exp\!\bigl(\ND\beta N\,\mathrm{Tr}\bigl[\tfrac{1}{N}\,\TMAT^{\top}\,\AECSN\,\TMAT\bigr]\bigr)}$
\\ \hline
    \LayerQualitySquared \GeneratingFunction & $\IZGINF := \ND\beta \sum_{\mu=1}^{\MECS}\int^{\LambdaECS_{\mu}}_{\LambdaECS_{\min}} dz R(z)$
 \\ \hline
    \GEN & $G_{A}(\gamma)=\int_{\LambdaECSmin}^{\LambdaECS}R_{A}(z)dz$ \\ \hline
    Eigenvalue for $\mathbf{X}=\tfrac{1}{N}\mathbf{W}^{\top}\mathbf{W}$ & $\lambda$ or $\lambda_{i}$ for $i=1\cdots M$ \\ \hline
    \PowerLaw ESD Tail for $\mathbf{X}$ & $\rho_{tail}(\lambda)\sim\lambda^{-\alpha}$ \\ \hline
    \EffectiveCorrelationSpace ESD Tail for $\mathbf{X}$ & $\rho^{ECS}_{tail}(\LambdaECS),\;\Trace{\ln\prod_{j=1}^{\MECS}\LambdaECS_{j}}=0$ \\ \hline
    Schatten Norm & $\Vert\mathbf{X}\Vert^{\alpha}_{\alpha}=\sum_{j}\lambda_{j}^{\alpha}$ \\ \hline
    ECS Tail Norm & $\tfrac{1}{\MECS}\sum_{i}^{\MECS}\LambdaECS_{i},\;\;\LambdaECS_{i}\in\rho^{ECS}_{tail}(\lambda)$\\ \hline
    Spectral Norm & $\Vert\mathbf{X}\Vert_{\infty}=\lambda_{max}$ \\ \hline
    \WW Start of PL Tail & $\lambda^{PL}_{min}=\lambda_{min}$ \\ \hline
    Start of~\ECS Tail & $\lambda^{ECS}_{min}=\lambda^{|detX|=1}_{min}$ \\ \hline
   ~\ECS-PL Gap between start of tails  & $\Delta\lambda_{min}:=\lambda^{ECS}_{min}-\lambda^{|detX|=1}_{min}$ \\ \hline            
    \WW~\ALPHA (layer) quality metric & $\alpha$ \\ \hline
    \WW~\ALPHAHAT (layer) quality metric & $\hat{\alpha}=\alpha\log_{10}(\lambda_{max})$ \\ \hline
  \end{tabular}
  \caption{Summary of various various symbols used throughout the text. 
          }
\label{tab:symbols}
\end{table}
\end{center}


\newcommand{\hthinline}{\rule{0pt}{2.5ex}} 


\begin{table}[ht]
  \hspace*{-2cm}
\begin{tabular}{|p{10cm}|p{6.2cm}|p{2.25cm}|}
\hline
\textbf{Explanation} & \textbf{Examples} & \textbf{Refs} \\
\hline
\textbf{\EnergyLandscape or NN Output function} & $\NNOUT$ & Sec.~\ref{sxn:htsr_setup}\\
\hthinline
The output of the NN given a single input data point. & &\ref{eqn:dnn_energy},\ref{eqn:T_ENN},\ref{eqn:S_ENN},\ref{eqn:nflow} \\
\hline
\textbf{Energy or \StudentTeacher (ST) Error} & $\DEL$, $\DELBF$ & Sec.~\ref{sxn:SMOG_main},~\ref{sxn:summary_sst92} \\
\hthinline
The squared between the output of a \Student NN and its prescribed Teacher label $\Yt$ for a single data point &$\DEL(\XImu):=(\Yt-\SOUT(\XImu))^2 $ & \ref{eqn:DEy} \\
And as the total error for a sample of $\ND$ data points  &  $\DELBF=\sum_{\mu=1}^{n} \DEL(\XImu)$ & \ref{eqn:detopxy}  \\
Or between the outputs of the \Student and the \Teacher NNs. & $
\DELBF:=\sum_{\mu=1}^{n} (\SOUT(\XImu) - \TOUT(\XImu))^2$ & \ref{eqn:DE_L},\ref{eqn:DE},\ref{eqn:DETOPNN} \\
\hline

\textbf{\AnnealedHamiltonian (and Potentials)} & $\HAN$ $(\mathcal{E}(R), \EPSL(R))$ & Sec.~\ref{sxn:mathP_annealed},\ref{sxn:summary_sst92} \\
\hthinline
The \EffectivePotential (for the Error) is defined as:  &
$\EPSL(R)=\tfrac{1}{\ND}\mathcal{E}(R)=\langle \DELBF\rangle_{\AVGNDXI}$ & ~\ref{eqn:epslR}\\
The \AnnealedHamiltonian (for the Error):  &
$\beta\HAN:=\tfrac{1}{\ND}\ln\langle e^{-\beta\DELBF}\rangle_{\NDXI}$ & ~\ref{eqn:Gan_def},~\ref{eqn:Gan0} \\
At high-T, the relation between $\HANHT$ and $\EPSL(R)$ is &
$\HANHT(R):=\EPSLR$ & ~\ref{eqn:Gan_highT} \\
\hthinline
The full ST model Hamiltonian: &
$\beta\HAN(\beta,R):= \tfrac{1}{2}\ln[{1+2\beta(1-R)}]$ & ~\ref{eqn:Gan2} \\
At high-T, the ST model Hamiltonian is: &
$\beta\HANHT(R):=\beta(1-R)$ & ~\ref{eqn:epslR},~\ref{eqn:Gan3} \\
\hthinline
The per-parameter ST model Hamiltonian: &
$\beta\HANPP:=\tfrac{1}{2}\ln[1 + \tfrac{2\beta N}{M}\Trace{\IM-\mathbf{R}}]$  & ~\ref{eqn:Gan_lnI_final_mwm} \\
At high-T, the matrix-generalized ST model Hamiltonian is: &
$\GANMATHT:=N(\IM-\OVERLAP)$&
~\ref{eqn:GANHTmatR} \\
The \LayerQualitySquared Hamiltonian (for the Accuracy): &
$\HBARE:=\OLAPTOLAP$ & ~\ref{eqn:RG},~\ref{eqn:HBARE} \\
\hline
\textbf{Different Average Model Errors} & $\AVGE$ & Sec.~\ref{sxn:mathP_errors}\\
Empirical Training, \Teacher, and Test Errors  & $\AVGEMPTE$, $\AVGE^{T}\approx\AVGEMPGE$ & \ref{eqn:Eg_train},\ref{eqn:Eg_test},\ref{eqn:emp_gen_error}\\
\hthinline
Empirical \GeneralizationGap & $\AVGE^{emp}_{gap}:=\AVGEMPTE-\AVGEMPGE$ & \ref{eqn:gen_gap} \\
\hthinline
\StudentTeacher Training and Generalization Errors  & $\AVGSTTE$, $\AVGSTGE$ & \ref{eqn:EtM2},\ref{eqn:EgCanonical}\\
\hthinline
Neural Network (MLP) Training and Generalization Errors, the (abstract) matrix generalization of ST error
& $\AVGNNTE$, $\AVGNNGE$ & \\
\hline
\textbf{Average Training and/or \GeneralizationError} & $\AVGTE, \AVGGE$ & Sec.~\ref{sxn:mathP_errors}\\ 
In the AA and at High-T, these are the same, 
and are just the \ThermalAverage of $\EPSL(R)$ & $\newline \AVGTE^{an,hT}=\AVGGE^{an,hT}=\THRMAVG{\EPSL(R)}$ &
\ref{eqn:avgte_anhT},\ref{eqn:avgge_anhT}\\
For the ST model, we always assume AA and High-T & $\AVGSTGE=\AVGGE^{an,hT}$ & \\
Likewise, when generalizing $\AVGSTGE$ to matrices,  & $\AVGSTGE\rightarrow\AVGNNGE=\AVGGE^{an,hT}$ & \\
\hline
\textbf{Layer Qualities} & $\Q$, $\QT$ & Sec.~\ref{sxn:matgen_quality_hciz_A},~\ref{sxn:quality} \\
\hthinline
For the ST \Perceptron, $\Q^{ST}$ is the generalization accuracy & $\Q^{ST}:= 1-\AVGSTGE = 1-\THRMAVG{\EPSL(R)}$ & \\
in terms of the \SelfOverlap $\ETA(R)$ & $\Q^{ST}:= \THRMAVG{\ETA(R)}$ & \\
In the AA, and at high-T, $\THRMAVG{\EPSL(R)}=1-R$ & $\Q^{ST}:= 1-\AVGGE^{an,ht} = \THRMAVG{R}$ & \ref{eqn:QST_final} \\
For an MLP / NN, we approximate the total accuracy as a product of layer qualities $\Q$ (in the AA, at high-T) &  $\Q^{NN}:=\prod \Q^{NN}_{L}$ &  \ref{eqn:ProductNorm}\\
For a matrix, the \LayerQualitySquared $\QT$ & $\QT:=\THRMAVGIZ{\OLAPTOLAP}$ &\ref{eqn:QT_1}\\
We approximate $\Q$ using the quality squared & $\Q:=\sqrt{\QT}\approx Q^{NN}_{L}$ & \ref{eqn:QT},\ref{eqn:QT_2}\\
\hline
\end{tabular}
  \caption{Summary of types of ``Energies,'' with simplified examples of the notation, and references to definitions.
          }
\label{tab:energies}
\end{table}

\clearpage
\renewcommand{\arraystretch}{1.0} 

\subsection{Summary of the \StatisticalMechanicsOfGeneralization (\SMOG)}
\label{sxn:summary_sst92}

In this section, we derive the Annealed Hamiltonian for two variants of the ST model, in the high-T limit:
in Appendix~\ref{app:st-gen-err-annealed-ham}, we derive an expression for $\GANR$ for the ST \Perceptron model, when the students and teachers are modeled as $N$-vectors $\WVEC$ (as in~\cite{SST92}); and
in Appendix~\ref{sxn:appendix_Gan}, we derive an expression for $\GANMAT$ for Matrix-Generalized case, i.e., when the students and teachers are modeled as $N \times M$ matrices $\WMAT$ (as our \SETOL requires).
From these, we will obtain expressions for the Average ST \ModelGeneralizationError $\AVGSTGE$ and the Average NN \ModelGeneralizationError $\AVGNNGE$, as well as for the corresponding data-averaged errors.
Although the functional form for these quantities will be the same for the vector case and the matrix case, there are several important differences in the derivation of $\GANMAT$, most notably having to do with a normalization for the weight matrix.

\subsubsection{Annealed Hamiltonian \texorpdfstring{$\GANR$}{H(R)} when Student and Teachers are Vectors}
\label{app:st-gen-err-annealed-ham}

In this section, we derive an expression for the Annealed Hamiltonian $\GANR$, 
in the AA and the high-T approximation,
when student and teachers are are modeled as $N$-vectors.
From this, we obtain an expression for the data-averaged ST error $\EPSL(R)$, 
which is the same as the expression given in \EQN~\ref{eqn:epslR}.

The procedure starts by computing the associated quenched average of
the \FreeEnergy, defined for the model error as
\begin{align}
\nonumber
\langle-\beta F \rangle_{\AVGNDXI} 
   :=& \langle \ln Z \rangle_{\AVGNDXI} \\ 
\nonumber
   =&\left\langle \ln \int d\mu(\SVEC)e^{-\beta\DETOPSTLL} \right\rangle_{\AVGNDXI}  \\ 
\label{eqn:qaF}
   =& \frac{1}{N}\int d\mu(\NDXI)\ln\int d\mu(\SVEC)e^{-\beta\DETOPSTLL}   ,
\end{align}
where $d\mu(\NDXI):=\prod_{i=1}^{N}d\XI_{i} P(\NDXI)$ and
where the data-dependent ST error, $\DETOPSTLL$, is defined in \EQN~\ref{eqn:DE_L},
with an $\mathcal{L}=\ell_2$ loss.  \footnote{Also, recall that the Teacher $T$ is fixed and is not learned, so we do not integrate over $d\mu(\TVEC)$. In fact, for the (vector) Perceptron model, the Teacher $T$ weights are simply subsumed into the overlap $R$, and even in the more general cases, such as non-Linear/Boolean Perceptron, in the full Replica calculations, etc.  See the original literature for more details. \cite{Opper01,SST92,EngelAndVanDenBroeck} as well as \cite{MM17_TR}.}
\charles{Do we need a $1/N$ to ensure the free energy is extensive ? I don't think we do because at the end we want to evaluate the
  generalization accuracy, so we need to take the partial w/r.t. $N$. But we need to check this and section 4.2 carefully.
Of course, with the HCIZ integral, we have to be also be careful since we move the $1/N$ from the L.H.S. to the R.H.S of Tanaka just for this reason}
If we apply the AA (see \EQN~\ref{eqn:AA} and \EQN~\ref{eqn:Jensens}) to \EQN~\ref{eqn:qaF}, then we obtain
\begin{align}
\label{eqn:qaFan}
\langle-\beta F \rangle_{\AVGNDXI} \simeq 
   \ln \frac{1}{\ND}\int d\mu(\NDXI) \int d\mu(\SVEC) e^{-\beta\DETOPSTLL}   .
\end{align}
Notice that we have interchanged the logarithm $(\ln)$ and
and the data average (the “disorder average”) 
$\langle\cdots\rangle_{\AVGNDXI}$ over the data; 
this is the essence of the AA, as it lets the disorder fluctuate rather than forcing the system to be quenched to the data.
We will now switch the order of integration in \EQN~\ref{eqn:qaFan}, giving
\begin{align}
\label{eqn:qaFan3}
\langle-\beta F \rangle_{\AVGNDXI} \simeq 
   \ln \int d\mu(\SVEC)\frac{1}{\ND}\int d\mu(\NDXI)e^{-\beta\DETOPSTLL)}   .
\end{align}

We now recall the definition of the \AnnealedHamiltonian, $\GANR$ (see \EQN~\ref{eqn:Gan_def} in Section~\ref{sxn:mathP}, which is analogous to \EQN~(2.31) of \cite{SST92}):
\begin{align}
\label{eqn:Gan0}
\beta\GANR = \beta\HAN(\beta,\SVEC,\TVEC):=-\frac{1}{\ND}\ln \int d\mu(\NDXI)e^{-\beta\DETOPSTLL}  .
\end{align}
where we have denoted the Hamiltonian as $\HAN(\beta,\SVEC,\TVEC)$ to indicate the explicit dependence on $\beta$,
and we have added $\beta$ to the R.H.S. to because the L.H.S. is unitless.
Using this definition, we can express the \Annealed \PartitionFunction, $Z^{an}_{\ND}$, in the AA in terms of the \Annealed Hamiltonian $\GANR$
(as in \EQN~\ref{eqn:Zan_def} in Section~\ref{sxn:mathP}, and as in \EQN~(2.31) of \cite{SST92}:
\begin{align}
 \label{eqn:Zan}
Z^{an}_{\ND}:=\ND\langle Z\rangle_{\AVGNDXI}=\int d\mu(\SVEC)e^{-\ND\beta\HAN(\beta,\SVEC.\TVEC)}  .
\end{align}

Following Section~\ref{sxn:mathP_errors}, we can write the \emph{Average Model \TrainingError} $\AVGSTTE$, in the AA,
in terms of \Annealed \PartitionFunction, $Z^{an}_{\ND}$:
\begin{align}
 \label{eqn:AVGSTTE_AA}
\AVGSTTE:= -\frac{1}{\ND}\dfrac{\partial }{\partial \beta} \ln Z^{an}_{\ND}
\end{align}

This now lets us write the Average Model \TrainingError $\AVGSTTE$ in the AA in terms of the \AnnealedHamiltonian $\GANR$:
  \begin{align}
  \nonumber
  \AVGSTTE
   =& \dfrac{1}{Z^{an}_{\ND}}\int d\mu(\SVEC)\dfrac{\partial \beta\HAN}{\partial \beta} e^{-\ND\beta\HAN(\beta,\SVEC,\TVEC)} \\ 
  \label{eqn:EtM2}
   =& \dfrac{1}{Z^{an}_{\ND}}\int d\mu(\SVEC)\langle  \DELBFell(\SVEC,\TVEC,\XI) \rangle_{\AVGNDXI} e^{-\ND\beta\HAN(\beta,\SVEC,\TVEC)}  ,
  \end{align}
  where $\DELBFell(\SVEC,\TVEC,\XI) \rangle^{\beta}_{\AVGNDXI}$
  is a \ThermalAverage but defined over the specific ST error in the AA for the chosen set of training data $\XItrain=\NDXI$, and is denoted by 
 \begin{align}
   \langle  \DELBFell(\SVEC,\TVEC,\XI) \rangle^{\beta}_{\AVGNDXI}:=\dfrac{\partial \beta\HAN}{\partial \beta}.
 \end{align}
 This is analogous to defining the average error $\DETOPXI$ as a \ThermalAverage, but as one over the data $\NDXIn$ instead of the weights.  This can be seen by expanding \EQN~\ref{eqn:Gan_def}, setting $\WVEC=\SVEC$, fixing $\TVEC$ (implicitly), and taking the partial derivative:
 \begin{align}
  \label{eqn:partial_Gan}
  \dfrac{\partial \beta\HAN}{\partial \beta}
  &=  \dfrac{\partial }{\partial \beta} \left(-\frac{1}{\ND}\ln \int d\mu(\NDXIn)e^{-\beta\DETOPXILL}\right) \\ \nonumber
  &=  -\frac{1}{\ND}\dfrac{\partial }{\partial \beta} \ln \int d\mu(\NDXIn)e^{-\beta\DETOPXILL} \\ \nonumber
  &=  -\frac{1}{\ND} \left( \int d\mu(\NDXIn)e^{-\beta\DETOPXILL}\right)^{-1}\dfrac{\partial }{\partial \beta} \int d\mu(\NDXIn)e^{-\beta\DETOPXILL} \\ \nonumber
    &=  -\frac{1}{\ND} \left(\int d\mu(\NDXIn)e^{-\beta\DETOPXILL}\right)^{-1} \int d\mu(\NDXIn)(-\DETOPXI)e^{-\beta\DETOPXILL}
 \end{align}

We can also write the \ModelGeneralizationError $\AVGSTGE$  as \BoltzmannWeightedAverage
of $\epsilon(R)$, weighted by $\HAN(\beta,S;T)$ (as in (2.32) in \cite{SST92}), as:
\begin{align}
\label{eqn:EgCanonical}
\AVGSTGE=\dfrac{1}{Z^{an}_{\ND}}\int d\mu(\SVEC)\epsilon(R)e^{-\ND\beta\HAN(\beta,\SVEC,\TVEC)} ,
\end{align}
where $\epsilon(R)=\epsilon(\SVEC)$ is the average ST error, for a fixed \Teacher T,
averaged over \emph{all} possible data inputs, i.e., not just over the specific training data.
(Note that we have dropped the subscript $train$ on $\XI$ since it is clear from the context.)



In the high-T (small $\beta$) limit, the two model errors become formally equivalent
(i.e., $\AVGSTTE=\AVGSTGE$ as $T\rightarrow\infty$).
To show this, consider the \AnnealedHamiltonian $\GANR$, for the \LinearPerceptron with the $\ell_2$ loss. 
As shown in \EQN~(C6) of~\cite{SST92}, this takes a simple analytic form--in the \LargeN limit in $\ND$--in terms of the ST overlap $R$:
\begin{align}
\label{eqn:Gan2}
\GANR = \dfrac{1}{2}\ln\left[{1+2\beta(1-R)}\right]  .
\end{align}
\EQN~\ref{eqn:Gan2} holds in the AA, but not in the High-T limit.

If we evaluate $\dfrac{\partial \HAN}{\partial \beta}$ in the High-T (small $\beta$) limit, 
then we can use the approximation $(\ln[1+x]\simeq x+\cdots)$ to obtain the High-T approximation:
\begin{align}
\label{eqn:Gan3}
\beta\GANR \simeq 
\beta\GANHTR:=\beta(1-R),\;\;\;\;\beta\;\text{small}  .
\end{align}
where we now see that $\GANHTR$ no longer explicitly depends on $\beta$.
By \EQN~\ref{eqn:Gan_highT}, this gives 
\begin{align}
\label{eqn:Gan4}
\EPSL(R) =
\langle  \mathbf{E}_{\ell_2}(\SVEC,\TVEC,\XI \rangle_{\AVGNDXI} \simeq 1-R\;\;\text{as}\;\ND\rightarrow\infty  ,
\end{align}
which we recognize as the same as the data-averaged ST error $\epsilon(R)$ in \EQN~\ref{eqn:epslR}.

\subsubsection{Annealed Hamiltonian \texorpdfstring{$\GANMAT$}{H(R)} for the Matrix-Generalized ST Error}
\label{sxn:appendix_Gan}

In this section, we derive an expression for our matrix generalization of the \AnnealedHamiltonian of the \LinearPerceptron,
in the AA and the high-T approximation, when student and teachers are are modeled as $N \times M$ matrices $\WMAT$,
i.e., $\GANR\rightarrow\GANMAT$, which has the same form as \EQN~\ref{eqn:Gan0} for the vector case.

From this, we obtain an expression for the data-averaged ST error $\EPSL(R)$, again when the student and teachers are are modeled as matrices.
There is a subtle normalization issue here, about which we need to be careful.
However, when we normalize appropriately, we will obtain an expression
for ``data-averaged ST error'' (i.e., \EffectivePotential) $\EPSL(R)$ that is of the same form as we obtained in the vector case (as given in \EQN~\ref{eqn:Gan4} and \EQN~\ref{eqn:epslR}).
The difference will be that in the vector case we take $R=\tfrac{1}{N}\SVEC^{\top}\TVEC$, while in the matrix case we take $R=\tfrac{1}{N}\SMAT^{\top}\TMAT$.

We will need to evaluate an average over the $\ND$ random $M$-dimensional training data vectors $\NDXI$,
which are i.i.d Gaussian with $0$ mean and $\sigma^{2}$ variance: 
\begin{equation}
  \label{eqn:XInorm}
  \Vert \NDXI \Vert^2 :=\sum_{\mu=1}^{\ND} \XI_\mu \XI_\mu^{\top} = \sigma^{2}\IM ,
\end{equation}
where each $\XI_{\mu}$ is a vector of length $M$, and $\IM$ is an $M \times M$ identity matrix.
The expected value of the squared norm is:
\begin{equation}
\mathbb{E}[\Vert \XI_{\mu} \Vert^2] = M \sigma^2 .
\end{equation}
If we let $\sigma^{2}\sim\tfrac{1}{M}$, 
then $\mathbb{E}[\Vert \XI_{\mu} \Vert^2]=1$, i.e., the data vectors can be normalized to $1$.
Let the probability distribution over the $N$ data vectors~be
\begin{align}
\nonumber
  P(\NDXI) &= \prod_{\mu=1}^{\ND} \left( \frac{1}{\sqrt{(2 \pi \sigma^2)^M}} \right) e^{-\frac{\|\XI_{\mu}\|^2}{2 \sigma^2}} \\ 
\nonumber
  &= \left( \frac{1}{\sqrt{(2 \pi \sigma^2)^M}} \right)^{\ND} \exp\left[-\frac{M}{2}\sum_{\mu=1}^{\ND}\|\XI_{\mu}\|^2\right] \\ 
  \label{eqn:pndx_vec}
  &= \XINORM \exp\left[-\frac{M}{2}\sum_{\mu=1}^{\ND}\|\XI_{\mu}\|^2\right] ,
\end{align}
where $M=N_f$ is the number of features in the data, where the normalization $\XINORM$ is
\begin{align}
\label{eqn:xinorm}
\XINORM 
:=\left( \frac{1}{\sqrt{(2 \pi \sigma^2)^M}} \right)^{\ND}
 =\left( \frac{M}{2\pi} \right)^{\ND M/2} .
\end{align}

\paragraph{The Total Data Sample Error $(\DETOPST)$ and the Matrix Normalization}
First, let us express the matrix-generalized \TotalDataSampleError, $\DETOPST$, for a single layer,
in operator form (for each  of the $\ND$ training examples)
\begin{align}
 \frac{1}{\ND}\DETOPST := N\Trace{\IM - \frac{1}{N}\SMAT^{\top}\TMAT} = NM - N\Trace{\OVERLAP} 
\end{align}

where $\IM$ is a diagonal matrix of dimension $M$.
Note that the matrices are by default data-averaged empirical quantities, so we can drop the $1/\ND$ on the RHS.

Also, notice that $\DETOPST$ scales as $N\times M$, the total number of parameters in the system.
Also,  importantly, when all the overlaps are perfect, then the error is zero, i.e. if $\Trace{\OVERLAP}=M$ then $\DETOPST=0$.
                                                                        
We can define the data-dependent form (i.e., in the basis of the data $\XI$) as
\begin{align}
\nonumber
\DETOPNN
   :=& N\sum_{\mu=1}^{\ND} (\XI^{\mu})^{\top} \left( \IM - \frac{1}{N} \SMAT^{\top}\TMAT \right) \XI^{\mu} \\
\label{eqn:DETOPNN}
    =&  N\sum_{\mu=1}^{\ND} \sum_{i,j=1}^{M} \XI_i^{\mu} \left( \delta_{ij} - \frac{1}{N} [\SMAT^{\top}\TMAT]_{ij} \right) \XI_j^{\mu}  .
\end{align}

\paragraph{The \AnnealedHamiltonian (per-parameter, $\HANPP)$}
The definition of the \AnnealedHamiltonian, $\GANMAT$, for the idealized case must be extended to account for the $N\times M$ parameters per training example.  We then have that the total energy is then the sum of the entries of $M$ (feature) vectors, as expected by 
\SizeExtensivity in $N$ and
\SizeConsistency in $M$:
\begin{align}
 \label{eqn:hanpp}
 \Trace{\GANMAT}=M\left(N\Trace{\HANPP}\right)
\end{align}
where the \AnnealedHamiltonian per-parameter, $\HANPP$, is obtained from \EQN~\ref{eqn:Gan0} as
\begin{align}
\label{eqn:Gan_lnI}
\beta\HANPP
   &:=-\frac{1}{\ND}\ln   \int\mathcal{D}\NDXI \, e^{-\beta \DETOPST} P(\NDXI) \\
\nonumber
   &=-\frac{1}{\ND}\ln \IH ,
\end{align}
where
\begin{align}
\label{eqn:InsideGan}
\IH := \int\mathcal{D}\NDXI \, e^{-\beta \DETOPST} P(\NDXI) .
\end{align}
That is, $\HANPP$ represents the Energy or Error that each of the $N\times M$ parameters contributes
(averaged over the $N$ training examples $\NDXI$).

The goal will be to derive the high-Temperature \AnnealedHamiltonian, $\GANMATHT$, which is now  defined such that:
\begin{align}
 \label{eqn:hanpp2}
  \Trace{\GANMATHT}:=MN\left(\Trace{\HANPPHT  }\right)
\end{align}

If examining the the trace of $\HANPPHT$, then we can infer the functional form necessary to define
the matrix-generalized \EffectivePotential for each parameter:
\begin{align}
  \label{eqn:EPSL_mat}
  \EPSL(\OVERLAP):=\Trace{\HANPPHT},
\end{align}
which would be like a mean-field potential, but we need something different for the matrix case.


To evaluate the integral, notice that $\IH$ is really just an average over i.i.d. data, and so it is just a product over $\ND$ independent terms ($1$ for each training example).
\begin{align}
\IH := \int\mathcal{D}\NDXI  e^{-\beta \DETOPST} P(\NDXI)  \rightarrow\left[\int\mathcal{D}\XI \;[\cdots]\; \right]^{\ND} ,
\end{align}
as in \EQN~\ref{eqn:I_4} below.
Moreover, when taking $\ln \IH$, the $N$ term pulls down and become a prefactor
\begin{align}
-\ln \IH = -\ln\left[\int\mathcal{D}\XI \;[\cdots]\; \right]^{\ND}= -\ND\ln\left[\int\mathcal{D}\XI \;[\cdots]\; \right] .
\end{align}
Thus, as with the vector case, $\GANMAT$ is like a mean-field average over the data $\XI$, indepedent of the sample size $N$.
Also, since the final result must scale as $N\times M$, the integral should scale as $M$, i.e.,
$\left[\int\mathcal{D}\XI \;[\cdots]\; \right]\sim M$.

If we substitute $\DETOPNN$, \EQN~\ref{eqn:DETOPNN}, into the integral $\IH$, \EQN~\ref{eqn:InsideGan}, then we obtain
\begin{align}
\label{eqn:I_1} 
\IH 
  & =  \int \mathcal{D}\NDXI \, \exp \left( -\beta\sum_{\mu=1}^{\ND} N(\XI^{\mu})^{\top} \left(\IM-\frac{1}{N} \SMAT^{\top}\TMAT \right) \XI^{\mu} \right) P(\NDXI)  \\
\nonumber
  & =  \int \mathcal{D}\NDXI \, \exp \left( -\beta\sum_{\mu=1}^{\ND} N(\XI^{\mu})^{\top} \left(\IM-\frac{1}{N} \SMAT^{\top}\TMAT \right) \XI^{\mu} \right) \NORM \exp\left( - \sum_{\mu=1}^{N} \frac{\|\XI^{\mu}\|^2}{2 \sigma^2} \right)  \\
    \nonumber
  &= \NORM \int \mathcal{D}\NDXI \, \exp \left(
    -\beta\sum_{\mu=1}^{\ND} N(\XI^{\mu})^{\top} (\IM-\tfrac{1}{N}\SMAT^{\top}\TMAT) (\XI^{\mu}) 
    - \sum_{\mu=1}^{\ND} \frac{\|\XI^{\mu}\|^2}{2 \sigma^2} \right) \\ 
\nonumber
  &= \NORM \int \mathcal{D}\NDXI \, \exp \left(
    -\frac{1}{2\sigma^2}\sum_{\mu=1}^{\ND}2\beta\sigma^{2} N(\XI^{\mu})^{\top} (\IM-\tfrac{1}{N}\SMAT^{\top}\TMAT) (\XI^{\mu}) 
    +  \Vert\XI^{\mu}\Vert^{2} \right) \\ 
\label{eqn:I_3} 
  &= \NORM \int \mathcal{D}\NDXI \, \exp \left(
    -\frac{1}{2\sigma^2}      
      \sum_{\mu=1}^{\ND}
          (\XI^{\mu})^{\top}[
      2\beta\sigma^{2}N (\IM-\tfrac{1}{N}\SMAT^{\top}\TMAT)+\IM] (\XI^{\mu})\right)  .
\end{align}
By combining the exponents, we obtain
\begin{align}
\nonumber
\IH
  &=  \NORM\int \mathcal{D}\NDXI 
  \exp\left[
    -\frac{1}{2\sigma^2}\sum_{\mu=1}^{N}(\XI^{\mu})^{\top}
    \left(\mathbf{M}
    \right)
    \XI^{\mu}
    \right ]\\ 
\label{eqn:I_4} 
  &=  \NORM\int \mathcal{D}\XI  
 \exp\left[
    -\frac{1}{2\sigma^2}(\XI)^{\top}
    \left(\mathbf{M}
    \right)
    \XI
    \right]^{\ND}   ,
\end{align}
where $\mathbf{M}=2\beta\sigma^{2}N(\IM - \tfrac{1}{N}\SMAT^{\top}\TMAT)+\IM$ is an $M \times M$ matrix.
We now use the familiar property of multi-variant Gaussian integrals,
\begin{align}
\label{eqn:det_M}
\int d\mathbf{x}  e^{-\frac{1}{2\sigma^{2}}(\mathbf{x})^{\top}\mathbf{M}(\mathbf{x}) } = (2\pi\sigma^{2})^{M/2}\frac{1}{\sqrt{\Det{ \mathbf{M}}}}
\end{align}
where $\mathbf{x}$ is an $m$-dim vector (with zero mean),
and $\mathbf{M}$ is a square positive-definite matrix, and $\Det{ \mathbf{M}}$ is the determinant of $\mathbf{M}$.
Using \EQN~\ref{eqn:det_M}, we can rewrite $\IH$ in \EQN~\ref{eqn:I_4} as
\begin{align}
\label{eqn:I_5}
\IH &=   \NORM\left[\frac{(2\pi\sigma^{2})^{M/2}}{\sqrt{\Det{ \mathbf{M}}}}\right]^{\ND} \\ \nonumber
    &=   \NORM(2\pi\sigma^2)^{NM/2}\left[
         \sqrt{\Det{ 2\beta\sigma^2N(\IM-\tfrac{1}{}\SMAT^{\top}\TMAT)+\IM}}\right]^{-\ND}  \\
\label{eqn:I_6}
   &=  \left( \frac{1}{2\pi\sigma^{2}} \right)^{N M/2}
       (2\pi \sigma^{2})^{M/2}
         \left[\sqrt{\Det{ \IM + 2\beta\sigma^{2}N(\IM-\tfrac{1}{N}\SMAT^{\top}\TMAT)}}\right]^{-\ND}  ,
\end{align}
where \EQN~\ref{eqn:I_6} follows by inserting $\XINORM$ from \EQN~\ref{eqn:xinorm}.
We can now identify $\sigma^{2}=\tfrac{1}{M}$ to obtain
\begin{align}
\nonumber
\IH &= \left[\sqrt{\Det{ \IM + 2\beta\sigma^{2}N(\IM-\tfrac{1}{N}\SMAT^{\top}\TMAT)}}\right]^{-\ND}   \\ 
\nonumber
    &= \left[\sqrt{\Det{ \IM  + \tfrac{2\beta}{M}N(\IM-\tfrac{1}{N}\SMAT^{\top}\TMAT})}\right]^{-\ND} \\ 
\label{eqn:I_7}
    &= \left[\Det{ \IM  + \tfrac{2\beta}{M}N(\IM-\tfrac{1}{N}\SMAT^{\top}\TMAT})\right]^{-\ND/2}  .
\end{align}

\paragraph{The High-Temperature Limit.}
In the high-T approximation, $\beta$ becomes small, giving the expression
$
\det(\IM+\epsilon\mathbf{\Omega})\approx1+\epsilon\Trace{\mathbf{\Omega}}  ,
$
which holds for an arbitrary matrix $\Omega$ for small $\epsilon$.
Using this, we can evaluate the determinant in \EQN~\ref{eqn:I_7} in the large-$N$ approximation, 
which gives
\begin{align}
  \label{eqn:I_8}
  \IH &\approx  \left[1  + \tfrac{2\beta}{M}N(\Trace{\IM-\tfrac{1}{N}\SMAT^{\top}\TMAT})\right]^{-\ND/2}   .
\end{align}
Inserting this into \EQN~\ref{eqn:Gan_lnI}, we obtain
\begin{align}
\beta\HANPP
   &=-\frac{1}{\ND}\ln \left[1  + \tfrac{2\beta}{M}N(\Trace{\IM-\tfrac{1}{N}\SMAT^{\top}\TMAT})\right]^{-\ND/2}  \\ \nonumber
      &=\frac{1}{2}\ln \left[1  + \tfrac{2\beta}{M}N(\Trace{\IM-\tfrac{1}{N}\SMAT^{\top}\TMAT})\right]
      \label{eqn:Gan_lnI_final_mwm}
\end{align}

This form of the Hamiltonian, $\GANMAT$, however, is not symmetric, and we will
eventually want a symmetric operator or matrix.
Fortunately, the high-T form, $\GANMATHT$, can be made symmetric, as
shown below.  But first, let show that this result is consistent with the our previous Percpetron result.
  
\paragraph{Matrix-Generalized ST Error $\GANMATHT$ for $N=1$.}
To start, observe that when $N=1$, \EQN~\ref{eqn:Gan_lnI_final_mwm} becomes
\begin{align}
\nonumber
\beta\GANMAT\vert_{N=1}
&= \beta\HANPP\vert_{N=1}  \\ 
  &=  -\frac{1}{\ND}\ln  \left[1  + 2\beta\tfrac{1}{M}\Trace{(M-\SVEC^{\top}\TVEC}\right]^{-\ND/2} \\ 
\nonumber
  &= \frac{1}{2}\ln  \left[1  + 2\beta\tfrac{1}{M}\Trace{M-\SVEC^{\top}\TVEC}\right] \\ 
\label{eqn:GANmat_m_equals_1}
  &=  \frac{1}{2}\ln \left[1 + 2\beta(1-R)\right]  ,
\end{align}
where we recall that $\SVEC$ and $\TVEC$ are implicitly normalized to $1/m$, where here $m=M$.  This result shows that \EQN~\ref{eqn:Gan_lnI_final_mwm} reduces to \EQN~\ref{eqn:Gan2}, as desired.

\charles{We want $\GANMATHT$ to represent the total energy per training example, so it should NOT include the $\tfrac{1}{M}$
FIX THIS}
This ensures the Hamiltonian scales as $M$ so the Free Energy scales as $N \times M$, the
number of free paramaters in the system.
Notice that for the final \LayerQualitySquared Hamiltonian $\HBARE$, this will change.
\begin{align}
  \label{eqn:I_9}
    \IH\approx  \left[1  + \frac{2\beta}{M}N\Trace{\IM-\tfrac{1}{N}\SMAT^{\top}\TMAT}\right]^{-\ND/2} ,
\end{align}
for any $M>1$.
Given this, it follows from \EQN~\ref{eqn:Gan_lnI_final_mwm} and \EQN~\ref{eqn:Gan_lnI} that we can define 
\begin{align}
\label{eqn:GANmat}
\beta\GANMAT \vert_{N=1}
  &:=  \frac{M}{2}\ln\det \left[1 + \frac{2\beta}{M}(\IM-\mathbf{R})\right]  ,
\end{align}
and reduces to the same functional form as \EQN~\ref{eqn:Gan2}, as desired (recalling that $\SVEC$ and $\TVEC$ are implicitly normalized by $M=m$).

To obtain the high-Temperature form, we use
\begin{align}
\ln\det\!\Bigl[\IM + \epsilon\mathbb{M}\Bigr]
\approx \mathrm{Tr}\!\bigl(\epsilon\mathbb{M}\bigr)
\quad(\epsilon\ll 1),
\end{align}
with $\epsilon=\tfrac{2\beta}{M}$ and $\mathbb{M}=\IM-\OVERLAP$.   
We now obtain the following result for the matrix-generalized high-T of $\GANMATHT$ using
\begin{align}
\label{eqn:GANHTmatRN1}
\Trace{\GANMATHT}\vert_{N=1} = \Trace{\IM-\OVERLAP} = M-\Trace{\OVERLAP}
\end{align}

The final expression for $\GANMATHT$ is
\begin{align}
\label{eqn:GANHTmatR}
\GANMATHT = N(\IM-\OVERLAP)
\end{align}


\subsection{Expressing the Layer Quality}
\label{sxn:quality}

In this section, we obtain an approximation expression for the \LayerQualitySquared from the IZ \FreeEnergy for the \GeneralizationError, 
given in \EQN~\ref{eqn:betaIZG_S} in Section~\ref{sxn:matgen_quality_hciz_A}.

For the required \FreeEnergy $\IZFE$, we will use the matrix-generalized Hamiltonian
from \EQN~\ref{eqn:GANHTmatR} for the
\LayerQuality, $\GANMATHT=N(\IM-\OVERLAP)$.
giving a Boltzmann distribution and the corresponding \ThermalAverage.  
Expanding this out, we have
\begin{align}
  \label{eqn:IZFE0}
  -\IZFE =& -  \ln \int d\mu(\mathbf{S}) \exp\left[-\ND\beta  \Trace{\HANHT(\OVERLAP)}  \right] \\
\end{align}
We could also express $\IZFE$ In terms of the matrix-generalized \EffectivePotential $\EPSL(\OVERLAP)$
(\EQN~\ref{eqn:EPSL_mat}), giving
\begin{align}
  -\IZFE =& -  \ln \int d\mu(\mathbf{S}) \exp\left[-\ND\beta N \EPSL(\OVERLAP)  \right] 
\end{align}
In analogy with \EQN~\ref{eqn:Gan_highT}, 
as $\HANHT(\mathbf{R})=M-\OVERLAP$,  write
\begin{align}
-\IZFE  =& -  \ln \int d\mu(\mathbf{S}) \exp\left[-\ND\beta N \operatorname{Tr}[M-\OVERLAP]  \right] 
\end{align}
Using the approximation $\operatorname{Tr}[\OVERLAP]\approx\sqrt{\OLAPSQD}$, we have
\begin{align}
  -\IZFE 
\approx& - \ln \int d\mu(\mathbf{S}) \exp\left[-\ND\beta N(M-\sqrt{\OLAPSQD} ) \right] \\ 
\label{eqn:IZFE1}
=& -  \ln \int d\mu(\mathbf{S}) \exp[-\ND\beta NM]\exp\left[\ND\beta N\sqrt{\OLAPSQD}\right], \\
\label{eqn:IZFE2}
=& -  \ln e^{-\ND\beta NM} \int d\mu(\mathbf{S}) \exp\left[\ND\beta N\sqrt{\OLAPSQD}\right], \\
\label{eqn:IZFE3}
=& -  \ln e^{-\ND\beta NM} - \ln \int d\mu(\mathbf{S}) \exp\left[\ND\beta N\sqrt{\OLAPSQD}\right], \\
\label{eqn:IZFE4}
=& \ND\beta NM - \ln \int d\mu(\mathbf{S}) \exp\left[\ND\beta N\sqrt{\OLAPSQD}\right], 
\end{align}

Notice that, as expected, the Free Energy scales $\IZFE$ as $\ND \times N \times M$, the total number of degrees of freedom of the theory.
\noindent
Since \EQN~\ref{eqn:IZFE0} equals \EQN~\ref{eqn:IZFE1}, we can write the \FreeEnergy in terms of $\OLAPSQD$. From \EQN~\ref{eqn:IZFE4}, 
we can identify a generating function ($\Gamma_{\Q}$) for the layer accuracy, or \Quality.
For example, to compute the average \Quality $\Q$, we would use
\begin{align}
  \label{eqn:IZG_Q}
  \beta\Gamma^{IZ}_{\Q} :=  \ln \int d\mu(\mathbf{S}) \exp\left[\ND\beta N\sqrt{\OLAPSQD}\right],
\end{align}
and to compute the average \Quality (squared) $\QT$, we would use
\begin{align}
    \label{eqn:IZG_QT2}
  \beta\Gamma^{IZ}_{\QT} :=  \ln \int d\mu(\mathbf{S}) \exp\left[\ND\beta N\OLAPSQD\right] .
\end{align}
We have recovered \EQN~\ref{eqn:betaIZG_S}.
We can now also define the \LayerQualitySquared \Hamiltonian as
\begin{equation}
      \label{eqn:HBARE}
  \HH_{\QT}:=\mathbf{R}^{\top}\mathbf{R}
\end{equation}
which is a symmetric operator, as desired.
Consequently, we may also write
\begin{align}
    \label{eqn:IZG_QT3}
  \beta\Gamma^{IZ}_{\QT} :=  \ln \int d\mu(\mathbf{S}) \exp\left[\ND\beta N \operatorname{Tr}[\HH_{\QT}]\right]  .
\end{align}

\subsection{Derivation of the \TRACELOG Condition}
\label{sxn:TraceLogDerivation}

\subsubsection{Setting up the Saddle Point Approximation (SPA)}
\label{sxn:TraceLogDerivation_A}
As in \EQN~\ref{eqn:IZG_dmuS}, 
we can write \EQN~\ref{eqn:IZG_QT} in terms of the $\AMATN = \SMAT\SMAT^{\top}$ form of the Outer \Student Correlation matrix, giving
\begin{align}
\IZG = \ln\INTS d\mu(\SMAT) \exp\left(\ND\beta N \Trace{ \tfrac{1}{N}\TMAT^{\top}\AMATN \TMAT } \right)
\end{align}
where $d\mu(\SMAT)$ is the measure over all $N \times M$ real-valued random matrices,
although we really want to limit this to all $N \times M$ real matrices that resemble the \Teacher $\TMAT$,
which we clarify below.

To transform $\IZG$ into a form we can evaluate using Tanaka's result~\cite{Tanaka2008}, 
we need to change the measure from an integral over all random $N \times M$ student weight matrices
$d\mu(\SMAT)$ to an integral over all $N \times N$
student correlation matrices $d\mu(\AMATM)$, i.e., $d\mu(\SMAT)\rightarrow d\mu(\AMATM)$.
To accomplish this, we can insert an integral over the Dirac Delta function
\begin{align}
  \label{eqn:I}
  \mathbf{I}:=
  \int d\mu(\AMATM)\delta(N\AMATM-\SMAT^{\top}\SMAT).
\end{align}

\noindent
(This is simply a resolution of the Identity.) This gives
\begin{align}
\label{eqn:IZG_3}
\IZG= \ln \INTS d\mu(\SMAT)\INTA d\mu(\AMATM)
           \delta\left( N\AMATM-\SMAT^{\top}\SMAT \right) 
           e^{ \ND\beta N Tr[\tfrac{1}{N} \TMAT^{\top} \AMATN\TMAT ] } ,
\end{align}
 where $d\mu(\AMAT)=\Probab{\AMAT}d\AMAT$ and $\Probab{\AMAT}$ is the
(still unspecified) probability density over the new random matrix $\AMAT$. 
Let us express \EQN~\ref{eqn:IZG_3} at \LargeN limit in $N$ as
\begin{align}
  \label{eqn:IZG_4}
  \lim_{N\gg 1}\IZG =
  \lim_{N\gg 1}\ln
  \int d\mu(\AMAT)
  \int d\mu(\SMAT)
  \delta(N\AMATM-\SMAT^{\top}\SMAT)
  e^{ \ND\beta N Tr[ \tfrac{1}{N}\TMAT^{\top}\AMATN\TMAT]) }  .
\end{align}
 
Now we assume we can first evaluate the term 
\begin{align}
  \lim_{N\gg 1} \int d\mu(\SMAT)    \delta(N\AMATM-\SMAT^{\top}\SMAT)
\end{align}
at \LargeN in $N$ using a \SaddlePointApproximation (SPA). Using the relation, 
\begin{align}
\delta(N\AMATM-\SMAT^{\top}\SMAT)
   =\mathcal{N}_M\INTAHAT  d\mu(\AHAT) e^{ iN Tr[\AHAT\AMATM] } e^{ -i Tr[\AHAT\SMAT^{\top}\SMAT] }  ,
\end{align}
where $\AHAT$ is an $M \times M$ auxiliary matrix, and the domain of integration $d\mu(\AHAT)$ is all $M \times M$ real-valued matrices, and where the normalization $\NORM_1$ is
\begin{align}
  \label{eqn:norm_1}
\NORM_1:=\frac{1}{(2\pi)^{M(M+1)/4}},
\end{align}
because $\AMAT$ is a symmetric matrix with $M(M+1)/2$ constraints.

This is simply the matrix generalization of
$\delta(x)=\dfrac{1}{2\pi}\int_{-\infty}^{\infty} e^{i\hat{x}x}d\hat{x}$,
so we can express the delta function as an exponential, giving
\begin{align}
\label{eqn:Q2}
\IZG = \NORM_1 \ln\INTS  d\mu(\SMAT) \INTA d\mu(\AMATM) 
                           \INTAHAT d\mu(\AHAT) e^{ iN Tr[ \AHAT\AMATM ] }
                           e^{ -i Tr[ \AHAT\SMAT^{\top} \SMAT ] }
                           e^{  \ND\beta N Tr[\tfrac{1}{N} \TMAT^{\top}\AMATN\TMAT ] } .
\end{align}

Rearranging terms, we obtain 
\begin{align}
\label{eqn:IZG_Gamma1}
\IZG =  \ln\INTA  d\mu(\AMATM) 
            e^{  \ND\beta N Tr[\tfrac{1}{N} \TMAT^{\top}\AMATN\TMAT ] } \times
           \Gamma_1,
\end{align}
where we define $\Gamma_1$ as 
\begin{align}
\Gamma_1 := \Gamma_1(\AMATM) 
         = \NORM_1 \INTS d\mu(\SMAT) 
                           \INTAHAT d\mu(\AHAT) e^{ iN Tr[ \AHAT\AMATM ] }
                                                           e^{ -i Tr[ \AHAT\SMAT^{\top} \SMAT ] } .
\end{align}
We can simplify the complex integral in $\Gamma_1$ with the Wick Rotation $i\AHAT\rightarrow\AHAT$.
The Wick rotation ensures that the Gaussian integral converges (although this has not been rigorously checked).
We may expect $d\mu(\AHAT)$ to be invariant to rotations in the complex plane,
so the Wick rotation does not introduce any complex prefactors. This gives
\begin{eqnarray}
\label{eqn:QWick}
\Gamma_1 = \NORM_1 \INTS d\mu(\SMAT)\INTAHAT d\mu(\AHAT) 
           e^{ N  Tr[ \AHAT\AMATM ] }
           e^{ - Tr[ \AHAT\SMAT^{\top} \SMAT] } \\
\label{eqn:QWick2}
         = \NORM_1 \INTS d\mu(\SMAT)\INTAHAT d\mu(\AHAT) 
           e^{ N  Tr[ \AHAT\AMATM ] }
           e^{ -Tr[ \SMAT\AHAT\SMAT^{\top} ] } ,
\end{eqnarray}
where the second line follows since the trace is invariant under cyclic permutations (i.e., $\Trace{ABC}=\Trace{BCA}=\Trace{CAB}$).
Swapping the order of the integrals yields
\begin{eqnarray}
\label{eqn:QWick3}
\Gamma_1 =\Gamma_1(\AHAT)  = \NORM_1
           \INTAHAT d\mu(\AHAT) 
           e^{ N Tr[\AHAT\AMATM ]}\times
           \Gamma_2  ,
\end{eqnarray}
where we define $\Gamma_2$ as
\begin{equation*}
\Gamma_2 := \Gamma_2(\AHAT)
         = \INTS d\mu(\SMAT)
           e^{ -Tr[ \SMAT\AHAT\SMAT^{\top} ] } .
\end{equation*}

To evaluate $\Gamma_2$, we will make several mathematically convenient approximations.
(These will yield an approximate expression which can be verified empirically.)
We first assume for the purpose of changing measure that the (data) columns of $\SMAT$ are
statistically independent, so that the measure $d\mu(\SMAT)$ factors into $N$ Gaussian distributions
\begin{align}
\label{eqn:dMuS}
d\mu(\SMAT) = \prod_{\mu=1}^{N}d\mu(\mathbf{s}_{\mu})=\prod_{\mu=1}^{N}d\mathbf{s}_{\mu} ,
\end{align}
where $\mathbf{s}_{\mu}$ is an M-dimensional vector.
The singular values of $\SMAT$ are invariant to random permutations of the columns or rows,
so the resulting ESD does not change.  
This is very different from permuting $\SMAT$ element-wise, which will make the resulting ESD Marchenko Pastur (MP).

Using \EQN~\ref{eqn:dMuS}, 
$\Gamma_2$ reduces to a simple Gaussian integral, which can be evaluated as a product of $N$ Gaussian integrals (over the $M\times M$ matrix $\AHAT$)
\begin{align}
\label{eqn:int-out-J2}
\Gamma_2
   =& \left[\INTsvec d\mathbf{s}e^{-\tfrac{1}{\sigma^{2}} \mathbf{s}\AHAT\mathbf{s}^{\top} }\right]^{N} \\
   =& \left[\NORM_2\;\Det{\AHAT}^{-1/2}\right]^{N}  ,
\end{align}
where the normalization term $\NORM_2$
\begin{align}
\label{eqn:norm_2}
\NORM_2 := \left(\pi\sigma^{2}\right)^{M/2}  ,
\end{align}
where $\sigma^{2}=\mathbf{s}^{\top}\mathbf{s}=1/M$
\charles{Need to be very careful here.  Is this $1/N$ or $1/M$ ?  See also A2. We pick $\sigma^{2}=1/M$ to ensure the normalization on $\XI$ is correct.
This needs to be double checked.}

For any square, non-singular matrix $\AHAT$,  $ \Trace{\ln\AHAT}=\ln \Det{\AHAT}$, so
it follows from \EQN~\ref{eqn:int-out-J2} that
\begin{align}
\nonumber
\ln\Gamma_2
   &=N\ln\NORM_2\left[( \Det \AHAT)^{-1/2} \right]  \\   
&= N\ln\NORM_2 -\tfrac{N}{2}\Trace{ \ln\AHAT }  ,
\end{align}
so that
\begin{align}
\label{eqn:log-Gamma}
\Gamma_2 = (\NORM_2)^{N} e^{ -\tfrac{N}{2} Tr[ \ln\AHAT ] } 
\end{align}

Substituting
\EQN~\ref{eqn:log-Gamma}
into \EQN~\ref{eqn:QWick3},
we can write $\Gamma_1$ as
\begin{eqnarray}
  \label{eqn:gamma1}
\Gamma_1(\AHAT)  =& C_{\Gamma_1}\INTAHAT d\mu(\AHAT)   e^{ N Tr[ \AHAT\AMATM ] }  e^{ -\tfrac{N}{2} Tr[ \ln\AHAT ] }  ,
\end{eqnarray}
where
\begin{equation}
    C_{\Gamma_1}:=\NORM_1 e^{\NORM_2}.
\end{equation}

We can now evaluate the integral in \EQN~\ref{eqn:QWick3} over the Lagrange multiplier $\AHAT$ (i.e., $\INTAHAT $). 
If we call this $\Gamma_1(\AHAT)$,
then (following Tanaka~\cite{Tanaka2008}) we can define the \emph{\RateFunction} $I(\AHAT,\AMATM)$ such that
\begin{align}
\label{eqn:LambdaA}
\Gamma_1(\AHAT)=\INTAHAT  d\mu(\AHAT) e^{-NI(\AHAT,\AMATM)}  ,
\end{align}
where
\begin{align}
\label{eqn:IAA}
I(\AHAT,\AMATM) = -\Trace{ \AHAT\AMATM} + \frac{1}{2}\Trace{ \ln\AHAT }  .
\end{align}

We can formally evaluate the integral in \EQN~\ref{eqn:LambdaA} in the large-$N$ limit using a \SaddlePointApproximation (SPA)
(see Section~\ref{sxn:mathP}, \EQN~\ref{eqn:SPA}), as
\begin{align}
\label{eqn:LAMBDA}
\Gamma_1(\AHAT)\rightarrow \sqrt{\dfrac{(2\pi)^{N/2}}{N\Vert I\Vert}}e^{-N I^{*}(\AHAT, \AMATM)}  ,
\end{align}
where $I^{*}(\AHAT,\AMAT)$ is the maximum value over all $\AHAT$ for fixed $\AMATM$, obtained using
\begin{align}
  \label{eqn:IAA-sup}
  I^{*}(\AHAT,\AMATM) :=
\underset{N\gg 1}{\lim} I(\AHAT,\AMATM) =
 \underset{\AHAT}{\sup}\left[-\Trace{\AHAT\AMATM}+\frac{1}{2}\Trace{\ln\AHAT}\right]  ,
\end{align}
where at the SPA we have stationarity,
\begin{align}
  \label{eqn:SP0}
  \dfrac{\partial}{\partial\AHAT}I(\AHAT,\AMATM) =& -\AMATM+\dfrac{1}{2\AHAT}=0  
\end{align}
and the Hessian of $I$ becomes
\begin{align}
  \label{eqn:Ixx}
\frac{\partial^2 }{\partial \hat{A}^2} I(\AHAT,\AMATM)= -\frac{1}{2} \left( \frac{1}{2} \hat{A}^{-1} \right) \otimes \left( \frac{1}{2} \hat{A}^{-1} \right) = -\frac{1}{8} \AMATM \otimes \AMATM
\end{align}
where $\otimes$ is the Kronecker product. Using \EQN~\ref{eqn:SP0}, we have that $1 / 8$ $\AMATM \otimes \AMATM$ is the saddle point Hessian of $I$ w.r.t. $\AMATM$. Solving the SPA equation, we find that 
the prefactor (i.e. Hessian) is given as
$ \Det{-\frac{1}{8} \AMATM \otimes \AMATM} = \left( -\frac{1}{8} \right)^{M^2} \left(  \Det{\AMATM} \right)^M$.   

Substituting for $\AHAT$ into \RateFunction (\EQN~\ref{eqn:IAA}), $I$ becomes
\begin{align}
\label{eqn:IAA_2}
I^{*}(\AHAT,\AMATM) = -\Trace{ \IM } + \frac{1}{2}\Trace{ \ln\AMATM }   \\ \nonumber
 = -M + \frac{1}{2}\Trace{ \ln\AMATM }  .
\end{align}

\noindent
In order for this result to be physically meaningful, 
we need that if $I^{*}(\AHAT,\AMATM)$ grows,
then it must grow slower than $N$, and,
more importantly, that $\Det{\AMAT}$ be non-zero.
Importantly, when $\Det{\AMAT}=1$ exactly, however, then $\Gamma_1$ becomes a constant,
and this simplifies things considerably!

\subsubsection{Casting the \GeneratingFunction \texorpdfstring{$(\IZG)$}{beta G} as an HCIZ Integral}
\label{sxn:TraceLogDerivation_B}

In this section, we express the \GeneratingFunction $\IZG$, 
given in \EQN~\ref{eqn:IZG_dmuS} (equivalently, in \EQN~\ref{eqn:betaIZG_S},) 
as an HCIZ Integral, 
as given in \EQN~\ref{eqn:IFA2_braket}.

Inserting $I^{*}(\AHAT,\AMAT)$ from \EQN~\ref{eqn:IAA_2} into $\IZG$, we obtain
\begin{align}
  \label{eqn:IZG_IAA}
  \IZG 
  & =  \ln \left[ C_{\Gamma_1} e^{-NM}\int d\mu(\AMAT)
  e^{\ND\beta N Tr[ \tfrac{1}{N}\TMAT^{\top}\AMATN\TMAT] }
  e^{\tfrac{N}{2}\ln(\Det{\AMATM})}\right]  \\ \nonumber
  & =
    \ln  C_{\Gamma_1}
  - NM
  +  \ln \left[ \int d\mu(\AMAT)
    e^{\ND\beta N Tr[ \tfrac{1}{N}\TMAT^{\top}\AMATN\TMAT] }
    e^{\tfrac{N}{2}\ln(\Det{\AMATM})}\right]  .
\end{align}

\noindent
So long as the second term $\Trace{\mathbb{I}_{M}}$ does not depend on $N$, 
it will vanish when we take the partial derivative of $\IZG$ to obtain the $\AVGNNGE$, in which case it is not important.  
We can then simply write the \GeneratingFunction $\IZG$  as in \EQN~\ref{eqn:IFA2_integral} as:
\begin{align}
  \label{eqn:IZG_integral}
  \IZG 
   =  \ln \left[ \int d\mu(\AMATM)
    e^{\ND\beta N Tr[ \tfrac{1}{N}\TMAT^{\top}\AMATN\TMAT] }
    e^{\tfrac{N}{2}\ln(\Det{\AMATM})}\right]  ,
\end{align}
or, in \BraKet notation, as
\begin{align}
  \label{eqn:IZG_braket}
  \IZG = 
   \ln\left\langle
  e^{\ND\beta N Tr[ \tfrac{1}{N}\TMAT^{\top}\AMATN\TMAT] }
  e^{\tfrac{N}{2}\ln(\Det{\AMATM})}
  \right\rangle_{\AMATM}   .
\end{align}

\subsection{MLP3 Model Details}
\label{sxn:appendix_MLP3details}
The empirical MLP3 Model implements the assumptions described in Section~\ref{sxn:matgen} used the following 
procedures:

A three-layer \MultiLayerPerceptron was trained for classification on the MNIST dataset\cite{MNIST1998}. The first Fully 
Connected (FC) hidden layer has 300 units, the second FC hidden layer has 100 units, and the third FC layer has ten 
units for classification, matching the ten digit classes of MNIST. Input images are grayscale, and were rescaled to the 
$[0, 1]$ range. Following the keras\cite{keras2015} defaults, the weights were initialized using the Glorot 
Normal\cite{GloBen10} method, and the biases were initialized to $0$. Each model was trained using Categorical Cross 
Entropy as the loss function. The loss function was {\em summed} over each mini-batch, which is the default behavior for 
Keras, rather than being {\em averaged}, which is the default for pytorch\cite{pytorch2019}. 

Optimization was carried out by either Stochastic Gradient Descent (SGD) 
without momentum, or the Adam algorithm \cite{kingma2014_TR}. The \LearningRate (LR) was set to 0.01 for SGD, and 0.001 
for Adam. The LR was held constant, i.e., there was no decay schedule. Each algorithm proceeded epoch by epoch until the 
value of the loss function did not decrease by more than 0.0001 for three consecutive epochs. At each epoch, 
the \WW~ tool was used to compute metrics for each layer. Loss values reported are the average loss per labeled example, 
and not the summed loss over each minibatch. Training loss is averaged over all batches in the epoch, whereas test loss 
is evaluated once at the end of the epoch.

In some experiments, only one layer was trained, while the others were left frozen. In other experiments all layers were 
trained. Models were trained using a series of mini-batch sizes ranging from $1$ to $32$. For each separate training 
run, the models were re-initialized to the same starting random weights, all random seeds were reset, and deterministic 
computations were used to train the models.

Separate notebooks are provided for keras and pytorch implementations of the experiments.

\subsection{Tanaka's Result}
\label{sxn:tanaka}

In this section, we will rederive the result by Tanaka~\cite{Tanaka2007,Tanaka2008} that we use in our main derivation,
and, importantly, explain how to address the missing Temperature term.
For completeness, we restate it here using the notation of the main text:
\begin{equation}
  \label{eqn:hciz}
  \lim_{N \gg 1} \frac{1}{N} \ln 
\underbrace{
  \Expected[\mathbf{\AMAT}]{
    \exp\left(\frac{\ND\beta}{2}
    \Trace{\mathbf{W}^{\top} \AMATN \mathbf{W}}
    \right)
  }
 }_{\text{HCIZ Integral}}
  = \frac{\ND\beta}{2} \sum_{i=1}^{M} \GNORM(\lambda_{i})
\end{equation}
where 
$\mathbf{W}$ is the $N\times M$ \Teacher weight matrix, 
$\AMAT=\AMAT_N$ is the $N\times N$ \Student (correlation) matrix, 
but $\beta$ is now the inverse-Temperature (because we are working with real matrices), $\ND$ is the size of the training set, 
and we have added the  $(\tfrac{1}{2})$ prefactor (which will be clear later).
$\GNORM(\lambda)$ is a complex analytic function of the eigenvalues $\lambda$ of (the \Teacher Correlation matrix) $\XMAT$, 
whose functional form will depend on the structure of the limiting form of (the \Student) ESD $\rho_{\AMAT}^{\infty}(\lambda)$.
We may also write it as $\GNORM(\XMAT)$ below.
We call it perhaps somewhat imprecisely a  \emph{\GEN} because the final results for the \LayerQuality  $\Q$ will take the form of a \emph{Tail norm} in many cases.

To apply this result, we note that
while the term $\beta$ is just a constant in~\cite{Tanaka2008}
($1$ or $2$, depending on whether the random matrix is real or complex),
it is not actually inverse Temperature $\beta=\tfrac{1}{T}$ in the original derivation.
Still, we seek a final result that is linear in $\beta=\tfrac{1}{T}$,
so that we can easily evaluate $\QT$ in the high-T limit, i.e.
$\QT
=\tfrac{\partial}{\partial \ND}\tfrac{1}{\beta}\IZGINF
=\tfrac{\partial}{\partial \beta}\tfrac{1}{\ND}\IZGINF$
(see \ref{eqn:IZG_QT}).
We can introduce the new term $\ND\beta$ by
simply changing the scale of $\AMAT_N$ since the final result is a sum of \RTransforms, which by definition
are linear, i.e., $\GNORM(\ND\beta\lambda)=\ND\beta \GNORM(\lambda)$, however, it is instructive
to rederive the final result, with $\ND\beta$ explicitly included.

\paragraph{Notation.}

We start by rewriting the Tanaka result, \EQN~(\ref{eqn:hciz}),
in our notation for the expected value $\Expected[\AMAT]{\cdots}$ operator, as follows:
\begin{equation}
\label{eqn:hciz2}
  \tfrac{1}{2N}\IZGINF = 
  \tfrac{1}{N}\lim_{N\gg 1} \ln \underbrace{ \int d\mu(\mathbf{\AMAT})\left[\exp\left(\frac{\ND\beta}{2}\Trace{\mathbf{W}^{\top}\AMATN\mathbf{W}}\right)\right] }_{\mbox{HCIZ Integral}} 
  = \ND\beta \tfrac{1}{2}\sum_{i=1}^{M}\GNORM(\lambda_{i})   .
\end{equation}
where we have added a $\tfrac{1}{2}$ for technical convenience (to make the connection with the LDP, below).
If we denote the internal HCIZ integral as 
\begin{equation}
\label{eqn:hciz_def}
  \HCIZ := \int d\mu(\mathbf{\AMAT})\left[\exp\left(\frac{\ND\beta}{2}\Trace{\mathbf{W}^{\top}\AMATN\mathbf{W}}\right)\right]  ,
 \end{equation}
then it holds that 
\begin{equation}
  \label{eqn:hciz_def2}
  \IZG := \tfrac{1}{N}\ln\HCIZ  ,
\end{equation}
from which it follows that 
\begin{equation}
\label{eqn:hciz_def3}
  \IZGINF := \lim_{N \gg 1}\tfrac{1}{N}\ln\HCIZ  .
\end{equation}

The SPA approximates the \PartitionFunction $\HCIZ$, which is now an HCIZ integral,  by its peak value.
For this, $\GNORM(\lambda)$ itself must either not explicitly depend on $N$ and/or at least not grow faster than $N$.

The trick here is we can choose an \RTransform of $\mathbf{\AMAT}$
that is a simple analytic expression based on the observed
empirical spectral density (ESD) of the $\mathbf{X}$.
And this can readily be done for the ESDs for a wide range of layer weight matrices
observed in modern DNNs because their ESDs are \HeavyTailed \PowerLaw\cite{MM19_HTSR_ICML}.
We can then readily express the \Quality $\Q$ of the \Teacher
layer in a simple functional form, (i.e  an approximate Shatten Norm).

Importantly, the matrices $\mathbf{X}$  and $\mathbf{\AMAT}$ must be well approximated
by low rank matrices since the derivation in Tanaka requires this.  Fortunately,
this appears to be generally true for the layers in very well trained DNNs,
which is what allows us to apply this withing the~\ECS.
In fact, technically we need to integrate over $d\mu(\AECS)$; this is straightforward
as this simply changes the lower bound on the integral from $0\rightarrow\LambdaECSmin$,
both above and in the subsequent derivation.

Finally, we note that $\GNORM(\mathbf{X})$ is kind of \emph{Generalized Norm} because 
it can be evaluated as a sum over a function of the $M$ eigenvalues $\lambda_{\mu}$ of the \Teacher
correlation matrix $\mathbf{X}=\frac{1}{N}\mathbf{W}^{\top}\mathbf{W}$.
$\GNORM(\mathbf{X})$ will turn out to be an expression similar to the Frobenius Norm or the
Shatten Norm of $\mathbf{X}$, depending on the functional form we choose to represent the
limiting form of the \Student ESD, $\rho_{\AMAT}^{\infty}(\lambda)$ and the associated \RTransform $R(z)$, and various approximations made thereafter,

\subsubsection{Setup and Outline}
\label{sxn:tanaka_setup}

\noindent
To evaluate \ref{eqn:hciz2},
we 
want to integrate over all \Student Correlation matrices $\mathbf{\AMAT}$
that ``resemble'' the \Teacher Correlation matrix $\mathbf{X}$.  
To 
formalize this idea,
we need to 
define the measure over ``all desired'' $\mathbf{\AMAT}$, $d\mu(\mathbf{\AMAT})$, 
in 
terms of the actual $M$ eigenvalues, $\left\{ \lambda_{i} \right\}_{i=1}^{M}$, of the \Teacher.  And WLOG, the final result can be restricted to the  $\MECS$ eigenvalues of the \EffectiveCorrelationSpace (ECS).

\paragraph{Randomness assumption.}
For real weights $\mathbf{W}$,  we assume an \emph{orthogonally invariant} ensemble,
$d\mu(\mathbf W)=d\mu(\mathbf U\mathbf W\mathbf U^{\top})$ for all $\mathbf U\in O(M)$,
mirroring the isotropic Gaussian initialisation widely used in neural networks.
Crucially, Tanaka’s large-$N$ analysis shows the resulting HCIZ exponent depends only on the eigenvalue spectrum, so the final integrated–$R$ expression should remain applicable even when full rotational invariance is later broken in training.

\paragraph{Using a source matrix $\mathbf{D}$ to represent $d\mu(\AMAT)$ with $d\mu(\WMAT)$.}

We consider all matrices $\mathbf{\AMAT}$ with the same limiting spectral density, $\rho_{\AMAT}^{\infty}(\lambda)$,
as the limiting (\emph{empirical}) ESD of the \Teacher.
That is, we want $\rho_{\AMAT}^{\infty}(\lambda)=\rho^{\infty}_{\WMAT}(\lambda)$, where $\TMAT=\WMAT$.
Of course, there are infinitely many weight matrices $\mathbf{W}$ with the same $M$ eigenvalues, $\left\{ \lambda_{i} \right\}_{i=1}^{M}$, as the \Teacher.
Let us specify these matrices with the measure $d\mu(\mathbf{W})$.
Doing this lets us then write the measure $d\mu(\mathbf{\AMAT})$ in terms of $d\mu(\mathbf{W})$ as:
\begin{equation}
\label{eqn:dmuA}
d\mu(\mathbf{\AMAT}) 
   := e^{- \frac{\ND\beta}{2} Tr[\mathbf{W}\mathbf{D}\mathbf{W}^{\top}]} d\mu(\mathbf{W})  ,
\end{equation}
where $\mathbf{D}$ is some $M \times M$ matrix, called the \SourceMatrix, to be specified below,
and the $\tfrac{1}{2}$ here as well.
Indeed, the key idea here will be to define $\mathbf{D}$ in such a way as to obtain the desired final result.
Notice also that we have added an $\ND\beta$ term; this will be factored out later.

We can now represent the partition function
$\ZD$, by inserting \EQN~\ref{eqn:dmuA} into \EQN~\ref{eqn:hciz_def}.
$\ZD$ is now defined as an integral over all
possible (\Teacher) weight matrices $\mathbf{W}$
\begin{equation}
  \label{eqn:ZD0}
    \ZD=\int d\mu(\mathbf{W})\exp\left[\frac{\ND\beta}{2}\left
    (\Trace{\mathbf{W}^{\top}\AMAT_N\mathbf{W}}-\Trace{\mathbf{W}\mathbf{D}\mathbf{W}^{\top}}  
    \right)\right]  ,
\end{equation}
Observe that this integral only converges when all the eigenvalues of $\DMAT$, 
$\left\{ \DeltaMu \right\}_{\mu=1}^{M}$, 
are larger than the maximum eigenvalue of $\mathbf{\AMAT}$, i.e., when $\DeltaMu >\lambda_{max}$, for $\mu\in[1,M]$
(although below this will become $\ND\beta\DeltaMu >\lambda_{max}$).
Later, we will place $\mathbf{D}$ in diagonal form, and we will obtain an explicit expression for its $M$ eigenvalues in terms of the $M$ non-zero eigenvalues of $\mathbf{X}$.
The eigenvalues of $\mathbf{D}$ will turn out to Lagrange Multipliers, needed later.

\paragraph{The Saddle Point Approximation (SPA) and the Large Deviation Principle (LDP).}

To evaluate the large-$N$ case of $\IZG$ (see \ref{eqn:hciz_def2},~\ref{eqn:hciz_def3}), 
we assume that the distribution of possible \Teacher correlation matrices,
$\mu(\mathbf{X})$, satisfies a \emph{Large Deviation Principle (LDP)}.
An LDP applies to probability distributions that take an exponential form,
such that $\mu(\mathbf{X})=e^{-N I(\mathbf{X})}d\mu(\mathbf{X})$,
where  $I(\mathbf{X})$ is Entropy or Rate function $I(\mathbf{X})$.

In applying an LDP, we effectively restrict measure of student correlation matrices $\mathbf{\AMAT}$
to those most similar to the empirically observed \Teacher correlation matrix $\mathbf{X}$.
We expect the measure over all \Teacher correlation matrices
follows an LDP because the ESD is far from Gaussian,
the dominant generalizing components reside in the tail of the ESD,
and at finite-size the tail decays at worst as an exponentially
Truncated \PowerLaw (TPL).


\paragraph{Three $3$ steps to evaluate $\Expected[\AMAT]{\ZD}$ in the \WideLayer large-$N$ approximation.}
The goal is to start with \EQN~\ref{eqn:ZD0} and obtain two separate, equivalent
relations, Eqns.~\ref{eqn:ZD_step1} and ~\ref{eqn:ZD_step2}:
\begin{enumerate}
   \item
   \textbf{Obtaining an integral transform of $\rho^{\infty}_{\AMAT}(\lambda)$.}
   First, we expand and reduce \EQN~\ref{eqn:ZD0} and evaluate the expected value of
   $\EZDA=\EZDATWO$ in the \LargeN limit in $N$ by expressing the $\rho_{\AMAT}(\lambda)$
   for the $N\times N$ matrix $\AMAT=\mathbf{A}_N=\tfrac{1}{N}\SMAT\SMAT^{\top}$
   in the continuum representation, i.e., as ]
   $\rho^{emp}_{\AMAT}(\lambda)\rightarrow \rho^{\infty}_{\AMAT}(\lambda)$, to obtain:
   \begin{equation}
      \label{eqn:ZD_step1}
      \lim_{N\gg 1}\dfrac{1}{N}
      \ln\EZDATWO =M\ln(\dfrac{2\pi}{\ND\beta})-\sum_{\mu=1}^{M}\int \ln(\delta_{\mu}-\lambda)\rho^{\infty}_{\AMAT}(\lambda)d\lambda  .
   \end{equation}
   This gives us an $\EZDATWO$ in terms of an integral transform $\rho^{\infty}_{\AMAT}(\lambda)$, which we can model.\footnote{This integral of $\rho^{\infty}_{\AMAT}(\lambda)$  is related to the \emph{Shannon Transform}, an integral transform from information theory that is useful when analyzing the mutual information or the capacity of a communication channel~\cite{Tanaka2007}. }
   \item
   \textbf{Forming the \SaddlePointApproximation (SPA).}
   We evaluate \EQN~\ref{eqn:ZD0} as the expected value of $\EZDA=\EZDAONE$
   for the $M \times M$ matrix $\AMAT=\mathbf{A}_M=\tfrac{1}{N}\SMAT^{\top}\SMAT$
   (but explicitly in terms of $d\mu(\mathbf{X})$).
   Then, taking in the large-$N$ approximation using the SPA,
    (and which can be done implicitly using the LDP), we obtain
   \begin{equation}  
  \label{eqn:ZD1} 
  \lim_{N \gg 1} \EZDAONE\simeq\int  \exp\left(\ND\beta N\Trace{\GFANCY}\right)d\mu(\mathbf{X}) \approx \exp(\ND\beta N\GMAX)
\end{equation}
  where  $\GFANCY$ depends on $\GNORM(\mathbf{X})$, and $\GMAX=\sup_{\XMAT}\GFANCY$.
  We can then write
   \begin{equation}
      \label{eqn:ZD_step2}
      \lim_{N \gg 1}\dfrac{1}{N}\ln\EZDAONE \approx \ND\beta\GMAX  ,
   \end{equation}

 \item
 \textbf{Finding the Inverse Legendre Transform.}
  To do this, we now equate
  \begin{equation}
  \lim_{N \gg 1}\frac{1}{N}\ln\EZDAONE=  \lim_{N \gg 1}\frac{1}{N}\ln\EZDATWO
  \end{equation}
 Then, we can form the
 inverse Legendre transform  which we will let us relate $\GNORM(\lambda)$ in \EQN~\ref{eqn:hciz} to the integrated \RTransform of $\rho^{\infty}_{\AMAT}(\lambda)$.
\end{enumerate}

\noindent

(See~\ref{sxn:tanaka_step3}.)

\subsubsection{Step 1. Forming the Integral Transformation of ESD 
\texorpdfstring{$(\rho_{\AMAT}^{\infty}(\lambda))$}{rho(lambda)}}
\label{sxn:tanaka_step1}
We first establish \EQN~\ref{eqn:ZD_step1}, in Steps $1.1-1.4$.
This is done by changing variables under a Unitary transformation, $\mathbf{W}\rightarrow\mathbf{\check{W}}$,
evaluating the resulting functional determinant,
and then taking the continuum limit of the ESD
$\tilde{\rho}_{\AMAT}(\lambda)\rightarrow\rho^{\infty}_{\AMAT}(\lambda)$.

\paragraph{Step 1.1}
To do so, let us first assume that \Teacher correlation matrix $\mathbf{X}$ and the source matrix $\mathbf{D}$
are simultaneously diagonalizable
(i.e., their commutator is zero: $[\mathbf{X}, \mathbf{D}]=0$).
In this case, we may write the generating function $\ZD$ in \EQN~\ref{eqn:ZD0} as
\begin{align}
\label{eqn:Z-diag}
\ZD &= \int d\mu(\mathbf{W}) \exp\frac{\ND\beta}{2}
 \bigg( 
\Trace{\mathbf{W}^{\top}\mathbf{U}^{\top}\mathbf{\Lambda}\mathbf{U}\mathbf{W}} 
- \Trace{\mathbf{W}\mathbf{V}^{\top}\mathbf{\Delta}\mathbf{V}\mathbf{W}^{\top}} 
\bigg)  ,
\end{align}
where we have defined
\begin{equation}
\label{eqn:diag-A-D}
    \AMATN=\mathbf{U}^{\top}\mathbf{\Lambda}\mathbf{U},\;\;
    \mathbf{D}=\mathbf{V}^{\top}\mathbf{\Delta}\mathbf{V}  ,
\end{equation}
%
where $\mathbf{U}$ ($N\times N$) and $\mathbf{V}$ ($M\times M$) are Unitary matrices.
%
Since $\mathbf{U}^{\top}\mathbf{U}=\mathbf{I}$ and $\mathbf{V}^{\top}\mathbf{V}=\mathbf{I}$,
we can insert these identities into $\ZD$ in \ref{eqn:Z-diag}, giving
\begin{align}\label{eqn:Z0-diag}
\ZD &= \int d\mu(\mathbf{W}) \exp\frac{\ND\beta}{2}\times  \\ \nonumber
&\bigg(\Trace{(\mathbf{V}^{\top}\mathbf{V})\mathbf{W}^{\top}\mathbf{U}^{\top}\mathbf{\Lambda}\mathbf{U}\mathbf{W}(\mathbf{V}^{\top}\mathbf{V})} 
-\Trace{(\mathbf{U}^{\top} \mathbf{U})\mathbf{W}\mathbf{V}^{\top} \mathbf{\Delta} \mathbf{V}\mathbf{W}^{\top}(\mathbf{U}^{\top} \mathbf{U})}\bigg)  .
\end{align}
We can identify the reduced weight matrix $\mathbf{\check{W}}$ as
\begin{equation}
   \label{eqn:Wcheck}
   \mathbf{\check{W}}=\mathbf{U}\mathbf{W}\mathbf{V}^{\top},\;\;
   \mathbf{\check{W}}^{\top}=\mathbf{V}\mathbf{W}^{\top}\mathbf{U}^{\top}  ,
\end{equation}
Rearranging parentheses, this gives 
\begin{align}
\ZD &= \int d\mu(\mathbf{W}) \exp\frac{\ND\beta}{2}\times  \\ \nonumber
&\bigg(\Trace{\mathbf{V}^T(\mathbf{V}\mathbf{W}^T\mathbf{U}^T)\mathbf{\Lambda}(\mathbf{U}\mathbf{W}\mathbf{V}^T)\mathbf{V}}  
-\Trace{ \mathbf{U}^{\top}(\mathbf{U} \mathbf{W}\mathbf{V}^{\top})\mathbf{\Delta}(\mathbf{V}\mathbf{W}^{\top} \mathbf{U}^{\top})\mathbf{U} }\bigg)  .
\end{align}
We can now express $\ZD$ in terms of $\mathbf{\check{W}}$ as
\begin{align}
\label{eqn:hciz-W-red}
\ZD &= \int d\mu(\mathbf{W})\exp\frac{\ND\beta}{2}
 \bigg(\Trace{\mathbf{V}^{\top}\mathbf{\check{W}}^{\top}\mathbf{\Lambda}\mathbf{\check{W}}\mathbf{V}} 
 -\Trace{\mathbf{U}^{\top}\mathbf{\check{W}}\mathbf{\Delta}\mathbf{\check{W}}^{\top}\mathbf{U}}\bigg)  .
\end{align}
Since the Trace operator $\Trace{\cdot}$ is invariant to Unitary (Orthogonal) transformations, we can
now remove the
$\mathbf{U}$ and $\mathbf{V}$ terms, giving the simplified expression
for our generating function $\ZD$ in terms of
the two diagonal matrices $\mathbf{\Lambda}, \mathbf{\Delta}$, 
the reduced weight matrix $\mathbf{\check{W}}$, and
the Jacobian $J(\mathbf{\check{W}})$ transformation for $d\mu(\mathbf{W})\rightarrow d\mu(\mathbf{\check{W}})$, as:
\begin{align}
\label{eqn:hciz-W-red2}
    \ZD & =\int d\mu(\mathbf{\check{W}})J(\mathbf{\check{W}})\exp\frac{\ND\beta}{2}
 \bigg( \Trace{\mathbf{\check{W}}^{\top}\mathbf{\Lambda}\mathbf{\check{W}}} 
       -\Trace{\mathbf{\check{W}}\mathbf{\Delta} \mathbf{\check{W}}^{\top}} \bigg)  .
\end{align}

\paragraph{Step 1.2}
We can now evaluate the
integral using the standard relation for the functional determinant for infinite-dimensional Gaussian integrals~\cite{EngelAndVanDenBroeck}

\begin{equation}
\label{eqn:hciz-det}
    \ZD=\left(\dfrac{2\pi}{\ND\beta}\right)^{NM/2}\det\left(\mathbf{\Delta}-\mathbf{\Lambda}\right)^{-1/2}
\end{equation}
where the Jacobian is unity for the Unitary transformation.

\begin{equation}\label{eqn:Jacobian}
    J(\mathbf{\check{W}})=1  .
\end{equation}
since $\mathbf{W} \mapsto \check{\mathbf{W}}$  is an orthogonal transfomation.
We now use the standard Trace-Log-Determinant relation~\cite{EngelAndVanDenBroeck}
\begin{equation}\label{eqn:tr-ln-det}
    \Trace{\ln\mathbf{M}}=\ln\det\mathbf{M}  .
\end{equation}
Let us insert $(\exp\ln)$ on the R.H.S. of \ref{eqn:hciz-det}, to obtain
\begin{align}
\nonumber
\ZD
  &=\exp\ln\bigg[\left(\dfrac{2\pi}{\ND\beta}\right)^{NM/2}\det\left(\mathbf{\Delta}-\mathbf{\Lambda}\right)^{-1/2}\bigg] \\ 
\nonumber
  & =\exp\bigg[\left(\dfrac{NM}{2}\right)\ln\dfrac{2\pi}{\ND\beta}-\dfrac{1}{2}\Trace{\ln\left(\mathbf{\Delta}-\mathbf{\Lambda}
\right)}\bigg] \\ 
\label{eqn:hciz-exp-ln}
  & =\exp\bigg[\dfrac{NM}{2}\ln\dfrac{2\pi}{\ND\beta}-\dfrac{1}{2}\ln\det\left(\mathbf{\Delta}-\mathbf{\Lambda}\right)\bigg]  .
\end{align}

\paragraph{Step 1.3}
We now want to express
the generating function $\ZD$ 
in \ref{eqn:hciz-exp-ln}
in terms of an integral over the continuous limiting spectral density
$\rho_{\AMAT}(\lambda)$ of the correlation matrix $\AMATN$.  

First, we express the Determinant of the matrix $\mathbf{\Delta}-\mathbf{\Lambda}$ in terms of discrete eigenvalues:
\begin{equation}
\label{eqn:det-discrete}
    \det\left(\mathbf{\Delta}-\mathbf{\Lambda}\right)^{-1/2}=\prod_{\mu=1}^{M}\prod_{i=1}^{N}\left(\DeltaMu-\lambda_{i}\right)^{-1/2}  .
\end{equation}
This gives the Log-Determinant in terms of the $M$ (non-zero)
eigenvalues of $\mathbf{D}$ and $\AMATN$, as
\begin{equation}
\label{eqn:ln-det-discrete}
    \ln\det\left(\mathbf{\Delta}-\mathbf{\Lambda}\right)^{-1/2}=-\dfrac{1}{2}\sum_{\mu=1}^{M}\sum_{i=1}^{N}\ln\left(\DeltaMu-\lambda_{i}\right)  .
\end{equation}
We can express 
the ESD, $\tilde\rho_{\AMAT}(\lambda)$, of the 
Student Correlation 
matrix
$\AMAT_N$ in terms of the Dirac delta-function, $\delta(x)$, as
\begin{equation}
\label{eqn:rho-emp}
    \tilde\rho_{\AMAT}(\lambda)=\frac{1}{N}\sum_{i=1}^{N}\delta(\lambda-\lambda_{i})  .
\end{equation}
Using this, the \ExpectedValue of the Log-Determinant 
in \ref{eqn:ln-det-discrete}
can be expressed in terms of the ESD of
$\AMAT_N$ as
\begin{align}
\nonumber
\Expected[\AMAT_N]{\ln\det\left(\mathbf{\Delta}-\mathbf{\Lambda}\right)^{-1/2}}
   & = -\dfrac{1}{2}\sum_{\mu=1}^{M}\sum_{i=1}^{N}\int
       d\lambda\ln(\DeltaMu-\lambda)\delta(\lambda-\lambda_{i}) \\ 
\nonumber
   & = -\dfrac{1}{2}\sum_{\mu=1}^{M}\int
       d\lambda\ln(\DeltaMu-\lambda)\sum_{i=1}^{N}\delta(\lambda-\lambda_{i}) \\ 
\label{eqn:ln-det-rho}
   & = -\dfrac{1}{2}\sum_{\mu=1}^{M}\int
       d\lambda\ln(\DeltaMu-\lambda)
       N\tilde\rho_{\AMAT}(\lambda)  .
\end{align}

Let us insert this back into our
expression for the generating function,
\ref{eqn:hciz-exp-ln},   
giving
$\EZDATWO$ in terms of the ESD $\tilde{\rho}_{\AMAT}$ as
\begin{equation}
\label{eqn:Z-rho}
    \EZDATWO=\exp\bigg\{\dfrac{N}{2}\big[M\ln\dfrac{2\pi}{\ND\beta}-\sum_{\mu=1}^{M}\int
        d\lambda\ln(\DeltaMu-\lambda)\tilde{\rho}_{\AMAT}(\lambda)\big]\bigg\}  .
\end{equation}

We can now replace
the sum over the $N$ eigenvalues $\lambda_{i}$ with an integral over the limiting
ESD, $\rho(\lambda)$, to obtain
\begin{equation}
\label{eqn:rho-}
\rho^{\infty}_{\AMAT}(\lambda)=    \lim_{N\rightarrow\infty}\tilde\rho_{\AMAT}(\lambda)  .
\end{equation}
Observe that this effectively means that we are taking a \LargeN limit in $N$, $N\gg 1$.
This lets us write the \ExpectedValue of the generating function $\ZD$
in \ref{eqn:Z-rho}
as
\begin{equation}
\label{eqn:Z-rho-2}
    \lim_{N\gg 1}\EZDATWO=\exp\bigg\{\dfrac{N}{2}\big[M\ln\dfrac{2\pi}{\ND\beta}-
    \sum_{\mu=1}^{M}\int
        d\lambda\ln(\DeltaMu-\lambda)\rho^{\infty}_{\AMAT}(\lambda)\big]\bigg\}
\end{equation}

\paragraph{Step 1.4}
Using the Self-Averaging Property,
\begin{equation}
   \ln \EZDATWO \simeq \Expected[\mathbf{A}_M]{\ln \ZD} ,
\end{equation}
It follows from \EQN~\ref{eqn:Z-rho-2}
that
\begin{equation}
   \lim_{N\gg 1} \ln \EZDATWO
   \simeq \dfrac{N M}{2}\ln\dfrac{2\pi}{\ND\beta}
         -\dfrac{N}{2}\sum_{\mu=1}^{M}\int d\lambda\ln(\DeltaMu-\lambda)\rho^{\infty}_{\AMAT}(\lambda)  .
\end{equation}
The $N$-dependence now cancels out,
and we are left an approximate expression due to the remaining dependence of the continuum limiting density
$\rho^{\infty}_{\AMAT}(\lambda)$ (for $\AMAT=\AMATN$)
\begin{equation}
\label{eqn:ln-Z-Nlim2}
    \lim_{N \gg 1}\dfrac{2}{N}\ln \EZDATWO
    = M\ln\dfrac{2\pi}{\ND\beta}-\sum_{\mu=1}^{M}\int d\lambda\ln(\DeltaMu-\lambda)\rho^{\infty}_{\AMAT}(\lambda)  .
\end{equation}
This completes the derivation of \EQN~\ref{eqn:ZD_step1}; 
we have an expression for the expected value of $\ZD$,
evaluated in the \LargeN (continuum) limit in $N$.

\subsubsection{Step 2: The Saddle Point Approximation (SPA): Explicitly forming the Large Deviation Principle (LDP)}
\label{sxn:tanaka_step2}
We now evaluate $\EZDA$ in \EQN~\ref{eqn:ZD1} as $\EZDATWO$ 
to  establish \EQN~\ref{eqn:ZD_step2}, \charles{in Steps $2.1-2.6$}.

Using the LDP (and following similar approaches in spin glass theory \cite{PP95}),
below we will show that we can write the expected value of $\ZD$ 
in terms of $d\mu(\mathbf{X})$ now (which is equivalent to $d\mu(\AMATM)$)
and in the large-$N$ approximation, as
\begin{equation}
  \label{eqn:LDP}
 \lim_{N \gg 1} \EZDAONE=
  \int\exp\left(\ND\beta N \Trace{\GNORM(\mathbf{X})}-NI(\mathbf{X})+o(N)\right)d\mu(\mathbf{X})
\end{equation}
where $I(\mathbf{X})$ is \RateFunction, defined below,
and $\GNORM(\mathbf{X})$ is what we are eventually solving for.

\paragraph{Step 2.0} We start with the \emph{expected} \PartitionFunction
\begin{align}
  \label{eqn:avg_ZD}
  \EZDATWO
  &=
  \int d\mu(\AMAT)\int  d\mu(\WMAT)
      \exp\Bigl[
         \tfrac{\ND\beta}{2}\Trace{\WMAT^{\TR}\AMATN\,\WMAT}
        -\tfrac{\ND\beta}{2}\Trace{\WMAT\DMAT\WMAT^{\TR}}
      \Bigr].
\end{align}
The average over $\AMATN$ affects only the first exponential; applying the SPA, we
\textbf{define} a matrix function $\GFANCY$, depending solely on
$\XMAT=\frac1N\WMAT^{\TR}\WMAT$, by
\begin{align}
  \label{eqn:def_GFANCY}
  \int d\mu(\AMAT)\,
        \exp\Bigl[\tfrac{\ND\beta}{2}\Trace{\WMAT^{\TR}\AMATN\,\WMAT}\Bigr]
  &=
  \exp\Bigl[\tfrac{\ND\beta N}{2}\Trace{\GFANCY}\Bigr].
\end{align}
which will be valid in the large-$N$ in $N$ approximation below.

We now note that given the duality of measures, we can assert 
\begin{equation}
\EZDA=\EZDAONE=\EZDATWO.
\end{equation}
This lets us insert \eqref{eqn:def_GFANCY} into \eqref{eqn:avg_ZD} and then write
\begin{align}
  \label{eqn:ZD_after_Aavg}
  \EZDA
  &=
  \int d\mu(\WMAT)
      \exp\Bigl[
         \tfrac{\ND\beta N}{2}\Trace{\GFANCY}
        -\tfrac{\ND\beta}{2}\Trace{\WMAT\DMAT\WMAT^{\TR}}
      \Bigr].
\end{align}

The now need to determine an explicit form for
$\GFANCY$. We introduce a new change of measure, 
 $d\mu(\mathbf{W})\rightarrow d\mu(\mathbf{X})$.
Then, we show this lets us express $\EZDA$ as $\EZDX$ and to express it using the LDP.
Next, we apply a SPA to solve for $\GMAX=\max\;\GNORM$.
Importantly, we also show how to incorporate the inverse-Temperature $\ND\beta$,
which is new.

\paragraph{Step 2.1}
To define the transformation $d\mu(\mathbf{W})\rightarrow d\mu(\mathbf{X})$,  where (recall) $\mathbf{X}=\frac{1}{N}\mathbf{W}^{\top}\mathbf{W}$,
we use the (again) the integral representation of the Dirac delta-function $\delta(x)$:
\begin{equation}
  \label{eqn:dirac}
  \delta(x):=\frac{1}{2\pi}\int_{-\infty}^{\infty} e^{i\hat{x}x} d\hat{x}.
\end{equation}
This lets us express the transformation of measure $d\mu(\mathbf{W})\rightarrow d\mu(\mathbf{X})$
(approximately) as
\begin{align}
\nonumber
  d\mu(\mathbf{W}) &:= \delta(\frac{1}{2}\Trace{N\mathbf{X}-\mathbf{W}^{\top}\mathbf{W}}) d\mu(\mathbf{X}) \\ 
  &= \frac{1}{2\pi}\int_{-\infty}^{\infty} e^{i\frac{1}{2}
    \Trace{\hat{X}(N\mathbf{X}-\mathbf{W}^{\top}\mathbf{W})}
  }
  d\mu(\mathbf{X})d\mu(\hat{X} ),
\end{align}
where $\hat{X}$ is a scalar (or really a matrix of scalars),
and we have 
$\frac{1}{2}$
term for mathematical consistency below.
\footnote{The full change of measure would require a delta function constraint
for each matrix element $X_{i,j}$, i.e., 
$\delta\left(\frac{1}{2}N\left(X_{i,j}-[\mathbf{W}^{\top}\mathbf{W}]_{i,j}\right)\right)$.
Here, we assume the Trace constraint is sufficient for our level of rigor.
}

\paragraph{Step 2.2}
Next, we take a Wick Rotation%
\footnote{The Wick rotation converts an oscillatory integral into an exponentially decaying one which should be well defined.  Technically, this is an analytic continuation which needs to be checked, but following standard practice in physics we will assume the resulting integral is analytic and therefore well defined and we will proceed onward. },
$i\XHAT\rightarrow -\XHAT$, so that the terms under the integral are all real (not complex), giving:
\begin{align}
  \label{eqn:dmuX}
  d\mu(\mathbf{W}) &= \NWICK\int_{-i\infty}^{i\infty} e^{\frac{1}{2}\Trace{\hat{X}(N\mathbf{X}-\mathbf{W}^{\top}\mathbf{W})}} d\mu(\mathbf{X})d\mu(\hat{X} ).
\end{align}
where $d\mu(\hat{X})$ is a measure over {\scalebox{0.7}{$\frac{M(M-1)}{2}$}} Lagrange multipliers, and the normalization is
\begin{equation}
\NWICK = \left(\tfrac{1}{2\pi i}\right)^{\scalebox{0.7}{$\scriptstyle \frac{M(M-1)}{2}$}}
\end{equation}

\paragraph{Step 2.3}
We now insert \ref{eqn:dmuX} into~\ref{eqn:ZD_after_Aavg}, which lets us
express $\EZDA$ as an integral over the \Teacher Correlation matrices.

\begin{align}
  \nonumber
  \EZDA&=
  \NWICK\int_{\XMAT}  \int_{-i\infty}^{i\infty}
  e^{N\frac{\ND\beta}{2} \Trace{\GFANCY}+\frac{N}{2}\Trace{\hat{X}\mathbf{X}} }
  e^{-\frac{1}{2}\Trace{\hat{X}\mathbf{W}^{\top}\mathbf{W}}}
  e^{\frac{\ND\beta}{2}\Trace{\mathbf{W}\mathbf{D}\mathbf{W}^{\top}}}
  d\mu(\hat{X} )
  d\mu(\mathbf{\XMAT}) \\ 
  \nonumber
  &=
  \NWICK\int_{\XMAT}  \int_{-i\infty}^{i\infty}
  e^{N\frac{\ND\beta}{2}\Trace{\GFANCY}+ \frac{N}{2}\Trace{\hat{X}\mathbf{X}}}
  e^{-\frac{1}{2}\Trace{\mathbf{W}\hat{X}\mathbf{W}^{\top}}+
  \frac{\ND\beta}{2}\Trace{\mathbf{W}\mathbf{D}\mathbf{W}^{\top}}}
  d\mu(\hat{X} )
  d\mu(\mathbf{\XMAT}) \\ 
  \label{eqn:ZD3}
    &=
  \NWICK\int_{\XMAT}  \int_{-i\infty}^{i\infty}
  e^{N\frac{\ND\beta}{2}\Trace{\GFANCY}+
  \frac{N}{2}\Trace{\hat{X}\mathbf{X}}}
  e^{\frac{1}{2}\Trace{\mathbf{W}(\ND\beta\mathbf{D}-\hat{X})\mathbf{W}^{\top}}}
  d\mu(\hat{X} )
  d\mu(\mathbf{\XMAT})  .
\end{align}

\paragraph{Step 2.4}
We can now rearrange terms to make this expression look like the \EQN~\ref{eqn:LDP}
In Large Deviations Theory, the \RateFunction is defined by the Legendre Transform,
\begin{equation}
\label{eqn:rate-fun}
    \mathcal{I}(\mathbf{X})=\underset{\mathbf{\check{X}}}{\sup}
    \left[Tr\dfrac{1}{2}\mathbf{{X}}^{\top}\mathbf{\check{X}}-\ln\mathbb{M}(\mathbf{\check{X}})\right]  ,
\end{equation}
where $\mathbb{M}(\mathbf{\check{X}})$ is the \MomentGeneratingFunction,
$\ln\mathbb{M}(\mathbf{\check{X}})$, is the \CumulantGeneratingFunction, and
and $\mathbf{\check{X}}$ is a (matrix of) \emph{Lagrange Multiplier}(s).  $\mathbb{M}(\mathbf{\check{X}})$ is defined in terms of the (unnormalized) density $p(\mathbf{x})$ as
\begin{equation}
\label{eqn:rate_function}
   \mathbb{M}(\mathbf{\check{X}})=\exp\left(\frac{1}{2}\mathbf{x}^{\top}\mathbf{\check{X}}\mathbf{x}\right)  ,
  p(\mathbf{x})d\mathbf{x}
\end{equation}
which, in turn, is defined in terms of the source matrix $\mathbf{D}$,
\begin{equation}
\label{eqn:X_density}
p(\mathbf{x})=\exp\left(-\tfrac{1}{2}\mathbf{x}^{\top}\ND\beta\mathbf{D}\mathbf{x}\right)  .
\end{equation}
The moment generating function $ \mathbb{M}(\mathbf{\check{X}})$ is then given by
\begin{equation}
\mathbb{M}(\mathbf{\check{X}}) 
   = \int \exp\left(-\frac{1}{2}\mathbf{x}^T(\ND\beta\mathbf{D} - \mathbf{\check{X}})\mathbf{x}\right) d\mathbf{x} 
   = (2\pi)^{\frac{M}{2}} \Det{\ND\beta\mathbf{D} - \mathbf{\check{X}}}^{-\frac{1}{2}}  .
\end{equation}

\paragraph{Step 2.5}
The \SaddlePointApproximation (SPA) can be used to solve for $\mathcal{I}(\mathbf{\check{X}})$ 
by solving for the stationary conditions
\begin{equation}
  \frac{\partial}{\partial \mathbf{\check{X}}} I(\mathbf{X},\mathbf{\check{X}}) = 0  .
\end{equation}
First, let us compute $ \ln \mathbb{M}(\mathbf{\check{X}}) $ as:
\begin{equation}
\ln \mathbb{M}(\mathbf{\check{X}}) = \frac{M}{2} \ln(2\pi) - \frac{1}{2} \ln \Det{\ND\beta\mathbf{D} - \mathbf{\check{X}}}.
\end{equation}
Substituting this into the expression for the Legendre transform, we obtain:
\begin{equation}
I(\mathbf{X},\mathbf{\check{X}}) 
   = \sup_{\mathbf{\check{X}}} \left[\frac{1}{2} \Trace{\mathbf{X} \mathbf{\check{X}}} - \frac{M}{2} \ln(2\pi) + \frac{1}{2} \ln \Det{\ND\beta\mathbf{D} - \mathbf{\check{X}}} \right].
\end{equation}
The supremum of this expression is attained at the value of $\mathbf{\check{X}}$ that satisfies:
\begin{equation}
\frac{\partial}{\partial \mathbf{\check{X}}} \left[\frac{1}{2} \Trace{\mathbf{X} \mathbf{\check{X}}} + \frac{1}{2} \ln \Det{\ND\beta\mathbf{D} - \mathbf{\check{X}}} \right] = 0.
\end{equation}
Taking the derivative, we obtain
\begin{equation}
\frac{1}{2} \mathbf{X} + \frac{1}{2} (\ND\beta\mathbf{D} - \mathbf{\check{X}})^{-1} = 0,
\end{equation}
which simplifies to:
\begin{equation}
\mathbf{X} = (\ND\beta\mathbf{D} - \mathbf{\check{X}})^{-1} \quad \Rightarrow \quad \mathbf{\check{X}} = \ND\beta\mathbf{D} - \mathbf{X}^{-1}.
\end{equation}
Substituting $ \mathbf{\check{X}} = \ND\beta\mathbf{D} - \mathbf{X}^{-1} $ back into the expression for $ I(\mathbf{X})$, we obtain:
\begin{equation}
I(\mathbf{X}) = \frac{1}{2} \left[\Trace{\mathbf{X} (\ND\beta\mathbf{D} - \mathbf{X}^{-1})} - \frac{M}{2} \ln(2\pi) + \frac{1}{2} \ln \Det{\mathbf{X}^{-1}}\right].
\end{equation}
\begin{equation}
\Trace{\mathbf{X}\ND\beta\mathbf{D} - \mathbf{I}} = \Trace{\mathbf{X}\ND\beta\mathbf{D}} - N,
\end{equation}
\begin{equation}
\ln \Det{\mathbf{X}^{-1}} = -\ln \Det{\mathbf{X}},
\end{equation}
we get:
\begin{equation}
I(\mathbf{X}) = \frac{1}{2} \left[\Trace{\mathbf{X}\ND\beta\mathbf{D}} - \ln \Det{\mathbf{X}} - M - M\ln(2\pi)\right].
\end{equation}

Finally, we express $I(\mathbf{X})$ in the form:

\begin{equation}
I(\mathbf{X}) = \frac{1}{2} \left[ -M(1 + \ln(2\pi)) + \Trace{\mathbf{X}\ND\beta\mathbf{D}} - \ln \Det{\mathbf{X}} \right].
\end{equation}

\paragraph{Step 2.6}
\begin{equation}
  \ND\beta\GFANCY=\mathbb{M}(1+\ln 2\pi)+\ND\beta\Trace{\GNORM(\mathbf{X})} - \Trace{\mathbf{X}\ND\beta\mathbf{D}} +  \ln \Det{\mathbf{X}}.
\end{equation}

We restrict our solution to those where $\XMAT$ and $\ND\beta\mathbf{D}$ can be diagonalized simultaneously.
In particular, this lets us write
\begin{equation}
\Trace{\mathbf{X}\ND\beta\mathbf{D}} = \sum_{\mu=1}^{M}\ND\beta\delta_{\mu}\lambda_{\mu}   ,
\end{equation}
where $\ND\beta\delta_{\mu}$ and $\lambda_{\mu}$ denote the eigenvalues of $\XMAT$ and $\ND\beta\mathbf{D}$, resp.

We can now write the maximum value of $\GNORM$, $\GMAX$, as
\begin{equation}
\label{eqn:gmax_final}
\ND\beta\GMAX = M \left( 1 + \ln \frac{2\pi}{\ND\beta} \right) - \sum_{\mu=1}^{M} \min_{\ND\beta\delta_{\mu}} \left[\ND\beta\delta_{\mu}\lambda_{\mu}
- \ND\beta\GNORM(\lambda_{\mu}) + \ln \lambda_{\mu} \right]   .
\end{equation}

\subsubsection{Expressing the~\GEN~
\texorpdfstring{$(\GNORM(\lambda))$}{GA(lambda)} as the Integrated~\RTransform~
\texorpdfstring{$(R(z))$}{R(z)} of the~\CorrelationMatrix~\texorpdfstring{$(\AMAT)$}{A}}
\label{sxn:tanaka_step3}
Having completed both steps, let us combine Eqns.~\ref{eqn:ZD_step1},~\ref{eqn:ln-Z-Nlim2}
with~\ref{eqn:ZD_step2} and~\ref{eqn:gmax_final}.
We follow the first arguments by Tanaka~\cite{Tanaka2007} (which follows Cherrier~\cite{Cherrier2003}).
\begin{align}
   M\ln(\dfrac{{2}\pi}{\ND\beta})-\sum_{\mu=1}^{M}\int \ln(\ND\beta\delta_{\mu}-\lambda)\rho^{\infty}_{\AMAT}(\lambda)d\lambda
      = M \left( 1 + \ln \frac{2\pi}{\ND\beta} \right) - \sum_{\mu=1}^{M} \min_{\ND\beta\delta_{\mu}} \left[\ND\beta\delta_{\mu}\lambda_{\mu}
      - \ND\beta\GNORM(\lambda_{\mu}) + \ln \lambda_{\mu} \right]   .
\end{align}
By canceling the $\ln \frac{2\pi}{\ND\beta}$ term from both sides, we obtain
\begin{align}
   -\sum_{\mu=1}^{M}\int \ln(\ND\beta\delta_{\mu}-\lambda)\rho^{\infty}_{\AMAT}(\lambda)d\lambda
   =
   M - \sum_{\mu=1}^{M} \min_{\ND\beta\delta_{\mu}} \left[\ND\beta\delta_{\mu}\lambda_{\mu}
   - \ND\beta\GNORM(\lambda_{\mu}) + \ln \lambda_{\mu} \right]   .
\end{align}
Since this is true for every $\mu$, we can solve this for any arbitrary eigenvalue $\lambda_{\mu}$.

Dropping the $\mu$ subscript, we have the following identity:
\begin{align}
\label{eqn:concave_id} 
 \min_{\delta} \left[\ND\beta\delta\lambda - \ND\beta\GNORM(\lambda) + \ln \lambda \right]
 = 1 -\int \ln(\ND\beta\delta-\lambda)\rho^{\infty}_{\AMAT}(\lambda)d\lambda   .
\end{align}

We need to invert \ref{eqn:concave_id} in order to find $\ND\beta\GNORM(\lambda)$.
If we choose the eigenvalues of $\mathbf{D}$ such that $\ND\beta\delta_{\mu}>\lambda_{max}$ for all $\mu$, then 
this relation is concave and therefore invertible via a Legendre transform.  

This gives
\begin{equation}
\ND\beta\GNORM(\lambda) = \ND\beta\delta(\lambda) \lambda - \int \ln[\ND\beta\delta(\lambda) - \lambda] \rho^{\infty}_{\AMAT}(\lambda) d\lambda - \ln \lambda - 1  ,
\end{equation}
where we need to  define  $ \ND\beta\delta(\lambda)$, which (not to be confused with the Dirac delta-function), describes
the functional dependence between the eigenvalues of the source matrix $\DMAT$ and the \Student \CorrelationMatrix $\AMAT$.

$\GNORM(\lambda)$ is computed by minimizing over $\delta$, ensuring the relationship holds for the entire spectrum.
So let us take the derivative of $\ND\beta\GNORM$ w.r.t. $\lambda$. 
Term by term, this gives:
\begin{equation}
\dfrac{d}{d\lambda} \ND\beta\delta(\lambda)\lambda = \ND\beta\delta(\lambda) + \dfrac{d \ND\beta\delta(\lambda)}{d\lambda} \lambda
\end{equation}
\begin{equation}
\dfrac{d}{d\lambda} \ln\lambda = \dfrac{1}{\lambda}
\end{equation}
\begin{align}
\dfrac{d}{d\lambda} \int \ln[\ND\beta\delta(\lambda) - \lambda] \rho^{\infty}_{\AMAT}(\lambda) d\lambda
&= \int \dfrac{d}{d\lambda} \ln[\ND\beta\delta(\lambda) - \lambda] \rho^{\infty}_{\AMAT}(\lambda) d\lambda \\ \nonumber
&= \int \dfrac{d \ND\beta\delta(\lambda)}{d\lambda}\dfrac{ \rho^{\infty}_{\AMAT}(\lambda)}{\ND\beta\delta(\lambda) - \lambda}d\lambda  \\ \nonumber
&=  \dfrac{d \ND\beta\delta(\lambda)}{d\lambda}\int\dfrac{ \rho^{\infty}_{\AMAT}(\lambda)}{\ND\beta\delta(\lambda) - \lambda}d\lambda
\end{align}

We can now simplify by defining $\delta(\lambda)$ implicitly by the integral relation
\begin{equation}
\lambda = \int \frac{\rho^{\infty}_{\AMAT}(\lambda)}{\ND\beta\delta(\lambda) - \lambda} d\lambda.
\end{equation}
Combining terms, this gives
\begin{equation}
\frac{d\ND\beta\GNORM(\lambda)}{d\lambda} = \ND\beta\delta(\lambda) - \frac{1}{\lambda},
\end{equation}

 Inverting the derivative, we obtain an integral equation for $\ND\beta\GNORM(\lambda)$ 
\begin{equation}
\ND\beta\GNORM(\lambda) = \int_0^\lambda \left(\ND\beta\delta(z) - \frac{1}{z}\right) dz.
\end{equation}

Notice since  $\ND\beta\delta(\lambda) \approx \frac{1}{\lambda}$ for $\lambda \ll 1$, then
as $\GNORM(0) = 0$ and we set the lower integrand to $0$ (for now).  Even though Tanaka’s original proof assumes an analytic continuation without branch cuts, a heavy-tailed spectrum merely shifts the lower limit of the $R$-transform integral, so the expression for $\ND\beta\mathcal G(\lambda)$ continues to hold.

To further connect these to the \RTransform $R_{\AMAT}(z)$, we recall that the \CauchyStieltjes (or just \Cauchy See~\ref{eqn:Cz}) transform $\mathcal{C}_{\AMAT}(z)$  is given by:

\begin{equation}
\mathcal{C}_{\AMAT}(z) = \int \frac{\rho_{\AMAT}(\lambda)}{z - \lambda} d\lambda.
\end{equation}

The relationship between the \Cauchy transform and the \RTransform is  then expressed as:

\begin{equation}
\mathcal{C}_{\AMAT}\left(R_{\AMAT}(z) + \frac{1}{z}\right) = z,
\end{equation}

which implies:

\begin{equation}
\ND\beta\GNORM(\lambda) = \int_0^{\lambda} R_{\AMAT}(z) dz.
\end{equation}

WLOG, as mentioned earlier, we can replace the lower bound on $\lambda$ from
$0\rightarrow\LambdaECSmin$ to obtain
\begin{equation}
\ND\beta\GNORM(\lambda) = \int_{\LambdaECSmin}^{\lambda} \Re[R_{\AECS}(z]) dz.
\end{equation}
where $\LambdaECSmin$ corresponds to the start of the \EffectiveCorrelationSpace (\ECS),
and, notably,  take the \emph{Real} part of $R(z)$.
We take the Real part because the imaginary parts coming from the upper and lower lips of
the cut cancel.\footnote{
Alternatively, one might try to replace $R(z)$ by its modulus;
doing so breaks the Legendre relation $G' = R$ and spoils the
additivity of free cumulants.}


\subsection{Existence of the Free \texorpdfstring{$R$}{R}–Transform for Power-Law Spectra}
\label{sxn:RTransformExists}

In this subsection, we examine when the \RTransform $R(z)$ does and does not exist
for power law tails.

\subsubsection{Analyticity criterion}
\label{sxn:RTransformExists:criterion}

Recall the Cauchy–Stieltjes transform defined in Eq.~\eqref{eqn:Cz} of
Sec.~\ref{sxn:r_transforms:elementary_rmt},
\begin{equation}
\label{eqn:RTransformExists_Gdef}
G_\mu(z)=\int_{\mathbb R}\frac{\rho(\lambda)}{z-\lambda}\,
         \mathrm d\lambda,
\qquad
z\in\mathbb C\setminus\operatorname{supp}\mu .
\end{equation}
If $G_\mu(z)$ is \emph{holomorphic} at $z=\infty$ one may invert the map
$z=G_\mu(z)$ in a neighborhood of $z=0$ and define
\begin{equation}
\label{eqn:RTransformExists_Rdef}
R_\mu(z)=G_\mu^{-1}(z)-\frac{1}{z}.
\end{equation}
Obstructions to the existence of $R_\mu$ therefore coincide with
non-analytic terms in the Laurent series of $G_\mu$ about $z=\infty$.
These can be easily removed, however, by considering that $\rho(\lambda)$ is always strictly bounded from above by $\LambdaMax$. 
\subsubsection{Model and notation}
\label{sxn:RTransformExists:model}

We first model a \emph{bare} power-law tail regularized only by a hard
lower cut-off $\LambdaMin>0$:
\begin{equation}
\label{eqn:RTransformExists_density_bare}
\rho_{\alpha}(\lambda)=
(\alpha-1)\,\LambdaMin^{\alpha-1}\,
\lambda^{-\alpha},
\qquad
\lambda\ge\LambdaMin,
\qquad
\alpha\in\{2,3,4\}.
\end{equation}
Normalization is immediate:
\begin{equation}
\label{eqn:RTransformExists_norm}
\int_{\LambdaMin}^{\infty}\rho_{\alpha}(\lambda)\,
      \mathrm d\lambda = 1.
\end{equation}

The choice $\alpha\in\{2,3,4\}$ mirrors the Heavy-Tailed exponents most
frequently observed in neural-network weight and Hessian spectra.

\subsubsection{Stieltjes (Green’s) transform}
\label{sxn:RTransformExists:greens}

For $z\in\mathbb C\setminus[0,\infty)$
\begin{equation}
\label{eqn:RTransformExists_Gbare_def}
G_{\alpha}(z)=
\int_{\LambdaMin}^{\infty}
\frac{\rho_{\alpha}(\lambda)}{z-\lambda}\,
\mathrm d\lambda .
\end{equation}
Carrying out the integration yields

\begin{equation}
\label{eqn:RTransformExists_G2}
G_{(2)}(z)=
\frac{1}{z}+
\frac{\LambdaMin\,\ln\!\bigl(1-\frac{z}{\LambdaMin}\bigr)}{z^{2}},
\end{equation}

\begin{equation}
\label{eqn:RTransformExists_G3}
G_{(3)}(z)=
\frac{1}{z}+
\frac{2\LambdaMin}{z^{2}}+
\frac{2\LambdaMin^{2}\,
      \ln\!\bigl(1-\frac{z}{\LambdaMin}\bigr)}{z^{3}},
\end{equation}

\begin{equation}
\label{eqn:RTransformExists_G4}
G_{(4)}(z)=
\frac{1}{z}+
\frac{3\LambdaMin}{2\,z^{2}}+
\frac{3\LambdaMin^{2}}{z^{3}}+
\frac{3\LambdaMin^{3}\,
      \ln\!\bigl(1-\frac{z}{\LambdaMin}\bigr)}{z^{4}}.
\end{equation}

The logarithmic pieces carry the entire heavy-tail fingerprint; their
placement in the expansion dictates whether $R(z)$ will be available.

\subsubsection{Moments and free cumulants}
\label{sxn:RTransformExists:moments}

Algebraic moments exist only up to order $\alpha-2$:
\begin{equation}
\label{eqn:RTransformExists_mk}
m_{k}=
\int_{\lambda_0}^{\infty}\lambda^{k}\rho_{\alpha}(\lambda)\,
      \mathrm d\lambda
=
\frac{\alpha-1}{\alpha-k-1}\,
\LambdaMin^{k},
\qquad
k<\alpha-1 .
\end{equation}
Hence
\begin{equation}
\label{eqn:RTransformExists_m1m2}
m_{1}=\frac{\alpha-1}{\alpha-2}\,\LambdaMin,
\qquad
m_{2}=\frac{\alpha-1}{\alpha-3}\,\LambdaMin^{2}
\quad\alpha>3).
\end{equation}
The first two free cumulants are $\kappa_{1}=m_{1}$ and
$\kappa_{2}=m_{2}-m_{1}^{2}$.

\subsubsection{R-transform for the bare tail}
\label{sxn:RTransformExists:Rbare}

Define $w=G_{\alpha}(z)$ and solve locally for $R(z)$.  
Using the finite cumulants one obtains

\begin{itemize}
\item \textbf{$\alpha=2$:}  
  the logarithm appears at order $z^{-2}$; $G(z)$ is \emph{not} analytic at
  infinity and the inversion fails.  
  \emph{Conclusion: no $R$–transform.  Truly $1/\lambda^{2}$ tails are
  outside the remit of free addition.}

\item \textbf{$\alpha=3$:}
  \begin{equation}
  \label{eqn:RTransformExists_R3_final}
  R_{(3)}(z)=2\,\LambdaMin
  \quad\text{constant}).
  \end{equation}
  Only a zeroth-order free cumulant survives, but that is \emph{enough}
  for free convolution.

\item \textbf{$\alpha=4$:}
  \begin{equation}
  \label{eqn:RTransformExists_R4_final}
  R_{(4)}(z)=\frac{3}{2}\,\LambdaMin+
             \frac{3}{4}\,\LambdaMin^{2}\,w.
  \end{equation}
  Here the series truncates after the linear term; higher cumulants
  diverge.
\end{itemize}

These three cases show explicitly how \emph{incremental} changes in the
tail exponent alter the analytic status of $G(z)$ and hence of $R(z)$. Since the  higher-order free cumulants diverge, the \RTransform cannot be a finite polynomial. It must either be an infinite series (where the terms beyond a certain point don't vanish) or, more strongly, exhibit non-analytic behavior (like the logarithmic terms present in $G(z)$) because its Taylor series coefficients (the cumulants) become infinite. Fortunately, $R(z)$ can be defined if we ensure that $\rho(\lambda)$ has compact support.
\subsubsection{Truncated \texorpdfstring{$\alpha=2$}{alpha=2} power law}
\label{sxn:RTransformExists:trunc2}

Introduce a cut-off $\LambdaMax>\LambdaMin$ and set
\begin{equation}
\label{eqn:RTransformExists_density_trunc2}
\rho_{\mathrm{tr}}(\lambda)=
C\,\lambda^{-2},
\qquad
\LambdaMin\le\lambda\le\LambdaMax,
\qquad
C=\frac{1}{\LambdaMin^{-1}-\LambdaMax^{-1}} .
\end{equation}
Exact integration gives
\begin{equation}
\label{eqn:RTransformExists_Gtrunc_exact}
G_{\mathrm{tr}}(z)=
C\Biggl[
\frac{\log\LambdaMax-\log(\lambda-z)
      -\log\LambdaMin+\log\LambdaMin-z)}{z^{2}}
-\frac{1}{\LambdaMax\,z}+\frac{1}{\LambdaMin\,z}
\Biggr],
\end{equation}
valid for $z\in\mathbb C\setminus[\LambdaMin,\LambdaMax]$.

\paragraph{Expansion at $z=\infty$.}
Using
$\log(\lambda-z)=\log z+\log\!\bigl(1-\LambdaMax/z\bigr)$ and
$\log(\LambdaMin-z)=\log z+\log\!\bigl(1-\LambdaMin/z\bigr)$,
the two $\log z$ terms cancel and we find the regular series
\begin{equation}
\label{eqn:RTransformExists_Gtrunc_Laurent}
G_{\mathrm{tr}}(z)=
\frac{1}{z}+\frac{m_{1}}{z^{2}}+\frac{m_{2}}{z^{3}}+\dots,
\qquad
z\to\infty,
\end{equation}
with finite moments of all orders.  Consequently
\begin{equation}
\label{eqn:RTransformExists_Rtrunc2}
R_{\mathrm{tr}}(z)=G_{\mathrm{tr}}^{-1}(z)-\frac{1}{w}
\end{equation}
is analytic for $|w|$ small.

\textbf{Interpretation.}  
A truncation at any physically reasonable $\LambdaMax$—for instance the
largest finite eigenvalue observed in a data set—instantly restores full
analyticity at infinity.  
From the point of view of free probability the system now behaves as if it
had \emph{all} moments, even though the raw tail is still $1/\lambda^{2}$
within $[\LambdaMin,\LambdaMax]$.

\subsubsection{Key points and implications for \SETOL}
\label{sxn:RTransformExists:summary}

\begin{enumerate}
\item Cutting the tail at $\LambdaMax=\LambdaECSmax$ \emph{removes} the non-analytic
  $\log z/z^{2}$ obstruction and turns the free-probability machinery back
  on.
\item Any model density $\rho(\lambda)$ with compact support has $G(z)$ analytic at
  $z=\infty$; hence its $R$–transform equals the usual free-cumulant
  series and is available for algebraic manipulation.
\item In all theoretical derivations and numerical experiments in
  \SETOL\ we therefore \textbf{model empirical spectra as effectively as \emph{truncated}
  power laws} (i.e with finite bounds, not necessarily exponentially truncated).
  This choice is both empirically justified (no spectrum is truly
  infinite) and mathematically essential: it guarantees that $R(z)$
  \emph{always} exists.
\end{enumerate}

\subsubsection{Explicit \texorpdfstring{$R$}{R}–transforms for the truncated tail}
\label{sxn:RTransformExists:explicit}

We can also provide exact expressions for the leading terms (free cumulants) in $R(z)$ for $alpha=2,3,4$.
The compact support $[\LambdaMin,\LambdaMax]=[\LambdaECSmin,\LambdaECSmax]$ guarantees that every
algebraic moment (and free cumulant) is finite. Hence the Voiculescu series 

\begin{equation}
R_{\alpha}^{\mathrm{tr}}(z)=
\sum_{n=1}^{\infty}\kappa{_\ND}\,z^{\,n-1},
\qquad |z|\text{ small},
\end{equation}

\noindent converges for every $\alpha>1$.  
Below we list the free cumulants $\kappa_{1}$, $\kappa_{2}$ and the resulting
$(R(z)$ for the three exponents most relevant to \SETOL.
Higher cumulants follow from the recursion in
Eqs.~\eqref{eqn:RTransformExists_mk}–\eqref{eqn:RTransformExists_m1m2} and
need not be written out.

---

\paragraph{$\alpha = 2$.}

\begin{equation}
C_{2}=\frac{\LambdaMin\LambdaMax}{\LambdaMax-\LambdaMin},
\end{equation}

\begin{equation}
\kappa_{1}=C_{2}\,\log\!\frac{\LambdaMax}{\LambdaMin},
\end{equation}

\begin{equation}
\kappa_{2}=C_{2}\,\LambdaMax-\LambdaMin)-\kappa_{1}^{2},
\end{equation}

\begin{equation}
R_{(2)}^{\mathrm{tr}}(z)=
\kappa_{1}+\kappa_{2}\,z+\mathcal{O}\!\left(z^{2}\right).
\end{equation}

\paragraph{$\alpha = 3$.}

\begin{equation}
C_{3}=\frac{2\,\LambdaMin^{2}\LambdaMax^{2}}{\LambdaMax^{2}-\LambdaMin^{2}},
\end{equation}

\begin{equation}
\kappa_{1}=\frac{2\,\LambdaMax\LambdaMin}{\LambdaMax+\LambdaMin},
\end{equation}

\begin{equation}
\kappa_{2}=C_{3}\,\log\!\frac{\LambdaMax}{\LambdaMin}
-\kappa_{1}^{2},
\end{equation}

\begin{equation}
R_{(3)}^{\mathrm{tr}}(z)=
\kappa_{1}+\kappa_{2}\,z+\mathcal{O}\!\left(z^{2}\right).
\end{equation}

\paragraph{$\alpha = 4$.}

\begin{equation}
C_{4}=\frac{3\,\LambdaMin^{3}\LambdaMax^{3}}{\LambdaMax^{3}-\LambdaMin^{3}},
\end{equation}

\begin{equation}
\kappa_{1}=
\frac{3\,\LambdaMax\LambdaMin\,(\LambdaMax^{2}-\LambdaMin^{2})}
     {2\,(\LambdaMax^{3}-\LambdaMin^{3})},
\end{equation}

\begin{equation}
\kappa_{2}=C_{4}\,
\left(\frac{1}{\LambdaMin}-\frac{1}{\LambdaMax}\right)
-\kappa_{1}^{2},
\end{equation}

\begin{equation}
R_{(4)}^{\mathrm{tr}}(z)=
\kappa_{1}+\kappa_{2}\,z+\mathcal{O}\!\left(z^{2}\right).
\end{equation}

---

\textbf{Interpretation.}  
$\kappa_{1}$ fixes the mean scale of the heavy tail;
$\kappa_{2}$ sets its leading spread.  
Because both depend only on the empirical cut‑offs
$\LambdaMin$ and $\LambdaMax$, the two–term truncation already delivers
an accurate $R$–transform for free‑probability manipulations inside the
\SETOL\ framework.
\subsection{The Inverse-MP (IMP) Model}
\label{sxn:IMP}
In this section, we rederive the integral $G(\lambda)[IMP]$ for the \InverseMP (IMP) model, focusing on the branch cut starting at $z = \kappa/2$ and extending to infinity. 
This branch cut corresponds to the support of the ESD in this region. 
We will:
\begin{enumerate}
\item Explain the presence of the branch cut and its implications.
\item Show that $R(z)[IMP]$ becomes complex along this branch cut because the term under the square root becomes negative.
\item Perform the integral $G(\lambda)[IMP]$, showing all steps.
\item Compute the \emph{Real} part  $\Re[G(\lambda)[IMP]]$ 
\end{enumerate}

\subsubsection{The Branch Cut in the IMP Model}

The R-transform for the IMP model is given by:
\begin{align}
\label{eqn:iw_r_transf}
R(z)[IMP] = \frac{\kappa - \sqrt{\kappa(\kappa - 2z)}}{z},
\end{align}
where $\kappa > 0$ is a parameter related to the dimensions of the random matrices under consideration.
The function $\sqrt{\kappa(\kappa - 2z)}$ introduces a branch point at $z = \kappa/2$ because the argument of the square root becomes zero at this point:
\begin{align}
\kappa - 2z = 0 \quad \Rightarrow \quad z = \frac{\kappa}{2}.
\end{align}
For $z > \kappa/2$, the argument $\kappa - 2z$ becomes negative, and thus the square root becomes imaginary. 
This leads to a branch cut starting at $z = \kappa/2$ and extending to $z = \infty$ along the real axis. 
This branch cut affects the analyticity of $R(z)[IMP]$, and it must be carefully considered in the integral $G(\lambda)[IMP]$.

\subsubsection{\texorpdfstring{$R(z)[IMP]$}{R(z)[IMP]} is Complex Along the Branch Cut}

For $z > \kappa/2$, we have:
\begin{align}
\kappa - 2z < 0 \quad \Rightarrow \quad \sqrt{\kappa(\kappa - 2z)} = \sqrt{-\kappa(2z - \kappa)} = i \sqrt{\kappa(2z - \kappa)}.
\end{align}
Therefore, $R(z)[IMP]$ becomes complex:
\begin{align}
R(z)[IMP] = \frac{\kappa - i \sqrt{\kappa(2z - \kappa)}}{z} = \frac{\kappa}{z} - i \frac{ \sqrt{ \kappa(2z - \kappa) } }{ z }.
\end{align}
This expression shows that $R(z)[IMP]$ has both real and imaginary parts when $z > \kappa/2$.

\subsubsection{Calculation of \texorpdfstring{$G(\lambda)[IMP]$}{G(lambda)[IMP]}}

We aim to compute the integral:
\begin{align}
G(\lambda)[IMP] := \int_{\LambdaCutoffMin}^\lambda \Re[R(z)[IMP]] , dz,
\end{align}
where $\LambdaCutoffMin \geq \kappa/2$.

Where the Real part of $R(z)[IMP]$ is
\begin{align}
\text{Re}[R(z)[IMP]] = \frac{\kappa}{z}.
\end{align}
The integral of this is
\begin{align}
\Re[G(\lambda)[IMP]] = \int_{\LambdaCutoffMin}^\lambda \frac{\kappa}{z} , dz = \kappa \left[ \ln z \right]_{\LambdaCutoffMin}^\lambda = \kappa \left( \ln \lambda - \ln \LambdaCutoffMin \right).
\end{align}


\end{document}